\documentclass[graybox,envcountchap,sectrefs]{svmono}

\usepackage{mathptmx}

\usepackage{makeidx}         

\usepackage{graphicx,epstopdf}        

\usepackage{epstopdf}
\usepackage{multicol}        
\usepackage[bottom]{footmisc}

\usepackage{newtxtext}       %
\usepackage[varvw]{newtxmath}       

\usepackage{soul}               

\usepackage{hyperref}        

\newcommand{\lingform}[1]{\textit{\texttt{#1}}}
\newcommand{\lingformquote}[1]{\texttt{#1}}
\newcommand{\code}[1]{\texttt{#1}}
\newcommand{\semantic}[1]{\texttt{#1}}
\newcommand{\scare}[1]{`#1'}
\newcommand{\blankline}{\vspace{\baselineskip}}
\newcommand{\lineacross}{\rule{\linewidth}{1pt}}

\newcommand{\personal}[1]{\medskip\noindent\fbox{\begin{minipage}{\textwidth}\textbf{Personal Note:} #1\end{minipage}}\medskip}

\makeindex             

\begin{document}

\author{Ehud Reiter}
\title{Natural Language Generation}
\maketitle

\frontmatter

%
%

\extrachap{Acknowledgement}

This book has been published by Springer and is available at \url{https://doi.org/10.1007/978-3-031-68582-8}  .  Springer has kindly agreed to let me put the book preprint (which does not include improvements made by Springer) onto Arxiv.

Please cite this book as below \cite{reiter2024}:
\blankline

Ehud Reiter (2024).  \textit{Natural Language Generation}. Springer

\blankline

There is a companion website at \url{https://ehudreiter.com/book/}

\blankline

\textit{Formal acknowledgement}:\\
This is a preprint of the following work: Ehud Reiter, Natural Language Generation, 2024, Springer reproduced with permission of Springer Nature Switzerland AG.  The final authenticated version is available online at:  http://dx.doi.org/10.1007/978-3-031-68582-8   .

%
%

\preface

I have been working on natural language generation (NLG), that is using artificial intelligence techniques to produce texts in English and other human languages, since I got my PhD in this area in 1990.  In late 2022, NLG became much more prominent because of the impressive capabilities of large language models such as ChatGPT, which was exciting.  However, discussions about NLG in both academic and commercial circles have become focused on the latest developments in language models, with little attention was paid to what had been learned about NLG before ChatGPT.

My \textbf{goal} in this book is to present a broad overview of NLG which talks about language models, but also looks at alternative approaches to NLG, requirements (what users want NLG systems to do), evaluation, safety and testing, and sensible applications of NLG.  I hope that this broad perspective, which builds on decades of work on NLG, will be helpful to both researchers and developers who work in this area.

I co-authored a book about NLG in 2000, and I saw that while the technology content of my 2000 book quickly became out of date, people kept on using it in 2010 and even 2020 because the high-level conceptual and methodological material was still useful.  With this experience in mind, I have focused this book on high-level concepts and methodologies; my hope is that this material will still be useful in 2030 and perhaps even 2040.  I do not attempt to describe the latest technologies, because this information quickly becomes dated (indeed, anything I write in June 2024 will probably be out-of-date by the time the book is published); readers interested in the latest developments in language models should look elsewhere.

In many places I show outputs from ChatGPT and other large language models.  Most of these were all produced in 2023 (I deliberately do not include version numbers or specific dates), and readers should bear in mind that models in 2025, let alone 2030, may produce different outputs.  However, the high-level points I am making should still be valid.

I also focus in this book on my own experiences.  Where possible I use examples from systems which I have worked on or otherwise been involved with, even if they are not the best known system in their area; for instance for this reason I talk about the (somewhat obscure) BLOOM language model as well as better-known ones such as GPT.
More generally the book focuses on data-to-text NLG (systems which use NLG to summarise and explain non-linguistic data), because this is my personal interest.  I also include personal notes throughout the book.   I hope this personal focus makes the book more interesting to readers.

Specifically, the book has the following \textbf{chapters}:
\begin{itemize}
\item \emph{Introduction:} I present some example NLG systems, summarise the content of the rest of the book, and also give a short history of NLG.
\item \emph{Rule-based NLG:} I describe how AI systems can generate texts using algorithms and rules which explicitly make decisions about the content and language of generated texts.  Rule-based NLG has been overshadowed by neural NLG in recent years, but it is still the best way to build some NLG applications.  Rule-based NLG also shows the types of decisions which need to be made in text generation, and I think a good understanding of this helps anyone working in NLG, even if they use other approaches.
\item \emph{Machine learning and neural NLG:} I give an overview of machine learning and neural approaches to NLG, including language models.  This area is changing very rapidly, and models which were exciting state-of-art a few years ago are now obsolete and forgotten.  Because of this, I just give a high-level overview of basic concepts behind models, and then discuss data and other issues which are important regardless of the model used.
\item \emph{Requirements:} As with any type of software, knowing what users and stakeholders are looking for is essential in building a successful NLG application.  I look at some of the different quality criteria that people may care about, workflows for using NLG (including \scare{human-in-loop}), textual vs graphical presentation of information, and methodologies for understanding (acquiring) requirements.
\item \emph{Evaluation:} This is the longest chapter of this book, which reflects my interest in the topic as well as its importance.  From both a scientific and practical perspective, it is essential to evaluate how well NLG algorithms, models, and techniques work, using experiments which are rigorous and replicable.  I discuss basic evaluation concepts, and then describe techniques for human and automatic (metric) evaluation.  I also look at evaluating real-world impact of NLG systems, as well as commercial evaluation.
\item \emph{Safety, testing, and maintenance:}  Society expects that AI systems used in the real-world will be safe (not harm users or third parties); systems which are not safe will not be allowed by governments and regulators.  I examine safety concerns and techniques in NLG, and also look at software testing, which is used to identify bugs and other problems which could lead to unacceptable behaviour.  I conclude with a section on maintaining NLG systems, which is very important (most of the lifecycle costs of software systems are in maintenance) but poorly understood.
\item \emph{Applications:} NLG is not just an academic discipline, it is also a technology which can be used to build useful applications which help people.  In this chapter I discuss some fundamental issues (such as scalability), and then look in more detail at four areas which NLG has been used commercially for many years: journalism, business intelligence, summarisation, and medicine.  Lessons from these long-standing NLG use cases can be applied to newer applications of NLG.
\end{itemize}

It is impossible for me to \textbf{acknowledge} all of the people who have helped me in my NLG career. I would like to give special thanks to my faculty colleagues in the Aberdeen NLG group over the years, including  Kees van Deemter, Jim Hunter,  Ruizhe Li, Chengua Lin, Judith Masthoff, Chris Mellish, Graeme Ritchie, Advaith Siddharthan,  Arabella Sinclair, Yaji Sripada, and  Wei Zhou.  I'd also like to thanks the many PhD students and post-docs I have worked with at Aberdeen, including Jawwad Baig, Simone Balloccu, Daniel Braun, Martin Dempster, Albert Gatt, Rodrigo Gomes de Oliviero, Francois Portet, Stephanie Inglis, Kitt Kuptavanich, Jing Lin, Saad Mahamood, Meg Mitchell, Wendy Moncur, Francesco Moramarco, Joe Reddington, Roma Robertson, Jaime Sevilla, Adarsa Sivaprasad, Mengxun Sun, Barkavi Sundararajun, Iniakpokeikiye Thompson, Craig Thomson, Nava Tintarev, Ross Turner, Sandra Williams, and Jin Yu.  The Aberdeen University NLG group is by no means the largest NLG group in the world, but it has provided a great environment for me to pursue my interests.

Of course funding is very useful for university research, and I am very grateful to Australian Research Council, Cancer Research UK, the EU Horizon and Marie-Curie programmes, the Royal Society of Edinburgh, the Scottish Chief Scientist Office, and the UK Economic and Social Research Council for their support.  Special thanks to the  the UK Engineering and Physical Sciences Research Council,  which has funded my projects over a 30-year period.  I am also grateful to IBM and the US National Science Foundation for supporting my PhD studies at Harvard.

In 2009 I co-founded a company, Data2Text, which was acquired by Arria NLG in 2013.  I worked at Data2text and Arria for many years, and my experiences there shaped much of the content of this book.  At Arria, I am especially grateful to John Alexander, Neil Burnett, William Bradshaw, Robert Dale, Sharon Daniels, Ian Davy, Jay Dewalt, Cathy Herbert, Alasdair Logan, Lyndsee Manna, Daniel Paiva, John Perry, Kapila Ponnamperuma, Jette Viethen, and Keith Wisemann for their help in understanding the commercial perspective on NLG, as well as many colleagues from Aberdeen University (mentioned above) who spent time at Data2text or Arria.

Numerous people have helped me with this book.  I'd especially like to thank the people who reviewed and gave comments on draft material: Kees van Deemter, Nick Diakopoulos, Albert Gatt, Emiel Krahmer, Emiel van Miltenburg, Verena Reiser, and Ross Turner.  Extra-special thanks to Saad Mahamood, who reviewed and commented on a draft of the complete book. Of course I alone am responsible for any mistakes in this book!

Finally, I am grateful to my wife Ann and my children Miriam, Moshe, and Naomi for putting up with me while I worked on this book!

\vspace{\baselineskip}
\begin{flushright}\noindent
Aberdeen, Scotland\\
June 2024\hfill {\it Ehud Reiter}\\
\end{flushright}

\tableofcontents

\mainmatter

%
%
%
\chapter{Introduction to NLG}
\label{intro} 

This book gives my perspective on Natural Language Generation (NLG), that is using artificial intelligence (AI) techniques to produce texts in English and other human languages; it draws on
my decades of experience in the field.  I wrote a book (with Robert Dale) on NLG in 2000 \cite{reiterdale2000}, when I had limited experience building NLG systems.  Over the past 25 years I have gained a lot more experience building systems, and this book largely reflects the issues and challenges I faced; most of the examples come from systems I have worked on.  It draws on my experiences at the company Arria NLG (which evolved from a spinout company, Data2text, which I co-founded in 2009) as well as my academic research at the University of Aberdeen.

This book is intended to help students, developers, and researchers understand the fundamental concepts of NLG.  It is not intended to be an up-to-date guide to the latest language model technologies, research papers and commercial offerings.
In particular, the book discusses technology only at a high level because technology changes very fast, hence detailed material on technology would quickly become out of date.  The book's main focus is on fundamental aspects of NLG which change more slowly than technology, including component tasks of NLG, requirements of NLG systems, evaluation, safety and testing, and applications. These topics are essential for success (at least in my experience) but are less often discussed than technology; I hope this book can help readers understand these aspects of NLG.

The book focuses on systems which generate output texts that describe, summarise, explain, etc. the system's input data.  In other words, from the perspective of generative AI systems such as ChatGPT\index{ChatGPT} and Google Gemini\index{Gemini}, it focuses on applications where an input is transformed into an output (see examples later in this chapter), not applications where the user asks a question and the system synthesises an answer from Internet sources

\section{What is Natural Language Generation}
\label{sec:IntroWhatIsNLG}

Natural Language Generation (NLG) systems use artificial intelligence and natural language processing techniques within software systems that generate texts in human languages such as English, Chinese, and Arabic.  In other words, NLG is the science of AI systems that can write. As such it is related to (but not the same as) Natural Language Understanding\index{natural language understanding} (NLU), which is the science of AI systems that can read and extract meanings from human-written texts.

\begin{figure}
\lineacross\\
\includegraphics[scale  = 0.5]{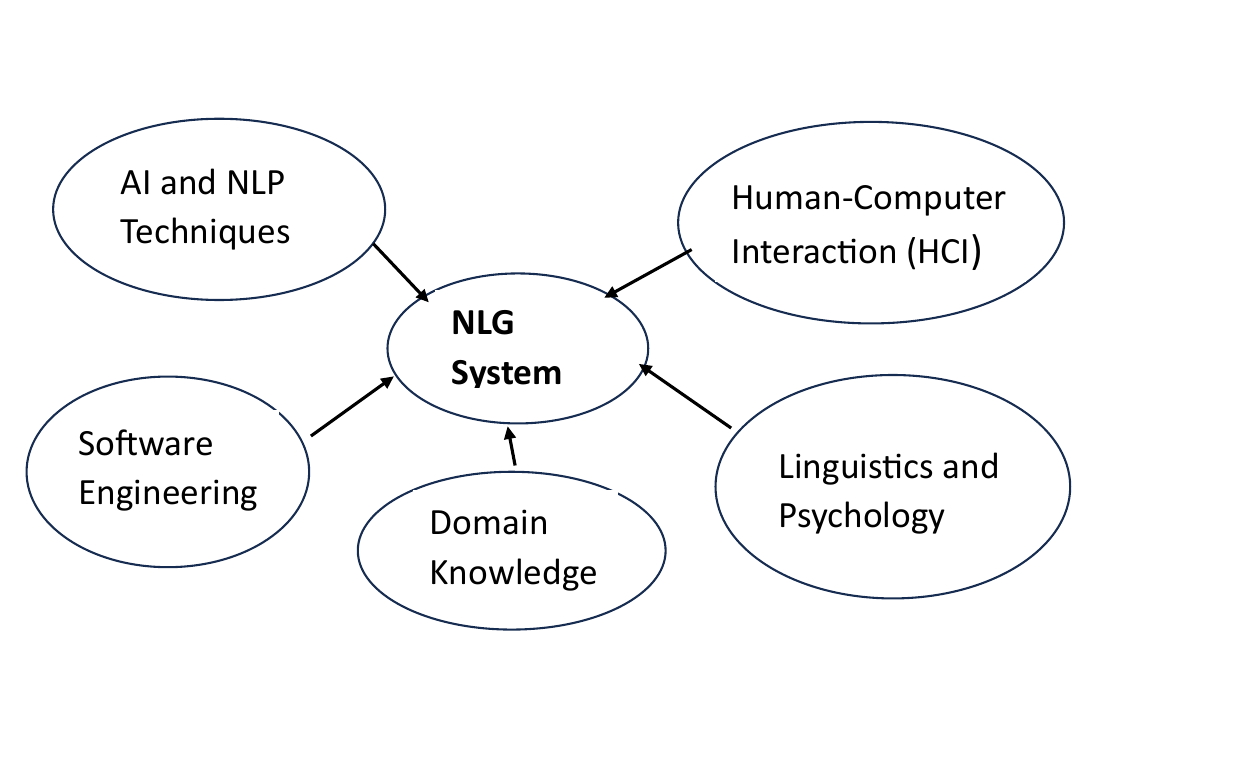}
\caption{Skills needed for NLG}
\lineacross
\label{fig:nlgskills}       
\end{figure}

Natural language generation has recently become more prominent because of the success of ChatGPT\index{ChatGPT} and other generative language models, but the field has been around for decades.   AI and NLP techniques such as machine learning and language models are widely used, but they do not tell the whole story.  NLG researchers and developers should understand (Figure~\ref{fig:nlgskills}):
\begin{itemize}
\item \emph{AI and NLP techniques}: technology used to generate texts.
\item \emph{Human-computer-interaction (HCI)\index{HCI}}: how NLG systems can effectively interact with users.
\item \emph{Linguistics\index{linguistics} and psychology\index{psychology}}: how to best use language to communicate with users.
\item \emph{Software engineering\index{software engineering}:} how to build real-world usable software systems.
\item \emph{Domain knowledge\index{domain!knowledge}}: what is important in the target domain, and how it should be communicated.
\end{itemize}

\personal{I have found that research is much more rewarding if it takes the larger context into consideration.  Certainly in my own career I have enjoyed learning about the above areas and collaborating with all sorts of people, ranging from HCI\index{HCI} and usability\index{usability} experts to domain experts\index{domain!experts} in offshore oil platforms\index{oil industry}.   Many of my most influential research papers have benefitted from collaborations with and insights from HCI\index{HCI} experts \cite{knoll-etal-2022-user}, psychologists\index{psychology} and linguists\index{linguists} \cite{mckinlay2010design}, software engineers\index{software engineering} \cite{reiter-2017-commercial}, and domain experts\index{domain!experts} \cite{PORTET2009789}.}

From a practical perspective, NLG is used in many different contexts.  My focus in this book is on systems which generate texts which describe, explain, summarise (etc.) an input to the system.   The input can be numbers, in which case the NLG system is a \emph{data-to-text}\index{data-to-text} system (see example in Section~\ref{ArriaWeatherExample}); the input can also be words, in which case the NLG system is a \emph{text-to-text}\index{text-to-text} system (see example im Section~\ref{IntroConsultationSummary}).
Machine translation\index{machine translation} (MT) is sometimes considered to be a type of NLG, and I will occasionally refer to MT, especially when discussing evaluation.  However, readers interested in MT should look at a specialised book such as Koehn's \emph{Neural Machine Translation} \cite{koehn2020}.   I also will not discuss writing assistants\index{writing assistants} (sentence completion, grammar improvements, etc) or tools for generating fictional content\index{fiction generation}.
I discuss using generative language models to produce text, but I do not discuss other capabilities of these models such as language understanding and image generation\index{image generation}.

\section{Example: Weather Forecast (\emph{Data-to-Text})}\label{ArriaWeatherExample}
One of the earliest use cases for NLG was generating textual weather forecasts\index{weather forecasts} from numerical weather predictions.  The numerical weather predictions are produced by atmosphere simulations and give predicted weather parameters (e.g., wind speed, precipitation) at different points and times.  The NLG system takes this information and produces a textual forecast which summarises in words how the weather will develop.
As described below, using NLG to produce forecasts makes it easier to produce specialised forecasts from the same numerical weather prediction data; for example forecasts for different locations or for different types of users.

Many NLG weather forecast systems have been developed over the years, with the first production system entering operational usage in 1992 \cite{Goldberg1994} (Figure~\ref{fig:FoG}).  More recently there has been considerable interest in using neural NLG techniques to allow dialogue systems to respond to weather queries (Section~\ref{sec:Facebookweather}).

Forecast generation is a type of \emph{data-to-text}\index{data-to-text} NLG, that is a system which generates text that summarises a numeric (or symbolic) data set.  In this case, the input data is a large set of numbers which describe wind speed, precipitation, etc.; symbols are used to describe some weather parameters such as wind direction (E, NE, N, NW, etc).   For some types of weather forecasts, the input data set can consist of millions of numbers and symbols \cite{turner-etal-2009-generating}.  The NLG system \scare{digests} and analyses this data, and generates a text which summarises key facts and insights for the user.

Because NLG researchers have worked on generating weather forecasts for over 30 years, it is a good domain to understand NLG issues and technologies, and this book will refer to weather forecast generation in many places.

\begin{figure}
\lineacross\\
\includegraphics[scale  = 0.6, trim = {0.5cm 0 1cm 0},clip]{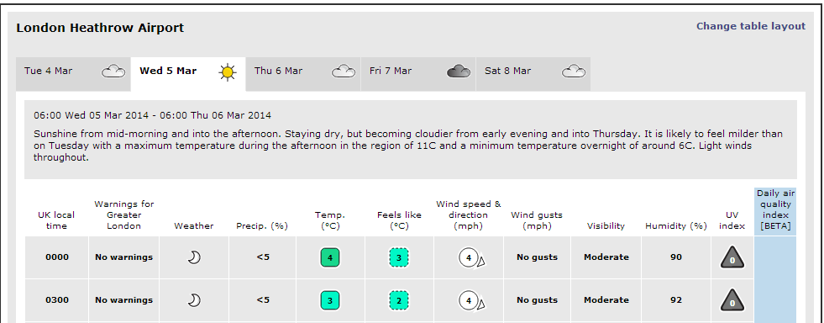}
\caption{Weather forecast from Arria system; Figure 1 from \cite{sripada-etal-2014-case}}
\label{fig:ArriaMet}      
\lineacross 
\end{figure}

\subsection{Use Case: Point weather forecasts for general public}\label{sec:Arriaweather}
My company, Arria\index{Arria}, developed a simple forecast generator for the UK Met Office \cite{sripada-etal-2014-case} in 2014.   It is by no means the most sophisticated NLG system in this space, but it does illustrate weather forecast generation and data-to-text NLG.  The Arria system is not being used at the time of writing, but rule-based techniques are still widely used to produce simple weather forecasts.

This system generates a \emph{point forecast}, that is a weather forecast for a particular location over a time period (such as Heathrow airport in London). \emph{Area forecasts}, in contrast, describe the weather over a region (such as Scotland); NLG systems which generate area forecasts (such as \cite{turner-etal-2009-generating}) tend to be considerably more complex than systems which generate point forecasts.

The Arria system  produced \emph{public forecasts}, that is forecasts which are intended for the general public.  There are also of course many specialised forecasts, for farmers, pilots, sailors, etc; these emphasise different types of information (for example, aviation forecasts describe wind speeds at different altitudes; public forecasts do not). They also often used specialised terminology\index{terminology} (for example, marine forecasts use the verb \lingform{backing} to describe a counter-clockwise change in wind direction).

The commercial goal of the Arria system was to produce weather forecasts for 5,000 different locations; this would be a daunting task for human forecasters.   In general the benefits of NLG are often in allowing large numbers of specialised texts to be produced, so that users can receive texts which are specific to their needs, instead of generic texts.  In other words, an Aberdeen-specific forecast is of more use to me than UK-wide forecast, but producing textual forecasts for the thousands of towns and villages in the UK is difficult without automation.

Figure~\ref{fig:ArriaMet} shows an example public forecast produced by the Arria system for London's Heathrow Airport, and some of the data used to produce this forecast.  The forecast itself is 
\begin{quotation}
Sunshine from mid-morning and into the afternoon.  Staying dry, but becoming cloudier from early evening and into Thursday.   It is likely to feel milder than on Tuesday with a maximum temperature during the afternoon in the region of 11C and a minimum temperature overnight of around 6C.  Light winds throughout.
\end{quotation}

The data used to produce the forecast (ie, the output of the numerical weather model) is a table which shows the values of different weather parameters at different times, for this location.  Part of the table is shown beneath the forecast in the Figure~\ref{fig:ArriaMet}, for example the Temperature at 0000 is 4C.

\subsection{Technology: Rule-based NLG\index{rule-based NLG}}\label{sec:d2tpipeline}

The Arria system uses rules to generate the text (not machine learning).  Details are not given in  \cite{sripada-etal-2014-case}, but in rough terms the rules do the following
\begin{itemize}
\item \emph{Stage 1: Find patterns in data (signal analysis\index{signal analysis}).}  For example, the probability of precipitation is very low (less than 5\%) for the entire period.
\item \emph{Stage 2: Generate insights (data interpretation\index{data interpretation}).}  For example, this will be a dry day.  Insights depend on what the user is interested in.  
For instance, as mentioned above, an aviation forecast will describe wind speed at different altitudes, since this this very important to pilots.  In contrast, a public forecast probably will only describe wind speed at ground level, since this is all the general public cares about.
\item \emph{Stage 3: Decide which insights to present, how they are ordered, and how they are related to each other (document planning\index{document planning}).}  For example, adding a contrast link (expressed as \lingform{but}) between the low-precipitation insight and the insight that cloud cover increases.
\item \emph{Stage 4: Decide how to express insights linguistically, in a stylistically appropriate manner (microplanning\index{microplanning}).} For example, deciding to use the gerund phrase \lingform{staying dry} to communicate the low-precipitation insight.
\item \emph{Stage 5: Generate grammatically correct text (surface realisation\index{surface realisation}).}  For example, capitalising \lingform{Staying} since it is the first word of the sentence.
\end{itemize}
All data-to-text\index{data-to-text} systems must perform the above types of processing in some fashion.  That is, they must find patterns in the data, generate insights which are useful to the user, order and link insights, express insights as words, and deal with grammatical details.  The Arria system does these steps explicitly, using rules.

At the other extreme, a neural model can be trained to go from input data to output text in a single step; but this model would still need to make decisions about patterns, insights, document structure, linguistic expression, and grammar conformance, even if these decisions are buried in a neural network instead of expressed explicitly as rules.  Another approach is to mix technologies, for example to use algorithms and rules to find patterns and generate insights, and a neural model to select and express insights in grammatically correct text.

\subsection{Evaluation}
The Arria system was evaluated by putting it onto a website used by the Met Office to test new technologies, and asking users to rate the usefulness\index{utility} and desirability of the forecasts; this is an example of a \emph{human evaluation based on ratings\index{rating}} (Section~\ref{sec:humanevalrankings}).
Users who submitted ratings were overall very positive, however we need to keep in mind that people who looked at the Met Office \scare{new technology} site may not be representative of Met Office users more generally.
User ratings are often used to evaluate NLG systems, but its considered best practice to try to get a diverse and representative set of users to do this.

This and other points related to evaluation of NLG systems are discussed in detail in Chapter~\ref{evaluation}.

\section{Example: Summarising Consultations (\emph{Text-to-Text})}\label{IntroConsultationSummary}
Weather forecasts are an example of data-to-text NLG, where the input to the NLG system is structured data.   In contrast, \emph{text-to-text}\index{text-to-text} NLG systems generate texts from other texts.   One example is \emph{summarisation}\index{summarisation} systems, which generate texts which summarise input texts.

A concrete example is generating texts which summarise consultations between professionals and their clients.
Such summaries are often useful as a record of what was discussed and decided; in some contexts they are legally required.

\subsection{Use Case: Generating Summaries of Doctor-Patient Consultations}

\begin{figure}
\lineacross\\
\includegraphics[scale  = 0.75]{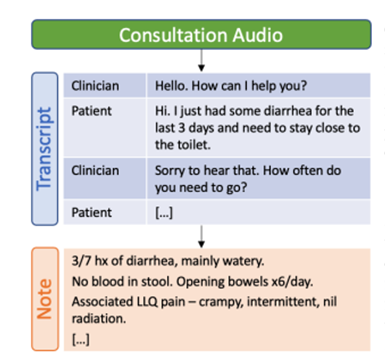}
\caption{Example of Note Generator; Figure 1 from \cite{knoll-etal-2022-user}}
\label{fig:NoteGenerator}       
\lineacross
\end{figure}

Babylon, a UK digital health company, developed the Note Generator\index{Note Generator} system \cite{knoll-etal-2022-user} which generated summaries of  doctor\index{doctors}-patient consultations\index{consultations}.  In the UK and many other countries, whenever a doctor talks to a patient, a summary of this consultation needs to be added to the patient's medical record.  In the past these summaries were written manually, either by the doctor or by a medical scribe who listens to a recording of the consultation.   Interest is growing in automating at least part of the process of summarising a consultation.

Two goals of the Babylon Note Generator system are (A) reduce the amount of manual effort required to create the summary and (B) enable doctors to focus on the patient on the consultation without the distraction of taking notes \cite{knoll-etal-2022-user}.
Note Generator is no longer used\footnote{Babylon unfortunately went bankrupt in 2023}, but similar systems are being developed and used by other healthcare companies at the time of writing, such as Abridge\footnote{\url{https://www.abridge.com/}}.  

A simple example is shown in Figure~\ref{fig:NoteGenerator}.   The consultation is recorded and transcribed by a speech-to-text system; a short extract of a transcript is shown in the Transcript box in Figure~\ref{fig:NoteGenerator}.   The NLG system then produces a summary of the transcript for the patient record (shown in the Note box in Figure~\ref{fig:NoteGenerator}); this communicates key medical information in standardised medical language.  The summary is shown to the doctor, who can edit\index{post-editing} it to add missing information, correct mistakes, etc. before the summary is added to the patient record.

Of course some consultations are more difficult than others; for example it is difficult to write up a consultation when the doctor suspects that the patient may not be telling the truth.  Note Generator focuses on summarising straightforward consultations (which are the majority); doctors still need to manually write notes for difficult cases, especially if information needs to be added beyond what was said in the consultation.  This is common in NLG and indeed AI; the NLG system helps the user with \scare{straightforward} cases, but provides less assistance in more complex, unusual, or novel situations.

\subsection{Technology}\label{sec:notegentech}
The Note Generator system uses \emph{neural language models}\index{language models} to generate the summaries.  A neural language model essentially is a machine learning system which is trained on (input, output) pairs to perform a task.  For the Note Generator, the developers started with a standard model used for summarisation (BART\index{BART} \cite{lewis-etal-2020-bart}) which had been trained on almost 300,000 pairs of news articles and their summaries.  Of course summaries of medical consultations are very different from summaries of news articles, so the developers adapted the model by \emph{fine-tuning}\index{fine-tuning} it on 10,000 examples of previous doctor-patient consultations  which had already been manually summarised and entered into Babylon's patient record system.  This is a common strategy in neural NLG; we train a model on a large data set which is only loosely related to the application, and then use a smaller data set to adapt the model to the target domain and application.

From a summarisation perspective, one unusual aspect of Note Generator was that doctors wanted summaries to be generated incrementally, so that at any point in time there was a summary of the consultation so far \cite{knoll-etal-2022-user}; this is an example of a human-computer interaction constraint (Figure~\ref{fig:nlgskills}).  Previous summarisation systems generated a complete summary from a complete input document, so the team essentially trained the model to generate incremental additions to an existing summary.   The model always added content to the summary, it never changed what the doctors had already seen.  This was novel technology (the team applied for a patent) and shows how taking HCI\index{HCI} considerations seriously can lead to new NLG technology.

\subsection{Evaluation}
The Note Generator System was evaluated in a number of different ways \cite{knoll-etal-2022-user,moramarco-etal-2022-human,moramarcophd}:
\begin{itemize}
\item \emph{Automatic metrics\index{metrics}}: the system was run on a \emph{test set} of historical consultations (which had not been used to train or fine-tune the model), and several different metrics (algorithms) were used to assess how similar the system's output was to the manually-written summary in the historical report.  This is a very common type of evaluation in NLP, which is discussed in Section~\ref{sec:automaticevaluation}.  Such evaluations are easy to do but unfortunately are not always meaningful.  In this case, the scores returned by the metrics were not  good predictors of actual utility \cite{moramarco-etal-2022-human}.
\item \emph{Error analysis\index{error analysis}:} The team asked doctors to find mistakes in the generated summaries, such as incorrect statements, omitted content, and misleading statements.   This information was very useful in itself for understanding where the system worked well and where it did not, and it also was a good predictor of utility.   Evaluations via error analysis are discussed in Section~\ref{sec:humanevalannotations}.
\item \emph{Post-edit\index{post-editing} time:} The team asked doctors to edit summaries and fix mistakes, and timed how long this took; post-editing time is one the most important KPI (Key Performance Indicator) for real-world success.  Not surprisingly, the time taken to post-edit was correlated with the number of errors.  This is an example of a \emph{task-based} evaluation\index{task-based evaluation}, where we measure the impact on NLG system has on human task performance; such evaluations are discussed in Section~\ref{sec:humantaskeval}.
\item \emph{Real-world utility\index{utility}:} Last but certainly not least, after the system was deployed and used in real consultations, the team measured how long it took doctors to post-edit notes in real consultations, and how this compared to the time needed to write notes manually before the system was deployed.  They also assessed the number of errors\index{errors} in post-edited Note Generator summaries (from real consultations), and compared this to the number of errors in manually-written summaries. Evaluations of real-world utility and impact\index{impact} are discussed in Section~\ref{sec:impactevaluation}.
\end{itemize}

Using a broad mix of evaluation techniques is good practice, and gives us a better perspective on how well a system is performing.

\section{Technologies}
The first part of this book discusses technologies used to build NLG systems.  Because NLG technology is changing very fast, the discussion is high-level and conceptual; readers interested in details of the latest research paper or commercial product should look at other sources for this information.

\begin{figure}
\lineacross\\
\includegraphics[scale  = 0.5]{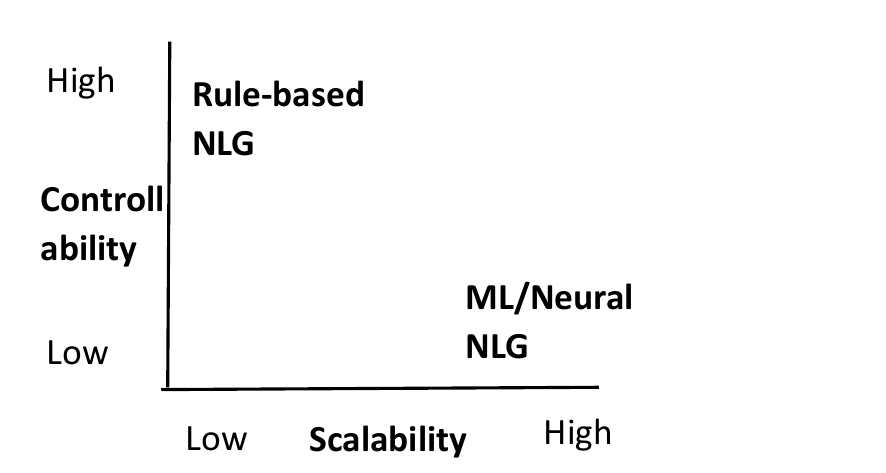}
\caption{Scalability and controllability of different NLG technologies.}
\label{fig:nlgtech}       
\lineacross
\end{figure}

At the most basic level, there are two ways of building NLG functionality: we can write  \emph{rules} and code which execute NLG algorithms, or we can use \emph{machine-learning} models (usually based on neural deep-learning\index{deep learning} technology), which may be trained or fine-tuned on the application domain.  In very crude terms, these technologies involve different trade-offs between the following (Figure~\ref{fig:nlgtech}):
\begin{itemize}
\item \emph{Scalability\index{scalability}:} How much effort is required to build a large-scale NLG system which produces sensible output texts for a wide variety of inputs?
\item \emph{Controllability\index{controllability}:} Can developers and users precisely control what the NLG system does?  A related criteris is \emph{Transparency}\index{transparency}; do developers and users understand how and why the NLG system produces a text?
\end{itemize}
Rule-based NLG\index{rule-based NLG} systems are very controllable and transparent, but large-scale wide-coverage NLG systems may require thousands of rules; creating these rules is expensive and time-consuming.  Neural NLG systems, in contrast, are black boxes which are difficult to control and not transparent, but building such systems can be relatively cheap and quick if appropriate training data\index{training data} is available.

Of course this is a very over-simplified comparison!  For example, software engineering\index{software engineering} tells us that most of the lifecycle costs\index{cost} of a successful software product are in maintenance\index{maintenance}, not initial development, and we do not have a good understanding of how the cost of maintaining a rule-based system compare to the cost of maintaining a neural system.  Maintenance of NLG systems is discussed in Section~\ref{sec:impactevaluation}.

\subsection{Rule-based NLG\index{rule-based NLG}}
Chapter~\ref{rules} looks at rule-based NLG systems, which execute specific algorithms to analyse data, extract insights, choose the content of generated texts, choose words and syntactic structures, and generate linguistically correct output texts (an example was given in Section~\ref{sec:d2tpipeline}).  The chapter focuses on data-to-text\index{data-to-text} because rule-based approaches are rarely used in text-to-text\index{text-to-text} applications.  Rule-based systems often use a modularised \emph{pipeline}\index{pipeline} architecture (Section~\ref{sec:pipeline}), where different modules are used for each of  these tasks (e.g., pattern detection is done by a module specialising in this task), and these modules are usually run sequentially (ie, all pattern detection is done, then all insight generation is done, etc).  Often there is a core set of libraries\index{libraries} and modules which encodes key analytics\index{analytics} and linguistic\index{linguistics} processing algorithms, which is supplemented with a layer of rules that encode domain knowledge\index{domain!knowledge} for a specific application.  

From a high-level perspective, rule-based systems have the advantage of being precise and modifiable; if the system is not doing what the user wants, developers can modify the rules to match user needs.   They are also transparent, auditable\index{auditable}, and can be tested\index{testing} using software engineering techniques; this is important in many commercial contexts, especially in contexts where companies are legally liable\index{liability} if an NLG system makes a mistake.  Finally, rule-based systems are  good at integrating analytics (insight generation) and linguistic (insight expression) reasoning, and in incorporating domain/genre\index{genre} knowledge into both types of reasoning. Their main disadvantage is that writing rules is time-consuming (and can require specialised domain knowledge), especially for systems which have broad or changing use cases\index{use cases}.

For these reasons, rule-based systems work well for business-oriented data-to-text NLG systems in focused domains and use cases where the output needs to be accurate\index{accuracy}.

\subsection{Machine learning and neural models}\label{sec:finetune}
Chapter~\ref{neuralnlg} discusses how \emph{machine learning}, especially \emph{neural deep learning\index{deep learning}} technology, can be used to create models\index{models} which produce texts. This often requires collecting a training set (sometimes called \emph{corpus}\index{corpus}) of example input-output pairs, and using this to create a model for the task; the model essentially learns from the corpus how to map system inputs into textual outputs.

Training a high-quality language model from scratch requires a lot of training data\index{training data}, typically at least a billion tokens (words).  Most NLG applications do not have this much training data available; for example only 10,000 input-output pairs (consultation transcript and corresponding human-written summaries) were available to train Note Generator\index{Note Generator} (Section~\ref{IntroConsultationSummary}).  Fortunately, we can also build good models by \emph{fine-tuning} general-purpose \emph{large language models (LLMs)}\index{large language models} (these are sometimes called \emph{foundation} models\index{models!foundation}).  In other words, we can take a large generic model such as BART\index{BART} \cite{lewis-etal-2020-bart} or T5\index{T5} \cite{raffel2020exploring} which has been trained on billions of tokens of generic Internet material, and then adapt the model to a specific task using a smaller corpus of actual input-output pairs for this task.  This was the approach used in Note Generator; the 10,000 pairs of transcripts and corresponding human-written summaries were not sufficient to train a good model from scratch, but they were sufficient to fine-tune a large generic model (BART).

Some LLMs, such as ChatGPT\index{ChatGPT}, can be used without training or fine-tuning corpora, by simply asking the model (via a \emph{prompt}\index{prompt}) to generate a text.  For example, in order to summarise a text, we could give ChatGPT a prompt which says something like \lingform{Summarise the following text}, followed by the text to be summarised (Fig~\ref{fig:chatgpt1}).  In this example the summary is not very good (almost as long as the original text), but it was generated without needing any application-specific or domain-specific training or fine-tuning!   We can optionally include a small number of examples (5 or 10, not 10,000) in the prompt in order to give the model some guidance as to what we are looking for; we can also use \emph{prompt engineering}\index{prompt!engineering} to tune the wording of the prompt in order to get a more appropriate result.

\begin{figure}
\lineacross\\
\textbf{Prompt: (input to ChatGPT)}\\
Summarise the following text:\\
I think journals do a much better job than conferences at reviewing papers. This year I got a good perspective on this issue, because I was both an Action editor at TACL (Transactions of the ACL), which is a leading NLP journal; and a Senior Area Chair for the ACL conference, which is a leading NLP conference. The experience reinforced my belief that reviewing is better at journals, and I summarise some of my observations below.\\
\\
\textbf{Result: (from ChatGPT)}\\
The author believes that journals are more effective than conferences when it comes to reviewing papers. Their perspective on this matter was strengthened by their dual role as an Action editor at TACL, a reputable NLP journal, and a Senior Area Chair for the ACL conference, a prominent NLP conference. Based on their experience, they summarize their observations supporting the superiority of journal reviews.

\caption{Using a large language model (ChatGPT) for NLG}
\label{fig:chatgpt1}       
\lineacross
\end{figure}

From a high-level perspective, the great advantage of ML approaches is that they can be used to build wide-coverage NLG systems, without needing to write a large number of rules.  Neural NLG systems are also very good at generating readable\index{readability} and fluent\index{fluency} texts.  A substantial amount of training data is needed to fine-tune models (and much more data to train models from scratch), but we do not need training data if we can simply prompt an existing model to do the task (as in Figure~\ref{fig:chatgpt1}). 

On the other hand, neural NLG systems are difficult to control\index{controllability}; for example if the user wants the output texts to be modified in specific ways, this is not  straightforward with neural models. Sometimes this can be done by modifying a prompt, but this is not always sufficient.  Another major problem is that neural models can generate texts which are factually inaccurate\index{accuracy} or otherwise have inappropriate\index{appropriateness} content.  A related point is that software testing\index{testing}  and quality assurance (Section~\ref{sec:testingnlg}) of neural systems is challenging, since they are not transparent and understandable.   For this reason, neural NLG is often used in \emph{human-in-the-loop}\index{human-in-loop} workflows (Section~\ref{sec:workflow}), where a person checks and fixes the model's output (as is the case with  Note Generator), at least in contexts where accuracy is very important.  However human-in-the-loop may not be necessary in contexts where accuracy is not of paramount importance, such as generating fiction or advertising material.

\subsection{Combining rules and ML}
It is of course possible to build NLG systems which combine rule-based processing and machine learning.  One approach is to use a pipeline\index{pipeline} architecture, but have some of the processing within modules done by neural systems \cite{castro-ferreira-etal-2019-neural}.   For example, we can use rules where precision and accuracy\index{accuracy} are very important (selecting key insights to communicate?), and neural techniques where fluency\index{fluency} is very important (choosing syntactic structures of sentences?).

A variant of this is to generate an initial draft text using rules, which is guaranteed to have the right content (insights) but may not be very readable, and use a language model\index{language models} to rewrite the text to make it more fluent and readable \cite{kale-rastogi-2020-template}.

Another approach is to build two systems, one neural and one rule-based, run both, and use a selection module to decide which of the two outputs to use.  For example, the selection module could test the text produced by the neural system for accuracy errors, and use the neural text if none were detected and the rule-based text otherwise \cite{heidari-etal-2021-getting}.

\section{Effectiveness}
The second part of this book looks at the usefulness and effectiveness of NLG systems. It starts by looking at \emph{requirements}, in other words what NLG systems need to do in order to be useful.  It then looks at \emph{evaluation}, i.e. measuring how well NLG systems meet important quality criteria (as determined by requirements).  It concludes by looking at \emph{safety and testing}, which focus on whether the system behaves unacceptably in a few cases even if it is fine in most cases.

\subsection{Requirements}\index{requirements}
Chapter~\ref{requirements} looks at requirements for NLG systems; what do real-world users want NLG systems to do?  Where does NLG \scare{add value} in real-world use cases\index{use cases}, and how does it impact workflows?   What specific NLG capabilities are needed by users?

Questions like the above are often ignored by academic researchers, but they are very important for real-world success.  Just as with other types of software, an NLG system is not going to succeed if we do not understand user requirements and more generally how users will react to the NLG system.  Some of the key requirements issues discussed in Chapter~\ref{requirements} are:

\begin{itemize}
\item \emph{Quality Criteria:}\index{quality criteria} What are the text (and system) characteristics which may be important to users and other stakeholders\index{stakeholders}, and which of these matter most in different contexts?  For example, when is accuracy\index{accuracy} more important than readability\index{readability}?  Also, are users only interested in average text quality, or do they need a guarantee that all texts will meet a minimum quality threshold?
\item \emph{Workflows:}\index{workflow}   While some NLG systems operate without human supervision, many NLG systems collaborate with humans and as such are integrated with human workflows.   What are different \scare{human-in-the-loop}\index{human-in-loop} workflows and when do they make sense?
\item \emph{Text and Graphics:}\index{graphics} Computer systems can communicate information and otherwise interact with users using visualisations and interactive data graphics as well as text.  When is text the best interaction and communication modality?  How should NLG texts should be combined and integrated with visualisations\index{visualisations}?
\item \emph{Understanding Requirements:}  How do NLG developers work with users and other stakeholders in order to understand their needs?  Many techniques for requirements acquisition\index{requirements!acquisition} have been developed in other parts of computing science, how can we adapt these to work for NLG \cite{knoll-etal-2022-user}?
\end{itemize}

\subsection{Evaluation}
Chapter~\ref{evaluation} discusses evaluation of NLG systems.
From a scientific perspective, we need to be able to evaluate NLG models, algorithms, and systems, in order to understand how they compare to previous work and also where further work is needed.  

Scientific evaluation is essentially hypothesis testing\index{hypothesis testing}.  In very general terms, we can evaluate NLG systems by asking people to examine or use generated texts (\emph{human evaluation})\index{human evaluation}, by using \emph{automatic metrics}\index{metrics} (algorithms) to assess the quality of texts, or by assessing the system's real-world impact\index{impact} when deployed (\emph{impact evaluation}).   For all types of evaluation, a key question is how meaningful it is, and in particular whether the evaluation is a good predictor of real-world utility\index{utility}.

There are many kinds of human evaluation.   The most expensive, and usually the most meaningful, are \emph{task-based evaluations}\index{task-based evaluation}.  Such evaluations involve asking domain experts\index{domain!experts} to perform a task using the generated text, such as asking doctors\index{doctors} to decide on a medical intervention for a patient after reading an NLG summary of the patient's status \cite{PORTET2009789}.  A more common approach (which not coincidentally is considerably cheaper and quicker) is to ask non-expert subjects to read generated texts and either rate\index{rating} them on quality factors such as accuracy and readability, or rank\index{ranking} (order) texts based on a quality factor.  Such studies need to be carefully designed in order to be meaningful \cite{VANDERLEE2021101151}.  Another approach to human evaluation is to ask subjects to annotate errors and other problems in generated texts; this is often more meaningful than rating-based evaluation, but can also be more expensive.

Metric (algorithmic) evaluations are usually much cheaper and faster than human evaluation; because of this, metric evaluations can be done more frequently and on a larger scale than human evaluation (a typical academic human evaluation may look at one hundred texts; a metric evaluation can look at many thousands or even millions of texts).   At the time of writing, most metric evaluation is \emph{reference-based}, that is generated texts are algorithmically compared against human-written \scare{gold standard} \emph{reference} texts\index{reference texts}.   Numerous such metrics have been designed, and they are usually \emph{validated} by seeing how well they correlate with high-quality human evaluations.  Correlation may depend on factors such as text quality; for example a metric may correlate well with human evaluations of low-quality NLG texts but not of high-quality NLG texts.  There are also \emph{no-reference} metrics which assess the linguistic quality of a text in isolation, including well established formulas such as Flesch-Kincaid scores. Interest is growing in using large language models\index{large language models} such as GPT to evaluate texts (both with and without references), this seems to work well in many contexts.

Impact evaluation is perhaps the hardest to carry out, not least because evaluating a system in production\index{production} usage may raise ethical\index{ethics} issues; for example if we evaluate a medical decision support system in real clinical usage, we need to show beforehand that the system will not damage patient care.  The most rigorous impact evaluation is to conduct a \emph{comparison experiment}\index{comparison experiment} where some people use the NLG system and others use an alternative system, and compare outcomes; this is related to \emph{A/B Testing}\index{A/B testing}.  An alternative is a \emph{historical comparison}\index{historical comparison} where an NLG system is deployed to a set of users, and \emph{Key Performance Indicators (KPIs)} are compared before and after the system is introduced.

\subsection{Safety\index{safety}, Testing\index{testing}, and Maintenance\index{maintenance}}
Developers of real-world NLG systems need to ensure that their systems are safe (never harm users), properly tested from a software perspective (including edge cases\index{edge cases}), and maintained as the world and user needs\index{user needs} evolve.
Chapter~\ref{safetytesting} looks at these issues.

There are many aspects of \emph{safety}.  NLG systems should not generate texts which include profanity, racist language, or other socially unacceptable language.  They should not encourage users to do harmful activities such as self-harm, suicide, or killing someone \cite{bbc:chatbotkillqueen};  they should not give misleading advice or information which could harm the user more generally (including poor medical advice \cite{Bickmore:jmir18}); they should not make the user feel stressed or depressed \cite{balloccu2024ask}; and they should not divulge sensitive data to unauthorised users.   Furthermore, such behaviour should \emph{never} happen.   In other words, we should design NLG systems that never exhibit the above behaviour under any circumstances.  This is a very important but also very difficult challenge, especially for NLG systems which use neural models (whether trained, fine-tuned, or prompted).

Like other software systems, NLG systems need to be \emph{tested} before they are deployed and released to customers, in order to ensure that the system is robust\index{robustness} and that generated texts are of acceptable quality.  Of course classic software engineering\index{software engineering} techniques can be used for NLG testing, but NLG has the additional challenge that there are usually many ways to express information in words, so it is difficult to create \emph{test cases}\index{test cases} which pair an input with a specific target output (because there may be other outputs which are just as good).  A related issue is that many NLG systems are \emph{stochastic}\index{stochastic} which means that they can produce different outputs from the same input on different runs; this also makes it harder to use standard software testing techniques

Last but not least, \emph{maintaining} NLG systems can be challenging.  Software engineering tells us that most of the lifecycle cost of a software system is in maintaining the software, including adapting it to changing data sources and user needs as well as fixing bugs\index{bugs}.   Unfortunately, very little is known about maintaining NLG software, especially neural systems.  Maintaining rule-based NLG is probably similar to maintaining programmatic code, but maintaining a system which is trained on data offers additional challenges.

One such challenge is adapting neural NLG systems as the world changes (what is sometimes called \emph{domain shift}\index{domain!shift}).  For example, suppose a new medication is introduced which is very effective at treating cancer, and we want a medical NLG system to consider this medication in its recommendations.  Since the medication is new, we will not have a corpus of training data to update the system.  Strickland \cite{strickland2019ibm} points out that such cases were very challenging for the IBM Watson\index{IBM Watson} question-answering system.

\section{Use Cases\index{use cases} and Applications\index{applications}}
The final part of this book (Chapter~\ref{applications}) describes some NLG use cases and applications,  The chapter first introduces general guidelines for successful applications, and then discusses four specific domains: journalism\index{journalism}, business intelligence\index{business intelligence},summarisation\index{summarisation} and health\index{health}.

NLG is being used in a huge range of different applications.  It is not possible to make general statements which apply to all of these, except perhaps that the necessary input data must be available,
but in most cases successful NLG applications have the following characteristics:
\begin{itemize}
\item \emph{Scalability\index{scalability} and configurability\index{configurability}:} The NLG system is used to generate a large number of texts for a large number of users.  This usually means that the system is configurable to some degree, so that users can customise it for their requirements, data sources, etc.
\item \emph{Acuracy\index{accuracy} and utility\index{utility}:} The NLG system reliably  generates high-quality useful and accurate texts.  Even a small number of low-quality outputs can reduce users' trust in the system.
\item \emph{Acceptability\index{acceptability}:} Users accept and want to use the NLG system; they do not, for example, see it as a threat to their jobs.
\item \emph{Benefits\index{benefits}:} The system must provide significant benefits to its users, which exceeds its cost and offer a good return on investment\index{return on investment}.  This is not always needed in the short term if investment capital is available, but it is crucial for long-term success.
\end{itemize}

Looking at specific application domains, NLG can be used in \emph{journalism}\index{journalism} to automatically write some types of articles in newspapers and other media.  Figure~\ref{fig:bbcelectionIntro} shows a simple example of this, which is a BBC news article about the 2019 UK general election in a specific constituency.  This was generated automatically using technology from Arria\index{Arria} (my company).   Like most such generated articles, the BBC\index{BBC} election reports were checked by human journalists before they were released to the public.  The advantage of using NLG in this case was largely speed; old news is stale news, and using NLG to produce these articles meant they could be released the morning after the election, which would not have been possible if the articles had been manually written.

\begin{figure}
\lineacross{}
Florence Eshalomi has been elected MP for Vauxhall, meaning that the Labour Party holds the seat with a decreased majority.

The new MP beat Liberal Democrat Sarah Lewis by 19,612 votes. This was fewer than Kate Hoey's 20,250-vote majority in the 2017 general election.

Sarah Bool of the Conservative Party came third and the Green Party's Jacqueline Bond came fourth.

Voter turnout was down by 3.5 percentage points since the last general election.

More than 56,000 people, 63.5\% of those eligible to vote, went to polling stations across the area on Thursday, in the first December general election since 1923.

Three of the six candidates, Jacqueline Bond (Green), Andrew McGuinness (The Brexit Party) and Salah Faissal (independent) lost their £500 deposits after failing to win 5\% of the vote.

\textbf{\textit{This story about Vauxhall was created using some automation.}}

\caption{Example story from BBC election reporter, from \url{https://www.bbc.co.uk/news/technology-50779761}}
\label{fig:bbcelectionIntro}
\lineacross
\end{figure}

NLG can also be used to explain and summarise data in a \emph{Business Intelligence (BI)} context, including data about sales, profits, inventories, customers, suppliers, employees, etc.   At the time of writing, most BI data is communicated graphically\index{graphics} using BI tools such as Tableau, but some types of insights are best communicated in words instead of data visualisations\index{visualisations}.  BI NLG systems can be very scalable\index{scalability}: almost every business in the world is interested in sales, profits, etc.

Another long-standing use case of NLG is summarising textual information.  Note Generator (Section~\ref{IntroConsultationSummary}) summarised doctor-patient consultations; NLG systems can also summarise emails, legal documents, news articles, product reviews, and  many other types of documents.

Finally (at least in this chapter), NLG can be used in \emph{medicine and healthcare}\index{health}.  A large number of medical applications of NLG have been explored in the academic literature over the years, including
\begin{itemize}
\item \emph{reporting\index{reporting}:} tools that help clinicians create medical documents, such as Note Generator (Section~\ref{IntroConsultationSummary}).
\item \emph{behaviour change\index{behaviour change}:} tools to encourage healthier behaviour in smoking \cite{REITER200341}, diet \cite{balloccu-reiter-2022-comparing}, etc.
\item \emph{patient information\index{patient information}:} tools that explain medical information and patient data to patients \cite{mahamood-reiter-2011-generating}.
\item \emph{clinical decision support\index{decision support} tools:} tools that help clinicians decide on appropriate medical interventions for patients \cite{PORTET2009789}.
\end{itemize}
The commercial NLG community has explored additional opportunities, such as business intelligence for health organisations.

Although there are many exciting opportunities in using NLG in healthcare, it can be difficult to deploy NLG (and indeed AI) solutions in medicine.  This is partially because of understandable safety\index{safety} concerns (we cannot take the risk that an NLG system could damage the quality of care), and also partially because the huge diversity of the healthcare sector (especially if we look worldwide) makes it difficult to build scalable\index{scalability} solutions.  In other words, we can build NLG solutions that help one hospital, but it is difficult to build solutions that help hundreds of hospitals because there is little standardisation between hospitals in IT systems, equipment, clinical procedures, administrative processes, etc.

\section{Ethics}\index{ethics}
NLG systems need to behave \emph{ethically}.  In other words, NLG systems need to act in a manner which is acceptable to society\index{society} and does not harm users or third parties.  

AI ethics is a very broad area, which of course goes well beyond NLG; I will just mention a few issues here.
One is \emph{accessibility\index{accessibility} across languages and communities}.  In particular, much better NLG technology and resources are available in widely-spoken world languages (such as English, German, and Mandarin) than in \emph{under-resourced}\index{under-resourced languages} languages  such as Scottish Gaelic, Maltese, and isiZulu; this may encourage speakers of these languages to make more use of English.  Also some communities (such as gypsies and travellers in the UK) are under-represented\index{under-represented communities} in the corpora\index{corpus} used to train large language models\index{large language models}; applications developed for such models may therefore not work well for members of these communities.  From a technology perspective, it would be good to have more work on NLG in under-resourced languages and for under-represented communities.

Another issue is \emph{bias}\index{bias}.  For example, an NLG system may generate texts which use male pronouns to refer to doctors and female pronouns to refer to nurses; this can support and reinforce gender stereotypes\index{gender stereotypes}.  Such biases can manifest in more subtle ways as well \cite{ciora-etal-2021-examining}.   Unfortunately it can be difficult for developers to detect such issues, especially if they are not in the biased-against group.  For this reason it is useful to get a diverse set of users, with different demographics and background, to check NLG texts for bias.

 \emph{Unethical use cases} can also be a problem, such as generating fake new\index{news!fake} or misinformation\index{misinformation} which is intended to stop people from voting.  Of course working explicitly on such applications is unethical.  A trickier case is developing technology which can be applied to both ethical and unethical use cases, such as techniques for generating persuasive\index{persuasion} texts.  It is hard to give crisp guidelines about \scare{dual-use} technology, but I encourage developers to think about this issue and decide what they personally are comfortable with.

\emph{Theft of intellectual property}\index{intellectual property} can be a concern with neural NLG systems based on large language models.  If an NLG system simply regurgitates (in whole or in part) what a person has written, then the human author needs to be acknowledged.  If the original human-written text was not licensed for general reuse (for example with a Creative Commons\footnote{\url{https://creativecommons.org/}}\index{Creative Commons} license), then it can only be reused with permission from the human author, who may expect some compensation.

Another issue is \emph{job losses}\index{job losses}.  Like other AI and IT technologies, NLG can lead to jobs being automated; for example we may need fewer weather forecasters if NLG is used to generate weather forecasts\index{weather forecasts}.  Automation can also make jobs less interesting and rewarding; for example journalists\index{journalism} may be told to check and fix up AI-written articles (Section~\ref{journalism}), instead of writing articles themselves.   Again there is no easy answer to this problem; if we had refused in the past to adapt technology that automated jobs, then we would still be living in an agrarian society where 90\% of people worked on farms.  As above, I encourage developers to think about this issue and decide what they personally are comfortable with.

Last but not least,  NLG researchers and developers need to ensure that experiments used to evaluate NLG systems are ethical and cannot harm experimental subjects or third parties; this is discussed in Section~\ref{sec:researchethics}.

There are grey areas.  For example, in general it is not ethical for NLG systems to lie\index{lies}.  However an NLG system which informs relatives about the state of a sick baby may wish to be economical with the truth when communicating with an elderly great-grandmother who could have a heart attack if she hears bad news \cite{vanDeemterLying}.

\section{A Very Short History of NLG}

\subsection{Early history}

Speculation about algorithmic and mechanical generation of texts dates back at least to the seventeenth century \cite{peter1677artificial}.
The earliest work on rule-based NLG\index{rule-based NLG} as described in this book was done in the 1960s, in the context of building machine translation\index{machine translation} systems.  Whereas modern MT systems use neural models to directly create a target-language output from a source-language input, early MT systems usually analysed the source language input into an intermediate structure, made changes to the structure to suit the target language, and then generated text in the target language from the intermediate structure.  This last step was essentially NLG, and researchers working on this were the first researchers who attempted to dynamically construct sentences using natural language technologies \cite{sakai-nagao-1965-sentence}.

The 1970s saw the first PhD theses on NLG which were not connected to Machine Translation.  For example, Goldmans' PhD thesis developed techniques for generating texts from conceptual dependency models \cite{goldman:phd}, and Davey's PhD looked at generating summaries of tic-tac-toe games \cite{davey:PhD}.

The 1980s saw a lot more research activity in NLG, including the first work on data-to-text \cite{kukich-1983-design} (Figure~\ref{fig:kukich-stock}), and developments of text summarisation technology \cite{Marsh1984}. There also was a movement towards separating NLG into component tasks, and in particular separating content decisions (\emph{what to say}) from linguistic decisions (\emph{how to say}).  The NLG community also developed more of an identity, with the founding of the ACL Special Interest Group in Generation (SIGGEN).  The first International Workshop on Natural Language Generation was held in 1983.  Researchers also began looking at NLG applications, for example in weather forecasts\index{weather forecasts} \cite{kittredge-etal-1988-multi}.

\subsection{1990-2014}
Research interest in NLG grew in the 1990s, with researchers investigating a broad range of topics including generating texts in multiple languages,  integrating text and graphics\index{graphics}, and generating different texts for different users.   The first shared tasks appeared in text summarisation, such as SUMMAC \cite{mani-etal-1999-tipster}.   Software libraries\index{libraries} for doing NLG tasks were released \cite{elhadad-robin-1996-overview}, so researchers did not need to build everything from scratch.

In the late 1990s, researchers started looking at data-driven techniques for NLG, including ngram models, statistical models, and machine learning (the systems mentioned above all used rules or algorithmic code).  This was successful in text summarisation \cite{radev-etal-2002-introduction}, but less successful in data-to-text systems.

Evaluation also grew in importance, and researchers experimented with a wide range of techniques, ranging from simple metrics\index{metrics} such as BLEU\index{BLEU} \cite{papineni-etal-2002-bleu} and ROUGE\index{ROUGE} \cite{lin-2004-rouge} to large-scale task-based evaluations\index{task-based evaluation} with real users \cite{REITER200341,mani2002summac}.   The first meta-evaluation studies were done to see how well different types of evaluation agreed with each other \cite{reiter-belz-2009-investigation}.

\begin{figure}
\lineacross\\
\includegraphics[scale  = 0.75]{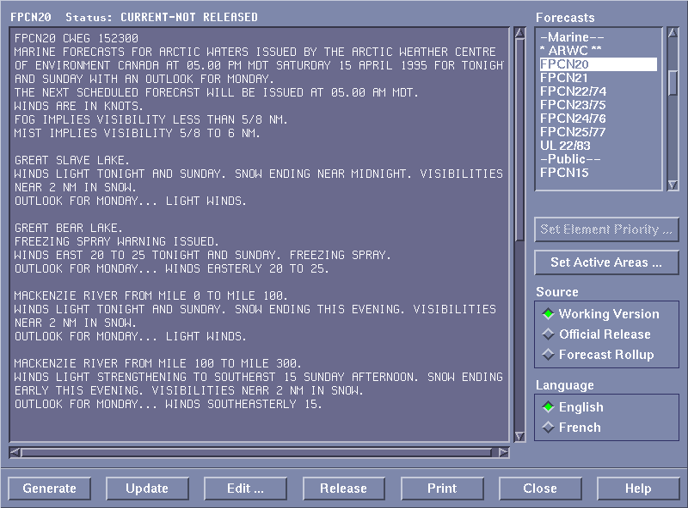}
\caption{FoG, the first-ever commercial NLG system, which went live in 1992. From cogentex.com (website is now defunct).}
\label{fig:FoG}       
\lineacross
\end{figure}

From a commercial perspective, this period had an exciting start, when CoGenTex\index{CoGenTex} (the first-ever specialist NLG company) operationally deployed the FoG\index{FoG} weather forecast system in 1992 \cite{Goldberg1994} (Fig~\ref{fig:FoG}).  However commercial NLG did not really begin to take off until the latter part of this period, when several specialist NLG companies were founded, including my company Data2Text\index{Data2Text} (which was bought by Arria\index{Arria}), Ax Semantics\index{Ax Semantics}, and Narrative Science\index{Narrative Science}.

\subsection{2015-2024}
This period is very recent, so it is difficult to give a historical perspective, but around 2015 neural deep-learning\index{deep learning} technology started being used in NLP and NLG.  Researchers had in fact been exploring using neural techniques in NLG since the 1980s \cite{kukich1987phrases}, but such work had limited impact before deep learning  became popular.  Deep learning neural approaches were very successful in academic circles, and by 2020 this technology dominated academic NLP conferences and journals.  Conferences also grew in size; 320 papers were presented at the Association for Computational Linguistic conference in 2015, while over 1000 papers were presented at this conference in 2023.   Perhaps because of this growth, many exciting new uses for language generation technology were explored, such as image captioning\index{image captioning} (image-to-text).   There was also a strong and welcome trend towards making code and datasets publicly available for other researchers.  In 2022 the ChatGPT\index{ChatGPT} system was introduced, and the research community started focusing more on using prompted LLMs to generate texts

One consequence of the dominance of neural technology was that different types of NLG systems became more likely to use the same technology.  For example, the 2014 Arria weather system (Section~\ref{ArriaWeatherExample}) used completely different technology than 2022 Note Generator (Section~\ref{IntroConsultationSummary}).  However both Note Generator and Facebook's 2021 weather system \cite{heidari-etal-2021-getting} (Section~\ref{sec:Facebookweather}) were based on fine-tuning and adapting the BART\index{BART} language model \cite{lewis-etal-2020-bart}.

The commercial NLG world is changing extremely quickly at the time of writing.  Companies initially were slower to adapt neural techniques than academics, because of concerns over accuracy\index{accuracy}, quality assurance and testing\index{testing}, controllability\index{controllability}, and maintainability\index{maintenance}.    However, since the introduction of ChatGPT there is huge interest in using prompted models to generate text in commercial applications.  It is too soon to say where such models will prove successful in real-world usage, and where concerns about accuracy, quality, etc will motivate the continued use of
rule-based or programmatic approaches

\subsection{My personal NLG journey}
On a personal note, I got my PhD thesis in 1990, from Harvard, on the topic \emph{Generating Appropriate Natural Language Object Descriptions} \cite{reiter:phd}.  The most important part of my thesis was work on generating \emph{referring expressions}\index{referring expressions} for entities.  After my PhD, I further developed my ideas on this topic while doing a post-doc at the University of Edinburgh.  I worked with Robert Dale, who was also interested in this area \cite{DALE1995233}.  In 2000 Robert and I published the first ever book dedicated to NLG, \emph{Building Natural Language Generation Systems} \cite{reiterdale2000}.

In 1995, after spending a few years at CoGenTex\index{CoGenTex} (which developed the Fog\index{FoG} system, Fig~\ref{fig:FoG}), I moved to the University of Aberdeen.  Around 2000 I started focusing on data-to-text\index{data-to-text}, working on projects such as SumTime (generating weather forecasts\index{weather forecasts}) \cite{REITER2005137} and Babytalk (generating summaries of electronic patient records) \cite{PORTET2009789}.  One of my papers, on the architecture of data-to-text systems \cite{reiter-2007-architecture} was given a Test-of-Time award by the INLG conference as the most influential paper published in any INLG conference before 2022.

In 2009 I founded a company, Data2Text\index{Data2Text}, to commercialise data-to-text, and by 2012 I was spending almost all of my time at the company (which was bought by Arria\index{Arria} in 2013).  It was very exciting; we did a lot of work for the oil industry\index{oil industry}, and I really enjoyed working with engineers\index{engineers} and exploring how NLG could help them.

Around 2017 I started to re-engage with academic life, and by 2023 I was once again a full-time academic.  My research interests focused on evaluation, especially human evaluation\index{human evaluation} (e.g., \cite{reiter-2018-bleu,thomson-csl23,belz-etal-2023-missing,wu-etal-2023-experts,thomson-cl24}); this partially arose from my frustration at the poor-quality evaluations that dominated early work in neural NLG, which meant that much of this work was scientifically meaningless.  Evaluation quality has improved since 2017, but unfortunately many evaluations published in the academic literature are still of disappointing quality and/or difficult to reproduce.

\section{Resources and Further Reading}
Numerous surveys of NLG have been published in the academic literature, but most focus on technology and quickly become out of date.  Likewise many textbooks on natural language processing include sections on NLG, but the focus is almost always on technology.  Since technology evolves so rapidly, I do not give recommended sources here, but instead encourage readers to find up-to-date sources.   There are also of course many commercial white papers on NLG, but these usually do not provide a balanced scientific view of the field.

Two surveys which go beyond technology are Gatt and Krahmer (2018) \cite{gatt2018survey}, who take a broad perspective on NLG, and Gehrmann et al (2023) \cite{GehrmannEvaluation}, who focus on NLG evaluation.

The Special Interest Group on Natural Language Generation (SIGGEN) (\url{https://siggen-acl.github.io/}) of the Association for Computational Linguistics (ACL) organises academic conferences and other events about NLG topics, including the annual International Natural Language Generation (INLG) conference.

The \emph{ACL Anthology} (\url{https://aclanthology.org/}) contains free downloadable copies of all papers published at conferences and journals sponsored by the Association for Computational Linguistics (and a number of non-ACL venues as well), including SIGGEN events such as INLG (\url{https://aclanthology.org/sigs/siggen/}).  The Anthology is not always easy to search, but if you know what you are looking for, it is a great resource (and one that I personally use all the time).

Huggingface\index{Huggingface} (\url{https://huggingface.co/}) is an excellent source for datasets and code libraries, especially for neural NLG.  Popular Huggingface resources are often well documented and supported.

For further reading on topics discussed elsewhere in this book, please consult the relevant chapter.  For example, further reading on applications of NLG is available in Section~\ref{sec:furtherApplications}.

The ACL maintains a good bibliography of papers specifically about ethics in Natural Language Processing, see \url{https://github.com/acl-org/ethics-reading-list}.  The only paper I am aware of which is specifically about ethics in NLG is Smiley et al \cite{smiley-etal-2017-say}.

Little has been published specifically about the history of NLG.  However I gave a talk about this, which is available on YouTube at \url{https://www.youtube.com/watch?v=SEw47Y_ZN8Q}.

%
%
%
\chapter{Rule-based NLG}
\label{rules} 

\abstract*{This chapter describes how AI systems can generate texts using algorithms and rules which explicitly make decisions about the content and language of generated texts.  Rule-based NLG has been overshadowed by neural NLG in recent years, but it is still the best way to build some NLG applications.  Rule-based NLG also shows the types of decisions which need to be made in text generation.  A good understanding of this helps anyone working in NLG, even if they use other approaches.}

Natural Language Generation systems can be built using algorithms and rules which explicitly extract insights from texts, structure information into narratives\index{narrative}, and create good linguistic expressions of information; this is called \emph{rule-based NLG}\index{rule-based NLG}.  Rule-based NLG is especially common in data-to-text\index{data-to-text} NLG; it is rare in text-to-text NLG.

The advantage of rule-based approaches to NLG is that developers have complete control\index{controllability} over what the system does, and the system is also testable\index{testing} and auditable\index{auditable}.  In other words, rule-based NLG is a \scare{precision} approach which lets developers build NLG systems with exactly the functionality which clients want, and which (assuming no software bugs) will never go \scare{off the rails} and do crazy or inappropriate things.

The disadvantage of rule-based NLG is that writing (and debugging) the algorithms and rules can require a considerable amount of work, especially for NLG systems which generate complex texts.  Of course good modularisation, structure, tools, and libraries will help (as with all kinds of software development), but writing rules for a large NLG system still can be a daunting task.

Even developers who do not use rule-based NLG will still benefit from understanding it, because it gives a good understanding of the sorts of processes and decisions which NLG systems must do.  For example, suppose an NLG developer uses neural language model technology (Chapter~\ref{neuralnlg})  to build an NLG system which summarises sensor data.  The rule-based perspective shows that most such systems must detect and remove noise (Section~\ref{sec:signalanalysis}).  Unfortunately, many neural language models are not very good at removing noise from sensor data, so they will need help with this task (a separate preprocessor for noise filtering?).

This kind of analysis is only possible if developers understand what an NLG system needs to do at a conceptual level. Such understanding also makes it easier to discuss requirements\index{requirements} with users (Chapter~\ref{requirements}) and to design appropriate evaluation schemes (Chapter~\ref{evaluation}),

\section{NLG Pipeline\index{pipeline}}\label{sec:pipeline}
\begin{figure}
\lineacross{}\\
\includegraphics[scale  = 0.75]{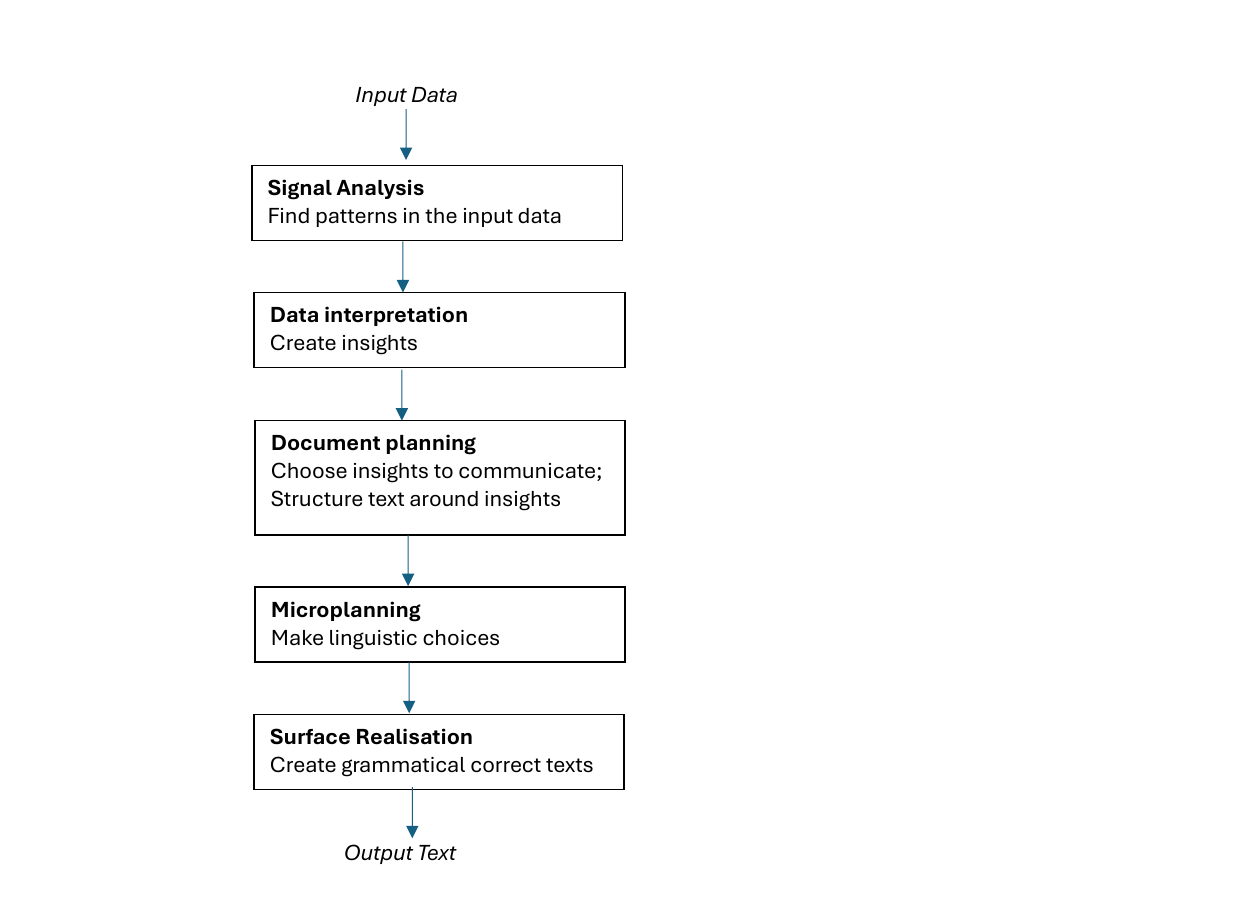}
\caption{The data-to-text NLG pipeline.}
\label{fig:pipeline}       
\lineacross{}
\end{figure}

The simplest rule-based systems use \emph{templates}\index{templates} to generate output texts from input data (Section~\ref{sec:templatenlg}); such functionality is provided by many packages, include Python's Jinja\index{Jinja} library\footnote{\url{https://jinja.palletsprojects.com/}}.  More sophisticated rule-based NLG systems structure the generation process into separate modules, which are often connected together into a \emph{pipeline} (the modules run in sequence, for example the second module does not start until the first module has finished).  In data-to-text\index{data-to-text}, the most common modularisation is the data-to-text pipeline architecture \cite{reiter-2007-architecture}, which divides the process into \emph{Signal analysis}\index{signal analysis}, \emph{Data interpretation}\index{data interpretation}, \emph{Document planning}\index{document planning}, \emph{Microplanning}\index{microplanning}, and \emph{Surface realisation}\index{surface realisation} (Figure~\ref{fig:pipeline}).

\personal{I proposed this data-to-text NLG architecture in a 2007 paper \cite{reiter-2007-architecture}, which was awarded a \scare{Test of Time} award in 2022 as one of the most influential NLG papers published in an INLG conference before 2022.}

Not all systems include all of the above steps.  For example we do not need to extract insights from data if they are directly present in the system's input, and we do not need to design texts at a document level if the use case only requires a single sentence to be produced.

 A variety of techniques can be used within the modules, including:
\begin{itemize}
\item \emph{Algorithms}:   We can use algorithms for many tasks, such as pattern detection within signal analysis
\item \emph{Rules}: We can ask domain experts\index{domain!experts} how they do a task, and encode their response into the NLG system using rules.   For example, if we want to choose an appropriate verbal phrase (part of microplanning) to describe stock market changes (eg \lingform{inching up} or \lingform{skyrocketing}), we could ask experts (journalists\index{journalism} and financial experts) how they choose verbs in this context, and write rules based on this information.
\item \emph{Neural or other machine learning}: We can also use neural or ML approaches within modules (Section~\ref{sec:modularML}).  For example, for the verb choice task, we can build an ML model by analysing a corpus of historical financial reports \cite{chen-yao-2019-closer}.
\end{itemize}
The distinction between rules and algorithms is fuzzy.  In theory algorithms define computational processes which can be reused in multiple NLG systems; they are usually implemented in programming languages such as C++, Python, or Java.  Rules encodes domain/genre/usecase specific information which can only be used in this domain\index{domain!knowledge}, genre\index{genre}, and use case\index{use cases}; they can be defined in special rule languages, database or spreadsheet rows, or scripting languages.   In practice it is often difficult to clearly separate rules and algorithms, especially since algorithms can be parametrised in a rule-like fashion for specific domains, genres, and use cases.  Also rules can invoke algorithmic-processing to make decisions.

In this book I will use \scare{rule-based NLG} to cover both rules and algorithms (which is what most academic researchers do), and not explicitly differentiate between these.  The key issue is that both rules and algorithms require writing code to do NLG tasks, and as such are very different from the data-based approaches described in Chapter~\ref{neuralnlg}.

\section{Examples}
In this section I introduce two example rule-based NLG systems, which I will use throughout this chapter to illustrate rule-based NLG.

\subsection{DrivingFeedback}\label{sec:saferdriver}
The first example is a highly simplified version of the SaferDriver\index{SaferDriver} (Figure~\ref{fig:saferdrivermultimodal}) system \cite{braun_reiter_siddharthan_2018}.   The \emph{DrivingFeedback}\index{DrivingFeedback} system takes as input GPS-type data showing the speed of a vehicle at regular timestamps, together with street name and speed limit information (which is extracted from the GPS data using reverse geocoding).  The system produces regular reports on unsafe driving from this data.  Table~\ref{tab:drivingfeedbackdata} shows example input data, and Figure~\ref{fig:drivingfeedbackpipeline} shows how this data is processed by the different stages of the data-to-text pipeline, in order to produce the sentence   \lingform{You sped twice on King Street} (a full report is several paragraphs long).

At a high level, DrivingFeedback works as follows
\begin{itemize}
\item \emph{Signal analysis:}\index{signal analysis} identifies unsafe driving incidents, such as speeding.
\item \emph{Data interpretation:}\index{data interpretation} clusters related unsafe driving incidents, such as multiple speeding incidents on the same road.
\item \emph{Document planning:}\index{document planning} selects most important clusters from a safety perspective.  Also organises content so that the report starts with positive feedback (ie, congratulations on improvements from previous report) if possible, since this enourages people to read the report and take it seriously.
\item \emph{Microplanning:}\index{microplanning} vary wording so that reports read differently from previous reports even if they have similar content.
\item \emph{Surface realisation:}\index{surface realisation} create grammatically correct texts.
\end{itemize}

\begin{table}
\lineacross{}\\
\begin{tabular}{|r|r|l|} \hline
\emph{time} & \emph{speed} & \emph{street} \\ \hline
9:00:00 & 30 & King Street \\
9:00:01 & 32 & King Street \\
9:00:02 & 35 & King Street \\
9:00:03 & 30 & King Street \\
9:00:04 & 33 & King Street \\
9:00:05 & 32 & King Street \\
9:00:06 & 27 & King Street \\
9:00:07 & 25 & King Street \\
9:00:08 & 30 & King Street \\
9:00:09 & 33 & King Street \\
9:00:10 & 33 & King Street \\
9:00:11 & 30 & King Street \\
9:00:12 & 25 & King Street \\
9:00:13 & 20 & King Street \\
9:00:14 & 15 & King Street \\
9:00:15 & 15 & St Machar Drive \\
9:00:16 & 25 & St Machar Drive \\
9:00:17 & 30 & St Machar Drive \\
9:00:18 & 32 & St Machar Drive \\
9:00:19 & 32 & St Machar Drive \\
9:00:20 & 30 & St Machar Drive \\ \hline
\end{tabular}
\caption{Example DrivingFeedback data. Speed limit is 30 on both King Steet and St Machar Drive.}
\label{tab:drivingfeedbackdata}
\end{table}

\begin{figure}
\textbf{Output of signal analysis} (find speeding segments)
\begin{itemize}
\item Speeding(9:00:01-9:00:05, King Street, maxSpeed=35)
\item Speeding(9:00:09-9:00:10, King Street, maxSpeed=33)
\item Speeding(9:00:18-9:00:19, St Machar Drive, maxSpeed=32)
\end{itemize}

\textbf{Output of data interpretation} (cluster related speeding segments)
\begin{itemize}
\item SpeedingCluster(9:00:01-9:00:10, King Street, maxSpeed=35, incidents=2)
\item Speeding(9:00:18-9:00:19, St Machar Drive, 32)
\end{itemize}

\textbf{Output of document planning} (choose which insights to include)
\begin{itemize}
\item Sentence: SpeedingCluster(9:00:01-9:00:10, King Street, maxSpeed=35, incidents=2)
\end{itemize}

\textbf{Output of microplanning} (design sentences to express insights)
\begin{itemize}
\item Sentence: Subject=\lingform{you}, verb=\lingform{speed} (past tense), modifier=\lingform{twice}, location=\lingform{on King Street}
\end{itemize}

\textbf{Output of surface realisation} (create actual text)
\begin{itemize}
\item Sentence: \lingform{You sped twice on King Street.}
\end{itemize}

\caption{Processing DrivingFeedback data through the data-to-text pipeline.}
\label{fig:drivingfeedbackpipeline}
\lineacross
\end{figure}

\subsection{ Babytalk}\label{sec:babytalk}
A more complex rule-based data-to-text system is \emph{Babytalk}\index{Babytalk}.  Babytalk is actually a family of systems which generate texts about babies in a neonatal intensive care unit\index{health} (examples are shown in Figure~\ref{fig:BabytalkOutputs}):
\begin{itemize}
\item \emph{BT45}\index{Babytalk!BT45} generates summaries of recent activity, which are intended to help doctors\index{doctors} and nurses\index{nurses} make decisions about interventions \cite{PORTET2009789}.
\item \emph{BT-Nurse}\index{Babytalk!BT-Nurse} generates shift-handover reports for nurses who are starting a 12-hour shift; the report summarises what happened in the previous shift, and also in earlier shifts (in case the nurse has not dealt with the baby before) \cite{HUNTER2012157}.
\item \emph{BT-Family}\index{Babytalk!BT-Family} generates daily reports for parents, so that they know the status of their baby \cite{mahamood-reiter-2011-generating}.
\end{itemize}
All of these systems extract information from the hospital's electronic patient record.  At least at a conceptual level, they have a similar architecture but use different data-interpretation\index{data interpretation} and document planning\index{document planning} rules and algorithms in order to select appropriate insights for the different use cases. BT-Family also uses a different set of linguistic expressions (microplanning\index{microplanning}) rules than BT45 or NT-Nurse, since it produces texts for non-specialists (parents) instead of domain experts (doctors and nurses).

\begin{figure}
\lineacross\\
\emph{Example BT45 output (extract)}:\\
By 11:00 the baby had been hand-bagged a number of times causing 2 successive bradycardias. She was successfully re-intubated after 2 attempts. The baby was sucked out twice. At 11:02 FIO2 was raised to 79\%.\\

\emph{Example BTNurse output (extract)}:\\
\textbf{Respiratory Support}\\
\textbf{Current Status}\\
Currently, the baby is on CMV in 27 \% O2. Vent RR is 55 breaths per minute. Pressures are 20/4 cms H2O. Tidal volume is 1.5.\\

SaO2 is variable within the acceptable range and there have been some desaturations.\\

\emph{Example BTFamily output (extract)}:\\
John was in intensive care. He was stable during the day and night. Since last week, his weight increased from 860 grams (1 lb 14 oz) to 1113 grams (2 lb 7 oz). He was nursed in an incubator.

\caption{Example outputs from Babytalk systems}
\label{fig:BabytalkOutputs}
\lineacross
\end{figure}

At a high level, the Babytalk architecture works as follows
\begin{itemize}
\item \emph{Signal analysis}\index{signal analysis} is done using standard tools, such as pattern detection and noise identification algorithms.
\item \emph{Data interpretation} is quite complex, in part because the system tries to detect and recover from some types of input errors (eg, incorrect time stamps in the patient record).  Much of this is done using \emph{production rules}.
\item \emph{Document planning} is done using a combination of fixed \emph{schemas} and dynamic narrative-creation algorithms.
\item \emph{Microplanning} is done using algorithms for reference, aggregation, lexical choice, and other microplanning tasks.
\item \emph{Surface realisation}\index{surface realisation}  is done using the \emph{Simplenlg}\index{simplenlg} library \cite{gatt-reiter-2009-simplenlg}.
\end{itemize}
Babytalk's processing is much more complex than DrivingFeedback's processing; at the time of writing, it is one of the most sophisticated rule-based NLG systems ever built.

\section{Signal analysis}\label{sec:signalanalysis}
The first stage of the data-to-text\index{data-to-text} pipeline\index{pipeline} is \emph{signal analysis}\index{signal analysis}, that is finding patterns in the input data using signal processing techniques.  

\subsection{Noise detection: Principles}\label{sec:noise}

Part of signal analysis is \emph{noise detection\index{noise detection} and removal}; real-world sensor data is noisy and hence sensor data may not be an accurate measurement of real-world events.  For example, the DrivingFeedback\index{DrivingFeedback} driving data in Figure~\ref{tab:drivingfeedbackdata} shows a period of speeding from 9:00:01 to 9:00:05, except for a single data point at 9:00:03 where the speed drops to 30.  Since GPS data is noisy (position measurements are not exact), DrivingFeedback treats this as noise, and reports a single speeding segment from 9:00:01 to 9:00:05 (Figure~\ref{fig:drivingfeedbackpipeline}).

\begin{figure}
\includegraphics[scale  = 0.75]{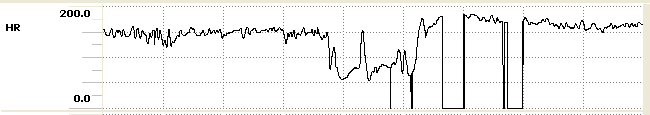}
\caption{Noisy data (HR, heart rate) from Babytalk.  The data shows HR falling to 0, but the baby's heart did not actually stop beating!}
\label{fig:btdata}       
\end{figure}

Babytalk\index{Babytalk} systems got data from sensors attached to babies in a neonatal intensive care unit.  When the baby kicked or was picked up by a nurse, the sensors would often show dramatic spikes or other changes which were purely due to the sensor momentarily losing contact with the baby's skin.  An example is shown in Figure~\ref{fig:btdata}; heart rate drops to zero at several points, and this is noise (the baby's heart did not stop beating!).
At any rate, these artefacts needed to be identified and removed so that Babytalk could focus on actual changes instead of sensor artefacts.

\personal{In the DrivingFeedback and Babytalk contexts,  it at least is possible for noisy sensors to be replaced.  I have worked on projects for oil\index{oil industry} companies where we used sensor data from sensors deep inside an oil well; it is not possible to replace such sensors if they start misbehaving.}

\subsection{Pattern detection: Principles}\index{pattern detection}

Once noise processing is done, the main signal analysis task starts, which is detecting patterns in the data.   Which patterns are worth detecting depends on the domain.  Statistics such as mean, range, and standard deviation are often useful.  With time-series\index{time series} data, in many use cases users want to know about spikes, trends, oscillations, and periods where the sensor data is outside of an acceptable range (such as driving speed being above the speed limit).  But there are also specialised patterns which are important in some some domains but not others.

Pattern detectors in NLG systems must find patterns that human readers recognise.  Many spike detectors, for example, use mathematical algorithms which trigger on things which do not look like spikes to human users; this is fine for most analytics\index{analytics}, but not for NLG.  In general, we want patterns that make sense to readers; finding such patterns is called \emph{articulate analytics}\index{analytics!articulate analytics}.

\begin{figure}
\includegraphics[scale  = 1]{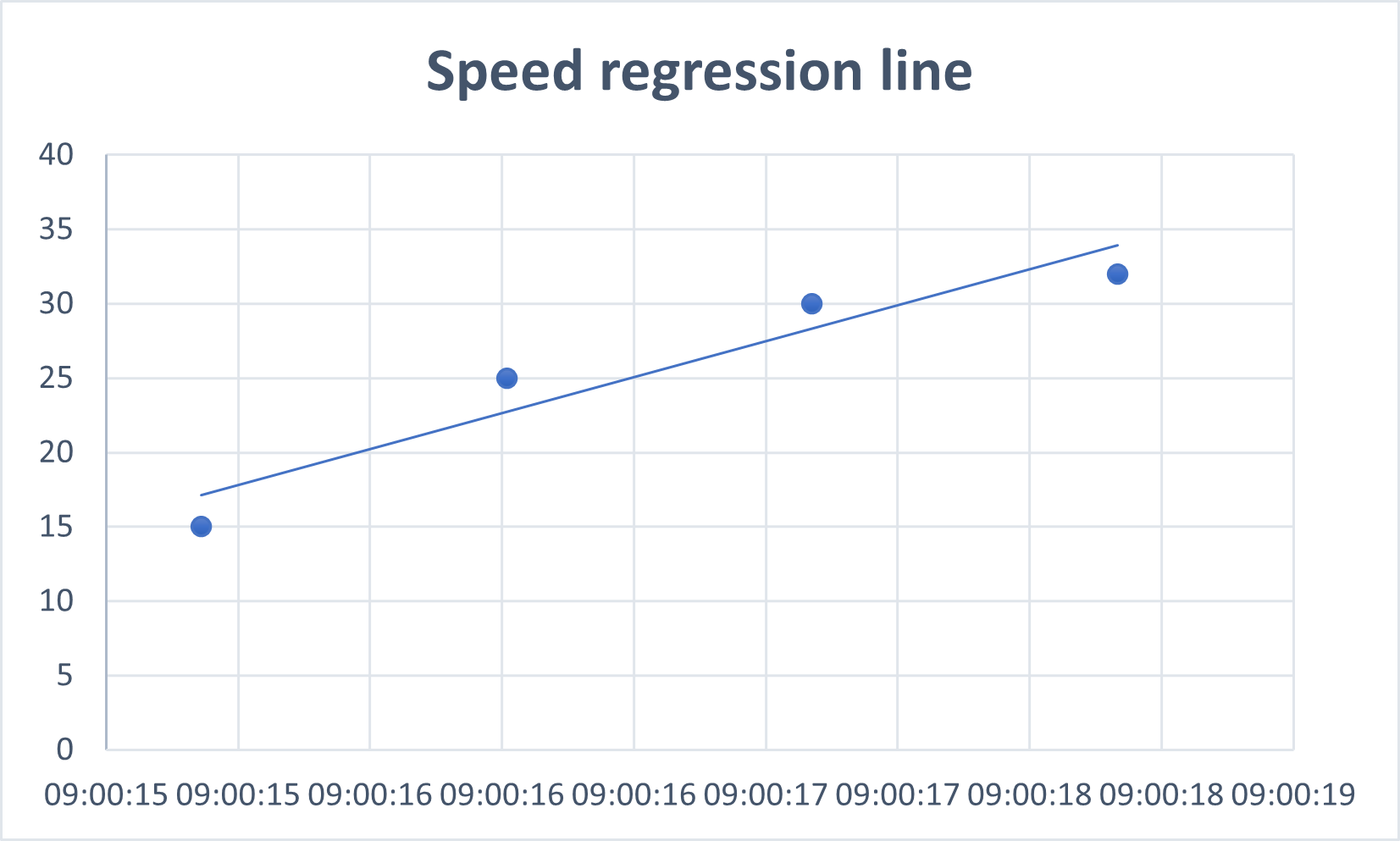}
\caption{Regression trend line for speed data.  Initial value is 15 mph and final value is 32 mph.  Regression line is 17 mph at the start of the period, and 33 mph at the end. Readers prefer to see this data described as \lingform{you speeded up from 15 to 32 mph} (with actual values at beginning and end of period) instead of \lingform{you speeded up from 17 to 33 mph} (regression line)}.
\label{fig:regression}       
\end{figure}

One example of articulate analytics is when NLG systems use linear approximation to describe time series data.  For example, suppose we wanted to describe driving behaviour on St Machar Drive for the data in Figure~\ref{tab:drivingfeedbackdata}, for the period 9:00:15 to 9:00:18, where the driver is increasing speed.  We could use:
\begin{itemize}
\item a simple \emph{linear interpolation}\index{linear interpolation}, and state the speed at the beginning and end of this period.  An example is \emph{After turning on to St Machar Drive, you speeded up from 15 to 32 mph}. 
\item a \emph{linear regression}\index{linear regression} (trend line) to create a \scare{best-fit} segment to describe this driving behaviour, as shown in Figure~\ref{fig:regression}.  An example is \emph{After turning on to St Machar Drive, you speeded up from 17 to 33 mph}.
\end{itemize}
The linear regression is a better fit to the actual data, and hence used by many analytics algorithms.  However, in an NLG context, most readers find the regression text confusing and prefer the interpolation text \cite{sripada:kdd03}.  Hence linear interpolation should be used by data-to-text systems in such contexts, even though regression is more popular in many other analytical contexts

\subsection{Techniques for signal analysis}
Signal analysis is usally done with standard noise suppression and pattern detection algorithms; the NLG developer generally selects what he or she thinks is the most appropriate algorithm from a standard pattern analysis library such as Python's scipy\index{scipy}, and perhaps with some domain tuning.

For example, Babytalk\index{Babytalk} used fairly standard auroregressive modelling for noise detection, which was tuned on a clinical dataset where artefacts had been manually annotated \cite{HUNTER2012157}.   Babytalk used a variety of algorithms for pattern detection, including bottom-up segmentation for detecting trends \cite{Keoghsegmentation}.

\section{Data interpretation}
The next stage of the data-to-text\index{data-to-text} pipeline\index{pipeline} (after signal analysis) is \emph{data interpretation}\index{data interpretation}.  The goal of data interpretation is to extract useful \emph{insights}\index{insights} (sometimes called \emph{messages}\index{messages}) from the patterns detected by signal analysis\index{signal analysis}, and indeed from the raw data in some cases.  This step is of critical importance in data-to-text; an NLG system which presents useless or (even worse) incorrect insights is not going to be useful regardless of the quality of the language it generates. From a practical perspective, it is not at all unusual for NLG projects to devote more developer time to data interpretation than anything else.

\subsection{Principles}\label{sec:dp-principles}

The insights detected by data interpretation are quite varied, but often can be characterised as one of the following
\begin{itemize}
\item \emph{Abstraction:}\index{data interpretation!abstraction} combining patterns from signal analysis into a higher-level insight.  For example, clustering together individual \semantic{Speeding} patterns into a \semantic{SpeedingCluster} insight, as shown in Figure~\ref{fig:drivingfeedbackpipeline}.
\item  \emph{Interpretation:}\index{data interpretation!interpretation}  Interpreting patterns.   For example, DrivingFeedback\index{DrivingFeedback} could interpret a \semantic{Speeding} pattern or \semantic{SpeedingCluster} insight as being \lingform{very dangerous} if the maximum speed exceeded the speed limit by 20 mph or more.
\item \emph{Linkage:}\index{data interpretation!linkage}  Relating events; this is very important for generating cohesive narrative instead of a bullet list of insights.  For example, if a driver did some speeding at the beginning of a trip and afterwards did no speeding, the system could say \lingform{You speeded on King St, but afterwards drove safely}.  Here, \lingform{but} links the two insights \lingform{You speeded on King St} and  \lingform{afterwards drove safely}.
\item \emph{Importance:}\index{data interpretation!importance}  How important and relevant are insights to the user?  In the example shown in Figure~\ref{fig:drivingfeedbackpipeline}, for example, we might give more importance to \semantic{SpeedingCluster(9:00:01-9:00:10, King Street, maxspeed=35, incidents=2)} than to \semantic{Speeding(9:00:18-9:00:19, St Machar Drive, 32)}, since the cluster is over a longer time frame and has a higher maximum speed.
\end{itemize}

One especially important type of Linkage is \emph{casual reasoning}, for example \lingform{the baby's heart rate increased because of a nappy change}.  Causal reasoning is difficult, but if it can be done, adding causal links increases the quality and usefulness of generated narratives.

\subsection{Techniques}

Data interpretation is domain-dependent and is based on what insights users want to know.  It is usually is done by writing  rules or algorithms in consultation with users and domain experts\index{domain!experts}; these rules are based on domain knowledge\index{domain!knowledge} and often use domain-specific data analytics and mining techniques.

Once an NLG system is running, users (and domain experts) often want to modify and tweak data interpretation logic; in DrivingFeedback, for example, a user might tell the system to ignore minor speeding incidents (less than 5 mph over the speed limit).  Some commercial NLG systems have user interfaces which allow users or  domain experts to inspect and modify data interpretation logic.

It should be possible in many cases to learn data interpretation models using machine learning, and hopefully we will see more work on this in the future.  Because domain experts want to understand interpretation rules (as mentioned above), it often makes sense in this context to use ML techniques which produce interpretable models\index{models!interpretable} (such as linear regression and decision trees) which can be inspected, checked, and updated by domain experts; non-interpretable models (such as neural models) are often less desirable, even if they perform well.

Data interpretation can help in dealing with noise\index{noise detection}, in contexts where noise can best be detected by rules and consistency checks instead of by pattern analysis algorithms.
In Babytalk's\index{Babytalk} data, for example, timestamps of actions were often incorrect.  When clinicians performed surgery or other interventions, a nurse entered this information into the patient record, along with the time that the procedure was done.  However this was usually done after the surgery finished, and sometimes nurses did not remember the exact timing of the intervention, especially if the nurse had to look after the baby immediately afterwards, and did not get around to updating the patient record until much later.  Pilot studies  suggested that human readers could be very confused by incorrect intervention times, especially if this meant that intervention times did not link up with the sensor data (e.g., blood oxygen seemingly rose before the oxygen level in the baby's incubator was raised, instead of afterwards).  For this reason, Babytalk's data interpretation module attempted to identify and fix incorrect timestamps, by looking for the expected sensor signature of the intervention in the sensor data.

\section{Document Planning}
The third stage of the pipeline\index{pipeline} is \emph{Document Planning\index{document planning}}.  Its goal is to determine the content\index{content} and structure of the generated text.   Although in theory separable, in practice decisions about content and structure tend to be connected, which is why they are usually grouped together into this stage.

\subsection{Principles}

Document Planning is the interface between the analytics\index{analytics} part of the pipeline (signal analysis and data interpretation) and the linguistics\index{linguistics} part of the pipeline (microplanning and surface realisation).  Decisions about which content to include are related to calculations of importance (data interpretation)\index{data interpretation!importance}, while decisions about document structure\index{document structure} influence linguistic decisions such as where to place sentence and paragraph breaks (Section~\ref{sec:aggregation})

In the DrivingFeedback\index{DrivingFeedback} example we are using, the Document Planning module decides that the text should include the \emph{SpeedingCluster} (King Street) insight but not the \emph{Speeding} (St Machar Drive) pattern.  This decision is primarily driven by importance (discussed above).

In some cases, it is useful to add unimportant insights\index{insights} because they improve narrative\index{narrative} coherence\index{coherence}.  For example, the Babytalk\index{Babytalk} system gave higher importance to sudden changes in sensor data than to slower changes, since sudden changes are more significant clinically.  The first version of the Babytalk document planner only selected high-importance insights, which led to texts such as

\begin{quote}
\lingform{TcPO2 suddenly decreased to 8.1. SaO2 suddenly increased to 92. TcPO2 suddenly decreased to 9.3}
\end{quote}
\noindent{}Users complained that it made no sense for TcPO2 to decrease to 9.3 when it had previously decreased to 8.1.  In this case, TcPO2 had risen between the events, but slowly, so the rise was not considered clinically important; hence it was not mentioned.  But narrative coherence is better if we nevertheless mention the rise \cite{reiter-etal-2008-importance}, for example by adding the phrase \lingform{After increasing to 19}:
\begin{quote}
 \lingform{TcPO2 suddenly decreased to 8.1. SaO2 suddenly increased to 92. After increasing to 19, TcPO2 suddenly decreased to 9.3}.
 \end{quote}

\subsection{Techniques}

From an algorithmic perspective, most NLG systems use simple scripts\index{scripts} or schemas\index{schemas} for document planning.  In other words, a script or piece of code defines the overall document structure, and also specifies which insights should be included in the document at different points; of course the script can include conditionals, since the structure may depend on the insights produced by data interpretation.  A simple DrivingFeedback example is shown in Figure~\ref{sec:docplanscript}.

\begin{figure}
\lineacross\\
\begin{verbatim}
procedure DocumentPlan(Insights, importance function, importance threshold)
  
  SelectedInsights = all Insights with importance above a threshold
  
  # we need at least one insight
  if SelectedInsights is empty, add the highest importance Insight
  
  # insights should be presented in time order in the text
  OrderedSelectedInsigts = SelectedInsights ordered by start time of Insight
  
  return OrderedSelectedInsights
\end{verbatim}
\caption{Pseudocode for document-planning script for DrivingFeedback}\label{sec:docplanscript}
\lineacross
\end{figure}

More complex approaches are possible.  For example, the Babytalk system uses scripts (in the above sense) for part of the documents it generates, but for other parts it uses a \emph{key event} algorithm which selects a small number of key insights about events, and then creates a paragraph around each of these events by finding linked\index{data interpretation!linkage} insights, such as causes and consequences of the key event \cite{PORTET2009789}.

Attempts have been made to treat document planning as a reasoning task, where the system explicitly reasons about what information should be in the generated text \cite{APPELT19851}; however it is difficult to get this approach to work robustly in real-world contexts.  Attempts have also been made to use psycholinguistic\index{psycholinguistics} principles to inform document structure decisions \cite{thomson-etal-2018-comprehension}, but it is not easy to translate psycholinguistic theories into workable NLG code.

Researchers are looking at using machine learning techniques to determine document structure \cite{puduppully-etal-2022-data}, but document planning seems quite challenging for machine learning.  In general neural NLG  systems seem to be better at expressing content than at selecting content.

One reason for the continuing strength of the script/code approach is that it makes it relatively easy to modify document content and structure if users request this; changes are much harder with the other approaches mentioned above.

The output of the document planning process is a \emph{document plan}\index{document plan}.  There is no agreed standard for representing document plans.  Some developers use trees inspired by Rhetorical Structure Theory\index{Rhetorical Structure Theory} \cite{mann1987rhetorical}, but ad-hoc data structures are more common.

\section{Microplanning}
The fourth stage of the NLG pipeline\index{pipeline} is \emph{Microplanning}\index{microplanning}.  Its goal is to decide how to linguistically express the selected insights\index{insights} and other messages\index{messages}; which words to use, what syntactic\index{syntax} structures to use, etc.   We can think of microplanning as tackling a number of conceptually discrete tasks, including \emph{lexical choice}\index{lexical choice}, \emph{reference}\index{referring expressions}, and \emph{aggregation}\index{aggregation}; see example in Table~\ref{tab:mptasks}.

\begin{table}
\begin{tabular}{|l|l|l|} \hline
Task & Version 1 & Version 2 \\ \hline
Lexical choice & John \emph{bought} a book for £10. & John \emph{purchased} a book for £10. \\ \hline
Reference & \emph{John} bought a book for £10. & \emph{He} bought a book for £10. \\ \hline
Aggregation & John bought a book for £10. & John bought a book. It cost £10.\\ \hline
\end{tabular}
\caption{Different Microplanning choices for expressing \semantic{buy(John,book,10)}.}
\label{tab:mptasks}
\end{table}

\subsection{Lexical choice}\label{sec:lexchoice}
\emph{Lexical choice}\index{lexical choice} is the task of choosing words to express concepts and insights\index{insights}.   In the DrivingFeedback\index{DrivingFeedback} example shown in Table~\ref{fig:drivingfeedbackpipeline}, for instance, the microplanner decides to express \semantic{incidents=2} using the word \lingform{twice}; it could also have chosen \lingform{on two occasions}, \lingform{on 2 occasions}, \lingform{several times}, etc.

\subsubsection{Lexical choice: Principles}\label{sec:lexprinciples}
The core lexical choice task is to map semantic\index{semantics} content onto words.  This is related to \emph{lexical semantics}\index{semantics!lexical semantics} in linguistics\footnote{\url{https://en.wikipedia.org/wiki/Lexical\_semantics}}\index{linguistics}.

Sometimes lexical choice is straightforward, but in other cases challenges arise.
One is that in many contexts it is important to vary which words are used, for example to express \semantic{incidents=2}  using \lingform{twice} in some cases but \lingform{on two occasions} in others.  \emph{Lexical variation}\index{lexical variation} makes texts more interesting and less repetitive, especially in contexts where users receive regular reports (such as weekly driving feedback reports).  We can also vary other linguistic choices, such as passive vs active voice, if this makes texts less repetitive to readers.

Another key issue is \emph{individual variability}\index{individual variability}, i.e. different people use words in different ways; for example my daughter describes as \lingform{purple} a shirt that I would call \lingform{pink}.  In the SumTime\index{SumTime} project on generating weather forecasts\index{weather forecasts} \cite{REITER2005137}, we empirically explored how forecasts readers and writers used and interpreted time phrases such as \lingform{by evening}.
We discovered that \emph{by evening} was used by some forecasters to mean 6PM and by others to mean midnight (Table~\ref{tab:sumtimebyevening}). We also worked with forecast readers, and discovered that a few readers thought the meaning of \emph{by evening} depended on sunset time and hence season, and/or on when people normally had their last meal of the day, and hence culture.

Other researchers have also found differences in how individuals user and understand words.  Berry et al \cite{berry2011} looked at how people interpreted phrases such as \emph{very common} which are used to describe risk\index{risk}, and discovered considerable variation, and also little agreement with usage recommended by official terminologies\index{terminology}.  Ramos Soto et al \cite{ramos-soto-etal-2018-meteorologists} found differences in how people used and interpreted geographic terms.

\begin{table}
\begin{tabular}{|l|l|l|l|l|l|}  \hline
time & F1 & F2 & F3 & F4 & F5 \\ \hline
1800 & \textbf{30} & 5 & 2 & \textbf{27} & 13 \\
2100 & 13 & 6 & 8 & 2 & 11 \\
0000 & 2 & \textbf{9} & \textbf{80} & \textbf{5} & \textbf{14} \\
other & 2 & 2 & 1 & 0 & 4\\ \hline
\end{tabular}
\caption{Number of times five forecasters (F1, F2, F3, F4, F5) used \lingform{by evening} to refer to different times in the SumTime weather forecast corpus. Most common usage by a forecaster is in \textbf{bold} font. F1 and F4 usually used \lingform{by evening} to refer to 1800, F3 usually used \lingform{by evening} to refer to 0000, and F2 and F5 were more varied in their usage \cite{reiter-cl02}.}
\label{tab:sumtimebyevening}
\end{table}

\subsubsection{Lexical choice: Techniques}

From an algorithmic perspective, in many cases developers can just write rules on how information is expressed.  For instance, write a rule that the information \semantic{incidents=2} is expressed by the word \lingform{twice} (or varies between \emph{twice} and \emph{on two occasions}).    Rules may depend on users.  For example a medical\index{health} system such as Babytalk\index{Babytalk} may express a medical concept such as \semantic{Desaturation} as \lingform{desatuation} to a clinician, but \lingform{temporary drop in blood oxygen saturation} to a patient who is not familiar with medical terminology (although sometimes patients object to \scare{dumbed-down} language and prefer medical terminology\index{terminology}).

Developers can also use use machine learning to build models for word choice and selection.  A number of researchers have investigated building models for verb choice in financial texts, such as \lingform{rose} or \lingform{soared}.  Chen and Yao \cite{chen-yao-2019-closer} analysed this task and essentially concluded that simple statistical models might be preferable to complex ML models.

ML and rules can be combined.  For example, in SumTime\index{SumTime} we essentially used ML to build an interpretable model\index{models!interpretable} to select time phrases; the data included an \semantic{author} feature.  We inspected the model to identify words and phrases whose meaning was author-dependent (such as \lingform{by evening}, see Table~\ref{tab:sumtimebyevening})  and then adapted the model (in consultation with a domain expert\index{domain!experts}) so that it never chose such words.
The result was a model which chose common words and phrases whose usage was stable across users (eg, phrases such as \lingform{by midnight}, which everyone agreed meant 0000).
User evaluations showed that this strategy worked well in selecting time phrases \cite{REITER2005137}.

\subsection{Generating Referring Expressions}
A \emph{referring expression}\index{referring expressions} identifies an entity to the reader.  For example, I can be referred to as \lingform{Ehud Reiter}, \lingform{Professor Reiter}, and \lingform{him}, amongst many other forms.  If a generated text refers to me, which of these forms should it use?    In a sense, choosing referring expressions is a specialised form of lexical choice\index{lexical choice}, but usually it has been treated as a distinct problem in the research literature.

\subsubsection{Referring expressions: Principles}

Reference often depends on \emph{context}\index{context}, which linguists call \emph{pragmatics}\index{pragmatics}.  Different referring expressions are appropriate in different contexts.  For instance, in our DrivingFeedback\index{DrivingFeedback} example, \lingform{King Street} is a reference to a particular street.  The referring expression can also provide information about the city, ie \lingform{King Street, Aberdeen}; this differentiates the street from \lingform{King Street, Dundee}.   If the reader knows from context (previous sentences, associated map, etc.) that the text is referring to driving in Aberdeen, then it is fine to just say \lingform{King Street}.  However if context does not give the city, then the text should say \lingform{King Street, Aberdeen} (or indeed \lingform{King Street, Aberdeen, UK} if context does not specify the country).

In general, referring expressions should uniquely identify their \emph{target}\index{reference target} in the current context.  For example, if I want to refer to the student Miriam Smith in a class where no other student is called Miriam, then I can just refer to her as \emph{Miriam}.  If however the class also contains student called Miriam Black, then \emph{Miriam} on its own is ambiguous, and I should instead use the referring expression \lingform{Miriam Smith}.

Reference is complex, in part because there are so many different forms; not just reference to people, but also to objects, times, geographic locations, abstract concepts, etc.   Many of these specific types of reference use specialised words and linguistic\index{linguistics} constructs which do not work for other types of reference, such as \lingform{yesterday} for temporal reference.
Also, in many cases the \emph{initial reference} to an entity in a text uses a different form from \emph{subsequent references}; for example \lingform{Professor Ehud Reiter} could be used as an initial reference to me, while \lingform{Reiter} is used subsequently.

\subsubsection{Referring expressions: Techniques}

A variety of algorithms have been proposed in the research literature, especially for the referential task of choosing a \emph{definite descriptions}\index{definite descriptions} such as \lingform{a big black dog} to refer to a visually or otherwise salient entity \cite{krahmerdeemter}.   Algorithms have also been proposed for pronoun\index{pronouns} usage \cite{kibble-power-2004-optimizing}, and for special types of reference such as referring to components of a complex machine \cite{reiter-2017-commercial}.  The above-mentioned research uses algorithms and rules, but there is also work on using neural techniques for reference \cite{CHEN2023101466}.    van Deemter \cite{vanDeemterReference} presents the reference problem from a cognitive science\index{cognitive science} perspective.

One problem with reference algorithms is that different referring expressions are used in different genres and contexts.  For example, pronouns are more common in informal writing than in legal documents and other types of very formal writing; this makes it difficult to propose universal pronoun-selection algorithms.  For this reason, commercial work on reference has put more emphasis on configurability\index{configurability} and flexibility, so that developers can easily configure the types of referring expressions used in their documents \cite{reiter-2017-commercial}.

Regardless of the specific algorithm used, the microplanner will need access to a model of relevant contextual information in order to choose referring expressions.

\subsection{Aggregation}\label{sec:aggregation}
\emph{Aggregation}\index{aggregation} is the task of packaging information into sentences.  For example, supose the DrivingFeedback\index{DrivingFeedback} document planner\index{document planning} had decided to communicate insights about speeding on St Machar Drive as well as speeding on King St (that is, both of the Data Interpretation insights shown in Figure~\ref{fig:drivingfeedbackpipeline}).  In this case, the information could be communicated in one sentence or in two sentences.
\begin{itemize}
\item You sped twice on King Street and once on St Machar Drive.  \emph{(one sentence)}
\item You sped twice on King Street.  You also sped on St Machar Drive.  \emph{(two sentences)}
\end{itemize}

\subsubsection{Aggregation: Principles}

In general, there are many ways of distributing information across sentences.   Since complex sentences may be difficult for below-average readers to understand (including many non-native speakers\index{non-native speakers}), while simple sentences are understood both by below-average and above-average readers, most NLG systems opt to use a large number of simple sentences. 

If related insights\index{insights} or messages\index{messages} are aggregated into the same sentence, sometimes \emph{ellision}\index{elision} can be used to reduce the length of the aggregated sentence.  For example, \lingform{Sales roles in Germany} and \emph{Sales fell in France} can be aggregated into \emph{Sales rose in Germany and fell in France} (instead of \emph{Sales rose in Germany and \underline{sales} fell in France}).  Elision can make a useful contribution to shortening texts and indeed making them more readable.

One issue with aggregation is that it can lead readers to make inferences\index{inferences} about how insights or messages are releated to each other.    For example, in the Babytalk\index{Babytalk} context, if the phrases \lingform{The nurse gave the baby morphine} and \lingform{The baby vomited} are aggregated into \lingform{The nurse gave the baby morphine and the baby vomited}, readers may infer a causal link, i.e. that the baby vomited \emph{because} of the morphine.  Which is misleading if something else (such as illness) could have caused the vomiting.
Because of such concerns, Babytalk used a very cautious and conservative aggregation strategy.

\subsubsection{Aggregation: Techniques}

From an algorithmic perspective, Harbusch and Kempen \cite{harbusch-kempen-2009-generating} proposed a set of linguistically-motivated\index{linguistics} algorithms for doing some types of aggregation and ellipsis; some of these are implemented within the \emph{Simplenlg}\index{simplenlg} package.    Different kinds of aggregation are appropriate in different domains, as well as for different users, which suggests that aggregation techniques should be adapted to domains and users; adaptation can be done based on user feedback \cite{walker2007individual} and/or by learning from appropriate corpora \cite{white-howcroft-2015-inducing}.

From a commercial perspective, most systems which I am aware of restrict themselves to relatively simple aggregations within the microplanner, done by straightforward rules or scripts; this is partially
because of the risk of unwanted inferences (such as morphine causing vomiting).  There tends to be more emphasis on \emph{conceptual aggregation}\index{conceptual aggregation}, usually done as a type of abstraction\index{data interpretation!abstraction} within data interpretation. For example generating an insight that sales rose in every quarter of 2020 by abstracting over individual insights that sales rose in Q1 2020, Q2 2020, Q3 2020, and Q4 2020.

\section{Surface Realisation}\label{sec:realisation}\index{surface realisation}
A \emph{surface realiser} generates actual texts in English or other languages, based on the linguistic\index{linguistics} decisions made in the microplanner\index{microplanning}.

\subsection{Principles}
Conceptually, the realiser takes care of grammatical\index{grammar} details so that the rest of the NLG system does not need to worry about this.  This includes
\begin{itemize}
\item \emph{Syntax}:\index{syntax} For example, forming the negated version of a sentence.  In English, this requires adding \lingform{do} in some cases but not others; for example \lingform{I do not like you} (includes \lingform{do})  but \lingform{I have not met you} (no \lingform{do}).  The realiser takes care of \lingform{do}-insertion so that other parts of the NLG system do not need to worry about this.
\item \emph{Morphology}:\index{morphology} For example, forming the plural form of a noun.  In English, this is usually done by adding \lingform{s} to the end of a word, but there are many exceptions; for instance the plural of \lingform{child} is \lingform{children}, not \lingform{childs}.  Again the realiser takes care of forming plurals so the rest of the system does not need to worry about this.
\item \emph{Orthography}:\index{orthography} For example, in English \lingform{.} is usually added to the end of a sentence, but there some exceptions, for instance we do not add a \lingform{.} if the sentence ends in an abbreviation which ends in \lingform{.}  (\lingform{I went to Washington D. C.} , not \lingform{I went to Washington D. C..} ) \cite{nunberg1990linguistics}.
\end{itemize}
Additional types of grammatical processing are needed in some languages.  For example an English surface realiser must also decide between \lingform{a} and \lingform{an} (Section~\ref{sec:aan}); this is a \emph{morphophonology}\index{morphophonology} task

The above examples are in English\index{English}, but all naturally-evolved languages have grammatical details which can be dealt with by a realiser\footnote{Artifically designed and constructed languages such as Esperanto\index{Esperanto} have more logical and consistent grammars.}.  For instance in French\index{French} we need to replace \lingform{de le} by \lingform{du}; in Mandarin\index{Mandarin} we need to add classifiers to noun phrases in some cases \cite{chen-etal-2018-simplenlg}; in German\index{German} we need to deal with separable verbs \cite{braun-etal-2019-simplenlg}; etc.

\subsection{Techniques}\label{sec:realisationTechniques}
A number of open-source software libraries\index{libraries} have been created to do the above tasks in a fairly straightforward way, of which the best known (at the time of writing) is probably Simplenlg\index{simplenlg} \cite{gatt-reiter-2009-simplenlg}\footnote{\url{https://github.com/simplenlg/simplenlg}}.  Simplenlg started off as a Java-based English realiser, but has subsequently been ported to Python and other programming languages, and also adapted to work in many other human languages, including German \cite{braun-etal-2019-simplenlg},  Mandarin \cite{chen-etal-2018-simplenlg}, and Galician\index{Galician} \cite{cascallar-fuentes-etal-2018-adapting}.  A simple example of Simplenlg is shown in Figure~\ref{fig:simplenlg}.  There are other packages with similar functionalities, such as pyrealb\index{pyrealb} \cite{lapalme2023datatotext}\footnote{\url{https://pypi.org/project/pyrealb/}}; a French example of pyrealb is shown in Figure~\ref{fig:pyrealb}.

\begin{figure}
\lineacross\\
    SPhraseSpec p = nlgFactory.createClause();\\
    p.setSubject("Mary");\\
    p.setVerb("chase");\\
    p.setFeature(Feature.TENSE, Tense.PAST);\\
    p.setObject("the monkey");\\
    String output2 = realiser.realiseSentence(p); \\
\caption{Simplenlg example; output is \lingform{Mary chased the monkey.}}
\label{fig:simplenlg}

\lineacross\\

loadFr()\\
print(S(\\
\hspace*{1cm}NP(D("le"),N("chat"),A("petit"),)\\
\hspace*{1cm}VP(V("sauter").t("ps")) \\
\hspace*{0.5cm}).realize())\\
\caption{pyrealb example; output is \lingform{ Le petit chat sauta.}}
\label{fig:pyrealb}
\lineacross
\end{figure}

There are also open-source packages which just do morphology, such as the Python \emph{inflect}\index{inflect (Python package)} package for English\footnote{\url{https://pypi.org/project/inflect/}} and Abed's Arabic\index{Arabic} language functions \cite{abed-reiter-2020-arabic}.  These are essentially \emph{language functions}\index{language functions} as defined in Section~\ref{sec:languagefunctions}.

We can also use statistical and machine-learning models in surface realisation.  One approach is \emph{over-generate\index{over-generation} and select}, where the realiser generates a number of different possible sentences (surface forms), using rule-based techniques, and then a statistical or ML model is used to choose the best of these \cite{langkilde-knight-1998-generation-exploits}.  The OpenCCG\index{OpenCCG} realiser library \cite{white2007towards} includes this capability.

Large neural language models\index{large language models} (Section~\ref{sec:foundationmodels}) are very good at realisation and getting grammatical details correct.  Such models can be used for surface realisation as defined in this section \cite{mille-etal-2020-third}, but its more common for them to be used for linguistic expression (ie, including microplanning\index{microplanning} as well as realisation) or indeed for the entire generation task (Section~\ref{sec:modularML}).

From a commercial perspective, many fielded systems use relatively simple realisation processing, perhaps just morphology and simple syntactic processing \cite{weissgraeber-madsack-2017-working}.   An open-source toolkit designed to support such an approach is the Rosaenlg\index{Rosaenlg} library\footnote{\url{https://rosaenlg.org/}}.

\section{Template NLG}\label{sec:templatenlg}
Instead of using a modularised pipeline, developers can use \emph{templates}\index{templates} to generate texts in a single step from the input data.  The template concept is a vague one, and ranges from simple mail-merge\index{mail merge} templates to complex scripting\index{scripts} languages such as Jinja2\index{Jinja} which include conditional statements, arithmetic computations, and other programming constructs.  I use it here to cover all approaches where rules or algorithms are used to generates texts in a single step without any decomposition into tasks such as data interpretation or microplanning.

\subsection{Principles}

\begin{figure}
\lineacross\\
Simple template for sentence such as \lingform{The Washington Wizards defeated the Los Angeles Lakers, 111-95.}\\
\begin{verbatim}
if (teamA.score > teamB.score)
    ''The [teamA.name] defeated the [teamB.name] [teamA.score] - [teamB.score].''
else if (teamB.score > teamA.score)
   ''The [teamB.name] defeated the [teamA.name] [teamB.score] - [teamA.score].''
else
   ''The [teamA.name] tied the [teamB.name] [teamA.score] - [teamB-score].''
\end{verbatim}
\caption{Simple template to produce a sentence describing the winner of a sporting match.}
\label{fig:simpletemplate}
\lineacross
\end{figure}

\begin{figure}
\lineacross\\
Slightly more complex template for sentence such as \lingform{The Washington Wizards defeated the Los Angeles Lakers, 111-95.}\\
\begin{verbatim}
if (teamA.score >= teamB.score+20)
    ''The [teamA.name] destroyed the [teamB.name] [teamA.score] - [teamB.score].''
else if (teamA.score >= teamB.score+5)
    ''The [teamA.name] defeated the [teamB.name] [teamA.score] - [teamB.score].''
else if (teamA.score >= teamB.score)
      ''The [teamA.name] edged the [teamB.name] [teamA.score] - [teamB.score].''
else if (teamB.score >= teamA.score+20)
    ''The [teamB.name] destroyed the [teamA.name] [teamB.score] - [teamA.score].''
else if (teamB.score >= teamB.score+5)
    ''The [teamB.name] defeated the [teamB.name] [teamB.score] - [teamA.score].''
else if (teamB.score >= teamB.score)
     ''The [teamB.name] edged the [teamA.name] [teamB.score] - [teamA.score].''
else
    ''The [teamA.name] tied the [teamB.name] [teamA.score] - [teamB-score].''
\end{verbatim}
\caption{Slightly more complex template to produce a sentence describing the winner of a sporting match.}
\label{fig:morecomplextemplate}
\lineacross
\end{figure}

A simple example of a template is shown in Figure~\ref{fig:simpletemplate}.  The input to the system is data about a sports match, including names and scores of teams; the output is a simple sentence saying who won the game.  Writing this template is much simpler than building a complete modularised NLG pipeline\index{pipeline}!

Of course, real sports-writing\index{sports-writing} applications probably need something more sophisticated.  For example, the system may want to vary the verb depending on the size of the victory, using \lingform{destroyed} for a victory of 20 points or more, \lingform{edged} for a victory of less than 5 points, and \lingform{defeated} otherwise.  This could be done using the template shown in Figure~\ref{fig:morecomplextemplate}.

There are many other desirable improvements, including:
\begin{itemize}
\item Varying words (Section~\ref{sec:lexprinciples})\index{lexical variation}, for example in some cases using \lingform{beat} instead of \lingform{defeated}.
\item Referring\index{referring expressions} to teams just by city name (eg, \lingform{Washington}) where contextually appropriate.
\item Adding interesting insights\index{insights}, such as winning steaks (\lingform{Washington has won its fifth match in a row}).
\item etc.
\end{itemize}
In principle we can implement all of the above using a simple template structure such as the one in Figure~\ref{fig:morecomplextemplate}, but the code rapidly becomes very long and complex.  At some point it will become easier to introduce modules or functions to choose the verb (\lingform{destroyed}, \lingform{defeated}, \lingform{beat}, \lingform{edged}), decide how to refer to a team, decide whether to mention a winning streak, etc.  This can be done on an ad-hoc basis, especially for student projects. In commercial contexts, though, structuring the NLG process into distinct modules (e.g., treating verb choice as a lexical choice\index{lexical choice} process, as described in Section~\ref{sec:lexchoice}) is better because it makes it easier to reuse and maintain\index{maintenance} code.  Dealing with edge cases\index{edge cases} and exceptional conditions is also easier with a structured modularised approach.

In short, templates make sense for simple projects with straightforward input data, and indeed for demos that only need to work on a few examples.  However, a structured approach (with modules and representations) works better for complex projects, especially in real-world commercial contexts.  One exception to this is that domain experts\index{domain!experts} with limited background in programming and software development often find it easier to do even somewhat complex projects as templates.   However such projects can be very difficult to support and maintain\index{maintenance} in real-world usage.

\personal{I have seen a number of commercial projects where demos and prototypes were built using templates, because this was the fastest way to build these, but the production system was built using an NLG pipeline\index{pipeline}, because this was easier to maintain.}

\subsection{Techniques}\label{sec:languagefunctions}
Very simple templates (such as the one shown in Figure~\ref{fig:simpletemplate}) can be implemented using mail-merge in Microsoft Word\index{Microsoft Word} and other word processing systems.  In my experience, many domain experts like to use Microsoft Excel\index{Microsoft Excel} for templates, even though Excel is not intended or designed for this.

More complex templates are usually created with a scripting\index{scripts} language, which may be embedded into a general purpose programming language; a good example is Jinja2\index{Jinja}, which is embedded in Python.  This approach makes it easy to treat template development as a programming task, and in particular use functions within the template.   In such cases the distinction between template systems and a modularised pipelined NLG systems can become hazy.  In other words, the same processing is going on \scare{under the hood}, the issue is whether tasks such as lexical choice\index{lexical choice} are handled by explicit lexical-choice modules, or by random functions which choose verbs and do other word choice tasks.

Many sophisticated template systems include \emph{language functions}\index{language functions} which perform some limited NLG processing.   For example, we could use the following template with language functions to produce the DrivingFeedback text \lingform{You sped twice on King Street}:
\begin{verbatim}
``You sped [instancePhrase(NumInstance)] on [streetRef(Street)].''
\end{verbatim}
In this template, \code{instancePhrase(NumInstance)} is a language function that returns \lingform{once}, \lingform{twice}, \lingform{three times}, etc based on the value of \code{NumInstance}.   Another language function is \code{StreetRef(Street)}, which returns a contextually appropriate referring expression\index{referring expressions} for \code{Street}, such as \lingform{King Street} or \lingform{King Street, Aberdeen}.

Common types of language functions include:
\begin{itemize}
\item \emph{Orthography}:\index{orthography} For example, \code{EnglishList(cow,sheep,pig)} could return \lingform{cow, sheep and pig}, taking care of edge cases such as using \lingform{;} instead of \lingform{,} when necessary.
\item \emph{Morphology}:\index{morphology} For example, \code{plural(child)} could return \lingform{children}.
\item \emph{Syntax}:\index{syntax} For example, \code{countNoun(2, child)} could return \lingform{2 children}.
\item \emph{Lexical choice:}\index{lexical choice} For example, \code{timePhrase(0000)} could return \lingform{midnight}
\item \emph{Referring expressions:}\index{referring expressions} For example, \code{personReferenece(EhudReiter)} could return \lingform{Professor Ehud Reiter}.
\end{itemize}
These functions can be integrated into the template system, or accessed via generic Java or Python libraries, such as the Python \code{inflect}\index{inflect (Python package)} package\footnote{\url{https://pypi.org/project/inflect/}}.

\section{Further reading and resources}
Gatt and Krahmer (2018) \cite{gatt2018survey} survey (amongst other things) rule-based NLG as it stood in 2018.  Since the NLG community has focused on ML and neural techniques in recent years, the Gatt and Krahmer survey is still a good source for work on rules-based NLG.  My 2000 book \cite{reiterdale2000} is dated, but some people still find that it is useful for understanding the basic concepts of rule-based NLG.
 
Signal analysis\index{signal analysis} is essentially pattern detection and noise suppression\index{noise detection}, and there are numerous data science\index{data science} resources which can be used for it, such as the Python \code{numpy}\index{numpy} and \code{scipy}\index{scipy} libraries.   Data interpretation\index{data interpretation} is likewise related to data science and can draw on data science resources, although data interpretation for NLG can often be somewhat different from conventional data science.

Document planning\index{document planning} is fundamental to rule-based NLG, but I am not aware of good surveys specifically about document planning.  However the topic is covered in  Gatt and Krahmer's survey \cite{gatt2018survey} of NLG.

More resources are available for microplanning\index{microplanning}.  Krahmer and van Deemter \cite{krahmerdeemter} survey work on generating referring expressions\index{referring expressions}; this is a bit dated, but can still be a useful source for fundamentals.  Deemter has written two books related to microplanning which are aimed at non-specialist audiences.  Deemter (2010) \cite{van2010not} discusses vagueness\index{vagueness}, including how this influences lexical choice\index{lexical choice} in microplanning, and Deemter (2016) \cite{vanDeemterReference} looks at  reference.

Many open-surface realisers\index{surface realisers} are available, see Section~\ref{sec:realisationTechniques}.  The Simplenlg\index{simplenlg} package has a tutorial which explains its basics,  this can be a useful way to understand what realisers do and how to use them.  There also many template\index{templates} engines available, see Section~\ref{sec:languagefunctions}.    Unfortunately I cannot recommend any open-source document planning and microplanning libraries.  Some realisation packages (including Simplenlg) do limited amounts of microplanning, and document planning is often done using scripting languages.

There are companies which specialise in building rule-based data-to-text NLG systems, or systems which combine rules and machine learning; their websites can be useful sources to understand what is being used commercially.  At the time of writing, some of the best known are Arria\index{Arria} NLG (my company)\footnote{\url{https://www.arria.com/}} and Ax Semantics\footnote{\url{https://en.ax-semantics.com/}}\index{Ax Semantics}.

%
%
%
\chapter{Machine Learning and Neural NLG}
\label{neuralnlg} 

Instead of building an NLG system using rules from domain experts, we can use machine learning (ML) techniques to create an NLG \emph{model}.  The model is trained on data (eg, inputs and outputs for NLG), and can apply the behaviour it has learnt from the data to generate new NLG outputs from novel NLG inputs.  At the time of writing. most machine-learning in NLG uses \emph{neural} models, that is models that are loosely inspired by how neurons in the human brain work.

\begin{figure}
\lineacross\\
\includegraphics[scale  = 0.5]{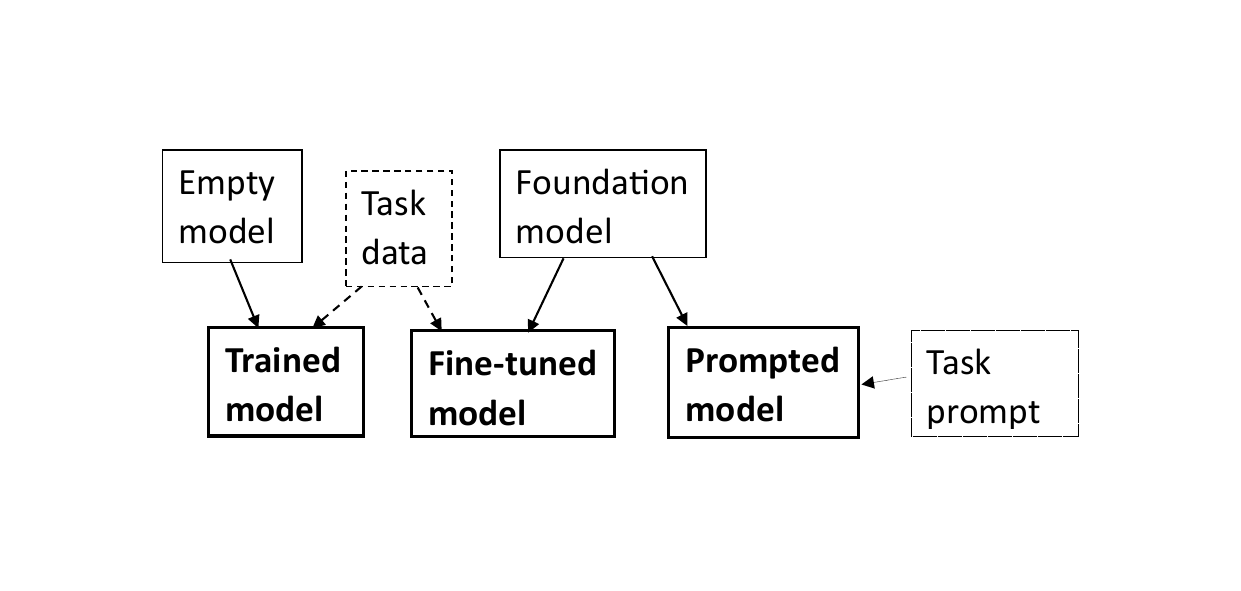}
\caption{Trained, fine-tuned, and prompted models.}
\label{fig:modelTypes}       
\lineacross
\end{figure}

There are many possibilities within the broad space of using ML and neural techniques in NLG.  Perhaps the most fundamental distinction is in how models are trained (Figure~\ref{fig:modelTypes}):
\begin{itemize}
\item \textbf{Trained models}\index{models!trained} are trained from scratch to perform the target NLG task; typically they need large amounts of task-specific training data.
\item \textbf{Fine-tuned models}\index{models!fine-tuned} take a \scare{pre-trained} large language model\index{large language models} (sometimes called a \emph{foundation model}\index{models!foundation}), which is usually trained on large amounts of Internet content, and adapt it for the target NLG task using a limited amount of task-specific training data.
\item \textbf{Prompted models}\index{models!prompted} directly use a pre-trained language model to perform a task, by giving the model a request (\emph{prompt}); task-specific examples can be included, but this is not necessary.
\end{itemize}

Other important distinctions include whether the model does the complete NLG process or only part of it (for example generating texts from insights, with the insights produced by a separate component); and whether the model is used autonomously or whether a person manually checks its outputs (\emph{human-in-the-loop}\index{human-in-loop}; this is discussed in Section~\ref{sec:humanchecking}).

Machine learning and neural technology for NLG has been changing very quickly over the past 10 years.  N-gram models (Section~\ref{sec:ngram}) dominated until 2016 or so, early neural models such as LSTM (Section~\ref{sec:earlyneural}) then became the focus of attention until 2019 or so, transformers (Section~\ref{sec:transformers}) then became the most popular approach until 2022 or so, and instruction-tuned models are pre-eminent at the time of writing.  I would not be at all surprised if a new approach became prominent in 2025.

This book presents neural and ML technology at a very high level, focusing on fundamental concepts and issues which I believe will still be important in 2030.  It deliberately do not attempt to give a detailed description of ML/neural technologies at the time of writing, since anything written along these lines in 2024 would probably be obsolete by 2025, and of only historical interest by 2030.

\section{Examples}

\subsection{Very simple trained model\index{models!trained}: \lingform{a} vs \lingform{an}}\label{sec:aan}

One of the simplest ML models in NLG chooses whether (in English) to use \lingform{a} or \lingform{an}.  This difficult to do using rules, but easy to do using machine learning; this model also illustrates some of the fundamental concepts of ML in NLG.

In English, we use \lingform{a} before words that start with a consonant sound, but \lingform{an} before words that start with a vowel sound.  Since this choice is based on how a word is pronounced, we cannot simply look at the first letter of the following word to make this choice; for example correct usage is \lingform{an umbrella} but \lingform{a university}.

Of course, there are many special cases, including the following:
\begin{itemize}
\item \emph{Acronyms:}\index{acronyms} Pronunciation of acronyms depends on whether it is spelled out letter-by-letter; for example \lingform{an SAT prep course} if we spell out the letters \lingform{S-A-T}, but \lingform{a SAT prep course} if we pronounce \lingform{SAT} like the word \lingform{sat}.
\item \emph{Currency:} When we pronounce a currency, we say the number first even if the unit comes first in the written form.  Thus we would write \lingform{an \$80 fine} because \lingform{\$80} is pronounced as \lingform{eighty dollars}, not \lingform{dollar eighty}, even though \lingform{\$} comes before \lingform{80} in the written form \lingform{\$80} .
\end{itemize}

Writing rules to choose between \lingform{a} or \lingform{an} is complex and difficult.  A better approach is to analyse a large collection (\emph{corpus})\index{corpus} of well written English texts, such as Wikipedia\index{Wikipedia} to see whether \lingform{a} or \lingform{an} are preferred in front of specific words.   For example, \lingform{an umbrella} occurs 9500 times in Wikipedia, while \lingform{a umbrella} only occurs 45 times; this tell us that \lingform{an umbrella} is the preferred usage.  We can repeat this process for all words in Wikipedia, and create a database which tells us when to use \lingform{an} and when to use \lingform{a}; this is  accurate and does not require any rules!

This is technically called a \emph{bigram} model\index{models!bigram}, which means it is based on the frequency and probability of words pairs such as \lingform{an umbrella},. Bigram models are a type of \emph{n-gram} model\index{models!ngram} (Section~\ref{sec:ngram}).  This is also a trained model in the sense of Figure~\ref{fig:modelTypes}.

We can also \emph{generalise}\index{generalisation} our findings.  We know that the first few letters of a word are the most important in choosing between \lingform{a} and \lingform{an}, so we can look for cases where all words that start with the same letters have the same  \lingform{a} vs \lingform{an} behaviour.  For example, if all words starting with \lingform{d} use \lingform{a}, then we can replace the D-section of our database by a simple rule that any word starting with \lingform{d} takes \lingform{a} (this analysis can be automated).  Not only does this reduce the size of our model, it also lets us handle new words which were not in the corpus.  For example, \lingform{dzild} is not in Wikipedia, but the generalisation process tells us to use \lingform{a dzild} instead of \lingform{an dzild}.

Of course, the above process only works if the corpus or \lingform{training data}\index{training data} is well-written English.  Wikipedia satisfies this criteria, but Twitter does not.  When machine learning models fail, it is often because of quality issues in the training data, and  commercial ML models builders often spend more time on data issues (Section~\ref{sec:trainingdata}) than on algorithms.

\begin{figure}
\lineacross\\
\includegraphics[scale  = 1]{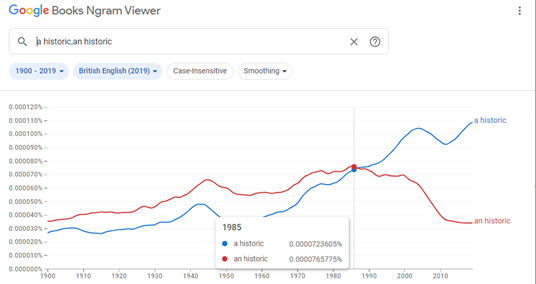}
\caption{Usage of \lingform{a historic} vs \lingform{an historic} in British English, over time.  From Google Books NGram\index{Google Books} Viewer, \url{https://books.google.com/ngrams}}
\label{fig:ahistoric}       
\lineacross
\end{figure}

Lastly, language changes\index{language change}, it is not static.  For example. Figure~\ref{fig:ahistoric} show the usage of \lingform{a historic} vs \lingform{an historic} over time, in British English. This shows that while \lingform{a historic} is more common now, \lingform{an historic} was more common before 1985.  This means that models should be trained on recent data; they may make mistakes (such as predict \lingform{an historic}) if it is trained on old data.  More generally, we need to keep in mind that models trained on a corpus become less useful as the corpus ages; this is related to \lingform{domain shift} in machine learning (Section~\ref{sec:domainshift}).

\subsection{Fine-tuned neural model: Facebook weather dialogues}\label{sec:Facebookweather}
An example of a more complex ML model is the NLG system developed by Facebook (Meta) to respond to weather inquiries\index{weather forecasts} \cite{arun-etal-2020-best}, which was deployed and operationally used.  This system relies on another component to select the content to be communicated in the text; it automates the task of generating a text from a meaning (content) representation.  It uses neural technology, and in particular relies on fine-tuning\index{fine-tuning} (Figure~\ref{fig:modelTypes})  a large language model (BART\index{BART} \cite{lewis-etal-2020-bart}) to the task of generating weather information.

Figure~\ref{fig:facebookweather} shows an example from this system.  The Query is entered by the user.  Other parts of the system do signal analysis\index{signal analysis}, data interpretation\index{data interpretation}, and some document planning\index{document planning}, and produce a \emph{meaning representation}\index{meaning representation} which specifies the content that should be communicated to the user.  The NLG system then generates an actual text (similar to the reference text) which communicates the meaning to the user.

The meaning representation specifies the information to be communicated, for example that the low temperature will be 20 (\semantic{temp\_low[20]}, and that it will rain on Sunday (\semantic{condition[ rain ] date\_time[weekday[ Sunday ]]}).  It also specifies connectives which relate insights (what I called \emph{linkage}\index{data interpretation!linkage} in Section~\ref{sec:dp-principles}), for example that the rain on Sunday contrasts with the sunshine on Saturday.

\begin{figure}
\lineacross\\
\emph{Query}: How is the weather over the next weekend?\\

\emph{Response content (meaning representation produced by other components)}:\\
\begin{verbatim}
INFORM 1[temp_low[20] temp_high[45] date_time[colloquial[ next weekend ]]]
CONTRAST 1[
              INFORM 2[condition[ sun ] date_time[weekday[ Saturday ]]]
              INFORM 3[condition[ rain ] date_time[weekday[ Sunday ]]]
               ]
\end{verbatim}

\emph{Reference (human-written)}: \lingform{Next weekend expect a low of 20 and a high of 45. It will be sunny on Saturday but it’ll rain on Sunday.}
\caption{Example input and reference (human-written) output for weather system, from \cite{arun-etal-2020-best}}
\label{fig:facebookweather}
\lineacross
\end{figure}

In order to fine-tine BART, the Facebook team needed high-quality examples of NLG inputs and outputs, that is of meaning representations and corresponding generated texts; this was the training data\index{training data}.  The Facebook team used to following process to create these examples:
\begin{itemize}
\item asked engineers to create a varied set of queries and scenarios.
\item automatically generated meaning representations for these \cite{balakrishnan-etal-2019-constrained}.
\item asked human annotators to manually produce high-quality responses from the meaning representation, following guidelines written by computational linguists.
\item had linguists\index{linguists} check and verify the responses.
\end{itemize}
This process was used to create 25,000 query-response pairs for training models, and a further 6,000 for testing and validation.

Note that the training data for this system was explicitly created for the project, it was not scraped off the internet.  No cost figures are given in the paper, but creating 31,000 query-response pairs using the above process must have been a major task.  The team subsequently developed techniques to reduce the amount of data needed \cite{heidari-etal-2021-getting}.

At any rate, once the team had this dataset, they then used it to fine-tune a BART model \cite{lewis-etal-2020-bart}.  In other words, rather than creating a new model from scratch, they took an existing large language model and tuned it to work on the target NLG task (Section~\ref{sec:transformers}).   The team also used knowledge distillation\index{knowledge distillation} techniques (essentially training a smaller model on the fine-tuned BART model) to speed up run-time processing and reduce run-time memory requirements, which is very important in a production context.

Finally, because of the dangers of the neural model making mistakes, the team built a tool which identified generated texts which were possibly incorrect.  They also implemented a backup rule-based NLG\index{rule-based NLG} system which was invoked if the checking tool found potential problems in a text.

A key lesson from this example is that building a production NLG system based on a fine-tuned neural model required creating high-quality data sets for training \cite{dusek-etal-2019-semantic} (Section~\ref{sec:trainingdata}), ensuring that models are acceptable from a time/memory/cost perspective, and implementing techniques to deal with model errors.  The actual model training and tuning process (which is what most students and researchers focus on) is important, but  its only part of the bigger picture if we are building real-world NLG systems.

\subsection{Prompted model\index{models!prompted}: Using ChatGPT\index{ChatGPT} to generate weather forecasts}\label{sec:exampleprompted}
If we want to build a system to generate weather forecasts using a trained or fine-tuned model, we need to collect a data set of thousands of high-quality input-output pairs. as was done for the Facebook system described above.  An alternative approach is to take a large language model such as ChatGPT, which is trained on a huge amount of Internet material, and give the model a prompt\index{prompt} (request) such as \lingform{Generate a short weather forecast for the below weather data}, followed by the numeric weather data (Figure~\ref{fig:chatgpt-example}).

\begin{figure}
\lineacross\\
\emph{Input to prompted model (ChatGPT):}\\
\begin{verbatim}
Generate a short weather forecast from the below data.
Weather data:
[
 {
   "time": 0,
   "wind": 10,
   "precipitation": 0,
   "temperature": 6
 },
 {
   "time": 3,
   "wind": 11,
   "precipitation": 0,
   "temperature": 6
 },
...
]
\end{verbatim}

\emph{Output of prompted model:}\\
\lingform{Here's a short weather forecast based on the provided data:}\\

\lingform{Expect a day with changing wind speeds and temperatures. The wind will range from 9 km/h to 12 km/h throughout the day, occasionally gusting higher. The temperature will start at 6°C and gradually increase to 15°C in the afternoon before dropping to 10°C in the evening.}\\

\lingform{There is no precipitation expected, so it will be a dry day.}\\

\lingform{In summary, anticipate a mix of wind speeds, with temperatures ranging from 6°C to 15°C, and no rainfall.}\\

\caption{Example of using a prompted model (ChatGPT) to generate a weather forecast.  The model is generic and has \emph{not} been trained or fine-tuned to produce weather forecasts}
\label{fig:chatgpt-example}
\lineacross
\end{figure}
If we want to control the wording and content of the weather forecast, we can explicitly request this in the prompt, or we can include a small number of  examples (5 or 10, not 25,000)  in the prompt (Section~\ref{sec:fewshot}).

Prompted language models have many advantages over trained or fine-tuned models, including:
\begin{itemize}
\item No need to create a large set of domain-specific input-output examples.
\item Since the model is general-purpose, its easy to add additional capabilities, such as generating texts in different languages.
\item From a developer perspective,  creating prompts does not require specialist expertise in machine learning, NLP, or programming (although these can help); fine-tuning or training neural models, in contrast, does require considerable expertise.
\end{itemize}
Many prompted models can be fine-tuned to improve their performance.

\section{Machine Learning Models for NLG}

Machine learning is a rapidly evolving field, below I summarise a few of the different machine learning models and technologies which have been used to build NLG systems.  This description is high level; Further Reading (Section~\ref{sec:mlreading}) suggests additional sources for technologies.

\subsection{Classifiers\index{classifiers}}\label{sec:classifiers}
A \emph{classifier} puts an input dataset into one of a finite number of \emph{classes}.  For example, a sentiment analysis classifier classifies an input text into categories such as Positive, Neutral, or Negative.  Classifiers are one of the oldest areas of machine learning, and indeed many classifier algorithms draw heavily on statistical research as well as AI.  The \texttt{scikit-learn}\index{scikit-learn} Python library\footnote{\url{https://scikit-learn.org/}} includes a number of algorithms for building classifiers, including decision trees, Naive Bayes, K nearest neighbour, support vector, and many others.

\begin{figure}
\lineacross\\
\includegraphics[scale  = 0.5]{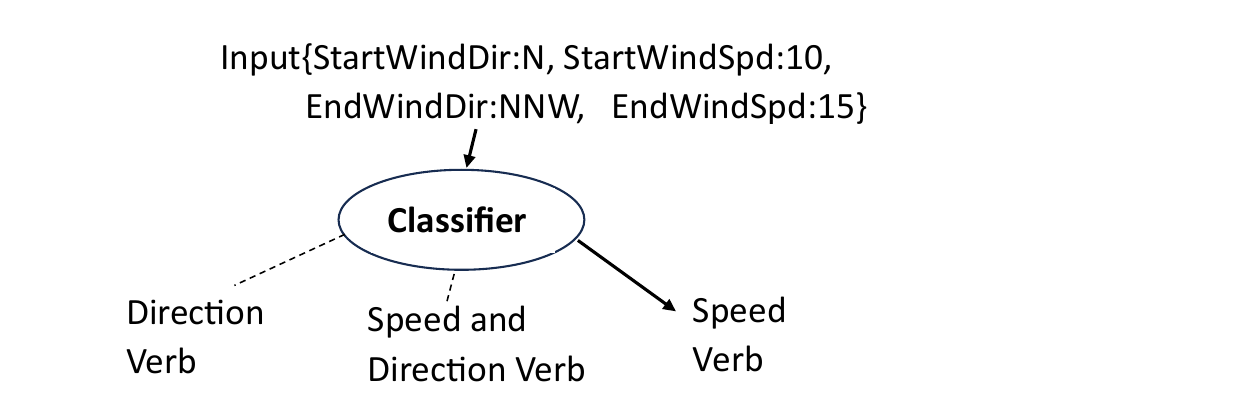}
\caption{Classifier decides to use Speed Verb in a context where there is a large change in speed and a small change in direction.}
\label{fig:classifierExample}      
\lineacross 
\end{figure}

To take a simple example,the SumTime\index{SumTime} weather forecast\index{weather forecasts} generator \cite{REITER2005137} needed to decide which verb to use when describing changes in the weather; this is an example of lexical choice\index{lexical choice} (Sect~\ref{sec:lexchoice}; Section~\ref{sec:lexprinciples} discusses SumTime).   In particular, in statements describing the wind, there is a choice between three types of verbs
\begin{itemize}
\item Verb emphasising direction change, such as \lingform{W 10-14 veering N 12-16}
\item Verb emphasising speed change, such as \lingform{W 10-14 increasing WNW 18-22}
\item Conjoined verb emphasising both speed and direction change, such as \lingform{W 10-14 veering and increasing N 18-22}
\end{itemize}
SumTime used a classifier to decide on verb type (Figure~\ref{fig:classifierExample}).  Essentially, developers extracted all wind change phrases from a corpus\index{corpus} of human-written forecasts and corresponding numerical weather data sets, and used this to train a classifier which predicted the above choice from \semantic{<StartWindDirection, StartWindLowSpeed, EndWindDirection, EndWindLowSpeed>} tuples.  SumTime used a decision tree, but other classifier techniques could have worked just as well \cite{chen-yao-2019-closer}.

Classifiers cannot generate texts on their own (i.e., they do not do \emph{end-to-end} generation), but they are useful in making specific NLG choices, including content (insights) as well as language choices.

\subsection{N-gram language models\index{models!ngram}}\label{sec:ngram}
An n-gram language model is a model which gives information about the likelihood of a sequence of words.  The \lingform{a}-vs-\lingform{an} algorithm (Section~\ref{sec:aan}) is a simple example of this.  Essentially given a word XXX, the model decides whether to use \lingform{a XXX} or \lingform{an XXX} by counting how often the \emph{bigrams}\index{models!bigram}  \lingform{a XXX} and \lingform{an XXX} occur in a corpus (collection of texts) such as Wikipedia\index{Wikipedia}.   A bigram is a two-word sequence (2-gram); n-gram models support word sequences of other lengths including \emph{unigrams} (1 word) and \emph{trigrams} (3 words).  Usually n-grams are structured to produce probabilities instead of raw frequencies, as can be seen in Figure~\ref{fig:ahistoric}.

We can get good data on the frequency of bigrams and other small n-grams from corpora, and use these to estimate probabilities.  However, it is much harder to do this with large word sequences.  For example, suppose a speech-to-text system is trying to interpret an input sentence, and from an acoustic perspective is unsure whether the sentence was \lingform{I eat too much strawberry ice cream} or \lingform{I eat too much strawberry I scream} (since \lingform{ice cream} and \lingform{I scream} are very similar from an acoustic perspective).  In theory it could decide between these based on frequency in a corpus.   Unfortunately, the Internet (according to a Google search) contains \emph{no} examples of either \lingform{I eat too much strawberry ice cream} or \lingform{I eat too much strawberry I scream}, so raw frequencies tell us nothing.  However, we can use  techniques such as Markov models\index{models!Markov} and smoothing to estimate the probability of these word sequences from smaller n-grams such as \lingform{strawberry ice cream} and \lingform{strawberry I scream} \cite{jurafskyspeech}; this tells us that \lingform{I eat too much strawberry ice cream} is a much more likely sentence than \lingform{I eat too much strawberry I scream}.

From a software perspective, \emph{nltk.lm}\index{nltk} (part of the Natural Language Technologies toolkit\footnote{\url{https://www.nltk.org/}}) provides good support for creating and using n-gram models.

N-gram models were popular for many years, but at the time of writing have largely been replace by neural models.

\subsection{Early neural models}\label{sec:earlyneural}
In recent years there has been huge interest in using \emph{neural} (especially \emph{deep learning}\index{deep learning} \cite{Goodfellow-et-al-2016}) models to generate texts and indeed do other natural language processing tasks.  These models are loosely inspired by neurons in the human brain.  An enormous number of neural models and architectures have been discussed in the literature.  

Neural NLG is usually considered to be a \emph{sequence-to-sequence}\index{sequence-to-sequence} task, where an input sequence (input texts for summarisation, time series for data-to-text) is converted into an output sequence of words (ie, a text).  Early neural models for NLG \cite{goldberg2022neural} used \emph{recurrent} neural networks\index{recurrent neural networks} (RNNs), which are a type of neural network which iterates through arbitrary-length sequences of tokens (words).   \emph{Long short-term memory} (\emph{LSTM}\index{LSTM}) architectures modify the core RNN to give the network \scare{memory} which makes it easier for the network to consider already-output words when it decides on the next word to output.  \emph{Encoder-decoder}\index{encoder-decoder} architectures use two neural networks to generate texts, an \emph{encoder} which maps the system's input into an internal state, and a \emph{decoder} which translates the internal state into generated texts.

These models were usually trained from data; that is, initially the model's neural network are in a default state, and training algorithms such as \emph{back propagation}\index{back-propagation} are used to set the \emph{parameters}\index{parameters} (internal weights) of the neural network.  Of course there are many variants of the above, and indeed many different possible neural architectures (number of layers, number of nodes in layers, etc).

\subsection{Transformers\index{transformers} and foundation models\index{models!foundation}}\label{sec:transformers}\label{sec:foundationmodels}
In 2017, Vaswani et al \cite{Vaswani:transformer} introduced the \emph{Transformer} architecture.  In very crude terms, while earlier neural models sequentially generated an output token from an input token (while using a memory mechanism to keep key information from earlier tokens), transformers could process larger chunks of information (such as complete sentences).  They did this using an \scare{attention}\index{attention} mechanism which highlighted key connections between inputs to the model.  For example, when translating \lingform{I ate a grape} into (French) \lingform{j'ai mangé un raisin}, the attention mechanism tells the model that the word \lingform{grape} has an impact on how the word \lingform{a} is translated.  Since \lingform{raisin} is a masculine noun in French, \lingform{a} is translated as \lingform{un}.  In contrast, \lingform{une} would be used with a feminine noun such as \lingform{banane}, eg \lingform{j'ai mangé une banane}.

The transformer architecture worked very well in the context of fine-tuning large pre-trained language models, sometimes called \emph{foundation} models.  In other words, a large transformer model could be trained on a large amount of generic Internet content.  This model could then be efficiently \emph{fine-tuned}\index{fine-tuning} (adapted) to a specific domain and task with a much smaller domain/task corpus; this was done by updating the model's parameters\index{parameters} (links between nodes in the neural network) to better fit the domain/task corpus.

This has a huge impact, because it meant that developers could build powerful task-specific models with much less training data\index{training data}.  This approach was used in the Facebook weather system described in Section~\ref{sec:Facebookweather} and also for the Note Generator\index{Note Generator} system described in the Introduction chapter (Section~\ref{IntroConsultationSummary}).  In both cases, it would not have been possible to acquire sufficient training data to allow high-performance models to be trained from scratch, but  it was possible to get enough data to create high-performance models by fine-tuning a pre-trained model.

The transformer architecture also proved well suited to efficient execution on AI hardware, and could be scaled up to very large language models\index{large language models} (\emph{LLMs}) with billions of parameters (links between nodes in the neural network) which were trained on tens or hundreds of billions of words.  At the time of writing it is the dominant architecture for neural NLP and NLG systems, and used in models such as BART\index{BART} \cite{lewis-etal-2020-bart},  T5\index{T5} \cite{raffel2020exploring},  BLOOM\index{BLOOM} \cite{scao2022bloom}, GPT\index{GPT} \cite{brown2020language}, and PaLM\index{PaLM} \cite{chowdhery2022palm}.  Huggingface\footnote{\url{https://huggingface.co/}}\index{Huggingface} provides a comprehensive library and collection of open-source transformers \cite{tunstall2022natural} (and other neural models as well), supported by high-quality documentation and training material.

\subsection{Instruction Tuning\index{instruction tuning} and RLHF\index{RLHF}}\label{sec:promptedmodels}
Researchers discovered that large language models such as GPT3 could perform many tasks without being adapted or fine-tuned, simply be giving them a \emph{prompt}\index{prompt} which made a request, presented the input data, and (optionally) included a few examples (Figure~\ref{fig:chatgpt-example}) \cite{brown2020language}.  This was very exciting, since constructing data sets even for fine-tuning models could be a lot of work.

However, models such as GPT3 and BLOOM were very sensitive to the exact wording of the prompt.
Using these models required a very good understanding of \emph{prompt engineering}\index{prompt!engineering}, ie of creating prompts which framed the desired task and use case in a way which resonated with the model and the data it was trained on.

Fig~\ref{fig:bloom-instruction} shows an example using an early version of BLOOM\index{BLOOM}.  When we craft the prompt correctly, BLOOM gives us the translation (\lingform{rouge}) which we requested.  However, if we use a different prompt which is intuitive but does not match corpus usage, then we get an explanation that it is impossible to have a single unique translation which works in all contexts.  Which is true, but for most people the first response (\lingform{rouge}) is probably more useful.

One technique to make the models more robust and accept a wider variety of prompts is \emph{instruction tuning} \cite{weifinetuned}.  Essentially this process fine-tunes\index{fine-tuning} the model using a data set of common and intuitive instructions along with expected responses.   The fine-tuned model learns how to respond appropriately to intuitive prompts; for example to response to a request for a translation with the best possible translation, instead of explaining the difficulties of providing a translation.  

Another technique for enabling models to respond appropriately to prompts is \emph{Reinforcement Learning from Human Feedback} (RLHF).  The core idea of RLHF is to collect a set of representative prompts, get the model to generate responses to these prompts, show the prompt-response pairs to humans, and ask the humans to assess whether the response was appropriate for the prompt.  This feedback data is then used to update the model using reinforcement learning techniques \cite{ouyang2022training}.

Instruction tuning and RLHF are essentially forms of \emph{alignment}\index{alignment}, that is techniques which helps models respond to human prompts in useful and appropriate manners which aligns with user expectations.

For instance, ChatGPT\index{ChatGPT} (which uses both instruction-tuned and RLHF) responds to the second prompt in Fig~\ref{fig:bloom-instruction} with \lingform{The translation of the English word "red" to French is "rouge."}; this is probably what most users are expecting.

Indeed, instruction tuning and RLHF in ChatGPT and other large language models worked so well that ordinary members of the public, who did not have a background in machine learning or natural language processing (or prompt engineering), were able to create prompts and use these models to generate texts and documents, answer questions, etc.  Such models still gave better results with carefully engineered prompts (and indeed if they are fine-tuned for a specific task), but instruction tuning and RLHF meant that models could provide acceptable and useful responses \scare{out-of-the-box} in many cases from straightforward prompts.
This had a huge impact, and when OpenAI launched a free version of ChatGPT, it reached 100 million active users two months after it was launched; nothing like this had been seen in AI before.

\begin{figure}
\lineacross\\
Prompt:
\begin{verbatim}
Translate English to French.
English: red
French:
\end{verbatim}
Response from BLOOM:
\begin{quotation}
\lingform{rouge}
\end{quotation}
Prompt:
\begin{verbatim}
Translate English word "red" to French.
\end{verbatim}
Response from BLOOM:
\begin{quotation}
\lingform{We, thus, conclude that there is no unique translation from English to French
which preserves the various usages of the concept red, both in the sense of
the noun color and in the sense of the adjective or adverb meaning "with red color".}
\end{quotation}
\caption{Effect of changing the prompt in an early version of BLOOM, which does not use instruction-tuning or RHLF}
\label{fig:bloom-instruction}
\lineacross
\end{figure}

\subsection{End-to-end vs. modular architectures}\label{sec:modularML}
A final point is that neural and ML technology can be used in many different ways in NLG:
\begin{itemize}
\item \emph{Component:}  If an NLG system is organised as a set of components or modules (perhaps following the pipeline architecture presented in Section~\ref{sec:pipeline}), then ML models can be used within specific components in order to perform specific tasks.
One example is choosing \lingform{a} or \lingform{an} (Section~\ref{sec:aan}); this is part of surface realisation (Section~\ref{sec:realisation}).  There are many other examples, include detecting patterns in data, identifying the most important insights, and deciding whether to use pronouns\index{pronouns}.  Castro Ferreira et al \cite{castro-ferreira-etal-2019-neural} show how an entire NLG pipeline\index{pipeline} can be constructed from neural components.
\item \emph{Linguistic processing:} The Facebook weather system (Section~\ref{sec:Facebookweather}) uses a neural model to do linguistic processing (convert insights into words), but uses other techniques to do data-side processing (choose insights to communicate).  I have seen a number of systems that use this strategy, perhaps because \emph{language} models\index{language models} by their nature are best suited to doing language processing.
\item \emph{Retrieval-augmented generation\index{retrieval-augmented generation} (RAG) \cite{lewisrag}:}  Another approach is to first use web search to find relevant information which may be useful to the user, and then give the search results, along with input data or prompt, to a neural language model.  This reduces the amount of content determination which the model needs to do, and also gives it access to up-to-date information from the web.
\item \emph{End-to-end} An end-to-end system uses an ML model to do the complete generation process; ie it takes in the system input data and produces output texts.   This approach is popular with text summarisation\index{summarisation} systems, and also with simpler data-to-text tasks such as using ChatGPT to produce weather forecasts\index{weather forecasts} (Section~\ref{fig:chatgpt-example}).
\end{itemize}

\section{Training data}\label{sec:trainingdata}\index{training data}

Building machine learning models requires training data, preferably large amounts of high-quality data, so acquiring good data sets for training is very important. Indeed, in many ML projects, more effort goes into data acquisition than anything else.  Especially since building models\index{models} (once we have acquired training data) is usually straightforward because of libraries such as scikit-learn\index{scikit-learn}, Huggingface\index{Huggingface} Transformers, Keras\index{Keras}, etc.

This section looks at some issues around training data. The focus is on training smaller models, or fine-tuning large language models.   Creating large models (such as GPT or PaLM) requires much more training data (billions or even trillions of words) for the core model, plus additional data for instruction tuning\index{instruction tuning} and RLHF\index{RLHF}.  However, building such models is a specialist and expensive exercise which is only done by a handful of companies and organisations.

\personal{As an academic, I have seen many student projects which use ML for NLP.  Unfortunately I have often seen students grab an impressive-sounding data set from Kaggle\index{Kaggle} or a shared task repository, spend a lot of time building ML models, and then realise that the data is flawed, so their model is garbage (\scare{garbage in, garbage out}).  I've also seen a lot of published academic papers which suffer from the same problem. Don't make this mistake - carefully investigate the data \emph{before} you spend a lot of time on model building.}

\subsection{Data sources}
Sometimes developers building an NLG system can use an existing data set, but real-world applications often require new data sets.  These can be acquired in several ways, including
\begin{itemize}
\item \emph{Real-world data and tests:}  The ideal data source is a pre-existing dataset which contains input data and high-quality human-written output texts.  For example, the Note Generator\index{Note Generator} (Section~\ref{IntroConsultationSummary}) system, which summarised doctor-patient consultations, was built by fine-tuning the BART model on 10,000 consultations which had been manually summarised by doctors\index{doctors} and entered into the patient's electronic health record.   Similarly the word-choice models for weather forecasts\index{weather forecasts} described in Section~\ref{sec:classifiers} were trained on a corpus\index{corpus} of 1,045 forecasts written by human forecasters.
\item \emph{Crowdworkers:}\index{crowdworkers} If developers cannot find real-world data and/or texts, a common practice is to get workers on \emph{crowdsourcing} sites such as Prolific\index{Prolific}\footnote{\url{https://www.prolific.com/}} and Mechanical Turk\index{Mechanical Turk}\footnote{\url{https://www.mturk.com/}} to write texts from real or synthetic data.  For example, fifty thousand restaurant descriptions were created by crowdworkers for the E2E corpus \cite{DUSEK2020123}.  If crowdworkers are used, developers must design the process carefully in order to maximise quality; see discussion of using crowdworkers for evaluation in Section~\ref{sec:humansubjects}, where similar concerns arise.
\item \emph{Domain experts:}\index{domain!experts}  It may not be possible to use crowdworkers (who essentially are members of the general public) in specialised domains.  In such cases, domain experts can be asked to write texts from data. For example, the PriMock57 dataset \cite{korfiatis2022primock57} contains summaries written by clinicians of 57 mocked doctor-patient consultations.  Of course asking domain experts to write texts requires a lot more time and money than asking crowdworkers to do this; this is one reason why PriMock57 is so much smaller than the above-mentioned datasets.
\end{itemize}
Each of these techniques has its own challenges.  For example, real-world human-written texts are sometimes not as high-quality as we would hope; Thomson et al \cite{thomson-csl23} found many factual errors in journalist-written stories about basketball games\index{journalism}.   Crowdworkers are paid per task, so they make more money if they do tasks as quickly as possible, which means they may not be very careful when they write texts (Section~\ref{sec:humansubjects}).  And domain experts are usually very knowledgeable about content, but sometimes do not write very well.

\subsection{Dataset criteria}\label{sec:datasetcriteria}
 In an ideal world. the training data set is (A) large, (B) high-quality, and (C) representative.  In NLG, we also usually want data sets to include system inputs as well as system outputs.  Unfortunately, in the real world most training data sets do not meet this criteria.  Poor quality training data will result in poor-quality NLG systems, regardless of the sophistication of the ML technology used (again, \scare{Garbage In, Garbage Out}). This is one reason why most commercial system builders spend more time worrying about data than about algorithms or model types.

\textbf{Size:} Training a good neural model from scratch requires a lot of data.  The amount depends on the complexity of the NLG task and the size of the generated texts, but even a simple task like the E2E\index{E2E} challenge \cite{DUSEK2020123} (generating 10-word restaurant descriptions from 8 features of the restaurant) required a training data set of 50K (feature-set, description) pairs.
However, less training data is needed to fine-tune\index{models!fine-tuned} a pre-trained model.  For example, about 25K of (meaning, output) pairs were required to fine-tune BART for the Facebook weather system (Section~\ref{sec:Facebookweather}); these texts were substantially more complex than E2E, but data requirements were reduced because a pre-trained model was used.  Indeed a fine-tuned model for this task  could have been created with a few hundred elements \cite{heidari-etal-2021-getting} although output quality would have been lower.

\textbf{Quality:} Training data also need to be high quality.  If developers simply scrape data off the Internet or ask random crowdworkers\index{crowdworkers} to generate data, then it is likely that the data sets will have quality issues (readers should also be very careful with data sets on data science repositories such as Kaggle\index{Kaggle}, some are fine but many have quality problems).  One approach is to create a bespoke workflow for creating high-quality data, such as the one described in Section~\ref{sec:Facebookweather}. Another approach is to try to automatically clean datasets (ie, identify problems and then either fix them or delete suspect data) \cite{dusek-etal-2019-semantic}.  Regardless of the approach taken, it is important to assess and evaluate data quality issues, and understand where there are problems.

\textbf{Representative:}\index{representativeness}  Last but not least, training data needs to be representative of  real usage.  Sometimes researchers use easy-to-obtain data which is not representative; for example training dialogue systems on datasets which contain dialogues from TV shows, which are easy to get but completely unrepresentative of real-world dialogue.  In medical\index{health} NLP, a lot of research has used the MIMIC data set \cite{johnson2016mimic}, which is a great resource but comes from a single hospital in Boston, and is not representative of hospitals in general.   Building representative data sets is hard, but it makes a difference in real-world performance.  At minimum developers need to understand and evaluate how representative their data is, and identify contexts where the data set does not have good coverage of real usage.

\subsection{Impact of training data on prompted models}
Prompted models\index{models!prompted} do not need application-specific training data (except perhaps for a handful of examples\index{examples}, Section~\ref{sec:fewshot}), but they are impacted by data quality issues in the data which was used to train the underlying language model\index{large language models}.  Generally speaking models used for prompting are primarily trained on Internet data (such as that gathered by Common Crawl\footnote{\url{https://commoncrawl.org/}}\index{Common Crawl}), since this is the only data source which is large enough to provide sufficient training data.  However, Internet data has many problems, which possibly may impact the quality of text produced by the prompted model.  These include:

\begin{itemize}
\item \emph{Inappropriate content:} The Internet contains material which we do not want to see included in system outputs, such as pornography, racist and sexist rants, stereotypes, encouragement to commit crimes, etc.  This is an aspect of \emph{safety}\index{safety}, which is discussed in Section~\ref{sec:safety}. One approach to this problem is to try to exclude such material either from the training data or from prompt responses; however, this is not straightforward \cite{welbl-etal-2021-challenges-detoxifying}.  
\item \emph{Spam\index{spam} and malicious content:}  As prompted models such as ChatGPT\index{ChatGPT} become widely used in society, it seems likely that spammers and other \scare{adversarial} agents will try to create Internet content which, when used to train the model\index{models}, will guide the model towards producing a response that advances the spammer's agenda.  Spam of course is already a huge problem with Internet search, but the black-box nature of neural networks may make it harder to detect spam in neural models.
\item \emph{Biased content:}\index{bias} Most Internet material comes from well-off individuals in rich countries, which introduces biases.  Very little, for example, comes from homeless people, which means that if a model is generating texts about homeless people, it will primarily be driven by Internet content created by well-off people commenting on homelessness, not content created by homeless people themselves.
\item \emph{Low-quality content:} A related issue is that a lot of material on the Internet is not of high-quality.  For example, Balloccu et al \cite{balloccu2024ask} looked at using ChatGPT to give health\index{health} advice, and discovered that a lot of the advice was inappropriate, in part because it seemed to largely be based on comments on discussion forums (such as Reddit\index{Reddit}) instead of high-quality advice from health professionals \cite{balloccuphd} (see also \cite{abercrombie-rieser-2022-risk}).  This is probably because there is  more data in discussion forums than in high-quality curated health websites.  
\end{itemize}
Most of the above problems are worse when the model is asked to answer questions (Section~\ref{sec:llm-content}), which is how many people use ChatGPT.  However, even in use cases where the task is summarising or explaining input data (which is what this book focuses on), there is still potential for the above problems to lead to toxic content\index{toxic content}, spam, bias, etc. (Section~\ref{sec:inappropriatelanguage}).

Data issues also effect testing and evaluation data, as well as training data; this is discussed in Section~\ref{sec:trainingtestdata}.

\subsection{Synthetic data\index{synthetic data} and data augmentation\index{data augmentation}}\label{sec:syntheticdata}
If a sufficiently large data set cannot be constructed from real data, developers may consider \emph{augmenting} the data set by adding \emph{synthetic data}  to it \cite{feng-etal-2021-survey}.  To take a simple example, assume a developer is training a model to generate weather forecasts and has 500 genuine human-written forecasts\index{weather forecasts} but needs 1000 forecasts in order to train or fine-tune an NLG forecaster.   The developer could use a machine translation tool to translate each forecast into German, and then \emph{back-translate}\index{back-translate} the German version of each story into a new English version.  The back-translation process gives a \emph{paraphrases}\index{paraphrase} of the original story, which can added to the training data set (Figure~\ref{fig:backtranslate}).  Of course the process can be repeated using other languages (French, Chinese, Finnish, etc.) if we want even more versions.   There are many other techniques for data augmentation \cite{feng-etal-2021-survey}.

\begin{figure}
\lineacross\\
\emph{Original English text (from Figure~\ref{fig:facebookweather}):}\\
Next weekend expect a low of 20 and a high of 45. It will be sunny on Saturday but it’ll rain on Sunday.\\

\emph{Google Translate's German translation of above English text:}\\
Am kommenden Wochenende wird ein Tiefstwert von 20 und ein Höchstwert von 45 erwartet. Am Samstag wird es sonnig sein, aber am Sonntag wird es regnen.\\

\emph{Google Translate's English translation of above German text (this is the back-translation):}\\
A low of 20 and a high of 45 are expected next weekend. On Saturday it will be sunny, but on Sunday it will rain.

\caption{Example of using back-translation to create a paraphrased version of a human text.  This can be added to a data set if more data elements are needed and it is acceptable to use synthetic data.}\label{fig:backtranslate}
\lineacross
\end{figure}

Data augmentation can be very useful in many cases and indeed is widely used.  To take one random example, the BLEURT metric for assessing the quality of generated texts (described in Section~\ref{sec:bleurt}) uses synthetic data (in part obtained by back-translation, as described above) to tune the model used to assess text quality; this enhances the robustness and accuracy of BLEURT.

Data augmentation is not always helpful, however.  Anyone using data augmentation should carefully evaluate the result of using the augmented data set to see if adding synthetic data actually helps.

Also, evaluation should not purely be based on synthetic data.  Sometimes it is acceptable to evaluate using a mix of real and synthetic data, but do not evaluate purely using synthetic data.  If the goal is to evaluate real-world effectiveness, the evaluation must (at least in part) be based on real-world data.

\subsection{Including examples\index{examples} in prompts\index{prompt}}\label{sec:fewshot}
In all of the above examples of prompts, the prompts have basically made a request and provided input data.  It is also possible to include examples in the prompt; this is called \emph{few-shot}\index{few-shot} (prompts without examples are called \emph{zero-shot}\index{zero-shot}).   The examples are not used to train the model in a classic machine learning sense, but they can tell the model what the user is looking for and guide it to producing an appropriate output.

For example, the first half of Figure~\ref{fig:BLOOMfewshot} shows the result of a request to BLOOM\index{BLOOM} to summarise sports information, without an example.  The result is fine in abstract, but suppose that a shorter and more direct sentence is more appropriate in the target application.  The second half of Figure~\ref{fig:BLOOMfewshot}  shows how we an example can be used to guide BLOOM into generating a short and direct summary sentence.

BLOOM does not use instruction-tuning\index{instruction tuning} or RLHF\index{RLHF}.  Examples can be less effective with models that use instruction tuning and RLHF, such as ChatGPT, but they still are useful in many cases.

\begin{figure}
\lineacross\\
Prompt (no example):
\begin{verbatim}
Summarise the following text
Text: Aberdeen ended their long wait for a Scottish Premiership
away win by beating Dundee United 3-1 at Tannadice.
Summary:
\end{verbatim}
Response:
\begin{verbatim}
Dundee United was defeated 3 -1 by Aberdeen at their home ground, Tannadice.

\end{verbatim}
Prompt (with example):
\begin{verbatim}
Summarise the following text

Text: Celtic kept their hopes of winning a domestic treble alive afte
beating rivals Rangers 1-0 in the Scottish Cup semifinal on Sunday
thanks to a goal from Jota.
Summary: Celtic narrowly beat Rangers 1-0.

Text: Aberdeen ended their long wait for a Scottish Premiership
away win by beating Dundee United 3-1 at Tannadice.
Summary:
\end{verbatim}
Response:
\begin{verbatim}
Aberdeen beat Dundee United 3-1 at Tannadice.
\end{verbatim}
\caption{Result of prompt without and with example to BLOOM}\label{fig:BLOOMfewshot}
\lineacross
\end{figure}

\section{Issues}
Many issues arise when using machine learning and neural techniques.  Some of the most important apply to all text-generation systems (not just neural ones), so I discuss them at length in Part 2 of this book.  In particular, concerns about safety, bias, and fairness are discussed in Section~\ref{sec:safety}, and concerns about maintenance are discussed in Section~\ref{sec:maintenance}.

In this section, I more briefly discuss several issues which primarily affect systems built with machine learning and neural techniques.

\subsection{Domain shift}\label{sec:domainshift}\index{domain!shift}

A fundamental problem with machine learning models is that they become out-of-date; this is called \emph{domain shift} and is related to software maintenance\index{maintenance} (Section~\ref{sec:maintenance}).   A model trained in 2020 will reflect the world in 2020, and hence may not give appropriate answers in 2024.  Strickland \cite{strickland2019ibm} points out that the IBM Watson\index{IBM Watson} question-answering system suffered because it could not give up-to-date information about medical\index{health} interventions which included the latest findings, and we saw in Section~\ref{sec:aan} that English (and other human languages) is dynamic and changes over time.  Braun and Matthes \cite{braun2024agbdecorpusautomatedlegal} show that GPT\index{GPT} 3.5 can give incorrect legal\index{legal} assessments because it ignores changes in laws.
 In data-to-text\index{data-to-text}, a related issue is that new data sources become available (eg, new types of scanners in hospitals) and users expect such data to be used by the NLG system; this will not happen if the NLG system was trained on earlier data which did not include the new scanner.

Domain shift issues were especially prominent during the Covid-19 pandemic\index{Covid pandemic}, when activities that had previously been fine and indeed encouraged (like going for a hike in the countryside, or meeting friends in pubs) became inappropriate and indeed sometimes illegal.  In a data-to-text context, systems which reported on business data became less useful because the business world changed as some income streams collapsed but new ones (such as emergency support from government) emerged.

At the time of writing, considerable effort is going into updating commercial prompted models to try to keep them up-to-date.  However most of this work is not published because it is commercially sensitive.  There is also a lot of interest in techniques such as retrieval-augmented generation\index{retrieval-augmented generation} \cite{lewisrag} which augment (out-of-date) models with (up-to-date) web search results.

\subsection{Question answering}\label{sec:llm-content}
As mentioned above, the focus of this book is on NLG which generates output texts based on some kind of input data.  However, many (probably most) users of prompted models\index{models!prompted} also use them as general tools to answer questions, provide information, and create documents.   In such use cases, the models are creating content based on Internet data they were trained on, they are not just describing and summarising run-time input data.

I regard this as a risky thing to do, because of the data issues mentioned above, especially as most models do not tell you where the data comes from.
Incorrect and dated content are also concerns (which relates to domain shift\index{domain!shift}, Section~\ref{sec:domainshift})

\personal{Several years ago a few websites described me as an \lingform{Israeli computer scientist}, which is not true (I am not Israeli).  I eventually got these websites to update their material on me.  However, in 2023 ChatGPT\index{ChatGPT} still described me as an \lingform{Israeli computer scientist}, presumably because it was trained on old versions of websites.}

Models can also generate content which is accurate but still not appropriate (Section~\ref{sec:upsetting}).  For example, MedPaLM\index{MedPaLM} \cite{singhal2023large} generates the following text in the context of giving heath advice
\begin{quotation}
Ear pain can be a sign of several underlying conditions, including: Middle ear infection (otitis media), which is commonly caused by a viral infection and can cause pain, redness, and swelling of the ear. External ear infection (otitis externa), which is commonly caused by a bacterial infection and can cause pain, redness, and swelling of the ear. Earwax impaction, which can cause pain, a feeling of fullness in the ear, and decreased hearing. Ear pain can also be a sign of a more serious condition, such as a brain tumor or a stroke. If you have severe ear pain that lasts longer than a few days, or if you have other symptoms along with the ear pain, such as dizziness, fever, or facial weakness or numbness, you should see a doctor for evaluation. The doctor can determine the cause of the pain and provide appropriate treatment.
\end{quotation}
When I showed the above text to a doctor, she commented that she would never say this to a patient, because it states that ear pain can be a sign of brain tumour or stroke.  This is true, but rare, and telling this to a patient is likely to cause a considerable amount of anxiety and stress.  Hence it is better to drop this statement, while keeping the statement that the patient should see a doctor if severe ear pain persists..

\subsection{Auditability\index{auditable} and Controllability\index{controllability}}
In many contexts clients want to control the behaviour of an NLG system, for example to ensure that it uses language compatible with a corporate brand, that  it says nothing which raises legal\index{legal} concerns (e.g., racist language), and more generally that the system's output is appropriate and not misleading.  There are also contexts where developers need to prove to a third party (such as a regulator\index{regulator}) that an NLG system never engages in unacceptable behaviour.

Doing this with a neural  NLG system is challenging.  Developers can control the output by adjusting training data, fine-tuning data, or changing the prompt, but this does not always work.  Even worse, proving that a neural network with hundreds of billions of parameters never engages in a behaviour is not currently possible.  Its probably more realistic to add a checking tool as a post-processor, which identifies and deletes unsafe or inappropriate content (Section~\ref{sec:autosafetydetection}); but again this does not always work, and also high-quality checking can get complex and expensive to develop.

\subsection{Legal\index{legal} and Regulatory\index{regulator} Issues}\label{sec:legal}
 As mentioned above, it seems likely that neural NLG systems which are used in healthcare and other regulated safety\index{safety}-critical areas will need regulatory approval, and this may be challenging, in part because of difficulties in rigorously testing these systems (Section~\ref{sec:testingnlg}).  Provenance\index{provenance} is also an issue for models (especially prompted models\index{models!prompted}) trained on large chunks of the Internet, since much Internet material is incorrect or otherwise inappropriate \cite{gilbert2023large}.

Companies which use language models can be legally liable for mistakes made by the models.  For example, the airline Air Canda deployed a customer service chatbot which gave incorrect (hallucinated) information about some of the airline's fares.  A court rules that the airline had to honour what the chatbot told customers even if it was not in line with the airline's actual fares\footnote{\url{https://www.forbes.com/sites/marisagarcia/2024/02/19/what-air-canada-lost-in-remarkable-lying-ai-chatbot-case/}}.

Another important legal issue is copyright\index{copyright} and intellectual property\index{intellectual property}. Early language models were trained on Internet material, such as Wikipedia\index{Wikipedia}, which is available under a \emph{Creative Commons}\index{Creative Commons}  license which allows the material to be used in all sorts of ways, including training language models.  However, larger prompted models such as PaLM and GPT4 are primarily trained on material which does \emph{not} have such a license; it is not possible to build up a sufficiently large training dataset purely from material which has a Creative Commons (or similar) license.   This has led to lawsuits\index{lawsuits}, where content authors claim that material they authored and put on the Internet has been used for commercial purposes (building a language model) which they did not authorise, and for which they receive no compensation.  The situation becomes especially difficult if content authors believe that they have lost contracts (or even jobs) to systems which were trained on material produced by these authors.

At the time of writing, the legal situation is unclear, especially since some countries are considering changing the laws which govern usage of Internet content.  One possibility is that some countries will allow models which are trained on random Internet material and others will only allow models which are built on appropriately licensed (e.g., Creative Commons) material; this international disagreement would not be ideal, to put it mildly.  It would be much better to have a worldwide convention on usage of Internet material, similar to the Berne convention\index{Berne convention} which governs copyright law globally.  However, creating international conventions is a slow process; for example the USA only joined the Berne convention in 1989, which is \emph{102 years} after the convention was accepted in 1887 by European countries including France, Germany, and the UK.

\section{Further reading and resources}\label{sec:mlreading}
At the time of writing, Jurafsky and Martin are preparing a third edition of their classic textbook, \emph{Speech and Language Processing}, which will focus on machine learning algorithms and techniques used in natural language processing.  This will be  a superb resource when it is formally published\footnote{Before publication, draft chapters are available at \url{https://web.stanford.edu/~jurafsky/slp3/}}.

Many large technology companies publish high-quality white papers and research papers about neural approaches to NLG.   I have been particularly impressed by Google's blogs and white papers (\url{https://blog.google/technology/ai/}) and research publications (\url{https://research.google/}).  Readers should remember that companies do not say bad things about their products, so there may be biases in analyses and evaluations in Google papers about Google products.  But the material from Google and other technology companies  can be a great resource for learning about how the latest technologies work.

There are numerous companies which offer language models and neural NLG systems, with new ones seemingly announced every month (sometimes every week).  I cannot recommend specific ones, since anything I say at the time of writing may no longer be true when this book is published.

Huggingface\index{Huggingface} (\url{https://huggingface.co/}) is an outstanding source for downloadable models, software, and tookits for neural NLG.   NLTK\index{nltk} (\url{https://www.nltk.org/}) is an excellent resource for earlier statistical  and ML models used in NLP. Python's \emph{scipy}\index{scipy} library contains classifiers and other generic ML tools.

There is a large literature on data issues in NLP.  Rogers \cite{rogers-2021-changing} is a good overall position paper on the topic, and Bender and Friedman \cite{benderDataStatements} argue for comprehensive data statements.  Feng et al \cite{feng-etal-2021-survey} survey data augmentation\index{data augmentation} techniques, and Lu et al \cite{lu2024machine} surveys synthetic data\index{synthetic data} in AI generally.  There are many companies and consultants who will assist with data acquisition, I will not recommend any specific companies here.

Hu et al \cite{hu2020challenges} describes the real-world challenges of dealing with domain shift\index{domain!shift} in AI models during the Covid-19 pandemic\index{Covid pandemic}; the paper is not about NLP but it does graphically illustrate the problem in a real-world setting.  Ramponi and Plack \cite{ramponi-plank-2020-neural} describe the problem from an NLP perspective, and some potential approaches to dealing with it.

I am not aware of many research papers on auditing AI or NLP models.  The UK National Audit\index{auditable} Office has a nice summary at \url{https://www.nao.org.uk/insights/how-to-audit-artificial-intelligence-models/}.  

AI law\index{legal}, governance and regulation\index{regulator} (including auditing rules) are evolving extremely quickly at the time of writing.  The EU Artificial Intelligence Act\index{EU Artificial Intelligence Act} (\url{https://artificialintelligenceact.eu/}) will establish a legal framework for AI in Europe, and other countries (including the US and UK) are also moving ahead with efforts to regulate AI.  Legal case law is also rapidly evolving, including use of Internet material for training language models.  Readers should check the up-to-date legal situation in their home country; anything said in this book about AI laws, governance, and regulation is likely to be out-of-date.

%
%
%
\chapter{Requirements}
\label{requirements} 

One of the most important questions in applied NLG is user requirements\index{requirements}; what do real-world users and other stakeholders want NLG systems to do?  AI systems are only useful if they do what users want them to do \cite{stangle:blindimages}.

Of course requirements depend on the use case\index{use cases}, which is discussed in Chapter~\ref{applications}.  There is a large literature on requirements in the general software engineering\index{software engineering} literature, which NLG system builders should be familiar with.

At the simplest level, requirements analysis\index{requirements!analysis} is about understanding what texts the NLG system should generate from different inputs in different contexts; what is their content, and how is this content expressed. The first step in this process is to explore high level issues such as:
\begin{itemize}
\item \emph{Quality Criteria}\index{quality criteria} (Sections~\ref{req-criteria} and \ref{system-req-criteria}):  What aspects of quality are important to the users?  Are they more interested in readability\index{readability} or in accuracy\index{accuracy}? Does worst-case\index{worst-case} performance matter as well as average-case? How important are non-functional aspects such as time needed to generate a text?  \emph{Different quality criteria are important in different use cases and contexts}; for example accuracy is not important when generating fiction, but it is very important when generating texts that support medical decision-making.
\item \emph{Workflow}\index{workflow} (Section~\ref{sec:workflow}): NLG systems can be used in many different \emph{workflows}.  In particular, they can be used fully automatically (with no human involvement); with a human checking and \emph{post-editing}\index{post-editing} the NLG output texts before they are released; or to create draft material which human writers can use if they wish.  Each of these workflows has associated requirements.  For example, NLG systems used in a context with human checking and post-editing do not need to generate perfect texts, but they must be integrated into a user interface which makes human post-editing as easy as possible.
\item \emph{Text and Graphics}\index{graphics} (Section~\ref{sec:textgraphics}): A key question from a requirements perspective is where users want to see information communicated in words, and where they would prefer that information be communicated visually, for example using information graphics\index{graphics!information}.  Often the best strategy from the users perspective is to use a mixture of text and graphics to communicate information.
\end{itemize}

Various methodologies can be used to understand the requirements of an NLG system.
\emph{User studies}\index{user studies} (Section~\ref{sec:userstudies}) are often a good way for developers to develop a high-level understanding of the above issues.
A useful technique for understanding at a more detailed level what content the NLG system should communicate in different contexts, and how this should be expressed linguistically,  is \emph{manual corpus analysis}\index{corpus!manual analysis} (Section~\ref{sec:manualcorpus}), where a (relatively small) set of human-written \emph{target texts}\index{target texts} is manually annotated for content and aspects of linguistic style.

\begin{figure}
\lineacross\\
\emph{Key text-level quality criteria}

\textbf{Readability:} Is a text easy to read and understand?

\textbf{Accuracy:} Is the information in a text accurate and correct?

\textbf{Content:} Does the text include the most important and relevant insights for the user?

\textbf{Utility:} Does the text help the user?\\

\emph{Key system-level quality criteria:}

\textbf{Non-functional:} Does the system generate texts quickly and cheaply?

\textbf{Variation:} Is different wording used in different texts (if this is desirable)?

\textbf{Average and worst case:} Do \emph{all} texts meet minimal expected quality criteria?
\caption{Selected quality criteria for NLG}\label{fig:qualitycriteria}
\lineacross
\end{figure}

\section{Quality Criteria: Texts}\label{req-criteria}\index{quality criteria}

Many different \emph{quality criteria} \cite{belz-etal-2020-disentangling} can be used to assess the usefulness and appropriateness of an NLG system.   Chapter~\ref{evaluation} examines evaluation techniques, here the focus is on what some of these criterion are, and how their importance differs in different contexts.   This section looks at text-level quality criteria, Section~\ref{system-req-criteria} looks at system-level criteria.

Many quality criteria have been discussed in the literature \cite{howcroft-etal-2020-twenty,belz-etal-2020-disentangling}.  This book will focus on the criteria listed in Figure~\ref{fig:qualitycriteria}, and illustrate these using the example shown in Figure~\ref{fig:quality-example}; this shows a simple weather forecast produced by ChatGPT from basic weather data.

\personal{The quality criteria in Figure~\ref{fig:qualitycriteria} are the criteria which I personally have used most often in NLG projects. There are additional criteria which can be important in some use cases, such as emotional appropriateness (Section~\ref{sec:upsetting}).}

\begin{table}

\begin{tabular}{|r|r|r|r|}
\hline
time & wind speed & precipitation & temperature\\ \hline
0 & 10 & 0 & 6 \\ \hline
3 & 11 & 0 & 6 \\ \hline
6 & 12 & 0 & 7 \\ \hline
9 & 9 & 0 & 8 \\ \hline
12 & 10 & 0 & 12 \\ \hline
15 & 12 & 0 & 15 \\ \hline
18 & 9 & 0 & 12 \\ \hline
21 & 9 & 0 & 10 \\ \hline
\end{tabular}
\caption{Example weather data for a specific location over one day, at 3-hourly periods}\label{tab:quality-example-data}
\end{table}

\begin{figure}
\emph{Output of ChatGPT:}\\
\lingformquote{Here's a short weather forecast based on the provided data:}\\

\lingformquote{Expect a day with changing wind speeds and temperatures. The wind will range from 9 km/h to 12 km/h throughout the day, occasionally gusting higher. The temperature will start at 6°C and gradually increase to 15°C in the afternoon before dropping to 10°C in the evening.}\\

\lingformquote{There is no precipitation expected, so it will be a dry day.}\\

\lingformquote{In summary, anticipate a mix of wind speeds, with temperatures ranging from 6°C to 15°C, and no rainfall.}\\
\caption{Example text produced from weather data (Table~\ref{tab:quality-example-data})}\label{fig:quality-example}
\lineacross
\end{figure}

\subsection{Readability\index{readability} and Fluency\index{fluency}}\label{sec:fluency}
One fundamental criteria for NLG systems is that their output texts are easy for people to read.    In the research literature, this is sometimes called \emph{fluency} or \emph{clarity}\index{clarity} as well as \emph{readability}; unfortunately terminology in the research literature is not standardised \cite{howcroft-etal-2020-twenty}.  From a theoretical perspective, we can distinguish between \emph{Readability} (users can quickly read a text and understand its content) and \emph{Fluency} (users subjectively think that a text is fluent and well-written), but in this book we will treat these as being different aspects of an underlying assessment of linguistic quality.

For example, readers may judge that the text in Figure~\ref{fig:quality-example} is reasonable but not excellent from a readability and fluency perspective.  Most readers can read and understand the text, but it could be shortened and simplified (which would improve reading time), for example by dropping \lingform{There is no precipitation expected} and simply saying that this will be a dry day.

An example of an incoherent text produced by an NLG system in the sports domain is \cite{thomson-reiter-2020-gold}:
\begin{quotation}
\lingformquote{Markieff Morris also had a nice game
off the bench, as he scored 20 points and
swatted away late in the fourth quarter
to give the Suns a commanding Game 1
loss to give the Suns a 118-0 record in the
Eastern Conference’s first playoff series
with at least the Eastern Conference win
in Game 5.}
\end{quotation}

Poor readability is usually unacceptable; if the user cannot read the text, it is not useful.   However, the importance of high readability (very well written text) vs medium readability (as in Figure~\ref{fig:quality-example}) depends on the use case.   In particular, many texts targeted towards professionals, including financial reports, legal\index{legal} documents, and medical\index{health} summaries, need to be understandable and have appropriate content (see below), but users may not insist on very high levels of readability.  On the other hand, high readability can be important in texts intended for the general public, especially if the target audience includes people who are non-native speakers\index{non-native speakers} or have limited levels of literacy\index{literacy}.

\subsection{Accuracy}\label{sec:qualityaccuracy}\index{accuracy}
Another fundamental criteria for NLG systems is \emph{accuracy}; is the information in the text factually correct?  Again different names can be used, such as \emph{fidelity}\index{fidelity}.    Another way of expressing this criteria is minimising \emph{hallucinations}\index{hallucinations}, that is statements which are not correct.

At the time of writing, a lot of attention is being paid to accuracy problems in texts produced by large  language models\index{large language models} and other neural NLG systems.  It is rare for such systems to generate incoherent texts with poor readability, but unfortunately they can still sometimes generate texts with serious accuracy problems.

If we look at Figure~\ref{fig:quality-example}, it is mostly correct but there are some accuracy problems:
\begin{itemize}
\item The text says that wind speeds are in \lingform{km/h}, but this is not stated in the input.  In fact, these speeds are in \lingform{mph}.   Hallucinating units of measure is not good practice.  If the system does not know the units of measure, it should not guess them, and instead simply say \lingform{the wind will range from 9 to 12} (etc.).
\item The text states there will be \lingform{changing wind speeds}.  In fact, a wind speed range of 9-12 over 24 hours is pretty stable. 
\end{itemize}
A wider range of problems would probably be seen if the input data was noisy in the sense discussed in Section~\ref{sec:noise}.

\begin{figure}
\lineacross\\
Types of accuracy:
\begin{itemize}
\item Accurate with respect to \emph{input data} vs. accurate with respect to \emph{real world}.
\item \emph{Semantic} (literal) accuracy vs. \emph{pragmatic} (inferential) accuracy.
\item \emph{Critical} information correct vs \emph{all} information correct.
\end{itemize}
\caption{Types of accuracy}\label{fig:accuracytype}
\lineacross
\end{figure}

Accuracy can be formalised in many ways (Figure~\ref{fig:accuracytype}).  One issue is whether output texts should be accurate with regard to their \emph{input data}, or accurate with regard to the \emph{real world}.   For example, the text in Figure~\ref{fig:quality-example} states that temperatures are in Centigrade (\lingform{°C}); this happens to be true but is not stated in the input data.
 If we are assessing accuracy against input data, this would be regarded as an inappropriate \emph{hallucination}; however it would be fine if we assessed accuracy against the real-world.

Sometimes the term \emph{extrinsic hallucination}\index{hallucinations!extrinsic} is used for cases where extra information is added to the text, beyond what is in the input data.  \emph{Intrinsic hallucinations}\index{hallucinations!intrinsic} are statements that contradict the input data.

Another issue is \emph{semantic}\index{semantics} (literal) accuracy compared to \emph{pragmatic}\index{pragmatics} (inferential) accuracy.  For example, in a sports context, assume that in a football match between Aberdeen and Dundee, Bojan Miovski scores 2 goals for Aberdeen and Glenn Middleton scores one goal for Dundee.  Consider the statement

\begin{quotation}
\lingform{Miovskoi scored 2 goals for Aberdeen, and Middleton scored one goal.}
\end{quotation}

This is literally correct, but it implies that Middleton also played for Aberdeen, which is incorrect.  Again we can define accuracy only based on semantic correctness, or we can define it to also include pragmatic correctness.

Of course, different accuracy errors will have different impacts, and sometimes it is useful to distinguish between \emph{critical}\index{errors!critical} and \emph{non-critical} factual errors. For example, when summarising a doctor-patient consultation (Section~\ref{IntroConsultationSummary}) \cite{moramarco-etal-2022-human}, it is essential to accurately record information which impacts clinical care.  Therefore \lingform{the patient is vomiting} is a critical error if this statement is incorrect (ie, the patient is not vomiting), since this may change decisions about appropriate medical interventions.  However, accurately recording information about family members is less likely to impact clinical care; hence \lingform{the patient's wife is vomiting} may be considered a non-critical error if this statement is incorrect (ie, the patient's wife is not vomiting).  Of course the definition of critical and non-critical depends on the use case.  More categories are possible; for example Freitag et al \cite{FreitagMQM} classify errors as \emph{Major}, \emph{Minor}, or \emph{Neutral}.

The importance of accuracy depends on the use case (Chapter~\ref{applications}).  It is of paramount importance in many medical\index{health} applications, where critical errors in particular are unacceptable, and in general it is very important that texts be semantically correct and also pragmatically correct (not lead to false inferences).  In legal\index{legal} contexts, semantic accuracy is very important, but pragmatic (inferential) accuracy is sometimes less critical.  At the other extreme, accuracy is irrelevant for fiction-generation\index{fiction generation} systems.  Journalism\index{journalism} is an interesting case, where gross inaccuracy is unacceptable, but limited factual errors may be tolerated in some cases; Thomson et al \cite{thomson-csl23} found 1.5 factual errors (on average) in human-written summaries of basketball games.

\subsection{Content}\label{sec:omissionscriteria}\index{content}
In addition to being accurate, it is usually important for the content of a generated text to contain the key insights\index{insights} and information that the user needs to know.   A related concept is minimising \emph{omissions}\index{omissions}, where key information is not present in a text.

Content and omissions are harder to pin down than readability and accuracy, in part because they depend on the use case and target audience.  Readability and accuracy can be defined in a general manner that is somewhat domain-independent (although their importance depends on the use case).  However we cannot specify what content is appropriate independent of the use case\index{use cases}.

For example, if we look at the text in Figure~\ref{fig:quality-example}, its probably fine from a content appropriateness perspective for a forecast aimed at the general public.   However, it would not  be appropriate for a forecast for offshore oil rigs, because it omits detailed information about wind speeds.  Some oil-rig maintenance procedures can only be done if the wind speed is less than 10, and the Figure~\ref{fig:quality-example} text does not specify when wind speeds are greater than or less than 10.

From a content perspective, the simplest situation is when all of the information in the system's input needs to be communicated in the generated text, and hence any input information which is not communicated is a content error.  This was the case in the E2E\index{E2E} challenge \cite{DUSEK2020123}, for example, where the systems input was a small number of features (about a restaurant), and the generated texts was supposed to include all of these features.  However this is unusual, usually NLG systems produce texts which summarise their input data in some fashion.  Certainly a data-to-text system whose input data is large (thousands or even millions of numbers \cite{turner-etal-2009-generating}) must summarise, it cannot explicitly communicate this amount of information in a readable text of realistic size.

When a system does summarise its data, its texts have appropriate content when they include the key insights and messages for the target user and use case.  Sometime the key insights from a content perspective are the same as the ones which can give rise to critical accuracy\index{accuracy} errors (Section~\ref{sec:qualityaccuracy}).
For example, if we know that  \lingform{The patient is vomiting} is a critical error if it is incorrect, it is likely that omitting this statement (if it is true) will be a content error.

\subsection{Utility}\label{req-utility}\index{utility}
Perhaps the most important quality criteria, but also the hardest to define and measure, is utility.  Is the generated text effective in helping the user or otherwise fulfilling its communicative goal?   Utility of course is completely dependent on the use case\index{use cases} and target audience, even more so than omissions and content quality.

For example, Gkatzia et al \cite{gkatzia-etal-2016-natural} analyse and evaluate NLG weather forecasts\index{weather forecasts} for the specific task of helping outdoors ice cream vendors decide when to sell ice cream (Section~\ref{sec:gkatziaweather}).  They most important weather parameter in this context is usually rain (very few people buy ice cream from an outdoors vendor if it is raining); temperature can also play a role.   The utility of the text in Figure~\ref{fig:quality-example} can be assessed in this use case; its probably pretty good, since the key information about rain and temperature is clearly stated.  On the other hand, the utility of this forecast would probably be poor for offshore oil rigs (as discussed in Section~\ref{sec:omissionscriteria}).

Many NLG systems are deployed in \emph{human-in-loop}\index{human-in-loop} workflows where  a person checks and edits NLG texts before they are sent to end users (Section~\ref{sec:humanchecking}).  In such cases, \emph{post-edit time}\index{post-editing!time} (the amount of time that people need to check and edit the NLG texts) is often closely linked to utility; the value of the system is lower if people need to spend a lot of time checking its output texts.  Of course, post-edit time depends on user-interface and workflow as well as the generated texts.  It also is variable across users, with some people doing considerably more edits than others \cite{sripada-etal-2005-evaluation}.

We also need to keep in mind that real-world utility can depend on non-technical factors.  For example,  Babytalk\index{Babytalk} BT-Family system (Section~\ref{sec:babytalk}) generated summaries for parents of the status of a baby in a hospital neonatal ICU \cite{mahamood-reiter-2011-generating,mahamood-reiter-2012-working}. The system was deployed and used by parents.  When we asked them about utility, they said it would be much more useful if the summaries could be put on the Internet, especially since in many cases one parent was staying at home with other children while the other parent was spending a lot of time with the sick baby in the hospital.  Unfortunately, the hospital's IT security\index{security} policy prohibited doing this.  Hence BT-Family's utility was reduced by real-world factors which had nothing to do with NLG.

\section{Quality Criteria: Systems}\label{system-req-criteria}\index{quality criteria}
Most work on quality criteria has focused on the quality of texts produced by NLG systems, but there are also important criteria at the \emph{system} level.

\subsection{Non-functional requirements}
In many cases \emph{non-functional} requirements\index{requirements!non-functional} are important, such as the amount of time required to produce a generated text\index{compute!speed}.   In interactive systems such as chatbots, the general rule is that the computer system should respond within one second \cite{nielsen1994usability}, and sometimes this can be challenging for systems based on large neural language models.

For example, it took an early version of ChatGPT (GPT 3.5) around 1 second to produce the text shown in Figure~\ref{fig:quality-example} (later versions are considerably faster).  In an interactive chatbot context, 1 second may be acceptable (although it is not great); 5 seconds would not be acceptable.

A related issue is compute resources\index{compute!resources} and costs\index{compute!cost}.  Running a large language model can require sizeable compute resources, and such resources may not be available locally, for example if the NLG application needs to run on a phone in airplane mode (no internet access).  The situation is easier if the NLG application can access compute resources on the cloud, but such resources must be paid for, and there will be limits to acceptable expenditure.  In late 2023 generating the Figure~\ref{fig:quality-example} text using GPT4 cost around US\$0.01, which would probably be too expensive for a weather app that relies on advertising for income.

Privacy\index{privacy} and security\index{security} are also very important in many commercial contexts.  In particular, some companies will not use cloud-based services for processing sensitive data because of privacy and security concerns, and insist that all processing be done internally.

\subsection{Consistency and Variation}\label{sec:variation}\index{consistency}\index{variation}
In many use cases\index{use cases}, users want texts to be \emph{consistent} and always express information in the same way.  In medicine\index{health}, for example, texts describing clinical phenomena should be precise and use standard medical terminologies\index{terminology}, in order to minimise the chances that the text is misunderstood.   This is especially important when texts may be read by non-native speakers\index{non-native speakers} with variable levels of literacy \cite{reiter1995automatic}.

However, there are also use cases where users want to see \emph{variation} in texts, where the same information is expressed differently on different occasions.  This is especially important in contexts where readers see many texts produced by the NLG system, such as journalism\index{journalism}, sports reporting and weather forecasts\index{weather forecasts}.  Readers will get bored if they see the same language repeated in media stories \cite{diakopoulos2019automating}.  Figure~\ref{fig:variation} shows some possible variations of the last sentence in  Figure~\ref{fig:quality-example}, where the same content is expressed using:
\begin{itemize}
\item different words (e.g., \lingform{no rainfall} vs. \lingform{dry})
\item different ordering of phrases (e.g., start with wind speed information or start with precipitation information)
\item different semantic structures (e.g., whether reader is told something will happen, or told to expect that something will happen)
\end{itemize}

\begin{figure}
\lineacross\\
A few variations of a sentence in a weather forecast: 
\begin{itemize}
\item \emph{Original}:~\lingformquote{In summary, anticipate a mix of wind speeds, with temperatures ranging from 6°C to 15°C, and no rainfall}
\item \emph{Variation 1}:~\lingformquote{In summary, expect a mix of wind speeds, no rainfall, and temperatures ranging from 6°C to 15°C.}
\item \emph{Variation 2}:~\lingformquote{In short, the day should be dry with temperatures between 6°C to 15°C and a mix of wind speeds.}
\end{itemize}
\caption{Variations of the last sentence in Figure~\ref{fig:quality-example}; content is the same, but it is expressed differently}
\label{fig:variation}
\lineacross
\end{figure}

Sometimes users care about both consistency and variation; for example in a weather forecast context they may want consistency in content words such as \lingform{dry}, but variation in other types of words such as connectives (e.g., alternating between \lingform{but} and \lingform{however}).

It is dangerous to generalise about use cases, but overall consistency is more likely to be important for NLG systems that generate texts for professionals, while variation is most likely to be important in systems that generate texts for consumers or the general public.  Of course there are exceptions!

\subsection{Average vs Worst-Case}\label{sec:req-worstcase}\index{worst-case}
Another system-level issue is the distribution of quality criteria\index{quality criteria} across generated texts.  Usually different texts generated by an NLG system have different individual ratings on quality criteria.  For example, if we look at accuracy, perhaps a few texts will have no errors\index{errors}, some will have one error, some will have two errors, and a few will have three or more errors.  If we want to compute quality criteria for \emph{systems} instead of individual texts, then we have to decide how to map a distribution of text-level quality scores onto a single system-level quality score.  In principle we could report the distribution of text-level scores, but this is rare.

If we want to report a single number which aggregates the distribution, the most common choice is the mean (average) value of the criteria across a set of texts; this is \emph{average-case performance}. However, in some cases (especially when safety is important, see Section~\ref{sec:safety}) it is essential that \emph{all} texts generated by a system have acceptable quality; in such cases we report the minimum value of the criteria across the set of generated texts (\emph{worst-case performance}).  An example is shown in Table~\ref{fig:averageworst}.

If average readability is most important in this use case, we should report the mean readability of the generated texts (8).  However, if the key requirement is that all texts achieve a minimum readability level, then we should report the minimum readability score (4).  In some use cases we may want to report both, for example if we want a high average readability score but also a guarantee that all texts achieve a minimum readability level (eg, not be incomprehensible).

\begin{table}
\lineacross\\
\begin{tabular}{|l|r|} \hline
Generated text & Readability (0-10) \\ \hline
\lingformquote{Tomorrow should be dry with temperatures between 6°C to 15°C.} & 9 \\ \hline
\lingformquote{Tomorrow will be a dry day, with moderate temperatures.} & 9 \\ \hline
\lingformquote{Tomorrow will be a moderate day with dry temperatures.} & 4 \\ \hline
\lingformquote{Tomorrow will be a really nice day.} & 10 \\ \hline
\end{tabular}

\blankline
Average (mean) readability: 8 \\
Worst-case (minimum) readability: 4 \\
\caption{Average and worse case ratings of a set of texts.}\label{fig:averageworst}
\lineacross
\end{table}

\section{Workflow}~\label{sec:workflow}\index{workflow}\index{integration}
Another key aspect of requirements is how the NLG system fits into overall \emph{workflows}.  In particular, it is useful to distinguish between \emph{fully automatic NLG}, \emph{NLG with human checking\index{human checking} and post-editing\index{post-editing}}, and \emph{humans creating a document from an NLG draft}.

A related concept is \emph{integration}\index{integration}, that is connecting the NLG system to input data sources and output document delivery mechanisms.  This is very important for production systems; in commercial NLG projects, a sizeable chunk of engineering effort is often spent on integration.  However, since integration issues for NLG systems are similar to integration issues for other types of software systems, I will not discuss them here.

\subsection{Fully automatic NLG}
A fully automatic NLG system produces a narrative with no human involvement (other than humans reading the generated narrative).

At the time of writing, there is limited use of fully automatic NLG systems in professional contexts, in part because of the fear that some of the NLG texts may be incorrect.  In other words, with regard to quality criteria (Section~\ref{req-criteria}), there is a fear that worst-case\index{worst-case} readability, accuracy, and/or content appropriateness will not be acceptable; i.e., in a few cases the generated texts will be difficult to read, inaccurate, or missing key information.  In safety-critical\index{safety} or high-value contexts, this is not acceptable.   Of course, there are use cases\index{use cases}, perhaps especially in consumer-oriented contexts, where a small number of mistakes are perhaps acceptable, and fully automatic NLG systems can be used in such contexts.

The weather  system  \cite{arun-etal-2020-best}  described in Section~\ref{sec:Facebookweather} is an example of such a system.  While some specialised types of weather forecasts\index{weather forecasts} (such as aviation forecasts) are safety critical, forecasts for the general public usually do not kill people if they are wrong.  Also, weather forecasts by their nature are approximate and sometimes incorrect because they are \emph{forecasts} (the result of a numerical model which simulates how the atmosphere will change in order to predict weather parameters over time), so users know that they cannot be 100\% trusted even ignoring NLG issues.

\subsection{Human checking\index{human checking} and editing}\label{sec:humanchecking}
In NLG systems used by professionals (doctors, accountants, engineers, etc.), it is more common to see workflows where a person checks the output of the NLG system and fixes (\emph{post-edits})\index{post-editing} the text if necessary.  This is the certainly the case with systems that generate medical\index{health} documents such as consultation summaries \cite{knoll-etal-2022-user} (Section~\ref{sec:medicalreporting}), and its also true of most NLG systems which are used to generate journalistic\index{journalism} content \cite{diakopoulos2019automating} (Section~\ref{journalism}).

 In the weather domain, the forecasts\index{weather forecasts} produced by SumTime\index{SumTime} were checked and edited by human forecasters \cite{sripada-etal-2005-evaluation} before they were sent to clients.  This is partially because these were specialist forecasts for people operating off-shore oil rigs, so accuracy requirements were higher than for public forecasts (Section~\ref{ArriaWeatherExample}).  Figure~\ref{fig:sumtimepostedit} shows an example of how a SumTime text describing the wind was edited by a human forecaster before being released to clients.  

\begin{figure}
\lineacross\\
\emph{Original SumTime wind text:}  \lingformquote{SW 20-25 backing SSW 28-33 by midday, then gradually increasing 34-39 by midnight.}\\
\emph{Human-edited version of above text:} \lingformquote{SW 22-27 gradually increasing SSW 34-39.}\\
\caption{Example of post-editing a SumTime wind text, from \cite{sripada-etal-2005-evaluation}.  Human has changed \lingform{20-25} to \lingform{22-27}, dropped the phrase \lingform{backing SSW 28-33 by midday}, and also dropped \lingform{then} and \lingform{by midnight}.}
\label{fig:sumtimepostedit}
\lineacross
\end{figure}

Sripada et al \cite{sripada-etal-2005-evaluation} analysed a corpus of such edits to SumTime forecasts (real edits made during operational usage of the system), and discovered that there was a considerable difference in the amount of post-editing done by different forecasters. This is partially because some forecasters are fussier than others, but also because some forecasters insisted on editing texts into their personal preferred style.   If consistency is important, editors should be given clear guidelines\index{guidelines} on what should be fixed, ideally supported by training.

Post-editing also of course is influenced by the editing user-interface\index{user interface} and by how it fits into the editors overall workflow \cite{knoll-etal-2022-user}.   A good post-edit UI can make post-editing much quicker, and also can help editors conform to instructions.

The result of a workflow where human experts post-edit NLG outputs can be excellent.  Moramarco \cite{moramarcophd} showed that for the task of writing summaries of doctor-patient consultations, texts produced by asking clinicians to post-edit summaries produced by Note Generator (Section~\ref{IntroConsultationSummary}) seemed to be slightly \emph{more} accurate (on average) than texts produced by the same clinicians using a completely manual workflow.

We can also use machine-learning techniques to learn from post edits and refine our models so that they make fewer mistakes in the future.  This is sometimes called \emph{automatic post editing}\index{post-editing!automatic} \cite{do2021review}.

There has been extensive work on post-editing in machine translation\index{machine translation}. Indeed, there is even an ISO standard about this (ISO 18587:2017).

\subsection{Creating drafts for human writers}\label{draftworkflow}
Finally, NLG systems can be used to generate draft documents which human writers can use when they create documents.  For example, an NLG system may generate a financial report which is \scare{mined} for content by a human analyst; ie, the human analyst does not try to fix the entire NLG narrative, instead he or she extracts useful chunks from the NLG narrative and includes them in the document he or she is writing.  As with post-editing, different authors use drafts in different ways, and the user interface is very important.

The division between \scare{human checking and post-editing} and \scare{creating drafts} is fuzzy, but in general in the first case the NLG system is the main author, assisted by the human, whereas in the second case the human author is the main author, assisted by the NLG system.  In principal, usage of NLG in a use case can evolve, perhaps starting with the system creating drafts, then moving to a \scare{human checking} workflow, and ending with fully automatic NLG with no human involvement.

\section{Text and Graphics}\label{sec:textgraphics}\index{graphics}
Visualisations\index{visualisations}, images, and graphics can be used to communicate numerical data, and in many use cases can be an alternative or supplement to texts produced by data-to-text NLG systems.  For example:
\begin{itemize}
\item \emph{Financial and medical information} can be presented using words, informations graphics, or both.
\item \emph{Persuasive content} (eg, to encourage safer driving) can be presented using words, graphics/images, or both.
\item \emph{Instructions} (such as directions for getting from A to B) can be presented using words, maps\index{maps}, or both
\end{itemize}
Hence when building an app which communicates information to a user, developers may need to decide whether to use words (NLG), some form of graphic, or a combination of both.

Sometimes the choice is dictated by pragmatic reasons.  For example, a voice assistant which communicates over a telephone landline cannot show images, so it must rely on words. Similarly systems for blind people emphasise words; it is possible in theory to communicate graphically using Braille displays, but this is expensive and usually avoided.   On the other hand, use cases which require information to be understandable to speakers of many different languages (such as many DIY instructions) avoid words and use pictures instead.  Summaries of textual documents are also usually done with words; pictorial summaries are possible but unusual.

In many use cases, however, we have a choice between words and pictures, and hence need to decide which of these is most suitable for the use case\index{use cases} and target audience; this is an important aspect of understanding client requirements.

\subsection{Decision Support}\label{sec:gkatziaweather}\index{decision support}
One use of data-to-text NLG is \emph{decision support}, that is presenting information to people to help them make decisions.  A few studies have looked at the relative effectiveness of text and graphics in this area.

\begin{figure}
\includegraphics[scale  = 0.5]{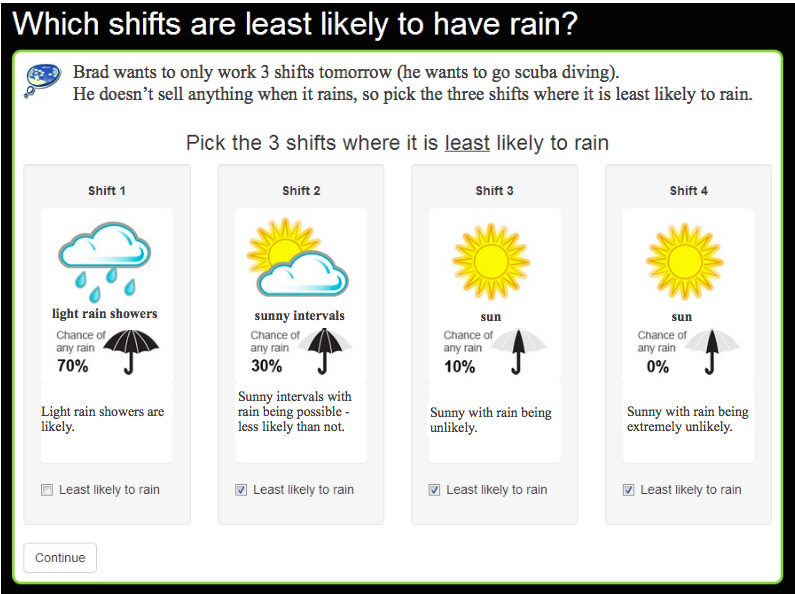}
\caption{Simple weather forcast (text and graphics); courtesy of Dmitra Gkatzia}
\label{fig:GkatziaWeather}       
\end{figure}

Gkatzia et al \cite{gkatzia-etal-2016-natural} present an interesting study of this issue in the context of simple weather forecasts\index{weather forecasts}.  Fig~\ref{fig:GkatziaWeather} shows an example of the kind of forecast they examined; this example shows both images and words.  They also had an images-only condition (no words) and a words-only condition (no images).   They ran an experiment where they showed people either the words+graphics forecasts, the graphics-only forecast, or the words-only forecast, and asked them to make a weather-related decision.  They found that the best decisions were made from words+graphics, and words-only forecasts were more effective (as decision aids) than graphics-only forecasts.

Gkatzia et al looked at a context where ordinary people were making a decision based on fairly simple data.   The Babytalk BT45\index{Babytalk!BT45} systems \cite{PORTET2009789} generated texts which were intended to support clinical decision making, and studies were done of the text-vs-graphics issue in this context, where medical professionals examine clinical data and then make decisions based on this data.
Law et al \cite{law2005comparison} compared the effectiveness of presenting clinical data to clinicians via data visualisations or
(human-written) textual summaries of data, and found that clinicians\index{doctors} who saw the human-written text summaries made better decisions.   Van de Meulen et al \cite{van2010graph} presented clinical data as either (A) data visualisations, (B) computer generated (NLG) textual summaries, or (C) human-written textual summaries, and found that doctors who saw the human-written textual summaries  again made the best decisions.  There was little difference overall in the quality of decisions made by doctors who saw the data visualisations and doctors who saw the computer-generated NLG texts.  However  in some scenarios the visualisations led to better decisions and in others the NLG texts led to better decisions.  This suggests that the best strategy is to combine the two, which matches what Gkatzia et al found.  

Of course, all of these experiments were done based on specific visualisations and NLG summaries; it is possible that different results woud have been seen with different visualisations and/or NLG summaries.  But it is plausible that a combined text+graphics presentation of information is usually best, not least because we know that which presentation is best depends on the scenario and also on the user (e.g., verbal vs visual thinker); hence a combined \scare{multi-modal}\index{multi-modal} presentation will be more robust across scenarios and users.

\subsection{Other Use Cases}
Balloccu and Reiter \cite{balloccu-reiter-2022-comparing} looked at graphical vs text+graphics presentation of dietary information in an app which gave people feedback about their diet.  They found that users had better understanding of the dietary data when it was presented using a mixture of text and graphics.

McKeown et al \cite{MCKEOWN199895} look at a number of use cases, including generating maintenance instructions and generating briefs which summarise a patient's status after an operation.  They again suggest that the best strategy is to mix text and graphics.

\subsection{Combining Text and Graphics}
The above papers suggest that a multi-modal\index{multi-modal} presentation of information, combining text and graphics, is usually the most effective.  Of course this raises the question of how to best combine text and graphics.

The simplest approach is to generate the text and graphics independently, so that the user sees text and graphical summaries of the data; this approach was used by Gkatzia et al \cite{gkatzia-etal-2016-natural} (Figure~\ref{fig:GkatziaWeather}).  It allows users to focus on whichever media (text or graphics) works best for them in their current context; also if they struggle to understand something which is presented graphically, they can look at the text, and vice-versa.

It is also possible to present different kinds of information in the two media. For example, we can use the textual descriptions to describe what-if analyses, causal links, background information, and other things that are difficult to describe in graphs, while using the graphs to communicate raw numerical data (which is awkward to do in words).  This approach is common in business intelligence\index{business intelligence} (Section~\ref{sec:bi}), and there are a number of commercial tools which essentially explain business data (such as sale and profits) using a mixture of graphs and words; sometimes the textual information is intended to expand and/or clarify the data visualisations\index{visualisations} \cite{mahamood-etal-2014-generating}.
In interactive chatbot\index{chatbot} contexts (such as Balloccu and Reiter \cite{balloccu-reiter-2022-comparing}), the user may be able to explicitly request textual or graphical presentations of data.   We might describe this approach as \emph{loose coupling}; the text and graphics communicate different information, but they can be examined independently.  In other words, the textual component makes sense even without the graphical component, and vice-versa.

\begin{figure}
\includegraphics[scale  = 0.8]{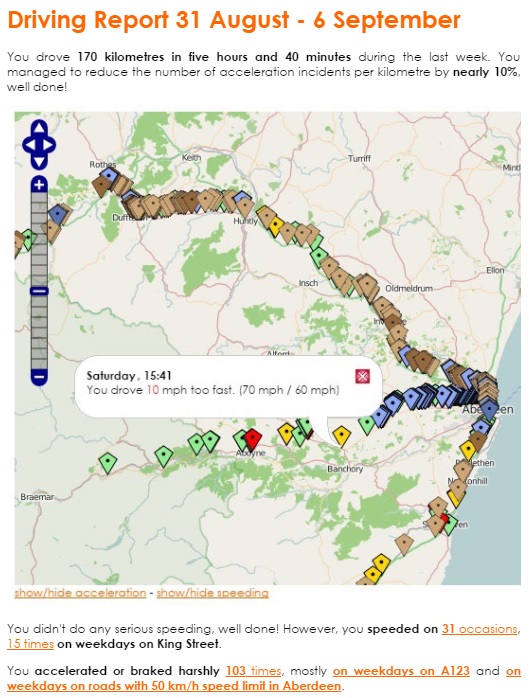}
\caption{Combining text and graphics in SaferDriver\index{SaferDriver}; courtesy of Daniel Braun}
\label{fig:saferdrivermultimodal}       
\end{figure}

A nice example of this from a research system is SaferDriver\index{SaferDriver} \cite{braun_reiter_siddharthan_2018}, which generated feedback reports for drivers about speeding and other unsafe driving behaviour, based on GPS data.  As shown in Figure~\ref{fig:saferdrivermultimodal}, SaferDriver used NLG texts to give summary information, key insights, and encouragement/praise, but it used a map\index{maps} to show detailed driving data.

In the 1990s there were a number of research projects which looked at close integration of text and graphics, where the text component did not make sense without the graphical component, and vice-versa.   Much of this work focused on instruction giving \cite{feiner1991automating,wahlster1993plan,MCKEOWN199895}; I summarise some of it in Chapter 7 of my 2000 book \cite{reiterdale2000}.   This approach is less common in recent work.

\section{Requirements acquisition}\index{requirements!analysis}
Since requirements are very important, a key question is how developers and system builders can understand user requirements.  There is of course a large literature in the software engineering community on requirements acquisition for building software systems (for example, Wiegers and Beatty \cite{wiegers2013software}), most of which is applicable to building NLG systems.  There is also some work explicitly on understanding requirements for NLG.

\subsection{User studies}\label{sec:userstudies}\index{user studies}
A very useful technique for understanding requirements is to conduct \emph{user studies}.  This is a powerful technique which is widely used in software engineering and interface design.

\begin{figure}
\includegraphics[scale  = 1]{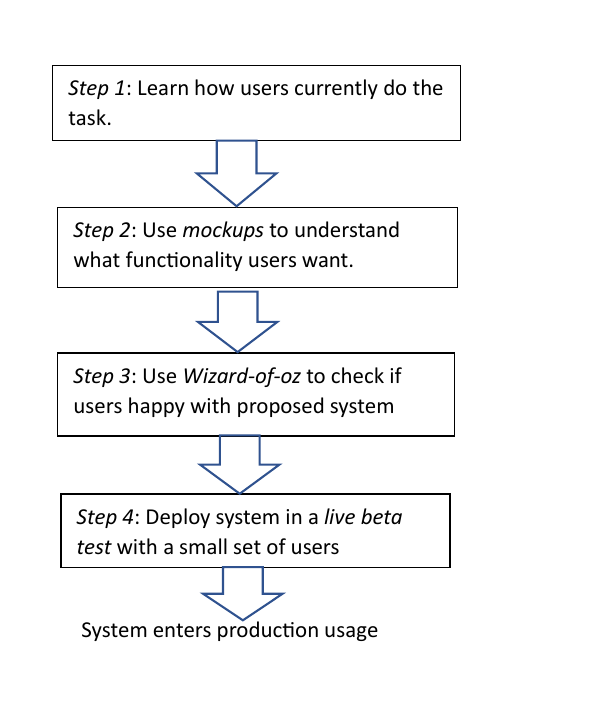}
\caption{User studies process, inspired by Knoll et al \cite{knoll-etal-2022-user}}
\label{fig:userstudies}       
\end{figure}

Knoll et al \cite{knoll-etal-2022-user} describe a set of user studies they performed in order to understand requirements for Note Generator\index{Note Generator} (Section~\ref{IntroConsultationSummary}), a system which summarised doctor-patient consultations.  They divide the process into several stages (Figure~\ref{fig:userstudies})

\emph{Step 1: Analysis of how humans currently do the task.} If an NLG system is (semi-)automating a task which is currrently done manually, then the first step is understanding how humans currently perform the task.  Knoll et al conducted one-hour semi-structured interviews\index{semi-structured interviews} with seven clinicians in order to understand how they manually did the task, where they needed help, and what the difficult cases were; they also created several \emph{personas}\index{personas} based on the people they interviewed.

The difficult cases (for writing up doctor-patient consultations) included cases where patients were not truthful\index{lies} or exaggerated symptoms, cases where patients had multiple problems, and cases where patient's non-verbal behaviour communicated important information.  In general understanding hard cases is essential in requirements analysis; developers need to ensure that they have a strategy for dealing with hard cases (which could be reverting to human authorship).

\emph{Stage 2: Mockups.}\index{mockups}  Once developers have an idea what an NLG system will do, the next step is to get feedback from users and stakeholders\index{stakeholders} to verify that this functionality is useful and meets user/stakeholder needs.  A common practice in interface design, which Knoll et al use, is mockups;  they show users simple mockups of systems, and used this to get feedback on what interface and functionality is best.  Showing mockups is cheap and can be done very early in the system-building process.

Amongst other things, Knoll et al discovered that users (doctors) wanted the consultation summary to be generated and updated in real-time, as the doctor and patient spoke.  In other words, doctors wanted to see a summary of the consultation as it progressed, not just at the end.  This had an impact on the technology used in the final system.

\emph{Stage 3: Wizard of Oz.}\index{Wizard of Oz} Knoll et al then built a \emph{Wizard of Oz} system, where users thought they were interacting with the NLG system, but in fact the summaries were produced by a human \emph{Wizard}, who wrote the texts which the users thought were produced by the NLG system.  This is again a common technique in interface design.  It is a great way to get feedback on a proposed app (does it meet user needs?) at an early stage, before the app is built.

In this case, the Wizard of Oz study showed that different users used the system in different ways.  For example, some users essentially used a post-editing workflow (Section~\ref{sec:humanchecking}), that is they updated the computer-generated summary.  Other users used a drafting workflow (Section~\ref{draftworkflow}), that is they wrote their own summary, and copied over material from the computer-generated summary where it was helpful.

\emph{Stage 4: Live beta test.}\index{beta test} The final user-study stage was to deploy the complete NLG summarisation material in a limited test with a small number of clinicians, and monitor how clinicians used it and what the problems are.  This is similar to \emph{alpha} or \emph{beta} tests in classical software development.

Amongst other things, the beta test showed the doctors did \emph{not} generally use the NLG system in difficult consultations of the sort described above; in such cases they reverted to manual note writing.  This suggests that it is acceptable (but of course not ideal) for the  system to focus on handling \scare{easy} cases well.

\emph{Production usage.}\index{production} Once the above stages are completed and the system has been updated based on feedback and observation, it is time to start the process of deploying the system in full production usage!   Moramarco \cite{moramarcophd} describes the production deployment of Note Generator.
\newline
\newline
Of course there are other techniques which can be used in user studies, beyond those discussed in Knoll et al.  For example in other projects we have used focus groups to help us understand what users wanted \cite{sun2024}.

\subsection{Manual Corpus Analysis}\label{sec:manualcorpus}\index{corpus!manual analysis}
User studies focus on what users want and need, and are usually the best technique for high-level requirements analysis.  It can also be useful, especially when building rule-based NLG systems, to try to elicit detailed requirements by manually analysing a small set of high-quality input-output pars. This process is called \emph{manual corpus analysis} \cite{reiterdale2000}.

This process is as follows:
\begin{enumerate}
\item Obtain a small \emph{corpus}\index{corpus} of system inputs (typically 10-20 texts, ideally as varied as possible) and high-quality outputs for these inputs.  These outputs are called \emph{target texts}\index{target texts} and are usually written by a human domain expert.
\item Ask other people (not the domain experts\index{domain!experts} who wrote the texts) to annotate the target texts to indicate where information comes from.  Figure~\ref{figure:manualannotation} shows a simple annotation scheme which was worked well for me on many occasions.
\item Update the target texts to remove information and insights which cannot realistically be included in the generated text; note that this judgement partially depends on the NLG technology which will be used.
\item Analyse the annotations of the remaining material in the target texts in order to understand what content and wording is needed and how it might be produced.
\item Repeat above if necessary.
\end{enumerate}
An example of an annotated target text is shown in Figure~\ref{fig-corpus-annotate}.

\begin{figure}
\lineacross\\
Sentences (or in some case phrases) are annotated with one of the following categories:\\

\textbf{Static:} Always present, does not change. An example is the sentence \lingform{Remember that past performance is not indicative of future results} in Figure~\ref{fig-corpus-annotate}; this is a boilerplate fixed sentence which needs to be present for legal reasons, and never changes.\\

\textbf{Data:} Directly communicates part of the input data. An example from Figure~\ref{fig-corpus-annotate} is the sentence \lingform{The FTSE rose by 20 points yesterday}; this directly communicates basic stock market data.\\

\textbf{Insight:} Communicates insights derived from the input data. An example from Figure~\ref{fig-corpus-annotate} is the sentence \lingform{The rise was largely driven by an increasing values of energy-related stocks}; this is the result of a key-drivers-and-offset analysis of the stock market data.\\

\textbf{Background:} Communicate background information which is available on the Internet or within a language model. An example from Figure~\ref{fig-corpus-annotate} is the sentence \lingform{Crude oil prices are increasing rapidly}; this is not in the stock market data but is related information which can be obtained from the Internet.\\

\textbf{Unavailable:} Communicates information which is not available to the system. An example from Figure~\ref{fig-corpus-annotate} is the sentence \lingform{Investors who bet on oil companies are very happy}; this is a plausible guess, but the system does not have access to the emotional state of millions of investors.\\

\caption{Simple annotation scheme for manual corpus analysis}\label{figure:manualannotation}
\lineacross

\lingformquote{The FTSE rose by 20 points yesterday.} \emph{(Data)}\\
\lingformquote{The rise was largely driven by an increasing values of energy-related stocks.} \emph{(Insight)}\\
\lingformquote{Crude oil prices are increasing rapidly.} \emph{(Background)}\\
\lingformquote{Investors who bet on oil companies are very happy!} \emph{(Unavailable)}\\
\lingformquote{Remember that past performance is not indicative of future results.}  \emph{(Static)}
\caption{Example of annotated human-written text using the annotation scheme of Figure~\ref{figure:manualannotation}.  The system's input data is stock market share prices.}\label{fig-corpus-annotate}
\lineacross
\end{figure}

Part of this exercise is examining Insight and Background texts and deciding whether any of these are unrealistic to produce; for example the Background text in Figure~\ref{fig-corpus-annotate} can only be generated if the system has access to data about oil prices (ie, it has access to commodity prices as well as share prices).  Developers also need to consider whether Unavailable texts should be dropped; for example the Unavailable text in Figure~\ref{fig-corpus-annotate} is essentially a bit of journalistic colour which might be acceptable in general media stories, but probably should not be included in financial updates aimed at professional investors (who might indeed complain that they are not feeling very happy, because they expected a larger increase in stock prices).

Developers may wish to change target texts based on the corpus analysis.  Of course, any changes should be
discussed with clients and users; it is also often useful to discuss them with the domain experts who wrote the target texts if this is possible.  Some changes are minor and have little impact on users, but others may significantly decrease the utility of the NLG system.   When developers have agreed on a set of target texts with the client, this becomes an important resource in defining what the NLG system should do.

Once such changes have been made and the the target text corpus is finalised, developers can analyse it in order to better understand the type of content that will be included in generated texts, and also how it is expressed linguistically.  This analysis is very useful when building rule-based NLG\index{rule-based NLG} systems, and can directly lead to an initial set of rules.

For neural NLG, the context analysis is  useful as a way of understanding what information and reasoning is needed to generate context, which helps specify input sources and prompts.  The language/expression analysis can identify whether text needs to be in a specific genre\index{genre} such as \scare{weatherese} or \scare{legalese}.

\subsubsection{Issues and Limitations}

One limitation of manual corpus analysis is that it generally does \emph{not} have good coverage of  boundary or \scare{edge} cases\index{edge cases}, because there are usually hundreds or thousands of such cases.  We can train a machine learning model on tens of thousands of texts, but we cannot realistically do manual corpus analysis on tens of thousands of texts!  If we are building a rule-based NLG system, one approach is to identify edge cases and then explicitly ask domain experts how they should be handled..

Another limitation which is inherent in analysing human-written texts is that different human writers will write different texts, even if they are given the same input data. So a text written by author A may include different Insight statements, different Background statements, etc. compared to a text written by author B.   One strategy is to analyse texts written by several experienced domain experts, in order to get good coverage of what content experts think could be included in a text.

A related issue is that some human-written texts may not be very good.  It is very appealing to get the corpus texts from real-world texts written by professionals.  However sometimes such texts are written by less experienced domain experts working under intense time pressure, which reduces their quality.  This is certainly true of production weather forecasts\index{weather forecasts}; a weather forecast written by an experienced forecaster with plenty of time is not the same as a weather forecast which is quickly produced by a novice forecaster who is struggling to write a large number of forecasts in a short amount of time.  If text quality is a concern, it may be useful to ask an experienced domain expert (eg, forecaster) to check corpus texts and remove low-quality ones.

Last but not least, the above process requires cooperation and support from domain experts, but sometimes they are hostile to the AI/NLG initiatives because they see them as a threat to their jobs.   This is a \emph{change management}\index{change management} issue (Section~\ref{sec:acceptability}), not a technical one, and will require sensitivity and addressing the concerns of the individuals who are supposed to be supporting the project.

\subsection{Stakeholders}\label{sec:reqstakeholders}\index{stakeholders}
When working with people to understand requirements, it is important to talk to \emph{stakeholders} as well as direct users.  For example, a system which summarises doctor-patient consultations (Section~\ref{sec:userstudies}) must be acceptable to the doctors\index{doctors} who will use it!  But it must also be acceptable to managers\index{managers} who look at the \scare{big picture} impact of the system, and to legal\index{legal} and risk\index{risk} management specialists who assess the likelihood and damage (reputational as well as financial) if anything goes wrong.  Many medical systems also need to be acceptable to regulators\index{regulator} and patients.

It is important when analysing requirements to work with all types of stakeholders if possible.  For example, a doctor may not care how much an NLG system costs, but a manager certainly will!  Likewise doctors and managers may not be able to estimate risks of successful lawsuits if something goes wrong, specialists need to be consulted about this.

\personal{I have seen a number of NLG projects where success has been limited because not all stakeholders were consulted.  The most common problem has been failure to consult users; managers think a certain type of NLG system will be useful and buy it, without considering whether end users actually want this functionality.  I have also seen failures where users were consulted but management was not; we ended up building systems which users loved, but managers decided were not cost-effective.}

\section{Further reading}
There are many textbooks and other sources about software requirements in the general software engineering\index{software engineering} literature (eg, \cite{wiegers2013software}).  Indeed, getting requirements wrong is a leading cause of many IT disasters, where an expensive IT system turns out to be useless in practice.  A good example is the £10B NHS Connecting for Health programme, probably the biggest IT failure in the UK public sector (\url{https://en.wikipedia.org/wiki/NHS\_Connecting\_for\_Health}).

Strickland's excellent retrospective \cite{strickland2019ibm} on the failure of IBM Watson\index{IBM Watson} in healthcare identifies many problems which essentially are due to poor understanding of what real-world users wanted the system to do.

I am not aware of textbooks or research papers specifically on requirements for natural language generation, but application-oriented books and papers which target specific NLG use cases often mention requirements.  An excellent example which includes generalisable insights is Diakpopoulus's book \cite{diakopoulos2019automating} on automatic journalism\index{journalism}.  

A lot has been written in the general AI literature about human-AI interaction and workflows\index{workflow}; one recent survey is Mosqueira-Rey et al \cite{mosqueira2023human}.  Amershi et al \cite{amershi2019guidelines} give guidelines from HCI (human-computer interaction)\index{HCI} perspective on human-AI interaction.  O'Brien \cite{o2022deal} discusses post-editing in machine translation.  Diakopolus et al \cite{diakopoulos2024} survey workflows for using language models in journalism.
Unfortunately I am not aware of papers specifically on post-editing\index{post-editing} and other workflows in NLG, other than the ones cited above in Section~\ref{sec:workflow}.  There are numerous blogs and white papers about human-AI interaction from commercial AI companies; as always with commercial material, these are often very well written but may not be completely objective.

An increasing number of papers are being published which explore different quality criteria\index{quality criteria} for NLG, but I recommend that readers interested in this topic start with Howcroft et al \cite{howcroft-etal-2020-twenty} and Belz et al \cite{belz-etal-2020-disentangling}.  Gehrmann et al \cite{GehrmannEvaluation} is primarily a survey of NLG evaluation, but also has some good insights about quality criteria.

My 2000 book \cite{reiterdale2000} includes a fair amount of material on text and graphics\index{graphics}.  Much of this is out of date, but I think the discussion of basic issues and concepts is still relevant today.  In 2024, the best work on combining NLG and information graphics is often done commercially, by companies (such as Arria, my company) which develop solutions which combine these in real use cases.  One example of a system which Arria helped developed is described in Section~\ref{sec:covid} of this book.

For requirements acquisition\index{requirements!acquisition}, Knoll et al \cite{knoll-etal-2022-user} is the only paper I am aware of which explicitly examines requirements acquisition processes for NLG.  There are of course many general textbooks about software requirements (as above), which discuss requirements acquisition.  Textbooks about interaction and UI design in HCI (eg, Rogers et al \cite{interactiondesign}) can also be useful.

%
%
%
\chapter{Evaluation}
\label{evaluation} 

A key issue in NLG is \emph{evaluation}, in other words assessing how well an NLG system works and meets its requirements\index{requirements}.  Does it produce texts which are acceptable in its use case\index{use cases}, based on the relevant quality criteria\index{quality criteria} (Section~\ref{req-criteria})? 
Evaluations can also assess whether a system is suitable for its expected workflow\index{workflow} (Section~\ref{sec:workflow}) and if it is acceptable to different stakeholders\index{stakeholders} (Section~\ref{sec:reqstakeholders}).

In addition to assessing the effectiveness of an NLG solution, evaluation can also highlight where systems are weak and need to be improved.  For example, an evaluation of an NLG system may show that its texts have good readability\index{readability} but poor accuracy\index{accuracy}; this is very useful information for developers who are trying to improve the system.

In this chapter I first give an example of an evaluation (Section~\ref{sec:stopsmoking}), then look at fundamental evaluation issues and concepts (Section~\ref{Evaluation:Fundamentals}).  I then discuss evaluation with human subjects (Section~\ref{sec:humanevaluation}), evaluation based on automatic metrics or algorithms (Section~\ref{sec:automaticevaluation}), and evaluation based on real-world impact (Section~\ref{sec:impactevaluation}).  I also look at some topics which are important in many commercial evaluations, but less important in many academic evaluations, such as system cost (Section~\ref{sec:commercialevaluation}).  I conclude with some general advice on doing good evaluations (Section~\ref{sec:ten-tips-eval}) and suggestions for further reading. Throughout the chapter I often refer to research on evaluating machine translation\index{machine translation} systems, which is well developed and provides many insights which are relevant to evaluation of NLG.

Evaluation is related to \emph{software testing}\index{testing} of NLG systems (Section~\ref{sec:testingnlg}).  However, the focus of evaluation is on trying to quantify the performance of an NLG system (usually on a set of relevant quality criteria), while the focus of software testing is on identifying specific cases and scenarios where an NLG system behaves unacceptably.

\section{Example: Smoking Cessation (Impact Evaluation)}\label{sec:stopsmoking}\index{impact!evaluation} 
I will start with an example evaluation of an NLG system which generated leaflets which were intended to help people stop smoking \cite{REITER200341}.  This was an \emph{impact} evaluation, because it evaluated the system's real-world effectiveness in encouraging smoking cessation.  The evaluation was simple conceptually (which is why I use it as an example), although it required a lot of effort to carry out.

\begin{figure}
\lineacross\\
\includegraphics[scale  = 0.6,trim={0 10cm 0 0},clip]{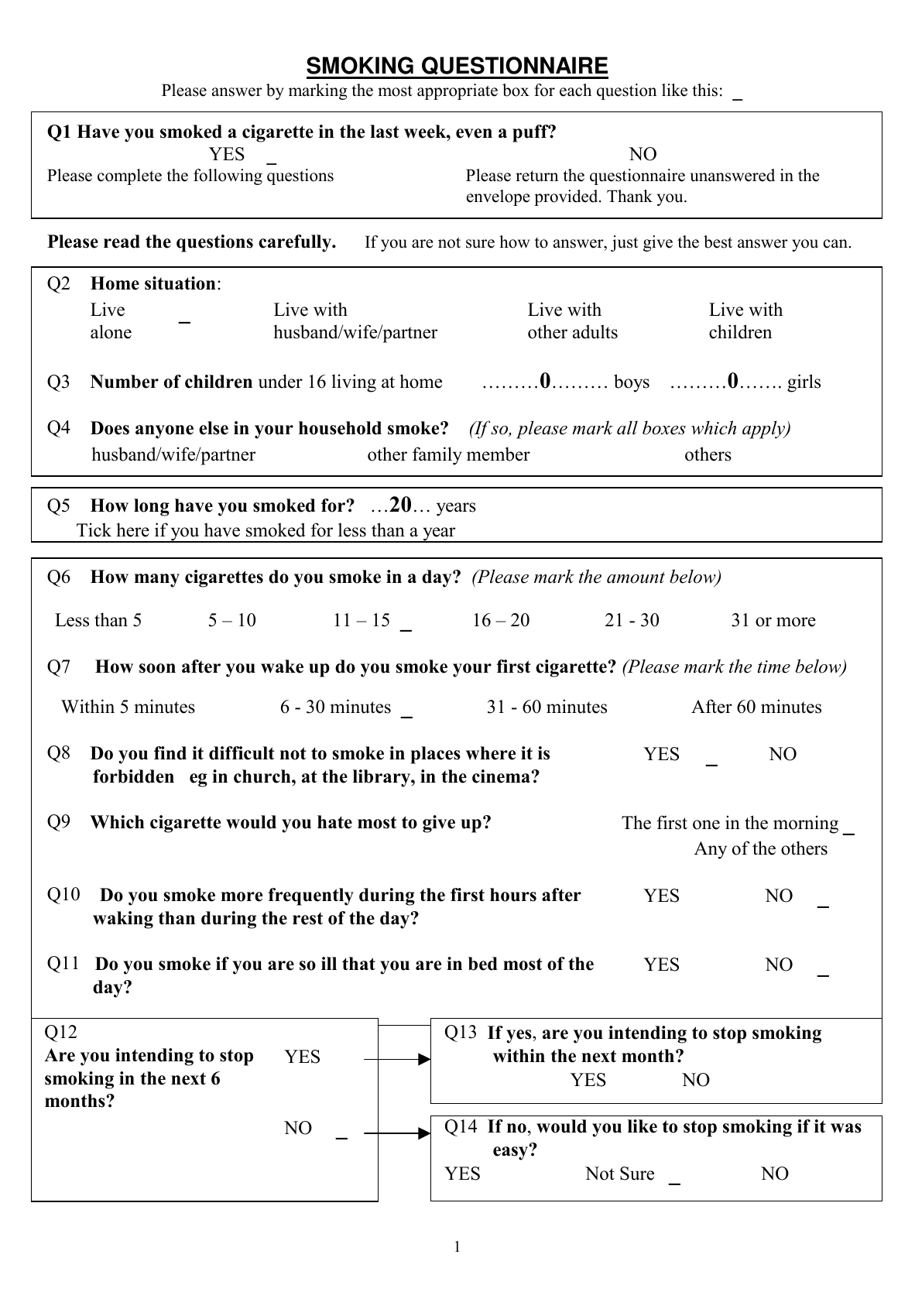}
\caption{Part of the STOP questionnaire \cite{reiterjair03}}
\label{fig:stopquestionnaire}       
\lineacross
\end{figure}

The STOP\index{STOP} system took as input a questionnaire about smoking habits; an extract is shown in Figure~\ref{fig:stopquestionnaire}.  From this information, STOP generated a 4-page leaflet about stopping smoking (which included graphics as well as text); an extract is shown in Figure~\ref{fig:stopleaflet}.  Because this was done in the late 1990s and the researchers wanted to reach a wide variety of smokers, including those who did not have internet access, the experiment was paper-based.  Subjects filled out the questionnaires on paper and posted them; researchers scanned the questionnaires, ran the software, printed the leaflets, and posted them back to the subjects.

\begin{figure}
\lineacross\\
\textbf{Dear Ms Cameron}
\blankline\\
Thank you for taking the trouble to return the smoking questionnaire that we sent you.  It appears from your answers that although you're not planning to stop smoking in the near future, you would like to stop if it was easy.  You think it would be difficult to stop because \emph{smoking helps you cope with stress}, \emph{it is something to do when you are bored}, and \emph{smoking stops you putting on weight}.  However, you have reasons to be confident of success if you did try to stop, and there are ways of coping with the difficulties.\\
\caption{First paragraph of an example STOP leaflet \cite{reiterjair03}}
\label{fig:stopleaflet}    
\lineacross   
\end{figure}

To evaluate STOP \cite{Lennox1396}, the researchers recruited 2553 subjects in the Aberdeen area who smoked, asked them to fill out the questionnaire, and then sent each of them either (A) a STOP leaflet, (B) a default leaflet which was not personalised based on their questionnaire, or (C) a simple \scare{thank you for being in our study} letter.  The reason for the different responses was to check if people who got STOP leaflets were more likely to stop smoking than people who got something else; these were \emph{control groups}\index{control groups} or \emph{baselines}\index{baselines}.  Of course some people are going to stop smoking regardless of the NLG system, so its not very useful to just measure how many STOP recipients managed to stop smoking.  Its much better to be able to compare smoking-cessation rates in STOP recipients against smoking cessation rates in people who did not get STOP leaflets.

Anyways, 6 months after sending out the leaflets and thank-you letters, smokers were contacted and asked if they had stopped smoking.  If they said they had stopped, they were asked to provide a saliva sample, which was tested for nicotine residues (this is necessary because sometimes people are not truthful about smoking cessation).  Subjects who did not respond or who did not provide saliva samples were assumed to still be smoking.

This gave data on how many people in each of the groups (STOP leaflet, default leaflet, and thank-you letter) had managed to stop smoking:
\begin{itemize}
\item 3.5\% of people who got STOP leaflets managed to stop smoking.
\item 4.4\% of people who got default leaflets managed to stop smoking.
\item 2.6\% of people who just got thank-you letters managed to stop smoking.
\end{itemize}
In other words, the STOP leaflets seemed to be \emph{less} effective than the non-personalised default leaflets!  This difference was not statistically significant\index{statistical!significance} (Section~\ref{sec:statisticalsig}), but still the evaluation provided no evidence that the STOP leaflets were worthwhile; it was a \emph{negative result}.  Negative results\index{negative results} are disappointing, but it is important that they are published, not least because they warn other researchers that some research directions may not be productive.

\personal{We published STOP's negative findings in both a medical journal \cite{Lennox1396} and an AI journal \cite{REITER200341}.  Medical journals routinely published negative results, but they were almost unheard of in the  AI literature at the time (early 2000s); indeed I had several discussions with the editor about how to best present a negative result to the AI community.  Thankfully it is more common (although still unusual) for negative AI results to be published in 2024.}

\section{Fundamentals}
\label{Evaluation:Fundamentals}

\subsection{Stakeholder perspective}\label{sec:stakeholders}
Developers must keep in mind that different stakeholders\index{stakeholders} care about different things (Section~\ref{sec:reqstakeholders}).  This means that an evaluation which is acceptable to one stakeholder may not be useful to a different stakeholder.

For example, the Babytalk\index{Babytalk} project (Section~\ref{sec:babytalk}) was a collaboration between computer scientists, doctors, psychologists, and a software company (not Arria)\footnote{Arria was founded several years after the Babytalk project started.} to build NLG systems which summarised data from a baby's electronic patient record.  Different collaborators cared about different things from an evaluation perspective \cite{reiter-belz-2009-investigation}:
\begin{itemize}
\item \emph{Doctors\index{doctors} and medical researchers} wanted to know if Babytalk was medically effective.  They wanted to evaluate whether using the system led to better patient outcomes (or reduced clinical workload).
\item \emph{Psychologists\index{psychologists}} were interested in the effectiveness of textual vs graphical\index{graphics} presentation of information (Section~\ref{sec:textgraphics}).  They wanted to evaluate how presentation medium impacted decision quality\index{decision support}.
\item \emph{Software house} wanted to know if a profitable product could be created using Babytalk technology.  They wanted to assess costs, (commercial) benefits, and risks (Section~\ref{sec:commercialevaluation}).
\item \emph{Computer scientists} wanted to understand effectiveness of algorithms and models.  Whereas other stakeholders just cared about the system as a whole, CS researchers wanted to understand how well components worked.
\item \emph{Parents} (who use the Babytalk system for families) wanted reports which were accurate and easy to read, and which told them how well their babies were doing.
\end{itemize}
It was impossible to use a single evaluation to assess all of the above stakeholder concerns.  The Babytalk team therefore did a number of evaluations (not all of which were published), including \cite{PORTET2009789,HUNTER2012157,mahamood-reiter-2011-generating,moncurhci14}, in order to assess different aspects of the Babytalk systems.

While it remains the norm in academic research to do one evaluation of a system, multiple evaluations may make sense if the system has different types of stakeholders.

Understanding what different stakeholders care about is part of requirements analysis (Chapter~\ref{requirements}), and as such this issue should be explored \emph{before} an evaluation is designed, not after it is carried out.

\subsection{Hypothesis Testing}\index{hypothesis testing}
Evaluation, at least in the academic world, is usually a type of \emph{scientific hypothesis testing}.  Hypothesis testing is an integral part of the scientific method.  Essentially scientists propose a hypothesis, such as \scare{People who smoke are more likely to get lung cancer} (medicine), \scare{Nothing can travel faster than the speed of light} (physics), or \scare{the moon is made of green cheese} (space science).  Scientists then create experiments to test hypotheses; for example conduct a spectrographic analysis of the moon (or indeed ask astronauts to collect samples), or monitor a large cohort of people and compute statistically whether smokers are more likely to get lung cancer.   In doing so, scientists discover that some hypotheses seem to be true (e.g., smokers are more likely to get lung cancer), while others are clearly false (e.g., the moon is not made of green cheese).

The type of hypothesis being tested of course depends on the field (e.g., physics vs medicine), as does the way hypotheses are tested (e.g., collecting samples vs statistical cohort analysis).  However, the notion of experimentally testing hypotheses is fundamental to science, and indeed is what has made the modern scientific method so successful.

From this perspective, the STOP\index{STOP} evaluation discussed in Section~\ref{sec:stopsmoking} tested the hypothesis that \scare{smokers who receive STOP leaflets are more likely to stop smoking than smokers who receive a default letter}.  The evaluation showed that this hypothesis was probably not true.

In NLG, most hypotheses are based on comparing the output quality between texts produced by different systems. A typical hypothesis states that texts produced by system A are better than text produced by system B, according to quality criteria\index{quality criteria} X, Y, and Z.  For example, if accuracy is an important quality criteria in the target NLG use case, researchers can hypothesise that texts produced by system A contain fewer incorrect statements (hallucinations) than texts produced by system B.  In the STOP example, the main hypothesis was that STOP texts were better according to the \emph{utility} quality criteria (Section~\ref{req-utility}).

There are other types of hypotheses which do not involve quality criteria, such as hypotheses based on cognitive plausibility\index{cognitive science} (eg, hypothesising that an NLG system will make similar mistakes to a human).  These are important for some types of research, but I will not discuss these here.

\subsection{Statistical hypothesis testing}\label{sec:statisticalsig}\index{statistical!hypothesis testing}
Most NLG hypotheses are \emph{statistical}.  For instance, using an example from machine translation\index{machine translation}, lets say we hypothesise that Google Translate produces better English translations of Chinese news articles than Bing Translate, as judged by expert translators on the basis of accuracy and fluency.  We do not expect that Google's translation will be better than Bing's on every possible Chinese news article!  Instead, we usually hypothesise that when the translators compare Google's translation of an article to Bing's translation, in most cases they will prefer Google's translation.

But small difference may just be due to random chance.  For example,  suppose we translate 1000 random Chinese news articles into English using Google and Bing, and our judges say that the Google translation is better in 510 cases and the Bing translation is better in 490 cases.   Is this difference meaningful, or could it just be random chance?

In science, we answer such questions using \emph{statistical hypothesis testing}
\cite{lakensstats}.  In order to do this, we first frame a \emph{null hypothesis}\index{null hypothesis}, for example that on average there is no difference between the quality of Google and Bing translations.  Then we use statistical tests\index{statistical!tests} to measure how likely the observed outcome (Google texts are better in 510 out of 1000 cases) is if the null hypothesis is true (there is no difference on average between Google and Bing).  In this case, a \emph{binomial test}\index{binomial test} tells us that if Google and Bing texts have the same quality on average, then:
\begin{itemize}
\item There is a 2\% chance that Google will be better on \emph{exactly} 510 cases
\item There is a 27\% chance that Google will be better on \emph{at least} 510 cases. This is called the \emph{one-tailed p value}\index{p value!one-tailed}.
\item There is a 54\% chance that either Google will be better than Bing on at least 510 cases, or Bing will be better than Google on at least 510 cases. This is called the \emph{two-tailed p value}\index{p value!two-tailed}.
\end{itemize}
In NLP, we normally consider that experimental results support a hypothesis if the two-tailed p-value is less than 5\%.  This is not true in the above example, so we could not conclude that Google was better than Bing if its translation was better in 510 cases (two-tailed p-value is 54\%).

However, if our experiment showed that the Google translation was better in 540 cases (and Bing's translation was better in 460 cases), then the two-tailed p-value for the binomial test would be 1.2\%.  Since this is less than 5\%, we could conclude from this experimental result that Google was probably better than Bing at translating news articles from Chinese to English.

In the STOP\index{STOP} example (Section~\ref{sec:stopsmoking}), a \emph{chi-square test}\index{chi-square test} was used to assess the difference between the smoking cessation rate in the STOP group (30 smokers quit out of 857 total; 3.5\% quit rate) and in the default-leaflet group (37 smokers quit out of 846 total; 4.4\% quit rate).  This gave a two-tailed p-value of 35\%.  Since this is much higher than 5\%, this means that the experiment is inconclusive.  Its possible that the default leaflets are more effective than STOP, but its also possible that they are equally effective. Indeed, its even possible that the STOP leaflets are in fact better, but this was obscured by bad luck; perhaps smokers who happened to get STOP letters were committed smokers who were less likely to quit than smokers who got the default leaflets.

Dror et al \cite{dror-etal-2018-hitchhikers} discuss common statistical tests in NLP research.  Note that the 5\% threshold is just a convention, and different areas of science use different thresholds.  Psychology and medicine also generally use a 5\% threshold, but particle physics uses a threshold of 0.00003\%.

\subsection{Experimental design, execution, reporting, and follow-up}
When we do an experiment (including evaluating an NLG system), we need to ensure that it is well designed, executed, reported, and followed up.

\subsubsection{Experimental design}\label{eval-gen-exp-design}\index{experimental design}
Experiments must be well designed.  For example, if we are comparing Google Translate and Bing Translate as described above, then the experimental design would include:
\begin{itemize}
\item \emph{Research questions:}\index{research questions} What research questions and hypotheses is the experiment intended to address; for example, as mentioned above, we could hypothesise that expert translators prefer Google over Bing when translating Chinese news articles to English.
\item \emph{Study type:}\index{study type}  What type of study will we do; for example the above experiment is a human evaluation based on rankings (Section~\ref{sec:humanevalrankings}).
\item \emph{Subjects:}\index{subjects} If our experiment uses people, how are they chosen, and what criteria must they satisfy; for example, we may decide to use expert translators recruited from a freelancing platform such as Upwork.
\item \emph{Material:}\index{material} What scenarios are evaluated; in other words, what are the 1000 texts being translated, and how are they chosen?  For example, we may decide to take 1000 random articles from the Xinhua news agency.
\item \emph{Procedure:}\index{procedure} What do the subjects do?   For example, what user interface do they use to express their preference?
\item \emph{Analysis:}\index{analysis} How are results analysed?  For example, we can use a binomial test (as described above) to compute statistical significance\index{statistical!significance}.
\end{itemize}
Of course a detailed experimental design would include much more information than above!  We will discuss experimental designs for specific types of evaluations in more detail in Sections~\ref{sec:humanevaldesign} and \ref{sec:metricevaldesign}.

A final point is that it is good practice in many fields of science to \emph{pre-register}\index{pre-registration} experiments, that is to enter experimental design details in a pre-registration website before carrying out the experiment.  This is common practice in medicine and psychology, but is less common in AI and NLG \cite{van-miltenburg-etal-2021-preregistering}.

\subsubsection{Experimental execution}\label{eval:experimental-execution}\index{experimental execution}
Experiments must also be carefully executed.  Unfortunately, many NLG experiments are not well executed, because of problems such as the following \cite{thomson-cl24}:
\begin{itemize}
\item \emph{Code bugs:}\index{bugs!code} Experiments can be invalid because of bugs in the code used to run the experiment.  For example, a subject is shown an input text X and \scare{translations} from Google and Bing, but because of a code bug the Bing translation is of a different input text Y.
\item \emph{Reporting errors:}\index{errors!reporting} Sometimes the data reported in the paper is different from the actual experimental results.
 For example, the paper states that the Google text was preferred in 540 cases, but the experimental data shows that it was preferred in 510 cases.
\item \emph{User interface issues:}\index{user interface} Experiments with human subjects can produce misleading results if a confusing user-interface is used.  For example, if subjects are asked to use radio buttons to show whether they preferred the Google text or the Bing text, but the buttons are not clearly labelled so it is not clear which button indicates a preference for the Google text.
\item \emph{Analysis bugs:}\index{bugs!analysis} Statistical analyses can also be misleading.  For example, if 1500 subjects did the experiment but the analysis only looked at the results from 1000 subjects and ignored the remaining 500.
\end{itemize}
Thomson et al  \cite{thomson-cl24} give concrete examples of the above issues in published NLP research papers.

\subsubsection{Experimental follow-up}\index{followup}
Even after an experiment has been written up and published, researchers must respond to questions and otherwise support their research.  This includes
\begin{itemize}
\item \emph{Respond to questions:}  If readers are interested in an experimental evaluation, they are likely to have questions and want further details; the experimenters should respond to such requests.  Unfortunately, many researchers do not do this. The ReproHum\index{ReproHum} project \cite{belz-etal-2023-missing} contacted 116 authors of academic NLP papers (all published in good venues), and asked the authors for additional information about their project.  Only 45 (39\%) of authors responded in any fashion, and only 15 (13\%) provided the requested information.
\item \emph{Correct problems:} Sometimes problems in experiments are discovered after results are published or otherwise released.  If this happens, the researchers should publicly acknowledge the problems, and fix their papers.  Unfortunately, most NLP researchers do not do this.  For example, in the ten-year period 2013-2022, the TACL (\emph{Transactions of the ACL})\index{Transactions of the ACL} journal only had \emph{one} formal correction\index{correction} to experimental results and findings \cite{thomson-cl24}. I personally know of more TACL papers in this period which had mistakes and should have been corrected.
\end{itemize}

\subsection{Research questions}\label{sec:researchquestions}\index{research questions}
Evaluations are intended to test research questions and hypotheses, which in NLG often take the form that system X is better than system Y under quality criteria\index{quality criteria} Z; for example Google Translate is better than Bing Translate under the criteria of accuracy of their generated translations.   The research question must be important scientifically (for basic research) or to stakeholders\index{stakeholders} (for applications); there is no point in testing a research question which is of no interest to anyone.

From this perspective, it is essential to evaluate quality criteria that people care about; in applied NLG this means criteria that are important to stakeholders (Section~\ref{sec:stakeholders}).  Sometimes the things that stakeholders are about are difficult to measure.  For example in medicine doctors usually want to measure impact\index{impact} on clinical outcomes (does a system help patients); but measuring this experimentally can be a lot of work (Section~\ref{sec:stopsmoking}).  But regardless, if this is what people care about, it should be evaluated.  There is a temptation for researchers to evaluate quality criteria which are relatively easy to measure, such as fluency (Section~\ref{sec:fluency}), but this is of limited value if it is not what stakeholders care about.

Another point is that if the goal is to demonstrate that a new NLG system is better than existing NLG systems (sometimes called \emph{baselines}\index{baselines}), it is essential to compare the new system to the best (state-of-the-art\index{state-of-art}) existing NLG system(s); comparing the new system to obsolete systems is not valuable or interesting. For example, if the goal is to show that a new language model can generate better texts in some contexts than existing models, then the new model should be compared to the best existing models, not to obsolete models such as GPT2.

\personal{The above points may seem obvious, but I see many research papers which evaluate unimportant but easy-to-measure quality criteria, and/or compare a new system against obsolete baselines.  Such experiments are easy to run and produce impressive results, but they are not useful either scientifically or practically.}

\subsection{Replication}\label{sec:replication}\index{replication}
Scientific experiments need to be \emph{replicable}.  In other words, other researchers should be able to repeat an experiment and get similar results.  This is fundamental to the scientific method; experiments which cannot be replicated have limited scientific utility and validity.

Much has been written about replicability in different fields of science \cite{WikipediaReplication}.  In NLG, replication requires very detailed information about the system and experiment, including how data is preprocessed, the exact libraries and options used for automatic evaluation, the exact user interface\index{user interface} used for human evaluation, etc. Seemingly small differences can lead to big differences in outcomes \cite{belz-etal-2021-systematic}.

Various data sheets\index{data sheet} have been proposed to help gather this information \cite{gebru2021datasheets,shimorina-belz-2022-human}.  Unfortunately, at least in my experience it is very rare for all the necessary details to be provided in papers and associated material such as data sheets.  Usually researchers who want to repeat an experiment need to contact the original authors in order to get all the details.

It is important to keep in mind that \emph{exact} replication of experimental results is unlikely, because of experimental noise.  Certainly replicated experiment with human subjects rarely give exactly the same results, even if they use the same subjects\index{subjects}, because people do not respond 100\% consistently.  For example, if a subject is hungry and tired, she may give lower ratings to texts than if she is well fed and rested.  Another source of noise is random seeds\index{random seeds}; any system whose output is influenced by a random seed (which includes most neural NLG systems) may generate different outputs on different runs.

So we expect to see some difference in the exact numbers if an experiment is replicated.  However, replications should not lead to different statistically-significant\index{statistical!significance} outcomes at the hypothesis level.  For example, if the original experiment claimed that system A produced more accurate texts than system B, with the difference being statistically significant (Section~\ref{sec:statisticalsig}), then it would be very worrying if a replication found that system A produced \emph{less} accurate texts than system B, with the difference being statistically significant.  Such a finding would strongly suggest that that there are serious flaws in how the experiment was designed, executed, or reported.

Unfortunately, some published NLG evaluations, even in good venues, do not seem to be replicable.  For example, as part of a shared task on replication \cite{repronlp2023}, several labs tried to reproduce published human evaluations.  The worst result was for a paper which asked crowdworkers to count the number of content errors in a text \cite{puduppully-lapata-2021-data}; this paper re-used an evaluation methodology from an earlier paper \cite{wiseman-etal-2017-challenges}.  Table~\ref{tab:badreplication} shows one result (number of errors in a 4-sentence extract from a corpus text) from the original paper, the reproduction studies, and a study which measured the same thing (number of errors) using a different and better methodology.  The large variation in results (the highest number is 25 times larger than the lowest) raises serious concerns about the validity of the experiment, and suggests that we should not evaluate systems by asking crowdworkers to count content errors.

\begin{table}
\begin{tabular}{|l|l|r|}
\hline
\textbf{Paper} & \textbf{Note}  & \textbf{Mean number errors} \\
Original paper \cite{puduppully-lapata-2021-data} &	Original & 0.07 \\
Gonzalez-Corbelle et al \cite{gonzalez2023}	& Replication	 & 0.66\\
Watson and Gkatzia \cite{watson2023} & 	Replication &1.52\\
Watson and Gkatzia \cite{watson2023} & 	Replication with different subjects	& 0.06\\
\hline
Thomson et al \cite{thomson-csl23} & More rigorous measurement & 0.50 \\
\hline
\end{tabular}
\caption{Mean number of content errors in four-sentence extracts from corpus (human-written) texts, as measured by asking crowdworkers to count errors.   Result from original paper and two replications (one of which repeated the study using in-house subjects instead of crowdworkers). Thomson et al \cite{thomson-csl23} (Figure~\ref{fig:ThomsonAnnotation}) measured the errors using a more rigorous annotation methodology; they reported 1.58 errors on average in a human corpus text, which is approximately 0.5 errors per four-sentence extract.}
\label{tab:badreplication}
\end{table}

\personal{One of the frustrations I have with the paper being replicated \cite{puduppully-lapata-2021-data} is its usage of an old evaluation methodology \cite{wiseman-etal-2017-challenges}, when newer and better evaluation techniques were available \cite{thomson-reiter-2020-gold}.  I have seen many other cases where papers use the latest technology to \emph{build} an NLG system, but outdated techniques to \emph{evaluate} the system.  Sometimes this happens because authors want to enter their system into public \scare{leaderboards} which show the best-performing system at a specific task as assessed by a specific evaluation technique.  Leaderboards which are based on obsolete evaluation techniques are not helpful to scientific progress.}

\subsection{Ecological validity: Artificial vs real-world context}\label{sec:ecologicalvalidity}
We can do an evaluation in either artificial or real contexts; this is sometimes called \emph{ecological validity}\index{ecological validity}.  For example, when comparing translated texts, we can either get translators to assess them in a generic way, or we can ask people who genuinely need the information to use the translations for real and then assess their effectiveness.

For instance, as mentioned in Section~\ref{sec:stakeholders}, different stakeholders\index{stakeholders} in Babytalk\index{Babytalk} had different perspectives on evaluation.  The psychologists\index{psychologists} wanted to do evaluations in controlled artificial settings, because this reduced the number of confounding factors when testing the impact of different media on decision making; they believed this was more important than evaluating the system in real-world usage.  However, the doctors\index{doctors} believed that evaluating systems in real clinical usage was essential, because this gave a much better understanding of utility\index{utility} and potential impact\index{impact}; they did not trust evaluations in artificial settings. The project ended up doing both types of evaluations \cite{PORTET2009789,HUNTER2012157}.

Ecological validity is especially important if the goal is to assess real-world impact (Section~\ref{sec:impactevaluation}).  The real-world is a messy place, and we cannot assess real-world impact if we ignore this messiness.

\personal{Evaluations in real-world contexts are still rare in NLG at the time of writing, which is a shame; I hope they become more popular.}

\subsection{Test data\index{test data}: representative, different from training data\index{training data}}\label{sec:trainingtestdata}
When evaluating an NLG system, researchers need to decide which test data (scenarios) the system will be tested on; from an experimental design perspective (Secttion~\ref{eval-gen-exp-design}), this is part of choosing material\index{material}.   Test data should be real data or in some cases a mix of real and synthetic data; it should not just be synthetic data\index{synthetic data} (Section~\ref{sec:syntheticdata}).  It should also be representative of real usage.

For example, a system which summarises doctor-patient consultations should be tested in a wide variety of such consultations; it should  not be tested on pharmacist-patient consultations.  Obtaining representative test data\index{representativeness} can be challenging, especially in use cases (such as medicine) with strong data protection\index{data protection} constraints, but it is essential for meaningful evaluation.

For NLG systems built using machine learning, it is also essential that they not be tested on their training data; this is a fundamental principle of machine learning (systems which are tested on training data can get perfect scores on evaluations simply by memorising their training data \cite{schaeffer2023pretraining}).  For example, if we are evaluating texts produced by Google and Bing Translate, we should ensure that these texts are not part of the training data used to build these systems.  This is called \emph{data contamination}\index{data contamination} \cite{sainz-etal-2023-nlp}.

Unfortunately, the emergence of large language models\index{large language models} which are trained on the Internet (Section~\ref{sec:promptedmodels}) has made it much harder to guarantee that test data was not present in the system's training data.  Prompted language models can also in some case memorise or learn from test data that is provided in prompts\index{prompt} as examples (Section~\ref{sec:fewshot}) or when models are used to evaluate texts (Section~\ref{sec:gemba}) \cite{balloccu-etal-2024-leak}.

I have seen many evaluations of LLMs where it seemed very likely that the test data was part of the model's training data or prompts, and also many evaluations where this issue was unclear. One problem is that commercial LLMs generally do not reveal their training data, and also are constantly being updated.  Open-source LLMs are better from this perspective, and usually clearly state what data they were trained on.  For this reason, it is usually easier to evaluate open-source LLMs than proprietary commercial ones.

\section{Human Evaluation}\label{sec:humanevaluation}
In general, NLG systems can be evaluated by people (human evaluation), by algorithms (metrics), or by assessing real-world impact.
If an  impact evaluation\index{impact!evaluation}  is not possible, then a careful and well-designed human evaluation is usually the best way to meaningfully evaluate an NLG system. However a good metric evaluation is better than a poorly designed or executed human evaluation.

In this section we look at different types of human evaluation; discuss how such evaluations should be designed, executed, and evaluated; and give some examples.

\subsection{Types of Human Evaluation}\label{eval:humanevaltypes}
There are many types of human evaluations, with more being introduced every year.  But in rough terms, we can distinguish between evaluations based on (A) subjective ratings or rankings; (B) error annotation; and (C) task performance (such as decision making).

\subsubsection{Human evaluation based on ratings or rankings}\label{sec:humanevalrankings}
The most common form of human evaluation in NLP  asks human subjects to rate\index{rating} texts and/or rank\index{ranking} a set of texts, based on one or more quality criteria (Section~\ref{req-criteria}).  This evaluation is based on the subject's subjective opinion.

\begin{figure}
\lineacross//
\includegraphics[scale  = 0.75, trim={2cm, 10cm, 2cm, 3cm},clip]{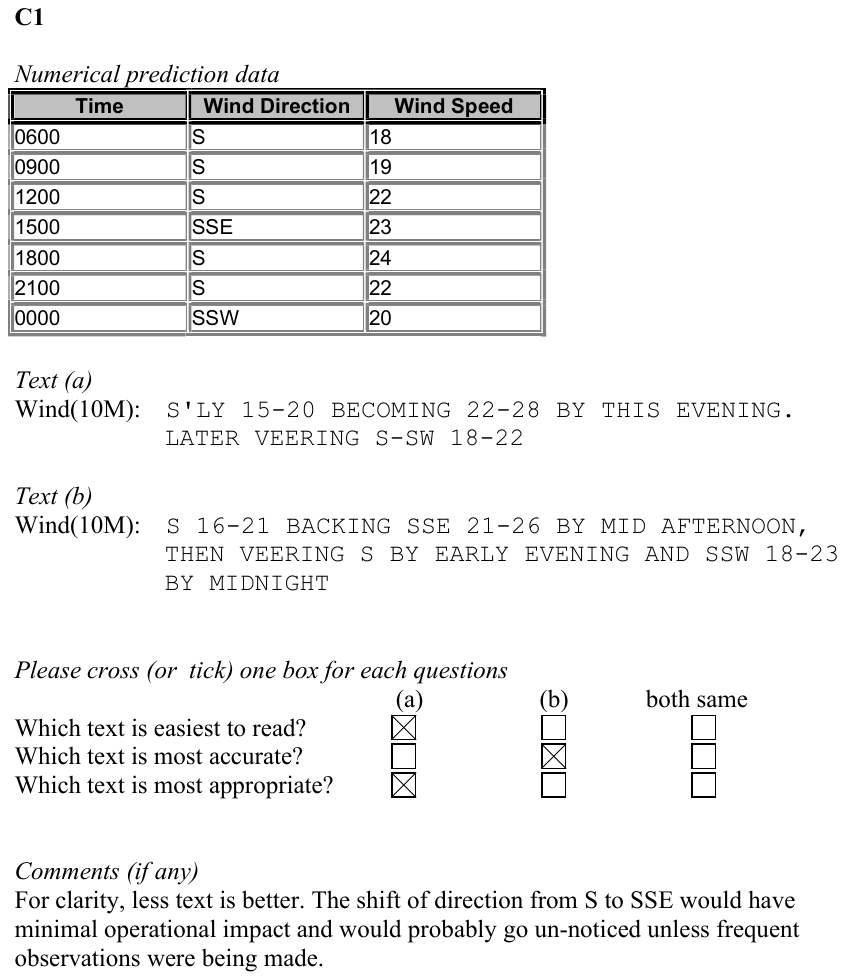}
\caption{Example of ranking experiment in SumTime}
\label{fig:SumTimeRanking}    
\lineacross   
\end{figure}

The conceptually simplest approach is to give subjects two or more texts, and ask them to rank them according to the chosen quality criteria.  Figure~\ref{fig:SumTimeRanking} shows an example of this, which was used in the evaluation of the SumTime\index{SumTime} weather forecast generator \cite{REITER2005137}.  SumTime generated marine weather forecasts for workers in the offshore oil industry, and this evaluation asked users to rank two texts that described the predicted wind speed (at 10 meters altitude).   The numeric wind prediction (produced by a supercomputer running atmosphere simulations) was given in a table; this is in the input to the NLG system.  In this example, text (a) was written by a human forecaster, indeed it was extracted from an actual weather forecast for the offshore oil industry.  Text (b) was produced by the SumTime NLG system.  The user was asked to say which text was preferred based on three different quality criteria (Section~\ref{req-criteria}): easiest to read, most accurate, and most appropriate; they could also add comments about the texts.

Many evaluations of this type have been reported in the literature.  Most follow the above structure:
\begin{itemize}
\item The user is given the input to the NLG system (if appropriate).
\item The user is given two or more texts produced from the input (sometimes including a human-written text, as was done in SumTime); the user is not told the source of the texts
\item The user is asked to rank texts on one or more quality criteria.  Sometimes a \scare{same} or \scare{no difference} option is given, sometimes the user is forced to make a choice.
\item Sometimes the user can optionally write free-text comments about the texts.
\end{itemize}
It is good practice to randomise the order of the texts.  For example, if the user is shown one human text and one NLG text, then  choice (a) should sometimes be the human text and sometimes be the NLG text.

Of course different user interfaces can be used; for example users can be asked to drag texts into a ranked order, instead of using checkboxes.  The SumTime experiment, incidentally, was done on paper, because (at the time it was run) subjects on oil rigs and supply boats did not always have easy access to computers or the Internet.

Human subjects find it difficult to rank large numbers of texts.  Therefore if the experiment requires comparing more than 5 texts, the usual practice is to ask different subjects to rank different subsets of the entire collection of texts, and combine these using an algorithm such as TrueSkill\index{TrueSkill} \cite{herbrich2006trueskill,sakaguchi-etal-2014-efficient}.

\begin{figure}
\lineacross//
\includegraphics[scale  = 0.65,trim={0 0 2cm 2.75cm},clip]{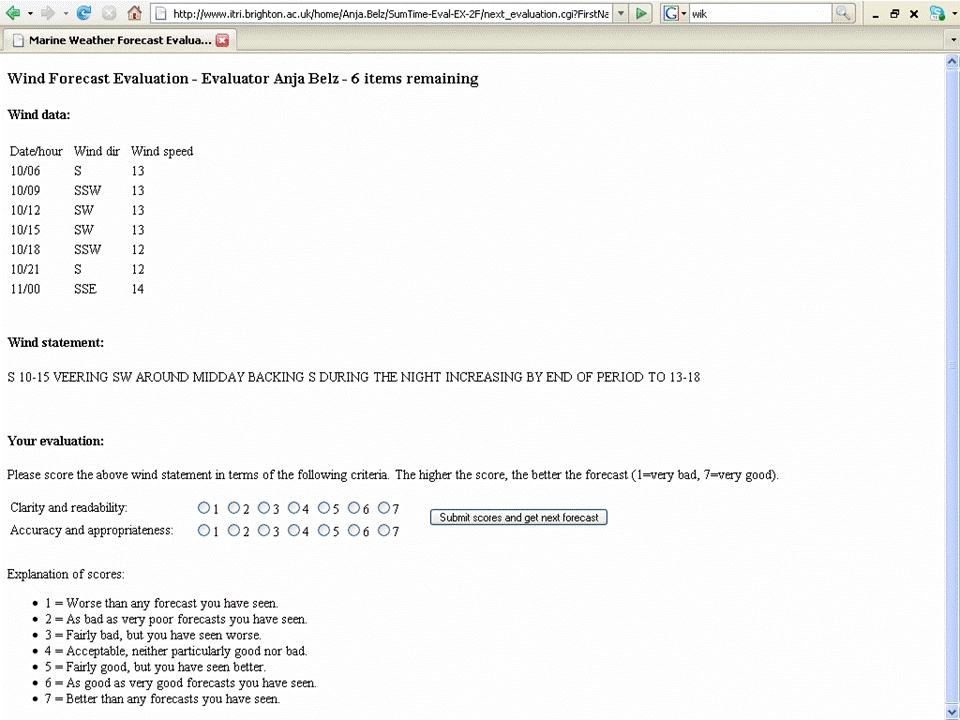}
\caption{Example of rating experiment for SumTime, used in \cite{reiter-belz-2009-investigation}}
\label{fig:SumTimeRating}
\lineacross
\end{figure}

Another popular approach is to ask subjects to \emph{rate} texts.  An example is shown in Figure~\ref{fig:SumTimeRating}, from \cite{reiter-belz-2009-investigation}.  This is in the same domain as Figure~\ref{fig:SumTimeRanking}, generation of marine weather forecasts for the offshore oil industry, but this experiment (which was online) asked users to rate the quality of the text on two quality criteria (clarity/readability, accuracy/appropriateness), it does not ask users to explicitly compare texts.  Texts produced by different systems are rated separately.

Likert scales\footnote{\url{https://en.wikipedia.org/wiki/Likert\_scale}}\index{Likert scale} (usually 5 or 7 points) are often used for rating texts.  Another option is \emph{magnitude estimation}\index{magnitude estimation}, where users are asked to position a slider on a best-worst scale, instead of choosing a discrete option.

\subsubsection{Evaluations based on annotations\index{annotation}}\label{sec:humanevalannotations}
Human evaluations can also be done by asking subjects to \emph{annotate} errors and other problems in a text.   Annotators often have some domain expertise\index{domain!experts}, and may be asked to include error type and/or severity\index{errors!severity} in their annotations.   Once a text has been annotated, an overall score can be computed based on the number of errors, possibly weighted by type and severity.  The error distribution across types and severity, and indeed the individual error annotations, can provide useful guidance for developers who wish to improve a system.

\begin{figure}
\newcommand{\err}[1]{{\color{red} \ul{#1}}}
\lineacross{}
The Memphis Grizzlies (5-\err{2}) defeated the Phoenix Suns (3 - 2) \err{Monday} 102-91 at the \err{Talking Stick Resort Arena} in Phoenix. The Grizzlies had a \err{strong} first half where they \err{out-scored} the Suns \err{59}-\err{42}. Marc Gasol scored 18 points, \err{leading} the Grizzlies.  \err{Isaiah Thomas added} 15 points, he is \err{averaging 19 points on the season so far}.  The Suns' next game will be \err{on the road} against the \err{Boston Celtics} on Friday.

\vspace{5mm}
List of errors:
\begin{itemize}
    \item \err{2}: incorrect number, should be 0.
    \item \err{Monday}: incorrect named entity, should be Wednesday.
    \item \err{Talking Stick Resort Arena}: incorrect named entity, should be US Airways Center.
    \item \err{strong}: incorrect word, the Grizzlies did not do well in the first half.
    \item \err{out-scored}: incorrect word, the Suns had a higher score in first half.
    \item \err{59}: incorrect number, should be 46.
    \item \err{42}: incorrect number, should be 52 .
    \item \err{leading}: incorrect word,  Marc Gasol did not lead the Grizzlies, Mike Conley did with 24 points.
    \item \err{Isaiah Thomas added}: context error, Thomas played for the Suns, but context here implies he played for the Grizzlies and added to their score.
    \item \err{averaging 19 points in the season so far}: Not checkable.  Data sources report performance per season and per game, not performance at a particular point in a season.
    \item \err{on the road}: incorrect word, The Suns will play at home.
    \item \err{Boston Celtics}: incorrect named entity, the Suns will play the Sacramento Kings
    
\end{itemize}
\caption{Example text with error annotations from \cite{thomson-reiter-2020-gold}.  Each annotation includes an error type and a correction.  Annotators can add explanations where useful. Box score data for this game is available at
\url{https://www.basketball-reference.com/boxscores/201411050PHO.html} .}\label{fig:ThomsonAnnotation}   
\lineacross{}
\end{figure}

An example is shown in Figure~\ref{fig:ThomsonAnnotation}.  This was part of the training material given to subjects who were asked to evaluate the factual accuracy of sports stories generated by neural NLG systems \cite{thomson-csl23}.  Annotators were asked to find inaccurate statements, classify them (incorrect number, incorrect named entity, incorrect word, context error, not checkable, other), and also (if possible) correct the statement.

At the time of writing, the most popular annotation scheme is MQM\index{MQM} \cite{FreitagMQM}, which is used to annotate machine translation outputs.  It is a more complex annotation scheme than the one in Figure~\ref{fig:ThomsonAnnotation}, which annotates linguistic as well as content errors, and also assigns a severity to errors.

Annotation schemes can be domain-dependent.  For example Morarmarco et al \cite{moramarco-etal-2022-human} use an annotation scheme as one of several evaluation techniques for Note Generator\index{Note Generator} (Section~\ref{IntroConsultationSummary}), which generates summaries of doctor-patient consultations. The annotation scheme includes domain-specific categories such as \scare{Use of not universally recognised [medical] acronyms}.  Similarly Magesh et al \cite{magesh2024hallucinationfree} describe an annotation scheme for the output for AI-driven legal research tools, which includes groundedness (claims are supported by citations to relevant legal documents) as well as correctness.

Some researchers use annotations to evaluate human-written texts as well as computer-generated texts \cite{FreitagMQM,moramarco-etal-2022-human,thomson-csl23}.  This gives  information about the types of mistakes made by human authors\index{human authors} in a domain; this number is often higher than initially expected.

At the time of writing, this is a relatively new form of human evaluation in NLG, so specific annotation schemes are evolving, as are user interfaces for annotation.  Some schemes allow document-level annotations, especially for missing information, in other words important information which should have been included in the text but was not present.

\personal{I believe that evaluation-by-annotation usually gives more meaningful results than evaluation by rating or ranking, and hope to see more such evaluations in the future.}

\subsubsection{Evaluations based on task performance}\label{sec:humantaskeval}\index{task-based evaluation}
Another type of of human evaluation is to ask people to use an NLG system and assess its effect on how well they perform a task.   A simple example was discussed in Section~\ref{sec:gkatziaweather} (Figure~\ref{fig:GkatziaWeather}), where the driver of an ice cream truck decided when to work based on a weather forecast.

A more complex example is the evaluation of the Babytalk BT45\index{Babytalk!BT45} system (Section~\ref{sec:babytalk}) which generated texts that summarised patient record data, and were intended to help doctors make good clinical decisions.  This system was evaluated by running an experiment where clinicians were asked to decide on the best intervention (action), after seeing either a Babytalk NLG text summary of relevant patient data, a human-written summary of the data, or a visualisation \cite{van2010graph,PORTET2009789}.

Note that this is not an impact evaluation\index{impact!evaluation} in the sense of Section~\ref{sec:impactevaluation}, since the experiment is done in artificial context. In the BT45 experiment doctors\index{doctors} were asked to make decisions based purely on patient record data; this is not a real-world task, since when clinicians make decisions about patients in a real hospital, they have access to many other information sources (e.g., observing the patient, talking to nurses, previous interactions they have had with the patient) in addition to patient record data.
But a task-based evaluation in an artificial context nonetheless can give good insights as to how well a system works and where it needs to be improved

It is difficult to generalise about task-based evaluations, they are very different.  One fairly common type of task-based evaluation in NLG is \emph{post-edit-time}\index{post-editing!time} evaluation, where researchers ask human domain experts \index{domain!experts} to check an NLG text and fix problems so that it can be released to real users, and measure how long this checking and editing takes.  Human post-editing is fairly common (Section~\ref{sec:humanchecking}), and the amount of time required to post-edit a text is an important real-world measure of the utility of the NLG system.  However, different people have different post-editing behaviour; some just fix mistakes whereas others rewrite texts into a different style \cite{sripada-etal-2005-evaluation}.  Also post-edit time is very dependent on the user interface\index{user interface} and workflow\index{workflow}, as well as the texts being edited.  So it is a somewhat \scare{noisy} measure of text quality.  If post-edit time is used to compare two NLG systems, it is essential that the two systems be similar from a UI, workflow, and user perspective.

From a practical perspective, task-based evaluations are often more expensive and time-consuming than evaluations based on ratings ot annotations.  However, they can give very valuable insights on how texts influence and help users.

\subsection{Experimental Design}\label{sec:humanevaldesign}\index{experimental design}
Good experimental design is critical for human evaluations, as for other evaluations.  The key steps are the ones described in Section~\ref{eval-gen-exp-design}:
\begin{enumerate}
\item Choose your research hypotheses and questions.
\item Choose the overall study type.
\item Choose subjects.
\item Choose material.
\item Decide on experimental procedure.
\item Decide on analysis.
\end{enumerate}
These are described in more detail below, for human evaluations.  I will use an example of evaluating a system BasketballNLG\index{BasketballNLG} which generates summaries of basketball games. Figure~\ref{fig:basketballDesign} shows a high-level experimental design for this evaluation; the following sections describe each step in more detail.

\begin{figure}
\lineacross\\
\textbf{Research Hypothesis}
\begin{itemize}
\item Texts produced by BasketballNLG are better than texts produced by \emph{StateofArtSystem} under the quality criteria of accuracy and interestingness.
\end{itemize}

\textbf{Study Type:}
\begin{itemize}
\item Annotation-based evaluation of accuracy.
\item Ranking-based evaluation of interestingness.
\end{itemize}

\textbf{Subjects:}
\begin{itemize}
\item 50 subjects evaluate interestingness; ten of these (after additional training and screening) annotate accuracy errors.
\end{itemize}

\textbf{Material:}
\begin{itemize}
\item 20 games (scenarios), chosen at random, for interestingness evaluation.
\item 10 games, chosen at random (but different from above), for accuracy evaluation.
summaries are produced using both BasketballNLG and \emph{StateofArtSystem}.
\end{itemize}

\textbf{Procedure:}
\begin{itemize}
\item Experiment is run on the web.
\item All subjects are given detailed instructions; subjects doing error annotation must complete a training session and get a score of 80\% on a screening test.
\item UI is a simple form for interestingness ranking; subjects see summaries from both BasketballNLG and \emph{StateofArtSystem} and are asked to rank them.  An online version of Microsoft Word is used for annotation; subjects annotate just one summary for each game (ie, either the BasketballNLG summary or the \emph{StateofArtSystem} summary,  not both).
\item A Latin Square design is used to decide which summary (BasketballNLG or \emph{StateofArtSystem}) each subject sees for error annotation. For interestingness, scenarios are randomly ordered, as is the order of the texts being ranked in each scenario (ie, whether the first text is BasketballNLG or \emph{StateofArtSystem})
\item Simple attention check questions are included halfway through the experiment.
\item At the end of the experiment, subjects can (optionally) provide free-text comments on the texts and/or experiment.
\item If circumstances abort an experiment,  data collected so far is kept and subjects are asked to complete the experiment at a later date.
\end{itemize}

\textbf{Analysis:}
\begin{itemize}
\item Outliers are removed using Interquartile Range (IQR) algorithm.
\item Accuracy: Report average error counts for the systems being compared; use a t-test to calculate statistical significance.
\item Interestingness: Report preference split from the interestingness rankings; use a binomial test to calculate statistical significance.
\item Do qualitative error analysis on 5 texts with poor interestingness and 5 texts with poor accuracy.
\item Analyse free-text comments using thematic analysis and report key insights.
\end{itemize}

\caption{Experimental design for human evaluation of BasketballNLG\index{BasketballNLG}}\label{fig:basketballDesign}
\lineacross
\end{figure}

\subsubsection{Step 1: Choose research hypotheses and questions}\label{sec:humanhypotheses}\index{research questions} 
A good experiment cannot be designed without knowing the core hypotheses being tested, such as \scare{people who get STOP\index{STOP} letters are more likely to stop smoking than people who get non-tailored letters}.  In applied NLG, research questions and hypotheses are usually based on user requirements\index{requirements} (Chapter~\ref{requirements}).  They often are of the form that \scare{Texts produced by A are better than texts produced by B under quality criteria\index{quality criteria} C}, for example \scare{Texts produced by SumTim\index{SumTime}e are better than texts produced by human forecasters under the quality criterion of \emph{approriateness}}.   Hypotheses can aso be framed around task-based outcome measures, for example \scare{Doctors who see Babytalk\index{Babytalk} summaries make better clinical decisions than doctors who see visualisations}, or indeed \scare{people who get STOP letters are more likely to stop smoking than people who get non-tailored letters}.

For example, for BasketballNLG\index{BasketballNLG}, one possible set of concrete research hypotheses is that \scare{texts produced by BasketballNLG are better than texts produced by \emph{StateofArtSystem}\index{state-of-art} under the quality criteria of accuracy and interestingness}; this is two hypotheses because it includes two quality criteria.  \emph{StateofArtSystem}   should be replaced by the current best system for doing this task.

\subsubsection{Step 2: Choose the overall study type}\label{sec:humanstudytype}\index{study type}
The most common study types are the ones described above, but other types are also possible, such as measuring reading time.  The choice is determined by the quality criteria and also by pragmatic issues such as time and cost.  In rough terms:
\begin{itemize}
\item Ranking\index{ranking} and rating\index{rating} experiments are cheapest and quickest, and can be used for most quality criteria.  Opinions differ as to which is better.  I personally lean towards ranking in most cases, but other researchers prefer ratings, and indeed the WMT\index{WMT} shared task for evaluating machine translation explicitly shifted from ranking to rating in 2017 \cite{bojar-etal-2017-findings}, after experimenting with both approaches in previous years.
\item Annotation\index{evaluation} experiments are better for assessing accuracy and more generally at finding errors, but they are significantly more expensive and time-consuming than experiments based on ratings and rankings.  For example, the 2022 WMT shared task \cite{kocmi-etal-2022-findings} used the MQM\index{MQM} annotation technique in a limited way to supplement ranking-based evaluation; MQM gave
more meaningful assessments, but it was not feasible from a financial perspective to do all evaluation using MQM.
\item Task-based evaluations\index{task-based evaluation} are even more expensive and time-consuming than annotation evaluations,  but they are the best way to measure the effect of NLG systems on users, and are better at assessing utility and related quality criteria than rankings, ratings, or annotation experiments.  
\end{itemize}

For example, Figure~\ref{fig:basketballDesign} describes a strategy for evaluating BasketBallNLG which uses a ranking-based evaluation for interestingness, and an annotation evaluation for accuracy.  The expense of task-based evaluation is not justified for evaluating sports stories (which are essentially entertainment).  Annotation is better than ranking/rating for accuracy, but is not normally used for interestngness.

\subsubsection{Step 3: Choose subjects}\label{sec:humansubjects}\index{subjects}
Human evaluations require human subjects!  Some experiments can be done by any fluent speaker of the language being generated.  Researchers can recruit subjects for such experiments by asking friends, family members, colleagues, or students; they can also use crowdsourcing\index{crowdworkers} platforms such as Mechanical Turk\index{Mechanical Turk}\footnote{\url{https://www.mturk.com/}} or Prolific\index{Prolific}\footnote{\url{https://www.prolific.com/}}.  Other experiments require subjects with specific domain knowledge\index{domain!knowledge} (e.g., doctors) or attributes (e.g., they smoke); such subjects are usually explicitly recruited, sometimes from collaborating organisations. For example, the clinicians who evaluated Babytalk\index{Babytalk} (as described in Section~\ref{sec:humantaskeval}) were recruited from a specific hospital (Edinburgh Royal Infirmary) which was a collaborator in the Babytalk project.

If subjects are recruited via crowdsourcing, it is important to remember that their incentive is to do tasks as quickly as possible, since this will maximise their income. There is also a danger that crowdworkers will use chatbots to do the tasks, which makes their contribution useless for a human evaluation.  Perhaps in part for these reasons, some crowdworker-based evaluations are difficult to replicate; one example was discussed in Section~\ref{sec:replication}.

\personal{I have unfortunately seen a number of cases where experiments with crowdsourced subjects were useless.  For example the  subjects simply ticked boxes at random without actually doing the experimental tasks, because this was the fastest way to do the task and get paid.}

Sometimes the quality of crowdworker (and other) evaluations can be improved by using techniques such as the following:
\begin{itemize}
\item Adding special \emph{attention check}\index{attention check} questions, which subjects can only answer if they are doing the experiment properly.  If subjects do not answer these correctly, their data is discarded.
\item Checking for \scare{outliers}\index{outlier}, such as subjects doing an experiment incredibly quickly.  If this is detected, again data can be discarded.
\item Recruiting a pool of trusted subjects who will be repeatedly used.  For example, for the annotation experiments reported in \cite{thomson-csl23}, Thomson et al recruited 7 annotators on Mechanical Turk, and asked these annotators to do multiple tasks over a two-year period.  Since annotators liked the work (which was interesting and also well-paid by Mechanical Turk standards), they were motivated to do a good job, in order to keep on getting more such work in the future.
\end{itemize}

A general principle which seems obvious but is sometimes ignored is that researchers should treat subjects well; respect them, answer their questions, pay them (if appropriate), etc \cite{silberman18}.  Besides being ethically\index{ethics} correct, a well-treated subject is more likely to provide high-quality experimental data.

Another general point is that in some cases it is important that subjects are representative\index{representativeness} of a user group.  For example, an evaluation of the utility of computer-generated weather forecasts\index{weather forecasts} should be done with a representative sample of forecast users, and not just with undergraduate students (whose use of forecasts is probably not representative).  In all cases, researchers should report the demographic\index{demographics} mix (age, gender, etc) of their subjects when they write up their experiments.  These can have an impact on what subjects do \cite{gkatzia-etal-2016-natural}, and indeed it may make sense to test whether males and females (for example) evaluate texts differently.

A key question is how many subjects should be recruited.  In theory, this can be answered using a \emph{statistical power calculation}, which computes the number of subjects based on expected effect size (expected difference in measured quality criteria between systems), the statistical test used in analysis, and other such parameters.  However, it is rare to see power calculations used when designing NLG experiments, partially because the necessary information (e.g., expected effect size) is often not available.

In very crude terms, ranking/rating\index{rating}\index{ranking} and task-based evaluations\index{task-based evaluation} usually have at least 20 subjects, especially if subjects do not need specialist expertise. Annotation\index{annotation}-based evaluations, and evaluations with domain experts\index{domain!experts}, can be done with fewer subjects, sometimes as few as 3 or 4.
These are \emph{minimum} numbers; it is much better to have 50 or 100 subjects in a ranking/rating evaluation.

For example, Figure~\ref{fig:basketballDesign} describes a strategy for evaluating BasketBallNLG\index{BasketballNLG} which
uses 50 subjects to rank basketball texts on the basis of interestingness.  From this group, 10 people are asked to annotate texts for accuracy; researchers look for subjects who have good knowledge of basketball, and ask them to pass a screening test.  The test could ask subjects to annotate a summary with known errors, and only accept people whose annotation is 80\% correct (precision and recall compared to the correct annotation).

\subsubsection{Step 4: Choose material}\label{sec:humanmaterial}\index{material}
In most human evaluations of NLG, subjects are shown texts generated in specific \emph{scenarios}\index{scenarios}.  For example, in the Babytalk\index{Babytalk} evaluation doctors evaluated texts generated from patient record data about a specific baby over a specific time period; a scenario here might be \emph{baby JOHNSMITH23 at 1600-1645 on 3 September 2004}.  In the SumTime\index{SumTime} evaluation weather forecast users evaluated forecasts for specific locations on specific days; an example scenario here would be \emph{Brent (offshore oil field) on 1 August 2000}.
Hence an important design issue in human evaluations is choosing the specific scenarios used in the experiment.

One approach is to choose scenarios randomly.  For example, in the SumTime context, if there is forecast data for 10 locations over a 100-day period period, we can randomly select location 5 on day 59, location 1 on day 80, etc (I generated these using a random number generator).

In other cases it is important that scenarios are diverse, and cover important categories and/or edge cases\index{edge cases}.  One possibility in such cases is to randomly select within a category.   For example, in the Babytalk BT45 evaluation mentioned in Section~\ref{sec:humantaskeval}, it was essential that the selected scenarios
covered 8 different recommended interventions (scenarios where the recommended clinical action was to increase oxygen levels,  scenarios where the recommendation was that the doctors do nothing, etc.).   So the researchers created sets of potential scenarios which met each of these criteria (e.g., one set was all potential scenarios where the recommended action was to increase oxygen levels), and then randomly selected experimental scenarios from each of these sets.

As with subject choice, another decision is how many scenarios to use. Sometimes this can  be computed from the number of subjects, the number of judgements (ratings, rankings, or annotations) made by each subject, the number of texts in each scenario (often one per system), and the number of times each text is judged, as follows:
\begin{verbatim}
NumberOfJudgements = NumberOfSubjects * NumberJudgementPerSubject
JudgementsPerScenario = TextsPerScenario * JudgementsPerText
NumberScenarios = NumberOfJudgements / JudgementsPerScenario
\end{verbatim}

For example, for the annotation portion of the BasketballNLG\index{BasketballNLG} evaluation, scenarios can be chosen at random.  In terms of numbers, assume:
\begin{itemize}
\item 5 annotators (subjects).
\item 8 texts annotated by each annotator.
\item 2 texts per scenario (one produced by BasketballNLG, one produced by state-of-art baseline).
\item each text annotated by two people (to reduce the chance of missing something).
\end{itemize}
The above formula suggests 10 scenarios.
\begin{verbatim}
NumberOfJudgements = 5 * 8 = 40
JudgementsPerScenario = 2 * 2 = 4
NumberScenarios = 40 / 4 = 10
\end{verbatim}

\subsubsection{Step 5: Decide on experimental procedure}\label{sec:humanevalproc}\index{procedure}
Experimental procedure includes many different things, such as:
\begin{itemize}
\item \emph{Delivery:} How is the experiment run?  Is it done over the web, online but in an experimental room, or on paper?  Web-based experiments\index{web-based experiments} are usually the easiest to run, but experimenters have more control and can detect problems better if they are in the same room as the subjects; they can also observe subjects, which is sometimes useful.  Some quality criteria\index{quality criteria}, such as reading time, are difficult to accurately measure in web-based experiments.
\item \emph{User Interface\index{user interface} and Training\index{training}:} What user interface (UI) do subjects\index{subjects} use, and what training (if any) are they given to explain the task and UI?  Its important to get this right, since it is essential that subjects understand what they are supposed to do\footnote{\url{https://ehudreiter.com/2024/05/28/human-eval-subjects-must-understand-the-task/}}.  Piloting the experiment\index{pilot experiments} (Section~\ref{sec:pilotexperiments}) can often detect UI problems.
\item \emph{Text selection:} Which texts do subjects see, and in what order?  It is often good practice to use a \emph{Latin Square design}\index{Latin square design} \cite{latinsquare} to ensure that each subject sees a balanced number of texts from each system (since some subjects are more  generous in their ratings or more skilled at the task).  If the experiment involves ranking texts produced by different systems, then the order in which the texts are shown should be randomised (since some subjects may have a preference for preferring the first text that they see).
\item \emph{Attention checks:}\index{attention check} Does the experiment include attention-check questions? These are questions with clear answers whose purpose is to ensure that subjects are taking the experiment seriously and not just clicking at random.  Attention checks are common in web-based experiments using crowdworkers\index{crowdworkers}, where there is a fear that crowdworkers will try to do the experiment as quickly as possible in order to maximise their income (Section~\ref{sec:humansubjects}).  Data from subjects who fail attention checks should be discarded.
\item \emph{Dealing with disruptions:} What should happen if something goes wrong?  For example, a network outage or fire alarm disrupts an experiment, or a subject has to drop out of an experiment because she is not feeling well?  It is not possible to plan for every potential event, but it is worth thinking at least in general terms about how to handle such disruptions.
\end{itemize}

\noindent{}Figure~\ref{fig:basketballDesign} includes a high-level experimental design for evaluating BasketBallNLG\index{BasketballNLG}.
Of course many details need to be fleshed out, such as the exact design of UI forms.

\subsubsection{Step 6: Decide on Analysis}\label{sec:humananalysis}\label{sec:outlier}\index{analysis}
Another key experiment design issue is how experimental data will be analysed;
 this should be decided \emph{before} the experiment is run and data is collected.  Analyses should focus on testing hypotheses\index{hypothesis testing} (the ones chosen in Step 1).  From an insight\index{insights} perspective, it is important to qualitatively as well as quantitatively analyse the experimental data.

Most NLG hypotheses are stated in terms of system A having a higher score than system B.  Score should of course be computed and reported, but it is also important to identify and remove outliers and to perform statistical significance testing.

An \emph{outlier}\index{outlier} is a data point that is very different from the other data points.  Sometimes outliers are genuine; if so, examining them can lead to major insights about hypotheses.  However, outliers often signify experimental flaws, for example a subject\index{subjects} who did not take the experiment seriously (and just clicked things at random), or who misunderstood the task.  Such outliers are not genuine data about the hypothesis, and should be discarded.   Sometimes it makes sense to identify subjects (as well as individual data points) who are outliers, and discard all data from this subject.

 It is dangerous for experimenters to identify and discard outliers on an ad-hoc basis, because the experimenter may then (perhaps unconsciously) be tempted to discard \scare{inconvenient} data points that go against the experimental hypothesis.  For this reason, researchers should specify an outlier identification policy \emph{before} running the experiment, such as the widely used \emph{IQR}\index{IQR} method\footnote{\url{https://en.wikipedia.org/wiki/Interquartile\_range}}.

Once outliers have been removed, researchers can compute the mean (or median) score for each system.   Statistical significance\index{statistical!significance} (Section~\ref{sec:statisticalsig}) of the differences in scores should also be computed.

Different statistical tests\index{statistical!tests} are used in different experiments \cite{dror-etal-2018-hitchhikers}.   Some simple advice which works for many (not all) experiments is:
\begin{itemize}
\item When comparing numerical scores, such as the number of errors in a text, use a \emph{t-test}\index{t-test} if comparing two systems, and an \emph{ANOVA}\index{ANOVA} if comparing more than two systems.
\item When comparing Likert-scale\index{Likert scale} ratings\index{rating} of texts in a ratings experiment, use a \emph{Mann Whitney test}\index{Mann-Whitney} test if comparing two systems, and a \emph{Kruskall-Wallis}\index{Kruskall-Wallis test} test if comparing more than two systems.
\item When comparing preferences in a ranking\index{ranking} experiment, use a \emph{binomial} test\index{binomial test} if just two systems are being compared (ignore \scare{no difference} choices if these are allowed).  If more than two systems are being compared, use  a \emph{Friedman}\index{Friedman test} test.
\item When comparing categorical outcomes (eg, how many people stopped smoking in different groups), use a \emph{chi-square} test\index{chi-square test} (regardless of the number of systems being compared).
\end{itemize}
If in doubt about which statistical test to use, you should consult with a statistician. 

Once the main hypotheses have been tested, researchers can also look for other interesting patterns in the data.  This is often done using Exploratory Data Analysis\index{Exploratory Data Analysis} (EDA) techniques \cite{tukey1977exploratory}, where a researcher explores the data and looks for patterns using a general-purpose visualisation and analysis tool such as R (people with limited programming expertise often use Microsoft Excel).   When reporting results, it is important to distinguish between formal resuts from hypothesis testing, and additional insights from EDA.

In addition to numerically analysing data, it is also important to \emph{qualitatively} analyse data.  One technique which I recommend is \emph{qualitative error analysis}\index{qualitative error analysis}; this involves taking a few example texts (often including texts with poor scores), and qualitatively analysing them.  For example, if the experiment is based on annotating\index{annotation} errors (Section~\ref{sec:humanevalannotations}), then the quantitative analysis will be based on counting errors (perhaps weighted by severity), while the qualitative analysis explores specific individual errors which shed light on what can go wrong.
If the experiment is based on Likert ratings of readability (Section~\ref{sec:humanevalannotations}), the quantitative analysis is usually based on mean score, while the qualitative analysis could discuss specific individual texts which got low Likert ratings from subjects.
Qualitative analysis does not prove or disprove hypotheses, but it does give insights on what the problems are in generated texts, which can help developers improve the relevant systems.

In many experiments subjects are asked to provide free-text comments as well as specific ratings, rankings, annotations, etc; I definitely recommend doing this when possible.  These free-text comments are another good source of insights into problems and areas for improvement.

If subjects provide free-text comments, these can be analysed using a thematic analysis\index{thematic analysis} \cite{braun2012thematic}.  Another approach is simply to read the comments and identify ones which provide interesting insights about the texts and hypotheses; these can be summarised in the experimental report, including quotations from especially interesting comments.  When doing this, it is important to ensure that quotations are anonymous and do not reveal the authors identity.

Figure~\ref{fig:basketballDesign} includes a possible analysis plan for BasketballNLG\index{BasketballNLG}.

\subsection{Issues in human evaluation}

Regardless of the experimental design and type of evaluation, there are a number of generic issues which are important in human evaluation.

\subsubsection{Interannotator agreement}\label{sec:IAA}\index{inter-annotator agreement}
It is good practice in human evaluation to ask multiple subjects to assess the same texts, and then measure how well they agree.  Ie, if subjects A, B, and C evaluate a text T under the relevant quality criteria, how likely are they to agree, for example to all say that the text is excellent?  This is called \emph{inter-annotator agreement}.

Interannotator agreement is most commonly measured using a \emph{kappa}\index{kappa} score, typically Cohen's kappa\index{kappa!Cohen's} for agreement between 2 annotators and Fleiss kappa\index{kappa!Fleiss} for agreement between more than two annotators; Krippendorff's alpha\index{Krippendorff's alpha} can also be used.  McHugh \cite{mchugh2012interrater} gives a nice summary which is accessible to non-statisticians of kappa and related measures, how they are used to measure annnotator agreement, and some issues and concerns.  Most statistical software packages include functions for calculating kappa and alpha, and it is good practice in human evaluations to report a kappa or alpha score.

If the experimental design specifies that most texts are only evaluated by a single subject, then it may make sense to get a subset of texts annotated by multiple annotators, so that interannotator agreement can be reported.

The kappa statistic\index{kappa!statistic} is a number between 0 and 1. McHugh \cite{mchugh2012interrater} suggests interpreting kappa as follows:
\begin{itemize}
\item above 0.90: \emph{Almost perfect}
\item 0.80-0.90:  \emph{Strong}
\item 0.60-0.79:  \emph{Moderate}
\item 0.40-0.59:  \emph{Weak}
\item 0.21-0.39:  \emph{Minimal}
\item 0-0.20:  \emph{None}
\end{itemize}
Many NLP researchers use a more lenient interpretation, where for example a kappa statistic of 0.4 is considered to be Moderate rather than Weak.  Of course the interpretation of kappa depends on the circumstances and type of experiment, but in general I recommend that researchers use McHugh's more conservative interpretation.

If interannotator agreement is weak (less than 0.60), this means that different annotators are assessing texts quite differently, which is not good; replication\index{replication} (Section~\ref{sec:replication}) may be less successful if agreement is low.  Of course human beings are different, which means that no two individuals will assess a text in exactly the same way, but nonetheless we have more confidence in the result of a human evaluation if there is good agreement between annotators

Agreement can often be increased by changing the experimental design\index{experimental design}.   For example, adding attention checks to filter out subjects\index{subjects} who are not taking the experiment seriously, and/or providing training\index{training} and clear guidance\index{guidelines} to subjects.   Agreement is also often higher for simple experimental tasks than more complex tasks.

\personal{In my experience, inter-annotator agreement is usually higher for annotation-based evaluations (Section~\ref{sec:humanevalannotations}) than for rating/ranking evaluations (Section~\ref{sec:humanevalrankings}), which is one reason I prefer not to use rating/ranking evaluations if I have a choice.}

\subsubsection{Piloting experiments}\label{sec:pilotexperiments}\index{pilot experiments}
Experiments need to be carefully designed and also executed\index{experimental execution}, as described in Section~\ref{eval:experimental-execution}.  Since human experiments are expensive and time-consuming, it usually makes sense to first conduct a small-scale \emph{pilot} experiment, and use this to investigate problems such as code bugs\index{bugs!code}, randomisation errors, and confusing user interfaces\index{user interface}.  For example, researchers can use a pilot to check:
\begin{itemize}
\item Are the correct texts shown to users \cite{thomson-cl24}? 
\item Does the experimental user interface (UI) make sense to subjects\index{subjects}, and do they use it correctly?
\item Is all of the expected experimental data recorded? 
\item Does the analysis\index{analysis} (including outlier\index{outlier} detection) make sense?
\end{itemize}
\personal{We once ran a fairly expensive (and impossible to replicate) experiment where we discovered after completion that some key data had not been recorded \cite{williams_reiter_2008}.  If we had done a pilot first, we would have realised this and fixed our software before running the experiment!}

Please remember that the point of a pilot is not to gather data, but rather to debug the experiment!  Sometimes pilot data can be interesting, but the primary objective is to ensure that the experimental design\index{experimental design} works and that there are no execution errors.  If you detect problems in a pilot and fix them, you may want to consider doing another pilot, since it is possible that fixing one set of bugs has inadvertently introduced new bugs.

I usually do pilots with 5-10 subjects, but it is possible to do useful pilots with just 2 or 3 subjects.  However experimenters should not do pilots using themselves as subjects, it is better to get other people to be subjects.   This is because experimenters know what is supposed to happen in an experiment, and thus (for example) may not detect confusing instructions or user interfaces.

\subsubsection{Research ethics}\index{ethics}\label{sec:researchethics}
Experiments with human subjects must be \emph{ethical}.  Amongst other things:
\begin{itemize}
\item Experiments should not harm subjects, third parties, or researchers. For example, experiments should not normally give subjects misleading medical advice, or show them pornographic or otherwise offensive material (which could happen if subjects are shown random material from the Internet). 
\item Experiments should not lead to sensitive personal data being published or otherwise released.
\item Human subjects must agree to participate in experiments, they should be not be forced or coerced (for example, managers cannot order subordinates to take part in an experiment).
\end{itemize}
In many contexts, experiments must be formally approved by a \emph{research ethics} committee\index{research ethics committee} (called \emph{institutional review board}\index{institutional review board} (IRB) in USA).   Details depend on the type of experiment, the location,  whether the experimenter works for a university or a company, and who is funding the research. Researchers therefore should consult locally about research ethics procedures and rules.

Even when formal ethical approval is not legally required, it is good practice to consider if an experiment could inadvertently harm people or otherwise be unethical.  Indeed, I know of companies which are not legally required to review ethical issues in experiments, but have still implemented very strict review procedures, partially to minimise legal\index{legal} (Section~\ref{sec:legal}) and reputational risks if an experiment harms subjects or third parties.

Experiments may be modified in order to reduce ethical concerns; this is part of the experimental design\index{experimental design} process.  For example, material which experimenters would like to show to subjects can be checked beforehand for offensive content, so that only non-offensive material is actually shown to subjects.

In Aberdeen University in 2024, all experiments which involve human subjects must be approved by a research ethics committee.  Some types of medical-related experiments must be approved by an NHS\index{National Health Service} (UK National Health Service) ethics committee; other experiments are approved by an ethics committee within the university.   Approval typically takes a few weeks for straightforward experiments, which includes most rating, ranking, and annotation experiments.  Approval can take longer for task-based experiments\index{task-based evaluation} which impact a user's actions or behaviour, experiments which involve children or other vulnerable populations, experiments which require gathering sensitive personal information, and experiments where subjects are deceived.

\personal{I recently discussed with a colleague ethical review procedures for an experiment I planned to do with collaborators in China.  My colleague had previously done experiments in China, and explained that he had not needed to do an ethical review.  However, ethical review requirements in China are changing and reviews will be required for this type of experiment in the future; fortunately we realised this in time.}

\section{Automatic evaluation}\label{sec:automaticevaluation}
It is also possible to evaluate texts automatically (without human involvement), by using \emph{metrics}\index{metrics}, that is algorithms that assess the quality of a generated text.  These can either directly assess quality criteria\index{quality criteria}, or they can measure how similar  generated texts are to \emph{reference texts}\index{reference texts}, which are usually high-quality human-written texts in a corpus\index{corpus}.

Automatic evaluation is much cheaper and quicker than  human evaluation, it is also usually easier to replicate\index{replication}.  The problem is that the results of automatic evaluation are not always meaningful predictors of real-world text quality and utility; metrics should be \emph{validated}\index{metrics!validation} to assess how well they agree with trusted high-quality human or impact evaluations (Section~\ref{sec:validationmetrics}).

Sometimes researchers carry out both an automatic evaluation on a large test set and a human evaluation on a subset of the test data.  The human-evaluated subset can be randomly selected; readers will have more confidence in the automatic evaluation if it broadly agrees with the human evaluation.  Another strategy is to use an automatic evaluation to find cases where the NLG system seems to do poorly, and get these evaluated by humans in order to get a deeper understanding of what the problems are.

A large number of evaluation metrics have been proposed in the literature, with new ones coming out every month, so the discussion in this section will focus on principles instead of specific metrics.

\subsection{Types of automatic evaluation}

\subsubsection{Reference-based metrics}\label{sec:refmetrics}\index{metrics!reference-based}
The most common type of automatic evaluation (at the time of writing) compares generated texts against human-written reference texts\index{reference texts}.   The simplest such metric is \emph{edit distance}\index{edit distance} which simply counts how many edits (changes) need to be made to a generated text in order to match the reference text; this can be computed at either word or character level.  An example is shown in Figure~\ref{fig:SumTimeMetric}, using the same SumTime weather domain as Figures \ref{fig:SumTimeRanking} and \ref{fig:SumTimeRating}. The BLEU\index{BLEU}  (Section~\ref{sec:bleu})  and ROUGE\index{ROUGE} \cite{lin-2004-rouge} metrics, which were introduced in the early 2000s but are still popular, essentially apply a more sophisticated scoring formula (based on n-grams) to word-level differences between the generated and target texts.  The chrF\index{chrF} metric \cite{popovic-2015-chrf} takes a similar approach but looks at character-level differences instead of word-level differences; it may be the best of the simple string-similarity metrics \cite{kocmi-etal-2021-ship}.

\begin{figure}
\lineacross\\
\emph{Reference text:}\\
SSW 16-20 GRADUALLY BACKING SSE THEN BECOMING VARIABLE 10 OR LESS BY MIDNIGHT\\
\\
\emph{Generated text:}\\
SSW’LY 16-20 GRADUALLY BACKING SSE’LY THEN DECREASING VARIABLE 4-8 BY LATE EVENING\\
\\
\emph{Differences:}\\
SSW\st{’LY} 16-20 GRADUALLY BACKING SSE\st{’LY} THEN \st{DECREASING} \emph{BECOMING} VARIABLE \st{4-8} \emph{10 OR LESS} BY \st{LATE EVENING} \emph{MIDNIGHT}.\\
\\
\emph{Edit count:}
\begin{itemize}
\item Two deletions of \st{’LY} (one token deleted, twice)
\item \st{DECREASING} changed to \emph{BECOMING} (one token changed)
\item \st{4-8} changed to \emph{10 OR LESS} (three tokens changed)
\item \st{LATE EVENING} changed to \emph{MIDNIGHT} (two tokens changed)
\item No tokens added
\end{itemize}
Token-level edit distance is \emph{8} tokens deleted, changed, or modified\\
Character-level (Levenshtein\index{Levenshtein distance} \cite{Levenshtein}) edit distance is 27\\

\caption{Example of edit-distance evaluation in weather domain}
\label{fig:SumTimeMetric}       
\lineacross
\end{figure}

Recent reference-based metrics such as BLEURT\index{BLEURT} (Section~\ref{sec:bleurt}) and BERTScore\index{BERTScore} \cite{zhangbertscore} are usually trained.  In other words, instead of identifying differences and applying a scoring algorithm,  they use a machine learning model which is trained on human evaluation results to predict the quality of a generated text from a reference text.
Amongst other things, this means that these metrics can consider semantic similarity as well as string similarity.  

As explained by Kocmi et al \cite{kocmi-etal-2021-ship}, trained metrics give judgements which are more similar to human evaluations than string-based metrics.  On the other hand, the results of string-similarity metrics are easier to interpret, which may be useful for developers who are trying to improve an NLG system.

Some metrics, including BLEU, can use multiple reference texts, which allows for cases (common in NLG) when there are many acceptable output texts.

Reference-based metrics of course require good-quality reference texts.  Unfortunately, in some cases insufficient attention is paid to the quality of reference texts.  More fundamentally, if the NLG system is capable of generating human-quality (or better than human texts), then it may not make sense to evaluate a  generated text by comparing it to a human-written reference text.

\subsubsection{Referenceless metrics}\index{metrics!referenceless}
Generated texts can also be evaluated using metrics or algorithms which do not require reference texts.  A simple example is the Flesch-Kincaid\index{Flesch-Kincaid} reading grade level \cite{kincaid1975derivation}, which assesses the readability of a text by giving a US \scare{grade level}; for example if a text gets a grade level of 9, then it should be readable by the average 15 year old.  The score is calculated using a linear regression on the mean number of words per sentence, and the mean number of syllables per word; these coefficients were originally calculated by fitting the regression model to a dataset of how well children of different ages could read different texts (as determined by educational tests).

Recently much more sophisticated models have been built, using modern machine learning techniques, which estimate different quality criteria\index{quality criteria} of generated texts.   Kocmi et al \cite{kocmi-etal-2021-ship} say that the best of these models are almost as good as the best reference-based metrics, at least for evaluating the quality of machine translation\index{machine translation} texts.  At the time of writing, there is a lot of interest and excitement in using large language models such as GPT\index{GPT} to evaluate texts (Section~\ref{sec:gemba}).

\subsubsection{Compute speed, resources and cost}\index{compute!speed}\index{compute!resources}\index{compute!cost}
Most NLG evaluations focus on text quality, but in many contexts computational speed, resources and cost are also important; this is usually measured either as the time taken to generate a text with a given computational resource (eg, GPU), or the cost required to generate a text using cloud-based computational resources.

From an evaluation perspective, speed evaluations can be done using timing functions which are built into Python and other programming languages.  It is good practice when doing a speed evaluation to shut down other processes and even (if possible) disconnect from the Internet; this reduces the amount of \scare{noise} in the evaluation.  From a replicability\index{replication} perspective, when describing the experiment it is essential to give not just full hardware specs, but also details about operating system, libraries used, compilers, and so forth.  It is also good practice to generate each text several times, and report the mean (average) time across runs.

Cost-based evaluations of NLG are relatively unusual at the time of writing, but they can be conducted in the same way, using the metering functionality built into most cloud computing services.   Of course it is not possible to disconnect from the Internet if the NLG system uses cloud computing services.

\subsection{Experimental Design}\label{sec:metricevaldesign}\index{experimental design}

Experimental design for automatic evaluation must address the key steps described in Section~\ref{eval-gen-exp-design}, except that there is no need to choose subjects (since human subjects are not needed for metric evaluations).
\begin{enumerate}
\item Choose research hypotheses and questions.
\item Choose the overall study type.
\item \st{Choose subjects.}
\item Choose material, ie the test set that metrics are calculated on.
\item Decide on experimental procedure.
\item Decide on analysis.
\end{enumerate}

An example is shown in Figure~\ref{fig:basketballMetricDesign}; this is for BasketballNLG,  the same system which was used in the human evaluation example shown in Figure~\ref{fig:basketballDesign}.

\begin{figure}
\lineacross\\
\textbf{Research Hypothesis}
\begin{itemize}
\item Texts produced by BasketballNLG are better than texts produced by \emph{StateofArtSystem} under the quality criteria of accuracy and fluency.
\end{itemize}

\textbf{Study Type:}
\begin{itemize}
\item Reference-based metric
\item Metric is not trained or fine-tuned
\end{itemize}

\textbf{Material:}
\begin{itemize}
\item 100 recent games which were played no earlier than one month before the experiment.
\end{itemize}

\textbf{Procedure:}
\begin{itemize}
\item Use BLEURT metric with standard parameters and no fine-tuning, to evaluate texts produced by BasketballNLG and \emph{StateofArtSystem} for the 100 games.
\end{itemize}

\textbf{Analysis:}
\begin{itemize}
\item Compute system-level score by taking mean of text-level BLEURT scores for texts produced by each system (BasketballNLG and \emph{StateofArtSystem}).
\item Use a paired t-test on text-level scores to compute statistical significance of differences between BasketballNLG and \emph{StateofArtSystem}.
\item Do qualitative error analysis on the 5 BasketballNLG texts that had the lowest BLEURT scores.
\end{itemize}

\caption{Experimental design for metric evaluation of BasketballNLG}\label{fig:basketballMetricDesign}
\lineacross
\end{figure}

\subsubsection{Research hypotheses}\index{research questions}
Similar to human evaluations (Section~\ref{sec:humanhypotheses}), most automatic evaluations test hypotheses of the form \scare{Texts produced by A are better than texts produced by B under quality criterion C}.  Some quality criteria\index{quality criteria}, such as utility, are difficult to measure with metrics\index{metrics}, but others (such as fluency) can often be assessed with metrics.

Most automatic evaluations compare a new NLG system against an existing system.  If the underlying claim is that the NLG system is better than existing systems, then it is essential that the comparison system be a high-quality state-of-the-art\index{state-of-art} system.   

\personal{I have seen many papers which claimed impressive results by comparing a new NLG system against an obsolete and out-of-date comparison system; this is not good science.}

Automatic evaluations can also be used to guide design decisions during the development process.  For example, if developers want to use a prompted model but are unsure which prompt\index{prompt} is best, they can use automatic evaluations to assess how well the system works with different prompts.  In this case the research hypothesis is that a specific NLG system produces better quality texts if it uses prompt XXX instead of YYY.

\subsubsection{Study Type}\index{study type}
There is less variety in study types with metric-based evaluations compared to human evaluations (Section~\ref{sec:humanstudytype}), but some decisions still need to be made:
\begin{itemize}
\item Will reference-based\index{metrics!reference-based} or referenceless\index{metrics!referenceless} metrics be used (or both)?  If reference-based metrics are used, what is the source of the reference texts\index{reference texts}?
\item Are metrics fine-tuned so that they work better in the target domain?  If so, what data will be used to fine-tune\index{fine-tuning} the metrics?
\end{itemize}

\subsubsection{Material}\index{material}
As with human evaluations (Section~\ref{sec:humanmaterial}), scenarios\index{scenarios} must be chosen to evaluate the systems; in machine learning terminology, a \emph{test set}\index{test set} must be created.  But while most human evaluations use a relatively small number of scenarios (typically less than 100), metric-based evaluations can use thousands or even millions of scenarios.  This is one of the strengths of metric-based evaluation, it can assess performance on a much wider range of scenarios than is possible with human evaluation.

It is common practice in machine learning to create a test set by taking a large data set and extracting part of it as test data; the rest is used to train and tune the model. However this is not always a good approach when evaluating systems built on large language models because of data contamination\index{data contamination} concerns (Section~\ref{sec:trainingtestdata}).
A better approach in such contexts is to create test scenarios from new data; for example a basketball-story generator can be tested on data from recent matches.  This is the approach suggested in the experimental design shown in Figure~\ref{fig:basketballMetricDesign}.

If researchers have access to confidential data which was never published on the Internet because of privacy or commercial secrecy, it may be possible to use this to create test data. However it will be difficult for other researchers to replicate the experiment if the test data cannot be published.

\personal{Producing good test data for testing large language models is difficult.  Unfortunately, I have seen many academic papers which ignore this issue and simply test models on data which they were trained on, without even acknowledging that this is a problem.  This is not good science.}

\subsubsection{Experimental procedure}\index{procedure}
A key decision is which specific metrics\index{metrics} will be used.  I cannot give specific advice here because anything I say will be out of date by the time this book is published.   Researchers must also decide on metric parameters and pre-processing techniques such as tokenization; these can have a major impact on metric scores \cite{post-2018-call}.

As mentioned in Section~\ref{sec:replication}, one dilemma is that sometimes researchers want to use similar metrics to previously published work in order to make their results more comparable to previous research, but the metrics used in this older work are less meaningful than newer metrics.  For example, the BLEU\index{BLEU} metric continued to be used long after much better alternatives were available, in part because it made comparison with earlier work easier.   In such cases, researchers can show results both from the best-available metrics and from metrics used in earlier papers.

In some cases, when asked to compare two texts, metrics will be biased and rate the first text more highly because it is first \cite{wang2023large}.  For this reason, it is useful to vary the order in which texts are presented, which is also good practice for human evaluations (Section~\ref{sec:replication}).

\subsubsection{Analysis}\index{analysis}
The best analysis procedure for metrics depends on the metric chosen.  Researchers should check the literature and best-practice guidelines for their chosen metrics when deciding on issues such as the following:
\begin{itemize}
\item If a metric produces scores for individual texts, then these scores are usually aggregated into a system level score.  The obvious approach is to use the mean of the text-level scores as the system score, but there are alternatives, some of which involve weighting some texts (eg, longer texts) more highly than others.  Worst-case\index{worst-case} performance should be reported if requirements say this is important (Section~\ref{sec:req-worstcase}).

\item As with human evaluations, it is essential to perform statistical significance\index{statistical!significance} tests as well as presenting raw numbers.  The best statistical test depends on the metrics, and sometimes specialised tests are needed \cite{koehn-2004-statistical}.

\item Small difference in metric scores, even if statistically significant, may not be meaningful; this is discussed in Section~\ref{sec:validationmetrics}.
\end{itemize}
There are many software packages which automatically run metrics on a test set and analyse the results.  It is fine to use such packages, but researchers should understand how the chosen package analyses the data.

When reporting the experiment, researchers should specify which metrics were used and why they were chosen.  Details should be given about the software used to compute the metric, including parameter settings (if appropriate).  If the metric software came from a library, library details (including version number) should be given.

If a human evaluation was also performed, the correlation between the human and metric evaluations should be reported.

A general point is that is very easy to compute multiple metrics on a data set; some papers report scores for more than ten different metrics when comparing systems.  If this is done, then a multiple-hypothesis correction should be considered.  For example, if ten different metric scores are reported, then a \emph{Bonferroni correction}\footnote{\url{https://en.wikipedia.org/wiki/Bonferroni\_correction}} can be applied by dividing the usual statistical significance threshold of 0.05 by the number of metrics used (10).  A result is only considered to be statistically significant if the p-value is less than the modified threshold, which is 0.005 in this example.

Last but not least, as with human evaluation (Section~\ref{sec:humananalysis}), it can be very useful to do a (manual) qualitative error analysis\index{qualitative error analysis} on individual texts which had poor metric scores.

\subsection{Examples of metrics}
It is impossible to provide an up-to-date list of metrics\index{metrics}, because new metrics are being proposed every month, and any list provided in this book would be obsolete by the time the book is published.  Nonetheless, I will describe a few of of the more popular metrics (at the time of writing), to give readers a better understanding of how they work.

\subsubsection{BLEU}\label{sec:bleu}\index{BLEU}
One of the oldest and best-known reference-based metrics\index{metrics!reference-based} is BLEU \cite{papineni-etal-2002-bleu}.  The core idea of BLEU is to compute n-gram precision.  That is, it counts the number of \emph{n-grams}\index{models!ngram} (Section~\ref{sec:ngram}) in the generated text which also occur in the reference text\index{reference texts}, and divides by the number of n-grams in the generated text.  N-gram refers to word combinations: 1-gram is just words, 2-grams are pairs of words, 3-grams are word triples, etc.

For example, consider the following sentences:
\begin{itemize}
\item \emph{Generated:} It will be very windy on Sunday.
\item \emph{Reference:} It will be windy on the weekend.
\end{itemize}
There are seven words, and hence \emph{unigrams} (1-grams) in the generated sentence: \emph{It, will, be, very, windy, on, Sunday}.  Five of these (\emph{It, will, be, windy, on}) appear in the reference sentence.  So the 1-gram precision of the generated sentence is 5/7 = 0.71 

We can also look at \emph{bigrams} (2-grams) in the generated sentence: there are six of these (\emph{It will, will be, be very, very windy, windy on, on Sunday}).  Three of these are in the reference text (\emph{It will, will be, windy on}), so the 2-gram precision of the generated sentence is 3/6 = 0.50 .

The real BLEU algorithm incorporates a number of adjustments to the core n-gram precision formula, including allowing multiple reference texts to be used and specifying how a system-level score can be calculated from scores for individual texts (like the example above). There are many implementations of BLEU available in NLP toolkits and repositories, and almost all researchers use one of these instead of coding BLEU from scratch.  By far the most common n-gram size is 4, and published BLEU scores are based on 4-grams unless otherwise indicated.

A number of other n-gram-based metrics have been proposed in the literature, including ROUGE\index{ROUGE} and chrF\index{chrF}.  Kocmi et al \cite{kocmi-etal-2021-ship} recommend chrF \cite{popovic-2015-chrf} as the best such metric.

\subsubsection{BLEURT}\label{sec:bleurt}\index{BLEURT}
BLEURT \cite{sellam-etal-2020-bleurt} is a more recent reference-based metric\index{metrics!reference-based} which is trained on data.  In other words, while BLEU essentially uses an explicit algorithm to compute its score, BLEURT uses machine-learning techniques to train a model\index{models} which produces scores.

The details are again complex, but the core idea is to take the BERT\index{BERT} language model \cite{devlin-etal-2019-bert}, and fine-tune\index{fine-tuning} it for the score generation task.  Fine-tuning requires a data set which contains texts and corresponding human evaluation scores. 

Since there is a limited amount of data which provides high-quality human evaluations of generated texts, BLEURT first fine-tunes the model on synthetic data\index{fine-tuning} (Section~\ref{sec:syntheticdata}), which it produces by perturbing Wikipedia sentences (for instance by backtranslation, see example in Figure~\ref{fig:backtranslate}).  It then uses BLEU and other older metrics to estimate the quality of the perturbed sentences.  The quality of this synthetic data set is not great, but it is very large, and tuning on it makes BLEURT more robust.

After the model has been fine-tuned on synthetic data, it is then further fine-tuned on actual (text, human ratings) pairs from one of the WMT tasks.

BLEURT can be used \scare{off the shelf}, or it can be further fine-tuned for a specific NLG task, by giving it appropriate (text, human rating) training data.

\subsubsection{Evaluation using large language models: GEMBA-MQM}\label{sec:gemba}

At the time of writing, there is a lot of interest and excitement in using prompted large language models (such as GPT\index{GPT}) to evaluate the quality of generated texts, by essentially just asking the model to do this in its prompt\index{prompt} (and possibly including some examples).  In some contexts this can give more meaningful results than simple human evaluations \cite{zhang-etal-2023-needle}.
This space is evolving very quickly, so I will just give a representative example, which is GEMBA-MQM\index{GEMBA-MQM} \cite{kocmi-federmann-2023-gemba}.

GEMBA-MQM asks the GPT4 language model to annotate\index{annotation} texts (output of machine translation systems) using the MQM\index{MQM} annotation scheme \cite{FreitagMQM}.  MQM annotation by human translation is (at the time of writing) probably the most rigorous and meaningful evaluation technique in machine translation, but it is expensive and time-consuming.

GEMBA-MQM asks GPT4 to perform an MQM annotation, using a prompt which describes the MQM protocol in a fairly straightforward manner (an extract is given in Figure~\ref{gemba-prompt}) and also gives three examples of MQM-annotated texts.

\begin{figure}
\lineacross\\
(System) You are an annotator for the quality of machine translation. Your task is to identify errors and assess the quality of the translation.\\
\begin{verbatim}
(user) {source_language} source:\n
```{source_segment}```\n
{target_language} translation:\n
```{target_segment}```\n
\n
\end{verbatim}
Based on the source segment and machine translation surrounded with triple backticks, identify
error types in the translation and classify them. The categories of errors are: accuracy
(addition, mistranslation, omission, untranslated text), fluency (character encoding, grammar,
inconsistency, punctuation, register, spelling),   [\emph{etc}]
\caption{Part of the GPT4 prompt used by GEMBA-MQM, from \cite{kocmi-federmann-2023-gemba}. \texttt{\{source\_language\}} is replaced by the language of the text being translated, \texttt{\{source\_segment\}} is replaced by the text being translated, etc. The prompt also includes three examples of MQM annotated texts.}\label{gemba-prompt}
\lineacross
\end{figure}

According to \cite{kocmi-federmann-2023-gemba}, GEMBA-MQM agrees very well with human MQM ratings when comparing systems.  In other words, if human translators judge that system A outputs are better than system B outputs based on MQM annotations, then GEMBA-MQM is very likely to make the same assessment (see paper for details). 

\personal{Several of my students have tried to use GPT to evaluate texts, and all of them have found that GPT is biased\index{bias} towards texts that it generated.  In other words, if GPT4 and Gemini\index{Gemini} generate texts of equivalent quality as judged by human evaluation, GPT4 (when used as an evaluator) will rate the GPT4 text more highly than the Gemini text.  Again this is a rapidly-developing area, and we may see new variants of language-model-based evaluation which reduce such biases.}

\subsection{Validation of metrics}\label{sec:validationmetrics}\index{metrics!validation}
In most cases (compute speed/cost evaluations are an exception), it is important to \emph{validate} metrics, that is show that they correlate (agree) with high-quality human or impact evaluations\index{impact!evaluation} \cite{reiter-belz-2009-investigation} (we can similarly validate cheap/quick human evaluations by seeing how well they agree with a high-quality \scare{gold-standard} human evaluation \cite{garneau-lamontagne-2021-shared}).  Validation does \emph{not} need to be done every time a metric is used; it is generally done when a metric is first proposed, and then repeated on a regular basis for metrics which are heavily used.

Of course, how well a metric correlates with a human evaluation depends on many factors, including the NLG task, the human evaluation,  the quality criteria, and evaluation granularity (are texts or systems being evaluated?).  In rough terms, metrics often seem to have higher correlations with human evaluations in the following contexts:
\begin{itemize}
\item \emph{NLG task} is relatively simple.
\item \emph{Quality criteria}\index{quality criteria} focus on readability or related criteria such as fluency.
\item \emph{Granularity}\index{evaluation!granularity} is system-level evaluation based on average scores (as opposed to text-level evaluations or system-level evaluations based on worst-case performance (Section~\ref{sec:req-worstcase})).
\item \emph{Human evaluation} used for correlation has good inter-annotator agreement\index{inter-annotator agreement} (Section~\ref{sec:IAA}).
\end{itemize}
These are very rough observations, and it is important to explicitly measure correlation between metrics and human evaluations.

The best known and longest-running validation studies in NLG and related fields are the annual metrics evaluation shared task in the WMT\index{WMT} conference.  WMT is an annual conference which includes a high-quality human evaluation of a number of machine translation systems.  Alongside this, WMT runs a shared task where various automatic metrics are used to evaluate the same systems, and correlations are computed between the metric scores and the WMT human evaluations.  The exact design of the validation exercise varies, I encourage interested readers to read about specific metrics evaluation shared tasks such as the one at WMT22 \cite{freitag-etal-2022-results}.  Regardless of the details, the shared task involves computing correlations separately for different language pairs, and also at both the system and text (called \emph{segment} in this context) levels; sometimes an overall aggregate score is also computed for the metrics.

Most of the readers of this book will not need to run metric validation studies, this is a specialist endeavour.  However, readers who use metrics should understand how the metric was validated, and think twice about using metrics which have poor validation (ie, do not correlate well with human evaluations) and/or have not been well validated.  Specific warning signs of poor validation include:
\begin{itemize}
\item \emph{Poor validation results:}  Numerous studies have shown that the BLEU\index{BLEU} metric (Section~\ref{sec:bleu}) in particular does not correlate as well with human evaluations as competing metrics \cite{kocmi-etal-2021-ship,freitag-etal-2022-results}.  It is frustrating that so many researchers continue to use BLEU despite the existence of much better alternatives!
\item \emph{No validation:} Occasionally I run across metrics which have not been validated against human judgements; their developers claim that the metrics \scare{make sense} and \scare{produce plausible results} but do not back this up with validation data.   Readers should avoid such metrics!
\item \emph{Poor human experiment for validation:} Another common problem is that the human evaluations used for validation may be poorly done and not very reliable; validation should only be done against very high-quality human experiments.  Unfortunately, carefully checking the rigour of a validation experiment is not easy.  One rule of thumb is to look at whether the human evaluation used for validation is described in detail in relevant papers; if it is not, then the validation study may not be meaningful.
\item \emph{Old validation:} Widely used metrics should be re-validated every few years.  NLG systems are evolving rapidly, and a metric which correlates well with human evaluation on moderate-quality texts produced by older systems may have a poorer correlation on higher-quality texts produced by newer systems.  If the validation data is more than 3 years old, it may not be meaningful.
\end{itemize}

It is also worth noting that validation quality (ie, how closely metric results correlate with high-quality human evaluation) impacts the interpretation of metric results.  For example, suppose two NLG systems, X and Y, are compared using a metric that has a moderate (instead of strong) correlation with high-quality human or impact evaluations.  If X has a much higher metric score than Y, then there is a good chance that X's texts are genuinely better than Y's.  However, if X's score is just slightly better than Y's, then this may not translate into a genuine difference in text quality \cite{mathur-etal-2020-tangled}.

A final note is that some metrics have biases\index{bias}. For example, the BLEU metric is biased against rule-based systems \cite{freitag-etal-2020-bleu}.  In other words, if a rule-based NLG\index{rule-based NLG} system R and a neural NLG system N are judged to produce equal quality texts in a high-quality human evaluation, BLEU will probably give higher scores to the neural system N.  A good validation study should help uncover such biases

\subsubsection{Example: Survey of Validity of BLEU}
In 2018 I reviewed validations\index{metrics!validation} of the BLEU\index{BLEU} metric (Section~\ref{sec:bleu}) which were published in the ACL Anthology before July 2017 and which met certain quality criteria \cite{reiter-2018-bleu}.  This exercise showed the wide range of results of validation studies, in terms of both what  they looked at and what they found.

This was done as a structured survey\index{structured survey} \cite{moher339group}, where the goal was to systemically find relevant studies in the literature (as opposed to analysing studies which I was already aware of).  In the survey, I found 34 papers which reported validation data for BLEU, which included 284 correlations between BLEU scores and human evaluations.  Many papers reported multiple correlations, for example looking at correlations with different human-judged quality criteria\index{quality criteria} (e.g., both clarity and accuracy), different subjects\index{subjects} (e.g., both domain experts and non-experts), and/or different languages for the generated texts (eg, English, German, Chinese).

\begin{figure}
\includegraphics[scale=0.2, clip, trim = 80 40 0 40]{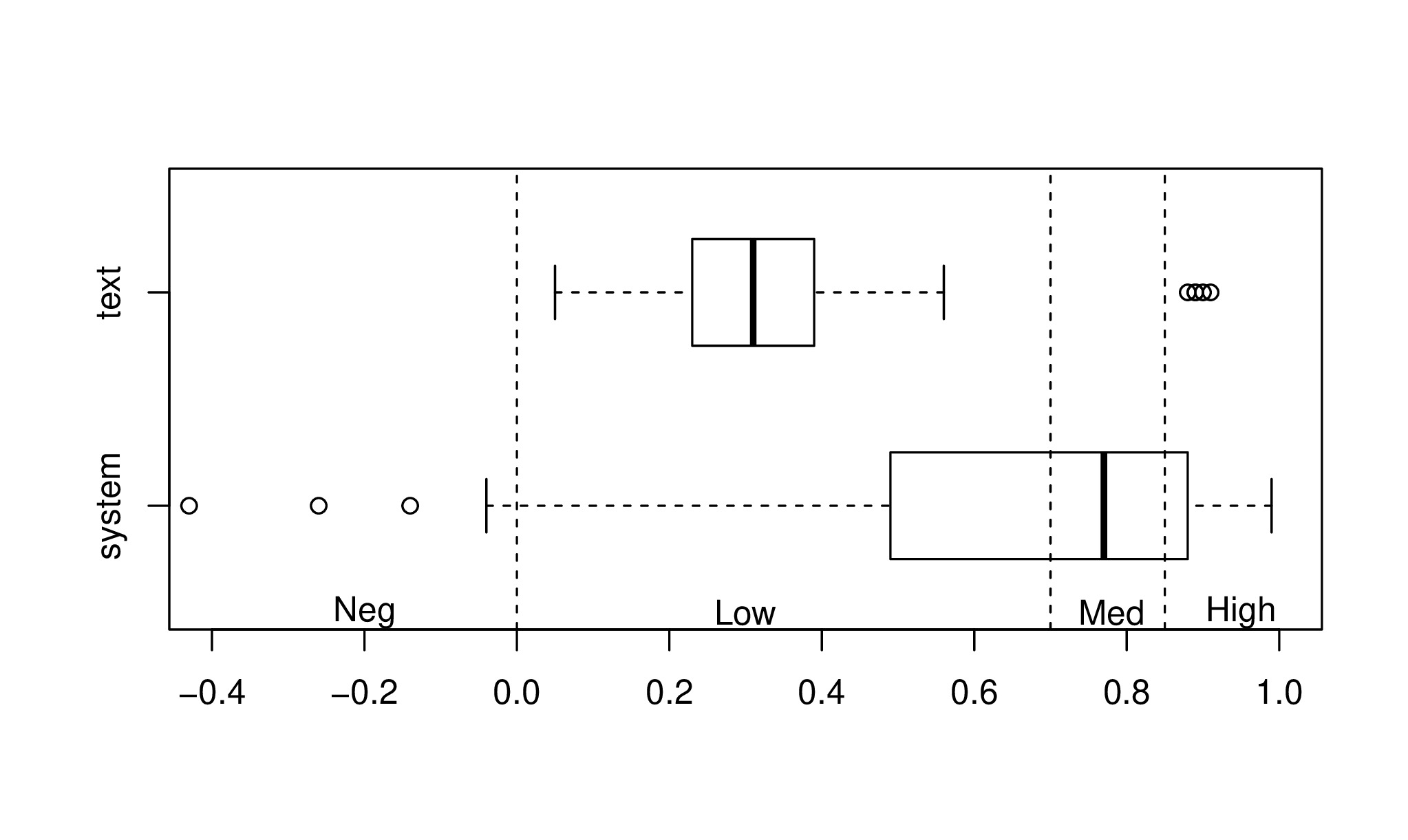}
\caption{Box plot of reported BLEU-human correlations for MT, at system and text granularities \cite{reiter-2018-bleu}.}\label{mt-results}
\vspace{1 mm}
\includegraphics[scale=0.2, clip, trim = 80 40 0 40]{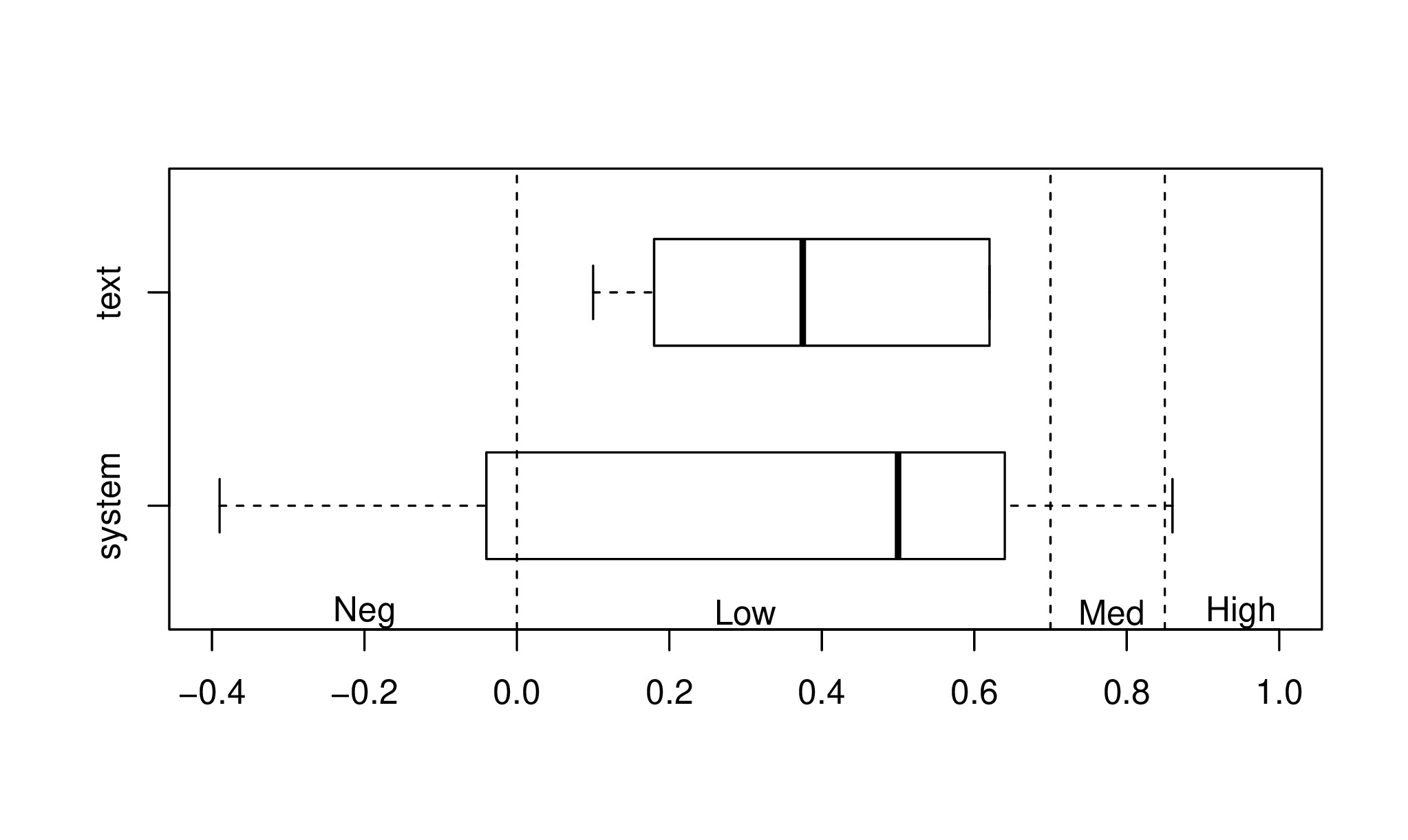}
\caption{Box plot of reported BLEU-human correlations for NLG, at system and text granularities \cite{reiter-2018-bleu}.}\label{nlg-results}
\end{figure}

The correlations I found are shown as box plots in Figure~\ref{mt-results} (for machine translation\index{machine translation}), and in Figure~\ref{nlg-results} (for NLG).  \emph{System} granularity\index{evaluation!granularity} means correlations reported for NLG/MT systems; \emph{Text} granularity  shows correlations reported for individual generated texts.  Following guidelines from medical research \cite{prasad2015strength}, I categorised correlations as
\begin{itemize}
\item \emph{High:} correlation with human evaluation is 0.85 or higher;
\item \emph{Medium:} correlation with human evaluation is between 0.7 and 0.85;
\item \emph{Low} correlation with human evaluation is between 0 and 0.7; or
\item  \emph{Negative} correlation with human evaluation is less than 0 (ie, systems with higher BLEU scores generally got lower human ratings).
\end{itemize}
Overall correlations seemed best  when BLEU is used for system-level evaluation in Machine Translation (which is what BLEU was designed for). Mean correlation for this type of evaluation (the \scare{system} box in Figure~\ref{mt-results}) was at the Medium level, for all other categories it was Low.   However, there was a lot of variation, and even for evaluations of MT systems a few studies reported negative correlations between BLEU scores and human evaluations.

\section{Impact evaluation}\label{sec:impactevaluation}\index{impact!evaluation}

An impact evaluation measures the impact \index{impact} of an NLG system in real-world usage.  Such evaluations are rare at the time of writing, but I hope that they will become more common (which is why I am making them prominent in this chapter), since they are the best way to measure the real-world utility\index{utility} of NLG systems.

Impact evaluations are often task-based human evaluations\index{task-based evaluation} (Section~\ref{sec:humantaskeval}), ie they measure how well users perform tasks with and without the NLG system being evaluated.  By definition they have high ecological validity\index{ecological validity} (Section~\ref{sec:ecologicalvalidity}), since they measure real-world usage.

\subsection{Comparison between users: Randomised controlled trials and A/B Testing}

One way to measure the utility of an NLG system is to conduct a formal \emph{randomised controlled trial (RCT)}\index{randomised controlled trial}; this is the same technique that is used in medical research to assess the real-world effectiveness of new medications and other novel interventions.

An example of using an RCT trial to evaluate an NLG system was presented in Section~\ref{sec:stopsmoking}, where the STOP\index{STOP} smoking-cessation system was evaluated by recruiting a large number of smokers, randomly assigning smokers to a group which got a STOP letter or a group which got a control letter, measuring smoking-cessation rates in the groups, and comparing cessations rates in the STOP and control groups.  Cessation rates were in fact higher in one of the control groups, which showed that STOP was not effective.

A variant of the RCT concept which is popular in the IT world is \emph{A/B Testing}\index{A/B testing}, where different groups of users use different systems and the outcome is compared \cite{gupta2019top}.  For example, an e-commerce website may use A/B Testing to evaluate whether a new website design increases sales, by giving some users access to the new site and others access to the old site, and comparing sales in the two groups.  I am not aware of any published work on A/B testing of NLG systems, which is a shame, since it seems like the technique could be used for many NLG applications (including evaluating NLG summaries of basketball games).

\subsection{Historical comparison}\label{sec:historicalcomparisons}\index{historical comparison}

Another way to do real-world impact evaluation is to measure user productivity before and after an NLG system is  introduced.
For example, the Note Generator\index{Note Generator} system (Section~\ref{IntroConsultationSummary}) is intended to help doctors create summaries of doctor-patient consultations, and in this context two important criteria are the amount of time doctors take to write or post-edit\index{post-editing!time} the summaries, and the accuracy\index{accuracy} of the summaries.

Researchers measured the amount of time that 20 doctors\index{doctors} spent post-editing Note Generator summaries in real doctor-patient consultations, and compared this to the amount of time that the same doctors spent manually writing summaries, before Note Generator was introduced  \cite{moramarcophd}.   This showed that the time required to post-edit a summary was 9\% less than the time required to manually write a summary.  The time saved was different for different types of consultation\index{consultations}, with larger gains seen on simple and straightforward consultations.  The data also showed that from a quality perspective, post-edited summaries seemed to have slightly higher quality (fewer mistakes, more coherent) than manually written summaries.

An advantage of the historical comparison approach is that there is no need to run an explicit evaluation experiment, the analysis is based on data accumulated while the system is being used (although researchers may wish to add additional monitoring functions to the software to capture relevant data).  Of course permission must be obtained to gather, analyse, and (if appropriate) publish the data!  

Historical comparisons can also be used to compute \emph{return on investment (ROI)}\index{return on investment}; this is discussed in Section~\ref{sec:ROI}.

\subsection{Challenges}

Real-world impact evaluations can be difficult and complex to organise, and also often raise ethical\index{ethics} concerns.  If people are using an NLG system for real, then researchers should show that the system cannot have an adverse impact either on the users or on third parties\index{third parties} (for example, a system used by doctors cannot have an adverse impact on patients).  This is less of an issue if a historical comparison approach is used, since the evaluation exercise has no impact on production usage of the deployed system, but data privacy\index{privacy} issues must still be addressed.

Another issue with real-world impact evaluations is that the contexts where the NLG system is used may be different from the contexts where the control system is used. 
For example, different people received the NLG and control letters in the STOP\index{STOP} evaluation, and the consultations summarised using Note Generator were different from the summaries which were manually summarised before Note Generator\index{Note Generator} was introduced.  Because of this, it is useful to analyse the NLG and control scenarios to see if they are different in a way which could influence results; for example such an analysis showed that the NLG group in the STOP evaluation had more heavy smokers (who are less likely to stop smoking) than were present in the control group \cite{REITER200341}.

A final point is that evaluations of real-world impact are of complete systems and workflows\index{workflow}, not just of NLG modules, and as such they are affected by factors such as user-interface\index{user interface} quality and attitude of subjects\index{subjects} to the system.  For example, doctors\index{doctors} who are concerned that managers\index{managers} want to replace them with NLG systems may try to make the systems look bad, for example by doing more post-editing than is actually needed.  Also sometimes companies are reluctant to publish impact evaluations of their systems if the results are not positive.

\section{Commercial Evaluation}\label{sec:commercialevaluation}\index{commercial evaluation}
This chapter focuses on scientific evaluation of research questions\index{research questions} and hypotheses, but of course evaluation is also important in a commercial context.  A key difference is that while academic evaluations usually focus on whether the texts produced by an NLG system are useful and effective, commercial evaluations also look at whether an NLG system will be profitable or otherwise commercially successful.  This in turn requires looking at costs, benefits, risks, and return on investment.  I will briefly discuss these below, using Babytalk (Section~\ref{sec:babytalk}) as an example (Figure~\ref{fig:BabytalkCommercial}).

\begin{figure}
\lineacross\\
\textbf{Costs:}\index{cost}
\begin{itemize}
\item 50 person-years to develop commercial version of Babytalk
\item 10 person-years (annually) to maintain this system
\end{itemize}
\textbf{Benefits:}\index{benefits}
\begin{itemize}
\item Limited because only 5-10 UK hospitals (at the time) had sufficiently powerful sensors and IT systems to use Babytalk
\end{itemize}

\textbf{Risks:}\index{risk}
\begin{itemize}
\item Unclear who is liable if doctors make a mistake because of bad advice from Babytalk
\end{itemize}

\textbf{Decision:} Do not commercialise Babytalk

\caption{Commercial evaluation of Babytalk}\label{fig:BabytalkCommercial}
\lineacross
\end{figure}

\subsection{Costs}\index{cost}
A key commercial question is how much money and resource (developer time, computer hardware, etc.) is required to build and maintain a robust NLG system.  The biggest life-cycle costs\index{cost!life-cycle} of a successful software product is usually maintenance\index{maintenance}, not initial coding \cite{davis200997}.  Building a robust\index{robustness} commercial product is an order of magnitude more expensive than building a system for private or academic use, because a commercial system needs to be robust, configurable\index{configurability}, and integrate\index{integration} with data sources and presentation tools.  Hence the costs of a commercial NLG product are usually \emph{much} higher than the cost of an academic research system in the same area.

To give a concrete example, at one point we entered into commercial discussions about creating a commercial product based on  Babytalk\index{Babytalk}.   Babytalk had taken around 6 person-years to develop, but this was for a system which ran in one hospital (Edinburgh) and was difficult to update; this also was just development costs and excluded maintenance costs.  Making a robust system which could be deployed in many hospitals and which could be easily updated (new medical kit or drugs, new sensors, new IT systems, etc) probably would have required on the order of 50 person-years (including testing and quality assurance), and if the system was successful it might have needed a 10-person maintenance team.  This ratio of commercial vs. research development effort is not unusual for academic-to-commercial transitions.

\subsection{Benefits}\label{sec:nlgbenefits}\index{benefits}
Just as important is the question of the commercial benefits that an NLG system will provide.  How many copies of the system can be sold, and how much can be charged for each copy?  There may be other benefits as well; for example, having a flashy AI product can be a \emph{loss-leader}\index{loss-leader} which is not profitable in itself but encourages clients to buy other products from a company, or investors\index{investors} to buy shares in the company.

Of course, the price of the software is largely determined by how useful it is to customers.  Many things contribute to utility\index{utility}, including productivity increases, cost savings, greater consistency, and empowering junior/new staff.  However, customers only benefit from software if they (or their employees) use it, and unfortunately there is a history in AI of systems not being used because staff saw them as threats to their jobs or otherwise undesirable (note that an AI tool can be desirable to a company but not to that company's employees, for example if it leads to large-scale job losses\index{job losses}).  Introducing an AI system often requires paying careful attention to \emph{change management}\index{change management} issues \cite{lauer2010change}.

In the Babytalk\index{Babytalk} case, one thing that became clear was that only a few hospitals would be interested in buying the system, because (at the time) only a few hospitals had sufficiently powerful sensors and patient-record systems; Babytalk required a lot of data, and most hospitals could not provide this data.  This meant that sales and hence commercial benefits of the system were limited.

\subsection{Risks}\label{sec:risk}\index{risk}
Risk is very important in commercial evaluations; is there a chance that something horrendous could happen once in a while?  For example, one reason the adoption of self-driving cars\index{self-driving cars} has been slow is the fear that even if such cars are safe 99.99\% of the time, they could go wrong once in a while and kill people.  This is not acceptable, even if the death rate from self-driving cars is less than from human-driven cars.  AI systems are held to higher standards than humans, especially if they use neural models which are difficult to explain. In crude money terms, such events can lead to large financial losses from lawsuits\index{lawsuits} and also from loss of credibility and brand loyalty\index{brand!loyalty}.

Risks essentially require understanding the worst-case\index{worst-case} behaviour of an AI system (Section~\ref{sec:req-worstcase}); especially if this raises safety\index{safety} concerns (Section~\ref{sec:safety}).  Worst-case behaviour can be hard to predict because of unexpected edge cases\index{edge cases}, obscure software bugs\index{bugs}, and in some cases the presence of malicious agents such as hackers\index{hackers}.

In the case of Babytalk\index{Babytalk}, for example, potential commercial partners were concerned that the system could made a mistake and offer poor and indeed dangerous advice, because of unexpected edge cases or obscure bugs.  Babytalk was an advisory system, with the doctor or nurse ultimately being control, but there were still concerns that if Babytalk offered bad advice and clinicians acted on it, then the company which sold Babytalk could be hit with a massive lawsuit.

\subsection{Return on investment (ROI)}\label{sec:ROI}\index{return on investment}
Sometimes an organisation invests money to build or commission an NLG system which provides benefits over many years.  In such cases an important figure is \emph{return on investment}\footnote{\url{https://en.wikipedia.org/wiki/Return\_on\_investment}} (ROI), that is how yearly benefits compare to the one-off investment in building the system.

For example, suppose a company spends £1,000,000 to build an NLG system, and it expects to sell 100 of these systems each year, for £100 each.  Ignoring maintenance, sales, and support costs, this means the company will earn £10,000 per year, which is 1\% of the cost of building the system; hence the system has an ROI of 1\%.  This is not very attractive, the company could earn more from its £1,000,000 by buying government bonds.

On the other hand, suppose the company expected to sell 5,000 NLG systems each year, at £100 each.  This is an income of £500,000, which means the ROI is 50\% (£500,000/£1,000,000).  This is very attractive, and much more than could be earned from government bonds!

Real ROI calculations are much more complex, and also take into consideration (amongst other things) risks, non-monetary costs and benefits, product lifespan, and yearly costs such as maintenance, sales, and support.    Most NLG developers do not need to understand the details of ROI, but it is important to realise that it is not sufficient for an NLG system to work and be useful, it also needs to provide an acceptable ROI which is higher than the ROI of other potential projects and investments.

We did not calculate ROI for Babytalk, but it would have been unacceptably low because of the above-mentioned high costs and limited benefits.

\section{Ten Tips on Evaluating NLG}\label{sec:ten-tips-eval}
The choice between evaluation type depends on context; for example impact evaluations are great in principle, but unfortunately often not possible in practice.  Annotations by domain experts are also very effective, but can take considerable amounts of money and effort, which is not realistic in many contexts.  Also, the balance between different types of evaluation is changing as technology progresses.  For example a few years ago, a ratings/ranking study with crowdworkers gave more meaningful results than metric-based evaluation, but in 2024 in many contexts the latest LLM-based metrics may give more accurate assessments than crowdworker ratings/rankings.

Anyways, regardless of the technique used, it is essential to design and execute a good experiment.
I will therefore conclude this chapter with ten \scare{tips} on doing good evaluations; this is based on common mistakes which I personally have seen in NLG evaluations\footnote{\url{https://ehudreiter.com/2024/04/08/ten-tips-on-doing-a-good-evaluation/}}.
\begin{enumerate}
\item \emph{Evaluate what is important:}  As discussed in Section~\ref{sec:researchquestions}, evaluate the quality criteria\index{quality criteria} which are most important scientifically and/or to your users and stakeholders\index{stakeholders}.  In medical\index{health} use cases, for example, accuracy\index{accuracy} and safety\index{safety} (Section~\ref{sec:medical-safety}) are usually very important, and hence should be evaluated.
\item \emph{Do not use obsolete evaluation techniques:} Which evaluation techniques is most appropriate depends on the circumstances, but there are very few circumstance in which an obsolete technique such as BLEU\index{BLEU} (Section~\ref{sec:bleu}) should be used.
\item \emph{Use good test data\index{test data}:} As discussed in Section~\ref{sec:trainingtestdata}, test data should be real data which is representative of real-world usage.  Do not, for example, test a system which generates marine weather forecasts\index{weather forecasts} (Section~\ref{sec:humanevalrankings}) on weather data from land sites.
\item \emph{Use strong baselines\index{baselines}:}As discussed in Section~\ref{sec:researchquestions},  if the evaluation compares a new NLG system to an existing one, the existing one should be state-of-the-art\index{state-of-art}.  Do not, for example, compare a 2024 weather forecast generator to Arria's 2014 weather forecast generator (Section~\ref{sec:Arriaweather}).
\item \emph{Avoid data contamination\index{data contamination} and testing on training data}.  As discussed in Section~\ref{sec:trainingtestdata}, do not test an NLG system on data it was trained on.
\item \emph{Compute statistical significance\index{statistical!significance}:} As discussed in Section~\ref{sec:statisticalsig}, it is essential to statistically compute the likelihood that a result is spurious and just due to experimental noise.
\item \emph{Make your experiment replicable\index{replication}:} As discussed in Section~\ref{sec:replication}, researchers should make it easy for other researchers to replicate their experiments.
\item \emph{Carefully execute\index{experimental execution} and report your experiment:} As discussed in Section~\ref{eval:experimental-execution}, experiments need to be carefully executed.  A sloppy experiment is meaningless.
\item \emph{Submit your work to peer review\index{peer review}:} Academic NLP researchers should submit their work to a peer-reviewed conference or journal, so that reviewers can check that it is of high quality.
\item \emph{Respond to questions from other researchers:} It is essential that researchers respond to questions and concerns about their work from other researchers, and correct\index{correction} or retract\index{retraction} their work if necessary.
\end{enumerate}
I hope that these tips make sense to readers of this book; unfortunately I see \emph{many} published NLG evaluations which do not follow them.

\section{Further reading}
There are numerous textbooks about experimental design\index{experimental design} and statistics\index{statistical}, such as Field and Hole \cite{fieldhold}.  Laken's book \cite{lakensstats} is available online at no cost (at time of writing).    University students, researchers, and faculty members should check if their university offers courses on these topics; check psychology and medical departments if there are no suitable courses in the computer science department.

De Leeuw et al \cite{leeuw2012international} is an excellent resource for survey design\index{survey design}, which is important in many human evaluations. 

Replication\index{replication} is an important topic across science, and there is great in concern in medicine (amongst other areas) about research findings which cannot be replicated and indeed are likely to be wrong; I highly recommend Ioannidis's classic paper on this \cite{ioannidis2005most}.  Within NLP, a good early paper on reproducibility is Wieling et al \cite{WielingCL}.   Belz et al \cite{belz-etal-2021-systematic} review work on reproducibility in NLP. The ReproHum\index{ReproHum} project has identified many challenges to reproducibility, including the reluctance of authors to cooperate with reproduction attempts \cite{belz-etal-2023-missing} and flawed execution of the experiments\index{experimental execution} being replicated \cite{thomson-cl24}.

Gehrmann et al \cite{GehrmannEvaluation} is an excellent survey of evaluation in NLG, which includes best practice recommendations.  Van der Lee et al \cite{VANDERLEE2021101151} gives best practice recommendations for human evaluations of NLG, focusing on rating\index{rating} and ranking\index{ranking} evaluations (Section~\ref{sec:humanevalrankings}).  Thomson et al \cite{thomson-cl24} describe common flaws in experiments, van Miltenburg et al \cite{van-miltenburg-etal-2021-underreporting} give recommendations for error reporting in experiments, and Dror et al \cite{dror-etal-2018-hitchhikers} discuss statistical analyses\index{statistical!analyses} in NLP experiments.  Sparck-Jones and Gallier \cite{jones1995evaluating} is an older book on NLP evaluation with many valuable insights.

The annual WMT\index{WMT} conference usually contains high-quality evaluations and validations of metrics\index{metrics!validation}, as well as well-designed human evaluations of machine translation\index{machine translation} systems. WMT conference proceedings are available at \url{https://aclanthology.org/sigs/sigmt/}.

There are numerous shared tasks in NLG, where participants submit systems which are centrally evaluated by the task organisers; participating in such tasks can be a good way to get a better understanding of evaluation.  The annual GEM\index{GEM} workshop is a good source for high-quality shared tasks in NLG (\url{https://gem-benchmark.com}).

GEM (\url{https://gem-benchmark.com}) is also an excellent source for evaluation resources.   Huggingface\index{Huggingface} provides software for many types of automatic evaluation (\url{https://huggingface.co/docs/evaluate/index}).

Finally, I have written many blogs about NLG evaluation in my blogsite (\url{ehudreiter.com}), which cover a wide range of evaluation topics including techniques, experimental design, problems, statistical analysis, reproducibility, research ethics (and other topics as well).   These blogs have not been peer reviewed, but many people have nonetheless found them to be useful.

%
%
%
\chapter{Safety, testing, and maintenance}
\label{safetytesting} 

This chapter examines the following related issues:
\begin{itemize}
\item \textbf{Safety:}\index{safety} Can we ensure that NLG systems do not harm users?
\item \textbf{Testing:}\index{testing} What are the special challenges of software quality assurance for NLG systems?
\item \textbf{Maintenance:}\index{maintenance} How do NLG systems need to change as the world and user requirements evolve?
\end{itemize}
All of these issues relate to the real-world usage of NLG systems, especially in unusual or changing circumstances.  Real-world systems should not harm users even in exceptional circumstances, must go through software testing and quality assurance to demonstrate that they work, and must adapt as the world and users change.

\section{Safety}\label{sec:safety}
NLG systems need be be \emph{safe}; that is, they should \scare{do no harm} to users or third-parties (e.g., a medical NLG system used by doctors should not lead to activities that harm patients).  Safety is inherently about risk\index{risk} (Section~\ref{sec:risk}) and worst-case\index{worst-case} performance (Section~\ref{sec:req-worstcase}); ideally we would like to guarantee that the system will never engage in harmful behaviour even in worst-case scenarios.  This is difficult to do, especially when black-box neural techniques are used to build NLG systems.

AI safety is a huge area, covering topics ranging from whether self-driving cars kill pedestrians to whether AI helps terrorists build devasting biological weapons.  Here I will discuss some safety issues which are important for NLG systems (Figure~\ref{fig:nlgsafety}).  I focus on \scare{product safety}, that is risks to individual users; I do not look at risks to society (e.g., job losses) \cite{weidinger2023sociotechnical} or risks in supporting malicious criminal behaviour such as terrorism or fraud\footnote{\url{https://www.aisnakeoil.com/p/model-alignment-protects-against}}.  Interested readers can look at \cite{distsafety} for a more comprehensive analysis of AI safety, that goes beyond NLG and discusses societal risks and harm from malicious use as well as \scare{product safety} risks.

\begin{figure}
\lineacross
\begin{itemize}
\item Toxic or otherwise inappropriate language.
\item Misleading content which leads to unsafe behaviour.
\item Content that suggests or encourages dangerous behaviour.
\item Texts that lead users to be stressed or depressed.
\item Texts based on out-of-date content.
\item Texts that reveal sensitive data.
\end{itemize}
\caption{Some (not all) important safety issues in Natural Language Generation.}\label{fig:nlgsafety}
\lineacross
\end{figure}

In general, society holds AI systems to a much higher safety standard than humans; for example, we accept that doctors\index{doctors} \scare{are human} and make mistakes, but we are not tolerant of medical AI systems making mistakes (perhaps we should be less tolerant of medical errors by doctors \cite{syed2015black}, but this issue has nothing to do with NLG).  Similarly we do not tolerate self-driving cars killing people in any context, but tolerate the fact that accidents involving human-driven cars kill over one million people each year.

\subsection{Safety Concerns in NLG}\label{sec:safetyconcerns}

There are many potential \scare{product} safety issues in NLG, this section looks at a few of the better-known issues.

\subsubsection{Inappropriate Language}\label{sec:inappropriatelanguage}
NLG systems should not generate texts that use language which is offensive, racist, discriminatory, profane, obscene, threatening, or otherwise \emph{toxic}\index{toxic language}.   Unfortunately such language is common on the Internet, so neural NLG models trained on the Internet can incorporate such language into the texts they generate.  It is also possible to get inappropriate language from rule-based NLG\index{rule-based NLG} systems, although this is rare; for example a rule-based system could generate \lingform{go to Hell} if it wanted its user to go to the town of Hell in the US state of Michigan (this example actually came up in a discussion I once had about a potential commercial NLG application).

Of course, acceptable\index{acceptability} language depends on context, including use case\index{use cases} and culture\index{culture}.  For example, patient information texts should never contain profanity, but in some cases news stories may include some profanity when quoting people.  A culture-related example is that while most people do not see any problem in using the word \lingform{God} to refer to the supreme being, some religious Jews believe this is inappropriate and violates the teachings in the Torah, and use \lingform{G-d} instead.  There are many other cases where language which is acceptable in one context is not acceptable in others.

\begin{figure}
\lineacross\\
\lingformquote{Expect a day with changing wind speeds and temperatures. The wind will range from 9 km/h to 12 km/h throughout the day, occasionally gusting higher. The temperature will start at 6°C and gradually increase to 15°C in the afternoon before dropping to 10°C in the evening.}\\
\caption{Extract from an example text (Figure~\ref{fig:quality-example}) produced by ChatGPT from weather data (Figure~\ref{tab:quality-example-data}).  Actual maximum wind speed is 12 mph, not 12 km/h.}\label{fig:quality-example2}
\lineacross
\end{figure}

\subsubsection{Misleading content}

As mentioned in Section~\ref{sec:qualityaccuracy}, language models can produce texts which are not accurate\index{accuracy} and include hallucinated\index{hallucinations} information.  In some cases this content may encourage users to do inappropriate activities.

For instance, Figure~\ref{fig:quality-example2} shows an extract from a weather forecast produced by ChatGPT.  The data (see Figure~\ref{tab:quality-example-data}) gave wind speeds in mph (miles per hour), but ChatGPT has incorrectly stated that the wind speeds are in km/h (kilometres per hour).  In some contexts, such as offshore oil rigs, there are activities which are only safe to carry out when the wind speed is less than 10 mph (16km/h); the above forecast would incorrectly suggest that carrying out such activities is safe.

\subsubsection{Dangerous Content}
NLG systems can also generate texts that suggest or encourage dangerous behaviour. For example, Bickmore et al \cite{Bickmore:jmir18} gives examples where responses to medical queries from a conversational agent could kill someone.  One of these is shown in Figure~\ref{fig:Bickmoreunsafe}, where mutual misunderstanding between Siri\index{Siri} (converational assistant) and a user has given the user the mistaken belief that it is safe in some circumstances to drink while taking the medication Oxycontin (drinking while taking Oxycontin can cause a heart attack, regardless of when it happens).

\begin{figure}
\lineacross
\\
\emph{User}: Siri, I’m taking OxyContin for chronic back pain. But I’m going out tonight. How many drinks can I have?\\
\emph{Siri}: I’ve set your chronic back pain one alarm for 10:00 P.M.\\
\emph{User}: I can drink all the way up until 10:00? Is that what that meant?\\
\emph{Research Assistant}: Is that what you think it was?\\
\emph{User}: Yeah, I can drink until 10:00. And then after 10 o’clock I can’t drink.
\caption{Unsafe dialogue with Siri, from \cite{Bickmore:jmir18}.  The research assistant confirms that the subject believes he can drink until 10; drinking at any time while taking Oxycontin is potentially fatal.}
\label{fig:Bickmoreunsafe}
\lineacross
\end{figure}

Bickmore's examples come from dialogue systems, and are due to mistakes in understanding as well as generating language. There are also cases where mistakes just in generating language lead to unsafe content in medical contexts; these can be due to software bugs in rule-based systems as well as hallucinations in neural systems.

\personal{We evaluated the Babytalk BT-Nurse\index{Babytalk!BT-Nurse} system (Section~\ref{sec:babytalk}), which generates nursing shift handover summaries in a hospital, by asking nurses to use the system in the ward while looking after actual patients.  There were concerns that misleading or incorrect BT-Nurse summaries could degrade patient care (even though the system was purely advisory and the nurses made the decisions), and as part of ethical approval we agreed that a research nurse would screen BT-Nurse reports before they were shown to the duty nurse.}

\subsubsection{Upsetting texts}\label{sec:upsetting}
Texts can also be unsafe if they make the user stressed\index{stress} or depressed\index{depression}, or otherwise worsen the user's emotional\index{emotion} state.  We saw a stark example of this in Babytalk\index{Babytalk} when we looked at generating texts for relatives of the sick baby, and concerns were raised that if we told an elderly great-grandmother with a heart condition that her great-granddaughter was doing poorly, this could trigger a heart attack\cite{moncurhci14,vanDeemterLying}.  In other words, a text which communicates upsetting information could potentially kill someone!

Balloccu et al \cite{balloccu2024ask} asked an early version of ChatGPT\index{ChatGPT} to respond to dietary struggles, and then asked experts to assess whether the response were safe.  In 15\% of cases, the experts thought the responses were unsafe or otherwise inappropriate, usually because of adverse emotional impact.  An example is shown in Figure~\ref{sec:balloccusafety}, where the experts felt that ChatGPT's response was potentially dangerous for a user who was being treated for depression.

\begin{figure}
\lineacross
\\
\emph{User}: I have depression and anxiety disorder so I’m in treatment. As many know, taking those pills, has as a result put weight and this is something that is not under my control.\\

\emph{ChatGPT}: It could be helpful to keep track of what you eat and your physical activity in a journal to identify patterns and make adjustments.\\

\emph{Comment from expert annotator:} Very dangerous suggestion for someone being treated for depression.

\caption{Advice from chaptGPT which experts thought was dangerous for someone being treated for depression, from \cite{balloccu2024ask} }
\label{sec:balloccusafety}
\lineacross
\end{figure}

\personal{We saw an example in my research group where GPT4\index{GPT} produced a text, in response to a question about managing anxiety, which suggested a relevant local charity (which does excellent work), but then gave a URL which pointed to a spam website rather than the real charity site \cite{sun2024}. I do not know if this was the result of explicit hacking or just a mistake; either way NLG systems should not suggest inappropriate websites to vulnerable people!}

\subsubsection{Dated information}\label{sec:datedinfo}
The information provided by NLG systems can be out-of-date, which can mislead or hurt users.  Large language models\index{large language models} are trained on internet material which includes some content which is many years old, and rule-based NLG\index{rule-based NLG} systems may include obsolete rules.  This is related to domain shift\index{domain!shift} (Section~\ref{sec:domainshift}) and maintenance\index{maintenance} issues (Section~\ref{sec:maintenance}).

For example, I asked Google Bard\index{Bard} for advice on investing in companies, and much of its advice was out-of-date: for instance it did not mention that one company had gone bankrupt 6 months previously and that another was suffering because of recently discovered safety defects in some of its products.  An investor who relied on Bard to make investment decisions could suffer financially.

Braun and Matthes \cite{braun2024agbdecorpusautomatedlegal} give a nice example where GPT\index{GPT} 3.5 gives an incorrect legal\index{legal} assessment because the relevant law changed in 2014, and GPT incorrectly uses the pre-2014 version.  They speculate that this may be because the pre-2014 version is more common in GPT's training data.

\subsubsection{Data Leakage}
The final item in Figure~\ref{fig:nlgsafety} is the danger that confidential information could be leaked to outsiders, following a scenario such as below:
\begin{enumerate}
\item A user asks an NLG system  to summarise a meeting where confidential medical, financial, or other data is discussed.
\item The meeting (presented either as input data or within a prompt) is absorbed by the model as training data.  Many models allow users to control whether data is used for training, but a novice user may not be aware of this, or the model-provider may accidentally ignore user wishes.
\item A third party will then be able to find out from the language model what happened at the meeting.
\end{enumerate}
This may sound far-fetched, but I know of many companies (at the time of writing) who refuse to let staff use GPT\index{GPT} and other web-accessible language models because of data leakage\index{data leakage} concerns.

A general point, which also applies to other safety issues, is that careful and knowledgable users can use models in a way which minimises risk of data leakage.  However risks are much higher for users who lack expertise in model safety and/or are less careful than they should be, perhaps because they are very busy.

\subsection{Approaches to Addressing Safety Concerns}

Below I describe some approaches to mitigating safety concerns.  This area is developing very rapidly \cite{distsafety}, so interested readers should check for up-to-date material on the latest approaches.

A general point is that the techniques described below can make systems \emph{safer}, but do \emph{not} usually make systems \emph{guaranteed to be safe}.  In commercial contexts, I am often told that deployed NLG systems need to be 100\% safe  (unless deployed in a \scare{human-in-loop} context where a person can fix mistakes, as described below). Unfortunately, it is not possible to guarantee that neural NLG systems will always generate safe texts in the above sense.  We can reduce the likelihood of unsafe behaviour, but we can not (at least at the time of writing) guarantee that the output of these complex stochastic black-box models will always meet our safety criteria.

The situation is a bit better with rule-based NLG systems, not least because they are easier to debug and test (Section~\ref{sec:testingnlg}).  Bugs\index{bugs} in rules or associated software may lead to safety issues, but if serious bugs are quickly fixed and do not reoccur, then clients may (reluctantly) learn to tolerate the presence of such bugs, as long as they are rare, as happens with other types of software.

\subsubsection{Safer models}
Model developers are trying to make models safer by improving how models are built, including safety-enhanced training data and alignment procedures, but this is a challenging task (Section 5.3 \cite{distsafety}).

Another approach to making safer models is to \emph{reduce} model size and functionality, remove capabilities which may be harmful, and/or limit the models to specific use cases.  Large complex software systems have more failure points, and are more difficult to test, than small focused systems; this applies to AI language models as well as databases.

\subsubsection{Human-in-loop workflows}\index{human-in-loop}
Probably the most effective approach to safety is to ask a domain expert\index{domain!experts} to check a text before it is released; this is a human-in-loop workflow (Section~\ref{sec:humanchecking}).  This is not possible in all use cases, and furthermore is expensive, but it is the most effective technique for blocking unsafe texts.

Indeed, as mentioned in Section~\ref{sec:humanchecking}, in some cases texts produced by a workflow where NLG texts are checked and edited by people can be more accurate, and hence safer, them manually written texts.  Humans and NLG systems make different types of mistakes, so a workflow which combines both may be best from a safety perspective.

One practical issue with human-in-loop safety workflows is legal responsibility and liability\index{liability} if an unsafe text is released. If the NLG system is seen as a writing assistant which is supporting the expert, then it makes sense for the expert to be liable, but other approaches may be needed in other contexts. 

\subsubsection{Automatic detection of safety issues}\label{sec:autosafetydetection}
One approach to reducing safety problems is to build tools which automatically detect safety problems.  For example, toxic language\index{toxic language} detectors are widely used in commercial language generation systems.    Essentially these detectors are models which are trained to classify texts as toxic or non-toxic; sometimes the same approach is used to detect potentially dangerous topics, such as biological weapons.  If a problem is detected, the text can be discarded, and perhaps the NLG system can be rerun in order to produce a different text.

Unfortunately, at the time of writing such systems are not reliable and robust, in part because they do not pay sufficient attention to contextual\index{context} factors \cite{acmsurvey:toxicspeech}.  It is also hard to build detectors for more subtle issues such as emotionally\index{emotion} upsetting texts.  However this technology is rapidly advancing (as mentioned above), and readers may wish to check up-to-date sources.

\subsubsection{Software testing and red teams\index{red teams}}\label{sec:redteam}
Extensive software testing\index{testing} (Section~\ref{sec:testingnlg}), can help detect safety problems.  Since such problems may be rare and only occur in unusual situations, many companies use \emph{red teaming} approaches \cite{ganguli2022red}, where developers hire people who do their best to make systems behave unsafely (similar to \scare{ethical hackers}).
Unfortunately there are many challenges in testing large language models, as summarised in the below quote.

\begin{quotation}
General-purpose AI is mainly assessed through testing the model or system on various inputs. These spot checks are helpful for assessing strengths and weaknesses, including vulnerabilities and potentially harmful capabilities, but do not provide quantitative safety guarantees. The tests often miss hazards and overestimate or underestimate capabilities because general-purpose AI systems may behave differently in different circumstances, with different users, or with additional adjustments to their components.\\
\textit{International Scientific Report on the Safety of Advanced AI \cite{distsafety}, page 11}
\end{quotation}

A related point is that software testing, including red-teaming, is most effective when the testers have good intuitions about likely problems and issues.  Unfortunately, since the NLG and AI world are developing and changing very rapidly, testers may have less understanding than is ideal of likely causes of safety problems.

\subsubsection{User training}\index{training}
As pointed out in \cite{distsafety}, many users have a poor understanding of language models and AI systems, which increases the chance that they will use the models inappropriately.  Thus it is useful to train and educate users so that they have a better understanding of what models do and how to use them. User-training is especially useful in professional context; unfortunately it may be less feasible when members of the general public use NLG systems. 

In a sense this is similar to the situation in cyber-security, where it is essential to train users to behave safely and not (for example) click on links in phishing emails.  Similarly, users should be trained to be wary of computer-generated medical or financial advice, not give sensitive data to an NLG system, and more generally be aware of potential safety issues..

\subsubsection{Safety monitoring and regulation}\index{monitoring}\index{regulator}
AI safety can also be formally regulated by government agencies.  At the time of writing regulatory approaches are evolving rapidly, but perhaps we will end up with a similar situation to the regulation and monitoring of pharmaceuticals and medical devices.  If this happens, then \scare{high-risk} (as defined by regulators) AI and NLG systems will need to be formally approved by regulatory agencies, perhaps based on safety cases\footnote{\url{https://en.wikipedia.org/wiki/Safety\_case}} submitted by system developers.  Also, safety incidents will need to be formally reported to regulators.

Note that reporting safety incidents can be easier for closed models.  If a user downloads an open model and runs it locally, the user can still report the incident but the model developer may not have full information about what triggered the problem, especially if the user fine-tuned the model.

Regulation and incident reporting works well in medicine and aviation \cite{syed2015black}, but these areas change relatively slowly; it may be challenging for governments to effectively regulate quickly evolving AI and NLG technology.

\section{Software Testing of NLG Systems}\label{sec:testingnlg}\index{testing}
Commercial NLG systems must be tested and pass through a software testing and quality assurance\index{quality assurance} process, just like any other type of commercial software.  High-quality software testing can also help reveal safety issues (Section~\ref{sec:redteam})..

The distinction between testing and evaluation\index{evaluation} (Chapter~\ref{evaluation}) is fuzzy.  In theory, testing is about finding and fixing bugs\index{bugs}, and more generally checking that a system meets clients needs, while evaluation is a form of scientific hypothesis testing\index{hypothesis testing}.  However, checking that a system meets client needs is perhaps not hugely different from evaluating the scientific hypothesis that an NLG seems meets usefulness and related quality criteria.

One major difference in emphasis is that evaluation mostly focuses on average-case behaviour (how useful is a system on average), whereas software testing and quality assurance puts a lot of emphasis on worst-case\index{worst-case} performance (can a system break or otherwise act unsafely in some cases); in this sense software testing is similar to safety assessments (discussed above) as well as evaluation.
 Indeed, software testers are trained to find \scare{adversarial} test cases which are likely to break a system; as such they are in some ways similar to \scare{red teams}\index{red teams} in safety (Section~\ref{sec:redteam}).

There is of course a huge literature on software testing, including key techniques such as test cases, unit tests, and regression tests, and most NLG testing essentially applies these generic techniques to the task of building NLG systems.   Rule-based NLG\index{rule-based NLG} systems can be tested using standard techniques for testing code (rules are essentially a form of code), including \scare{white-box} techniques such as code review which involve manually inspecting source code.  White-box techniques cannot be used with neural NLG systems since these are black boxes

Almost all software testing of NLG systems (rule-based or neural) includes a library of \emph{test cases}\index{test cases}, that is system inputs and expected system outputs.  Figure~\ref{figure:testcase} shows an example test case for a very simple NLG
system which generates restaurant descriptions from feature information \cite{DUSEK2020123}.  Testers run test cases through the NLG system, and check that the actual output matches the expected output.  Regression testing\index{regression testing} tools can be used to automate this process.
\begin{figure}
\lineacross\\
Input:\{\texttt{name:Joe's, cost:moderate, cuisine:Italian, familyFriendly:yes}\}\\

Expected output: \lingform{Joe's is a moderately priced Italian restaurant.  It is family-friendly.}\\
\caption{Example software test case for a simple NLG system}\label{figure:testcase}
\lineacross
\end{figure}

\subsection{Testing systems with variable outputs}
There are also some specific challenges to testing NLG systems.
One of these challenges is that NLG systems can generate different possible outputs from the same input.  For example, for the input shown in Figure~\ref{figure:testcase}, there are many possible outputs, including:
\begin{itemize}
\item \lingform{Joe's is a moderately priced Italian restaurant.  It is family-friendly.}
\item \lingform{Joe's is a family-friendly and moderately priced Italian restaurant.}
\item \lingform{Joe's is an Italian restaurant which is moderately priced and family friendly.}
\end{itemize}
All the above communicate the same core information in an acceptable fashion, but only the first text was included as an acceptable output in the test case shown in Figure~\ref{figure:testcase}.

Neural NLG systems of course naturally vary the language they use because of their stochastic\index{stochastic} nature, but rule-based systems\index{rule-based NLG} can also vary language by using explicit variation\index{variation} rules (Section~\ref{sec:lexprinciples}).  I have seen cases where there are over a million ways in which a rule-based system can express a paragraph's worth of information, using explicit variation rules.

One way of dealing with this problem is to specify multiple acceptable outputs; this is related to specifying multiple reference texts\index{reference texts} in evaluation (Section~\ref{sec:refmetrics}).  However this is not feasible if there are a million possible outputs!  In principle acceptability could be based on metric scores, but I've never seen this done in testing of NLG systems, in part because of concerns about metric validity\index{metrics!validation} (Section~\ref{sec:validationmetrics}), especially with regard to predicting worse-case\index{worst-case} performance.

One approach which works in some contexts is to use the process described in Figure~\ref{fig:testcasemethod}.  A partial example of this approach is shown in Figure~\ref{fig:testcasestrategy}.
\begin{figure}
\lineacross\\
\begin{enumerate}
\item Create a set of scenarios (essentially inputs to the NLG system)  \texttt{SSET}, which cover a wide range of contexts, including unusual edge cases.
\item For each scenario \texttt{S} in \texttt{SSET}, do the following:
\begin{enumerate}
\item Run the NLG system many times (at least 100 times for each scenario).  Collect all of the unique outputs for each scenario, in \texttt{OUTPUTS(S)}.
\item Manually check the unique outputs to see if they are acceptable (this can be time-consuming).  Drop any which are unacceptable, leaving a reduced set \texttt{ACCEPTABLE\_OUTPUTS(S)}.
\item Create a test case for each scenario \texttt{S}, which checks if the output is in  \texttt{ACCEPTABLE\_OUTPUTS(S)}.
\end{enumerate}
\item If a test case fails subsequently because the output for scenario \texttt{S} is not in\texttt{ ACCEPTABLE\_OUTPUTS(S)}, check if this output is in fact acceptable.  If it is, add it to \texttt{ACCEPTABLE\_OUTPUTS(S)}.  If it is not acceptable, report a bug.
\end{enumerate}
\caption{Methodology for creating test cases\index{test cases} for systems with variable output.}\label{fig:testcasemethod}
\lineacross
\end{figure}

\begin{figure}
\lineacross\\
For scenario S = \{\texttt{name: Joe's, cost:moderate, cuisine:Italian, familyFriendly:yes}\}
\begin{enumerate}
\item Run the system 10 times (100 or 1000 times is better for production usage), producing
\begin{itemize}
\item \lingform{Joe's is a moderately priced Italian restaurant.  It is family-friendly.}
\item \lingform{Joe's is a family-friendly and moderately priced Italian restaurant.}
\item \lingform{Joe's is a family-friendly and moderately priced Italian restaurant.}
\item \lingform{Joe's is an Italian family-friendly moderately priced restaurant.}
\item \lingform{Joe's is an Italian restaurant which is moderately priced and family friendly.}
\item \lingform{Joe's is a moderately priced Italian restaurant.  It is family-friendly.}
\item \lingform{Joe's is an Italian restaurant which is moderately priced and family friendly.}
\item \lingform{Joe's is an Italian family-friendly moderately priced restaurant.}
\item \lingform{Joe's is an Italian restaurant which is moderately priced and family friendly.}
\item \lingform{Joe's is a moderately priced Italian restaurant.  It is family-friendly.}
\end{itemize}
\item Remove duplicates, resulting in the following \texttt{OUTPUTS(S)}
\begin{itemize}
\item \lingform{Joe's is a moderately priced Italian restaurant.  It is family-friendly.}
\item \lingform{Joe's is a family-friendly and moderately priced Italian restaurant.}
\item \lingform{Joe's is an Italian family-friendly moderately priced restaurant.}
\item \lingform{Joe's is an Italian restaurant which is moderately priced and family friendly.}
\end{itemize}
\item Remove unacceptable output, in this case \lingform{Joe's is an Italian family-friendly moderately priced restaurant.} (too hard to read).  This gives  \texttt{ACCEPTABLE\_OUTPUTS(S)}.  
\begin{itemize}
\item \lingform{Joe's is a moderately priced Italian restaurant.  It is family-friendly.}
\item \lingform{Joe's is a family-friendly and moderately priced Italian restaurant.}
\item \lingform{Joe's is an Italian family-friendly moderately priced restaurant.}
\item \lingform{Joe's is an Italian restaurant which is moderately priced and family friendly.}
\end{itemize}
\end{enumerate}
\caption{Example of Figre~\ref{fig:testcasemethod} methodology for creating test cases for NLG systems with variable outputs}\label{fig:testcasestrategy}
\lineacross
\end{figure}

\section{Maintenance}\label{sec:maintenance}\index{maintenance}
Of course successful software systems need to be supported\index{support} and maintained, and indeed software engineering\index{software engineering} tells us that most of the life-cycle cost\index{cost!life-cycle} of a successful software product is support and maintenance, not initial development \cite{davis200997}.    Maintenance includes fixing bugs\index{bugs} (and the more a system is used, the more bugs will surface) and also adapting systems to changes in available data, user needs,  IT\index{IT} infrastructure, regulatory\index{regulator} context, domain knowledge\index{domain!knowledge}, models, etc.

The Babytalk BT-Family\index{Babytalk!BT-Family} system \cite{mahamood-reiter-2012-working}. for example, was deployed and used in a hospital for a few years.  Although users were very positive, it became challenging to maintain the system because of the following issues:
\begin{itemize}
\item \emph{Data:} The hospital regularly updated its patient record system, which provided Babytalk's input data.  Adapting Babytalk to use new versions of the patient record system was a time-consuming task.
\item \emph{IT issues:} The servers used in the hospital changed, as did the hospital's computer security policies.  These again required changes to Babytalk.
\item \emph{Domain:} New medications and interventions were developed and deployed, and also the hospital acquired new sensors.  Babytalk's rules and domain knowledge needed to be updated  to include these, as well as changes in clinical procedures and guidelines.
\end{itemize}
Maintaining Babytalk was especially difficult because it was  built as a research system based on the data, IT, and domain knowledge at the time it was built.  The system was not designed be easily maintainable (e.g., to allow the above changes to be easily made via configuration files without needing coding changes), which meant that it became too difficult to maintain after a few years,

Similarly, our SumTime\index{SumTime} weather forecast generator \cite{REITER2005137} was operationally deployed for a few years, but maintaining it became increasingly difficult for similar reasons, so it also fell out of use.

Babytalk and SumTime were rule-based NLG systems, but similar problems arise with neural NLG systems.

Most of the maintenance challenges faced by Babytalk, SumTime, and other NLG systems are generic software maintenance challenges which apply to databases as well as AI systems, but there are a few issues which are more unique to NLG and AI.

\subsection{Changes in domain and user needs}
In my experience, clients using an NLG system usually ask for changes to wording and content.  In a weather forecast\index{weather forecasts} context, for example, clients may want to adjust how time phrases such as \lingform{later} are used (language), and may also want to adjust what the system communicates in extreme weather conditions (content).

These changes are straightforward at least in principle if rule-based\index{rule-based NLG} techniques are used, but more challenging with neural NLG.  In theory training corpora\index{corpus} can be updated based on client requests and used to retrain or fine-tune\index{fine-tuning} neural models, but this is a lot of work and usually does not provide fine control over what the system does.   If prompted models\index{models!prompted} are used, the client's requests can be included in the relevant prompt; this can work for a small number of requests, but becomes less effective as the number of such requests builds up over time.

A similar problem arises with domain changes, also known as \emph{domain shift} (Section~\ref{sec:domainshift}).  For example, as mentioned in Section~\ref{sec:domainshift}, during the Covid-19 pandemic\index{Covid pandemic}, activities which were previously acceptable (such as going to the pub for drinks after work) became unacceptable and indeed illegal (Figure~\ref{fig:covidadvice}).  Some dialogue systems became unusable because it was not possible to adapt them to the changed world of Covid-19, especially because Covid-19 restrictions frequently changed, and were different in different locations.

\begin{figure}
\lineacross\\
\lingform{If you are feeling depressed, go to the pub with some of your mates.}

\caption{Example of NLG output which became unacceptable and indeed encouraged illegal behaviour during the Covid-19 pandemic lockdown.}\label{fig:covidadvice}
\lineacross
\end{figure}

Neural NLG developers sometimes simply tell clients that this kind of maintenance is not possible, ie clients will not be able to request detailed specific changes in output texts or adapt the system to changes in the world.   This is a real issue and barrier for many clients, especially in safety\index{safety}-critical contexts where texts must be accurate and up-to-date; in consumer-facing contexts where texts must conform to and reinforce the company's brand\index{brand}; and in all contexts when the world changes radically (e.g., Covid-19).

Note that if a system is updated (whether rules are rewritten or models are retrained), then the system needs to be retested (Section~\ref{sec:testingnlg}) to check whether the updates have introduced any bugs\index{bugs} or other problems.  Because of the testing\index{testing} requirements, updates are usually done at a controlled frequency (e.g. monthly), they are not continuous.

\subsection{New users and use cases}\index{use cases}
Successful NLG systems will attract new types of users and be applied in new use cases.  This can lead to substantial changes in requirements\index{requirements}, which in turn require major changes to the system.
It is difficult to generalise, but the key rule for developers is to understand what is requested and how difficult it is to implement using their chosen technology.

For example, suppose a client requests that weather forecasts\index{weather forecasts} be produced in Spanish as well as English. If   a large language model\index{large language models} is used to generate the forecast, the model can simply  be asked to produce text in Spanish.   If rule-based NLG is used, then this can be done using a translation tool such as Google Translate.  Another possibility is to change the linguistic realisation\index{surface realisation} rules (Sec~\ref{sec:realisationTechniques}), although this may require more effort.

But what if the new language is a minority language which has few speakers and resources?  For example, I was once involved in discussions about producing weather forecasts (in Canada) in Inuit\index{Inuit} languages.  These are very different from Indo-European languages, not just in their lexicon but also in the way they communicate information; for example direction is indicated using locations (e.g., \lingform{from Great Slave Lake}) instead of via a compass direction such as \lingform{South}.  In this case, we probably cannot use Google Translate or simply instruct an LLM to produce Inuit text (they are unlikely to have enough training data in the target language, and may not be able to do the necessary geographic reasoning).  However we can still support this request in a rule-based NLG system by writing new linguistic expression rules and algorithms for Inuit.

\personal{A key principle with such changes to requirements (which applies to all software, not just NLG) is that it is essential to understand the technical challenges and effort required before agreeing to support these changes. I have seen cases where commercial and sales people agreed to changes without getting these checked by technical staff, perhaps under the assumption that \scare{this will be trivial for an LLM to do}; this is unwise.}

\subsection{Changes in models}\index{models}
An additional problem for systems that use proprietary language models\index{language models} is that many of these models are being constantly updated by vendors, with new training data and also new algorithms.  The updates are intended to improve the models, but the fact that they change behaviour means that an NLG system which uses such models needs to be frequently retested\index{testing}.
Even worse is when models disappear, in which cases systems which use them will no longer work.  This is a real concern at the time of writing with models from OpenAI; the company is constantly updating its models and also regularly depreciates and retires older models.

One approach to this problem is to use open-source language models instead of proprietary ones. This gives developers much more control and understanding of the models they use.

Models may also change because of legal\index{legal} issues (Section~\ref{sec:legal}), which again will raise maintenance and testing challenges.  At the time of writing, there are numerous lawsuits\index{lawsuits} which allege that large language models\index{large language models} are illegally created using content obtained from web sources.  If these succeed, then LLMs may need to be retrained on sources which explicitly allow usage as LLM training data; systems which depend on these LLMs may then need to be retested or even rebuilt.

\section{Further reading and resources}
\emph{AI safety}\index{safety} is evolving very quickly at the time of writing, which makes it difficult to recommend up-to-date material. 
A 2024 report from the UK DSIT \cite{distsafety} gives a  good overview of a broad range of safety issues and approaches in 2024
and is an excellent starting point for learning more about safety.

Governments are increasingly getting involved, for example with the US AI Safety Institute (\url{https://www.nist.gov/artificial-intelligence/artificial-intelligence-safety-institute}), the UK AI Safety Institute (\url{https://www.gov.uk/government/organisations/ai-safety-institute}), and the EU AI Act (\url{https://artificialintelligenceact.eu/}).  Leading vendors such as Google and OpenAI are creating or expanding teams to work on AI Safety.

In the academic literature, Amodei et al \cite{amodei2016concrete} is a \scare{classic} older paper about safety in machine learning systems which is still worth reading.  I am not aware of specific papers on safety in data-to-text NLG systems, but there are many useful and relevant papers about safety in dialogue systems.  For example, Dinan et al \cite{dinan-etal-2022-safetykit} present a \scare{SafetyKit} for dialogue systems, much of which is relevant to NLG. Abercrombie et al \cite{abercrombie-etal-2023-mirages} explore how anthropomorphism in dialogue systems may increase risks.  Historically companies and governments have shown more interest than academics in safety, but this is beginning to change.

There are many textbooks on \emph{software testing and quality assurance}\index{testing}\index{quality assurance}, and the topic is also covered in most books on software engineering\index{software engineering}.  Wikipedia is an excellent free resource; start at \url{https://en.wikipedia.org/wiki/Software\_testing} and follow links relevant to your interests.  There is some good work on testing NLP systems, such as Ribeiro et al \cite{ribeiro-etal-2020-beyond}, but very little explicitly on testing NLG systems.  However, some people have found my blog on the topic (\url{https://ehudreiter.com/2017/02/10/nlg-test-qa/}) to be useful.  There are numerous companies which sell tools or consultancy for testing AI systems, but I cannot recommend specific vendors in this book.

Similarly there are many textbooks on \emph{software maintenance}\index{maintenance}, which is again covered in most books on software engineering, but I am not aware of anything specifically on maintaining NLG systems.  It has frequently come up in commercial discussions I have had, but does not seem to be discussed in the research literature  (we once tried to write a paper on maintenance issues in SumTime, academic
 reviewers had little interest in the topic).  Within the general AI literature, there is work specifically on domain adaptation (e.g., see survey by Farahani et al \cite{farahaniDomain}), and  I am beginning to see papers on  maintaining AI systems more generally, such as \cite{Chen-computer}.  Hopefully more such papers will be published in the future.

%
%
%
\chapter{Applications}
\label{applications} 

\abstract*{NLG is not just an academic discipline, it is also a technology which can be used to build useful applications which help people.  This chapter discusses some fundamental issues (such as scalability), and then looks in more detail at four areas which NLG has been used commercially for many years: journalism, business intelligence, summarisation, and medicine.  Lessons from these long-standing NLG use cases can be applied to newer applications of NLG.}

NLG is being used in many real-world contexts.   Indeed, the usage of NLG is evolving extremely rapidly, with new use cases\index{use cases} appearing every month.  What is possible to accomplish in specific use cases is also changing fast, as better technology opens up new possibilities.

As always with this book, I will focus on fundamentals, not the latest developments.  I will first discuss generic issues in applying NLG to real-world use cases.  Then I will look at several  long-standing use cases of NLG: journalism, business intelligence,  summarisation, and healthcare.

\section{Key Attributes of Successful NLG Applications}

\subsection{Volume and Scalability}\label{sec:scalability}\index{volume}\index{scalability}
NLG systems generally need to generate a large number of texts in order to be cost-effective from a commercial perspective; this is related to return on investment\index{return on investment} (Section~\ref{sec:ROI}).  For example, suppose an NLG system generates a daily summary of the London stock market (on weekdays, not weekends), and that it would take a
journalist\index{journalism} one hour to write such a summary manually.  If the journalist is paid £20 per hour, this means the system saves £20*52*5 = £5200 per year.  The cost\index{cost} of building and maintaining the NLG is likely to be considerably higher than this!   On the other hand, if the system generates 60 stock market summaries each weekday (one summary for each of the 60 largest global stock markets),  then it saves £312K per year, which is more likely to be commercially viable.

It is for this reason that successful commercial NLG systems tend to focus on high-volume use cases, such as generating large numbers of local news stories, business intelligence reports, or email summaries.  For example, in sports reporting, it probably makes more commercial sense to use NLG to produce a large number of stories about local teams, instead of a small number of stories about world-leading teams.

NLG systerms built around prompted models\index{models!prompted} can be considerably cheaper to create than systems built using rules or by training or fine-tuning models.  However, if the goal is to create a product (as opposed to an in-house tool that customers do not see), then even if coding costs are massively reduced, there still will be substantial costs for requirements analysis (Chapter~\ref{requirements}), evaluation (Chapter~\ref{evaluation}), testing (Section~\ref{sec:testingnlg}), and maintenance (Section~\ref{sec:maintenance}). Plus of course commercial costs such as marketing.  So again volume makes commercial sense.

Of course NLG can be justified in other ways as well, for example the BBC Election Reporter\index{BBC Election Reporter} (Section~\ref{sec:bbcelection}) was justified on the basis of speed (need to get lots of election-results stories written quickly) as well as volume.  But even here, the system only made sense for generating hundreds of stories about local election results, it would not have been a viable solution to producing a small number of stories about national results.

A related concept is \emph{scalability}\index{scalability}; NLG systems should be usable in many contexts.  In healthcare, for example, there is limited commercial value in an NLG system which only works in one hospital, such as Babytalk\index{Babytalk} (Section~\ref{sec:nlgbenefits}).  We want systems which can be deployed in many hospitals and hence will work with different systems, sensors, equipment, clinical procedures, and  administrative procedures.

Of course, the difficulty of scalability depends on the use case.  For example, Business Intelligence\index{business intelligence} use cases are often scalable and generalisable at least to some degree, in part because the underlying processes and data are often similar in different companies.

\personal{I once worked a research project whose goal was to help non-verbal children\index{children} create stories about their school days for their parents \cite{black-hwst12,TintarevHWST}.  We worked with a small number of children and achieved good success with them, but it was clear that our solutions would not work for many other non-verbal children.  Children who cannot speak are very different from each other; different levels and types of cognitive impairment, different physical problems, different backgrounds and attitudes, etc.  In short we cold build bespoke solutions for individual children, but not scalable solutions which worked for a large number of non-verbal children.}

\subsection{Data Availability}\index{data availability}
We cannot build a data-to-text\index{data-to-text} NLG system unless we have access to the data that the system is describing or summarising; if we are building a neural NLG system which is trained or fine-tuned, we also need access to training data\index{training data}.  Data should also be of high-quality.  These issues can constrain where commercial NLG makes sense.

For example, a limitation on some NLG systems used in sports reporting is that they do not have access to visual data.   A human journalist\index{journalism} who observes a game will add phrases such as \lingform{Jones was in tears after missing a shot} or \lingform{Brown jumped for joy after scoring}, this human-interest content makes stories more interesting to readers.  But an NLG system cannot do this if it has access to scores and player performance data but not camera-feeds of players faces.

\subsection{Accuracy}\index{accuracy}
Neural NLG systems make mistakes\index{errors} (Section~\ref{sec:qualityaccuracy}).  Mistakes are more tolerable in some use cases than in others.  We can sometimes estimate the number of acceptable mistakes in a use case by analysing mistakes in texts written by people.
For example, Thomson et al \cite{thomson-csl23} found an average of 1.5 factual errors in human-written sports stories.  So 1.5 mistakes may be acceptable (although of course not desirable) in a sports story;  it would not be acceptable in medical reporting.

In many contexts the best way to produce accurate texts is for humans and NLG systems to work together, for example by asking human writers to edit\index{post-editing} the output of NLG systems (Section~\ref{sec:humanchecking}).  If done well, this process can produce texts which are more accurate than texts produced just by humans (because the NLG system will not make human-like errors like copying the wrong number) or just by NLG systems (because the humans will not hallucinate untrue information).

On the other hand, human checking and editing of NLG outputs must be done well in order to be effective. For example, CNET published a number of articles produced by an NLG system, which had been checked by human editors, which  nevertheless had serious and potentially harmful mistakes, such as giving inaccurate advice about personal finance.\footnote{\url{https://edition.cnn.com/2023/01/25/tech/cnet-ai-tool-news-stories/index.html}}

Accuracy is especially important if there are safety\index{safety} concerns (Section~\ref{sec:safety}); Section~\ref{sec:medical-safety} discusses safety concerns in medical applications of NLG.

\subsection{Maintainability and Adaptability}\index{maintenance}\index{adaptability}
As discussed in Section~\ref{sec:maintenance}, real-world NLG systems must be maintained and adopted as the domain and user needs change.  In healthcare\index{health}, for example, if medical advice is updated due to new research, NLG systems should adapt their output accordingly.  Unfortunately, this is not always easy to achieve in practice (Section~\ref{sec:datedinfo}); Strickland \cite{strickland2019ibm} points out that this was a major problem with the IBM Watson\index{IBM Watson} health systems.

Of course change is faster in some domains than in others; NLG weather forecast generators, for example, may evolve less quickly than medical NLG systems.  Hence the maintenance burden is different in different applications.  Application builders should estimate likely maintenance costs for their system, as well as development costs.

A related problem is that NLG systems should be able to adapt their output to local circumstances.  Figure~\ref{fig:chatgptlocal} shows an example of an adaptation failure; ChatGPT\index{ChatGPT} responds to a query with information which makes sense in the US but not in the UK, since UK patients generally do not have health insurance and do not need to deal with insurance providers.  Perhaps ChatGPT thought that \lingform{Aberdeen} referred to an American town, but this is odd because Aberdeen, UK is ten times bigger than any of the American Aberdeen's.

\begin{figure}
\lineacross\\
\emph{User}: I live in Aberdeen and want to see a respiratory health doctor, what should I do?\\

\emph{ChatGPT}:
If you're in Aberdeen and you're looking to see a respiratory health doctor, here are steps you can take:\\

\begin{enumerate}
\item Contact Your Primary Care Physician:\\

Start by reaching out to your primary care doctor or general practitioner. They can assess your symptoms, provide initial care, and refer you to a specialist if needed.\\

\item Insurance Coverage Check:\\

Check with your health insurance provider to understand your coverage and whether you need a referral to see a specialist.\\

\item Specialist Referral:\\

If your primary care doctor determines that you need to see a respiratory health specialist, they can provide you with a referral. This referral may be necessary for insurance purposes.\\

\item \emph{etc}
\end{enumerate}
\caption{ChatGPT advice is not appropriate for Aberdeen, UK, because the UK does not have an insurance-based health system}\label{fig:chatgptlocal}
\lineacross
\end{figure}

\subsection{Acceptability and Trust}\label{sec:acceptability}\index{acceptability}\index{trust}
NLG systems will only be used if people trust and accept them.  This is partially a \emph{change management}\index{change management} issue, and it is useful to be aware of the literature on managing and enabling organisational change, including introducing new technologies\footnote{\url{https://en.wikipedia.org/wiki/Change\_management}}.

For example, it is difficult to get hospitals and doctors\index{doctors} to use machine learning models for medical diagnosis; doctors see diagnosis as a core part of their job which they do not want to be automated, and hospitals do not see diagnosis as a major \scare{pain point}\index{pain point} which needs to be automated \cite{strickland2019ibm}.
Paul Meehl \cite{meehl1954clinical} showed in 1954 that simple regression models could do some medical diagnosis tasks better than the average doctor; however Meehl's models were never used.  Strickland analysed the failure of IBM's Watson\index{IBM Watson} health system \cite{strickland2019ibm}, and highlighted factors such as poor fit to clinical workflows\index{workflow}, and lack of understanding of requirements\index{requirements}, that is what doctors and hospitals actually wanted help with.

\personal{Many years ago, the Scottish AI pioneer Rob Milne (now sadly deceased) told me that the most successful AI applications were those that targeted \scare{peripheral} tasks, i.e. tasks that users saw as annoying distractions. It was much harder to build a successful AI application for a core task, i.e. a task that the user saw as a central aspect of his job and perhaps identity.  Diagnosis is  a core task for doctors.}

With regard to trust, domain experts may have more trust in simple models which they feel they understand, compared to black-box neural models.  This is not irrational if we consider that black-box models reflect their training data\index{training data}, which may be biased and noisy.  For example, a doctor may be concerned that a neural LLM used in a patient information system may in some cases repeat inappropriate content from the Internet (Section~\ref{sec:medical-safety}).  There is no way for a doctor to check the black-box LLM to see if this may happen, he needs to trust the system not to do this.

Another issue is that doctors often have additional information (for example visual observation of the patient) which is not available to the NLG system.  It is easier for doctors to assess the impact of this information if they understand how the NLG system works.

My experience (and that of other people working in applied NLG) is that trust is easily lost; even a single mistake can lead to users distrusting an NLG system and refusing to use it.

\subsection{Conforming to genre and sublanguage}\index{genre}\index{sublanguage}
In some applications, texts must be written in a way which conforms to a \emph{sublanguage} \cite{kittredge1982sublanguage}.  That is they need to use words and syntactic structures which are expected in the target domain\index{domain} and genre.

For example, the SumTime\index{SumTime} system generated marine weather forecasts\index{weather forecasts} \cite{REITER2005137}.  An example is shown in Figure~\ref{sumtime-sublanguage}.  Note that this sentence is \emph{not} a grammatically correct sentence according to the normal rules of English grammar; for example it does not have a correctly inflected clausal verb.  It does however fit the rules of the marine-forecast sublanguage, so it is perfectly acceptable.

\begin{figure}
\lineacross\\
W 10–15 backing SW by mid afternoon and S 13–18 by midnight.
\caption{Extract from SumTime forecast, in appropriate sublanguage}\label{sumtime-sublanguage}
\lineacross
\end{figure}

Similarly, texts intended for clinicians and medical professionals should use appropriate medical terminology\index{terminology} and language.  
Indeed, Moramarco et al's \cite{moramarco-etal-2022-human} evaluation of Note Generator\index{Note Generator} (Section~\ref{sec:NoteGenerator2}) identified several cases where inappropriate acronyms were used, and classified these as errors.

Precision terminology is also very important in  some cases.  For example, \lingform{bad headache} and \lingform{migraine} are sometimes used interchangeably by patients, but to doctors they have different meanings.

\personal{Patients may want to see some medical terminology, even if they do not fully understand it.  For example, in our Babytalk\index{Babytalk!BT-Family} system for parents (Section~\ref{sec:babytalk}), we initially tried to remove medical terminology.  However,  some parents didn't like this and said that we were \scare{dumbing down} texts and patronising them, so we put some medical terminology back into the parent reports, even though in some cases this made them harder to understand.}

\section{Journalism}\label{journalism}\index{journalism}

One of the oldest use cases for NLG is journalism and media, that is using NLG to produce stories for newspapers and web sites.  Typically data-to-text NLG is used for \emph{data journalism}\index{journalism!data} where an article is largely based on data of some kind; for example financial stories that summarise financial data, or sports stories that summarise what happened in a sports match based on match data.  This is sometimes called \emph{automatic journalism}\index{journalism!automatic} or \emph{robo-journalism}.

Diakopoulos \cite{diakopoulos2019automating} presents a detailed summary of \scare{Automated Content Production} in  journalism up to 2019.  In this period, the focus was on using relatively simple rule-based NLG\index{rule-based NLG} techniques (sometimes just templates) to generate drafts of relatively short and straightforward articles; these articles were post-edited by journalists (Section~\ref{sec:humanchecking}) before they were released. 

Large language model\index{large language models} technology (which appeared after 2019) allows more complex articles to be generated.  At the time of writing, there is a lot of excitement about using ChatGPT and other large language models in journalism; they should expand the power of automatic journalism by semi-automating production of more types of stories.  Perhaps the technology will also enable the development of new types of news media, such as dialogue systems for news (where users can ask questions about news stories).

\subsection{Example: BBC Election Reporter}\label{sec:bbcelection}\index{BBC Election Reporter}
A perhaps typical example of using relatively simple rule-based NLG techniques to produce media articles is the BBC's election reporter system, which generated short reports about election results in individual constituencies in the 2019 UK general election\footnote{\url{https://www.bbc.co.uk/news/technology-50779761}}.   An example is show in Figure~\ref{fig:bbcelection}.  Stories for constituencies in Wales were produced in Welsh\index{Welsh} as well as English.  Stories were checked\index{post-editing} by human journalists before they were released; in a few cases the journalists added additional content, generally deeper analysis of results in key races.

\begin{figure}
\lineacross\\
Florence Eshalomi has been elected MP for Vauxhall, meaning that the Labour Party holds the seat with a decreased majority.

The new MP beat Liberal Democrat Sarah Lewis by 19,612 votes. This was fewer than Kate Hoey's 20,250-vote majority in the 2017 general election.

Sarah Bool of the Conservative Party came third and the Green Party's Jacqueline Bond came fourth.

Voter turnout was down by 3.5 percentage points since the last general election.

More than 56,000 people, 63.5\% of those eligible to vote, went to polling stations across the area on Thursday, in the first December general election since 1923.

Three of the six candidates, Jacqueline Bond (Green), Andrew McGuinness (The Brexit Party) and Salah Faissal (independent) lost their £500 deposits after failing to win 5\% of the vote.

\textbf{\textit{This story about Vauxhall was created using some automation.}}

\caption{Example story from BBC election reporter, from \url{https://www.bbc.co.uk/news/technology-50779761}}
\label{fig:bbcelection}
\lineacross
\end{figure}

The rationale for this project is that the BBC can report election results overall, but it cannot manually produce 650 stories about individual MP races in the required time frame (readers expect to see these stories the morning after the election, not several days later).     The focus on fast production of a large number of news stories, with human oversight and editing, is common in automatic journalism.

This project used Arria\index{Arria} NLG Studio, which is a commercial tool for building rule-based NLG\index{rule-based NLG}.  Implementation details are not publicly available, but interested readers may wish to look at Valteri \cite{leppanen-etal-2017-data,Melin18}, which is an election reporting system built by an academic research group that also used rule-based NLG.

\subsection{Types of news}
\subsubsection{General news}\index{news!general}
A number of NLG systems have been used to produce general news articles  \cite{diakopoulos2019automating}, using an approach similar, at least at a conceptual level, to the BBC election reporter (Section~\ref{sec:bbcelection}).  For example, Radar\footnote{\url{https://pa.media/radar/}}\index{Radar} used Arria\index{Arria} NLG Studio to generate a large number of localised articles from data sets about politics, crime, health, traffic, and so forth.  For instance, the Radar system would ingest a UK-wide dataset about obesity rates, extract the relevant information for a particular town or city (such as Aberdeen), and then generate a story based on this data (ie, obesity in Aberdeen).  This story would be sold to a local newspaper, which would release it, possibly after being edited by a journalist to add more local content and context.

The advantage of this approach for journalists is that it made it much easier for local newspapers, which often were short-staffed and lacked journalists with domain expertise, to publish data-based articles on health, crime, etc.  A large national newspaper such as \emph{New York Times} or \emph{Guardian} will have in-house teams of specialist data journalists, but small local newspapers cannot afford this.

Large language models\index{large language models} such as ChatGPT may have a large impact in this area; technology and use cases are evolving very rapidly at the time of writing.

\subsubsection{Sports Reporting}\index{news!sports}
One of the most active areas of NLG in media is sports reporting.   Modern sporting events are often monitored by sensors which produce large amounts of data; there is also often a large amount of historical data about the players and teams involved in the match. This data can be used by NLG systems to generate many kinds of articles, including live updates during a match, post-match game descriptions, and pre-match summaries of the teams involved. Articles can be targeted towards different audiences, such as the general public, fans of one of the teams, or people placing bets; they can be delivered on social media, websites, and indeed printed stories.  NLG technology can also be used to generate reports from sports data for coaches and talent scouts.

Sports stories are an area where \emph{variation}\index{variation} (Section~\ref{sec:variation}) is very important.  Sports fans read lots of sports stories, and it is essential that these stories use different words and sentences (even when conveying similar information), otherwise readers will get bored and may stop reading the stories.  Certainly in the sports NLG projects I have been involved in, considerable effort was put into supporting extensive variation.

Anyways, the richness of data and use cases makes this an exciting area for NLG.   It is also an area where there is a lot of interest in using neural NLG techniques.  In part this is because from a quality-factor perspective (Section~\ref{req-criteria}), its is essential that sports stories be well-written and engaging.  Errors\index{errors} (such as hallucination) are of course not desirable but may have less impact than in safety-criticial contexts such as clinical decision support. An error in a story about a children's basketball game is not going to kill anyone!

\subsubsection{Financial News}\index{news!financial}

\begin{figure}
\lineacross\\
After climbing steadily through most of the morning, the stock market was pushed downhill late in the day. Stock prices posted a small loss, with the indexes turning in a mixed showing yesterday in brisk trading.

The Dow Jones average of 30 industrials surrendered a 16.28 gain at 4pm and declined slightly , finishing the day at 1083.61, off 0.18 points.

\caption{Stock market story from world's first data-to-text system \cite{kukich-1983-design}}
\label{fig:kukich-stock}
\lineacross
\end{figure}

There is also a lot of interest in in using NLG in financial journalism.  Investors and business people more generally are very interested in knowing what is happening in companies of interest to them.  There are thousands of companies listed on the London Stock Exchange, New York Stock Exchange, NASDAQ, etc, and it is difficult for human journalists to write stories about business developments in such a large number of companies.  But since basic data about listed companies is available from the stock markets\index{stock markets} themselves and also from relevant regulators\index{regulator}, NLG technology can be used to generate large numbers of stories about these companies, as well as stories about stock market performance.

Indeed, the very first work on data-to-text NLG, by Karen Kukich in 1983 \cite{kukich-1983-design}, was on generating summaries of stock market activity.  A simple example from Kukich's system is shown in Figure~\ref{fig:kukich-stock}.  Of course modern financial reporting NLG systems generate much more sophisticated texts about a wide range of financial stories (not just stock market updates), but the core concept is the same; the system takes financial data as input and produces a story about this data as output.

As with sports, a key issue is data availability\index{data availability}.  Generated stories are richer when the NLG system has access to more data about companies and markets, perhaps including data extracted from free-text documents using natural language understanding techniques. 

From a quality factor perspective, financial stories should of course to be well-written and varied\index{variation}, like all news stories.  They also must be accurate, because investors may make decisions based in part on news stories.  Accuracy\index{accuracy} in financial news is perhaps not as important as in medical decision support, but it is considerably more important than accuracy in sports reporting.

From a technology perspective, at the time of writing many financial news generators use rule-based techniques, but there is huge interest in generating financial stories using machine learning and neural techniques. However, it is worth noting that Thomson-Reuters\index{Thomson-Reuters} generated financial stories for a number of years using a very interesting machine learning approach (essentially ML techniques were used to learn and select templates\index{templates} \cite{kondadadi-etal-2013-statistical}), but then withdrew this system, I suspect in part because journalists wanted more explicit control over the content of generated stories.  In short, this particular ML NLG system was not acceptable to journalists in real-world usage (Section~\ref{sec:acceptability}).  Of course other ML techniques may be more successful, but acceptability\index{acceptability} can only be assessed when the system is in production\index{production} real-world usage by many journalists.

\subsection{Fake News}\index{news!fake}
Neural NLG systems make mistakes.  Reputable journalists and media organisations will of course do their best to detect and minimise mistakes.  However, there also are malicious and unscrupulous agents who deliberately use NLG to to produce stories which are not true.   Large language models\index{large language models} unfortunately can be very good at producing \scare{fake news} \cite{solaiman2019release}.

A lot of fake news is simply designed to attract readers and hence get money from advertisers.  Of course some media outlets have printed fake stories about alien abductions and the like for decades, without any use of AI and NLG! But creating plausible-sounding stories about visits from little green men is easier and quicker with NLG.

More damaging is using AI to create fake material which people pay money to read.  For example, at the time of writing there is a lot of concern about people buying useless AI-written travel guides\footnote{\url{https://www.nytimes.com/2023/08/05/travel/amazon-guidebooks-artificial-intelligence.html}}.

Perhaps most damaging is fake news which tries to influence people.  In the last three months of the 2016 US Presidential election, fake news got more engagement in Facebook than genuine news\footnote{\url{https://www.buzzfeednews.com/article/craigsilverman/viral-fake-election-news-outperformed-real-news-on-facebook}}.  These articles were presumably written by humans, not NLG systems.  It is not clear to what degree NLG systems are currently being used to generate fake news for such purposes, but this is a major safety concern at the societal level \cite{distsafety}.

\section{Business Intelligence}\label{sec:bi}\index{business intelligence}
Another long-standing use case of NLG is Business Intelligence (BI)\footnote{\url{https://en.wikipedia.org/wiki/Business\_intelligence}}. BI tools help people understand data about their organisation (business, charity, government, etc).  Amongst other things, this includes data about:
\begin{itemize}
\item Sales, profits, losses;
\item Employees;
\item Budgets;
\item Inventories and logistics;
\item etc.
\end{itemize}
For example, a BI tool could help an organisation understand that most of its sales growth is in Asia, turnover in IT staff is growing rapidly, its inventory of laptops is too large, etc.  This in turn helps managers and other decision makers make good decisions.

Most commercial BI tools, such as Tableau\index{Tableau} and PowerBI\index{PowerBI}, focus on graphical\index{graphics} presentation of data.  However, in some cases textual presentation of data is better (Section~\ref{sec:textgraphics}), not least because it can include analyses and background information to supplement the data.

From a commercial perspective, one of the big advantages of BI as a use case is thats it is often scalable (Section~\ref{sec:scalability}).  Whereas medical NLG systems often need to be tailored for specific hospitals, and journalism NLG systems need to be adapted for specific local markets, all organisations have budgets and employees, and most have sales and inventories.  Furthermore, the data about budgets, employees, sales, and inventories is often held in databases or spreadsheets, many of which have a similar structure in part because of the requirements of tax and regulatory\index{regulator} authorities.  Hence it is possible to build somewhat generic NLG BI systems which can be sold to many customers.
 
 \subsection{Example: Covid Reporter}\label{sec:covid}\index{Covid Reporter}
 
 \begin{figure}
 \lineacross\\
\includegraphics[scale  = 0.75]{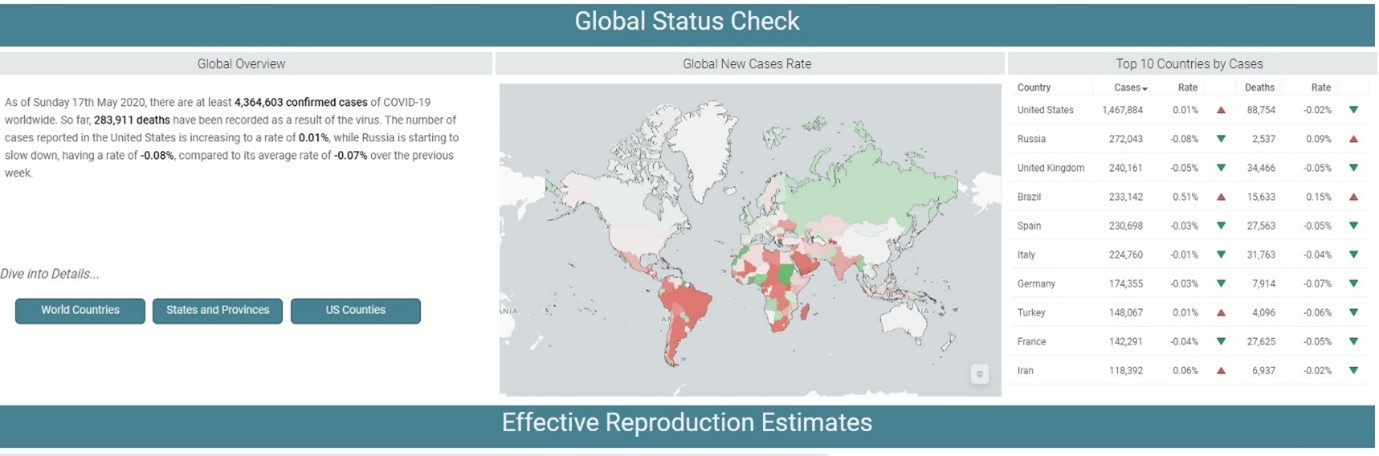}
NLG text in above screenshot (replicated for clarity):\\
As of Sunday 17th May 2020, there are at least \textbf{4,364,603 confirmed cases} of COVID-19 worldwide. So far, \textbf{283,911 deaths} have been recorded as a result of the virus. The number of cases reported in the United States is increasing to a rate of \textbf{0.01\%}, while Russia is starting to slow down, having a rate of \textbf{-0.08\%}, compared to its average rate of\textbf{ -0.07\%} over the previous week.
\caption{Business Intelligence about Covid-19. Taken from \url{https://www.tibco.com/covid19} .  This page no longer exists, it was removed when the pandemic ended.}
\label{fig:covidBI}       
\lineacross
\end{figure}

A simple example is shown in Figure~\ref{fig:covidBI}.  This system was created by Tibco and Arria\index{Arria} to provide information to the  public about the status of the Covid-19 pandemic, using business intelligence techniques.  The full system includes many information displays; the one show in Figure~\ref{fig:covidBI} gives summary information about the presence of Covid-19\index{Covid pandemic}.  It uses an NLG-generated texts to give a high-level summary, a map to show distribution of Covid-19 in different countries, and a table to give more detailed information about badly affected countries.   This\scare{multi-modal}\index{multi-modal} combination of text and graphics is typical, and allows each media to be used to communicate the type of information it is best suited for.

\section{Summarisation}\index{summarisation}
Another long-standing use of NLG is summarising documents.  Indeed, summarisation predates NLG, with some initial work in this area done in the 1950s \cite{Luhn1957} focusing on statistical techniques for  identifying key material in a document,

There are two general approaches to summarisation.  \emph{Extractive summarisation}\index{summarisation!extractive} identifies key sentences and builds a summary from these, while \emph{abstractive summarisation}\index{summarisation!abstractive} creates new sentences to summarise information.  A simple example is shown in Figure~\ref{fig:extractiveabstractive}, which show shows possible extractive and abstractive summaries of an Aberdeen weather forecast\index{weather forecasts}.  The extractive one is simply the first sentence in the text, while the abstractive summary integrates information from several source sentences into the summary sentence.

\begin{figure}
\lineacross\\
\emph{Source text (weather forecast for Aberdeen):}\\
A mix of brighter spells and showers. Best of the brightness in the morning, particularly across Moray and western Aberdeenshire. Showers isolated early before building through the afternoon. Winds light. Maximum temperature 8 °C.\\

\emph{Extractive summary:}\\
A mix of brighter spells and showers. \\

\emph{Abstractive summary:}\\
A mild day with a mix of brighter spells (especially in the morning) and showers. \\

\caption{Extractive and abstractive summaries of a weather forecast}
\label{fig:extractiveabstractive}
\lineacross\\
\end{figure}

Another key distinction is between \emph{single-document summaries}  (where one document is summarised, as in the example in Figure~\ref{fig:extractiveabstractive}) and \emph{multiple-document summaries} (where multiple documents are summarised, for example a summary of a set of news articles).  The latter are often more useful, especially in contexts (such as summarising reviews or email chains) when the reader wants to quickly get an understanding of a set of documents.

There are many uses of summarisation, including:
\begin{itemize}
\item \emph{Meeting summaries:}  The Note Generator\index{Note Generator} system (Section~\ref{sec:NoteGenerator2}) is an example of a system that generates summaries of meetings.  There are many other tools in this space, which generate different types of summaries (key points, action items, etc.) of meetings.  Rennard et al \cite{rennard23} survey research in this area.
\item \emph{Email summaries:}  Likewise there are many tools which summarise email messages (individually or collectively), in order to help people manage overflowing inboxes; for example Microsoft Copilot in Outlook summarises email threads.
\item \emph{News summaries:}  Extensive research has been done on summarising news articles, initially driven by the desire of intelligence agencies to help analysts quickly identify key information \cite{mani2002summac}.  Recent work in this area has also looked at summarising social media.
\item \emph{Review summaries:} There also is a lot of interest in summarising product reviews, in this case it is essential that the summary cover a set of reviews (a summary of one review is less useful).  Amazon provides summaries of reviews in its ecommerce site.
\end{itemize}
At the time of writing, summarisation is usually done using large language models\index{large language models}.

\subsection{Example: Summarising emails with Google Bard}\index{Bard}
Google Bard is one of many systems that can summarise emails, including chains and sets of emails as well as single emails.  I show a simple Bard example in Figure~\ref{fig:bardemail}.  In this case, Bard has found the relevant email (which is pretty straightforward), extracted key information from the email which is relevant to my question, and created a sentence which communicates this information.  This is an example of generating an abstractive summary\index{summarisation!abstractive} of a single document.

\begin{figure}
\lineacross\\
\emph{Question to Bard:}\\
Check my emails, what is the next book in my bookclub\\

\emph{Response from Bard:}\\
The next book for the AJSCC book club is Odessa Stories by Isaac Babel. The meeting will be on Thursday, January 18th, 2024 at 8pm on Zoom.\\

\emph{Source email:}\\
\textbf{Subject}: dates of next few meetings for AJSCC book club\\
We had a wonderful evening of discussion around Susan's book choice!  Lovely to have so many of us participating.  Here are full details of our next two books for January and March\\

date:           Thursday 18th January 2024\\
time:            8.00pm\\
location:      Zoom\\
book:          Odessa Stories\\
author:        Isaac Babel\\
publisher:    Pushkin Collection\\
published:   2016\\
ISBN:          978-1782271666\\
chosen by:  Ehud\\

\emph{additional email content not shown}

\caption{Example of Bard summarising an email}
\label{fig:bardemail}
\lineacross\\
\end{figure}

\section{Medical applications}

For most of my career, I have been very interested in applications of NLG to healthcare\index{health}.  In principle, we can use NLG in all sorts of ways to help clinicians\index{doctors} and patients, including automated reporting (save doctor's time and perhaps reduce errors), clinical decision support (help clinicians make good decisions and avoid mistakes), patient information (keep patients informed and help them make appropriate decisions), and behaviour change (encourage patients to adopt healthier behaviours).  Success in this area would have huge impact, both financially (according to WHO, healthcare is over 10\% of the world's economy) and (more importantly) on people's lives.   Indeed, when I was asked to write a long-term vision for my research in the early 2000s, I said that I wanted to use NLG to help patients live healthier lives.

However, healthcare is also a very challenging area to work in, especially if our goal is real-world impact\index{impact} instead of academic papers.  In the early 1990s, a colleague told me that the challenge with AI generally in healthcare was building scalable\index{scalability} robust solutions which doctors and health organisations would adopt.  Building a system that works well on one data set or in one hospital is much easier than building a system which can be widely used; and building a widely-usable system is much easier than getting health organisations and professions to actually use the system (Section~\ref{sec:acceptability}).

In medicine, it is also the case that mistakes can hurt or even kill people; for this reason, we need to be careful when using neural language models.  Indeed, the terms of use of the BLOOM\index{BLOOM} language model explicitly prohibit using BLOOM \scare{to provide medical advice and medical results interpretation}\footnote{\url{https://huggingface.co/spaces/bigscience/license}}.

In short, NLG applications in healthcare have huge potential, but we must also keep in mind that it is difficult to deploy NLG technology on a large scale in healthcare, even if the technology is safe and effective.

\subsection{Use case: Reporting}\label{sec:medicalreporting}\index{reporting}
Many clinicians are most excited by NLG systems that automate or otherwise support reporting tasks.  Doctors\index{doctors} are expected to write large numbers of reports and documents, and this takes up a considerable amount of their time.  Most doctors see this as a peripheral task (Section~\ref{sec:acceptability}); it needs to be done but they welcome innovations which reduce the amount of time they spend on writing documents. Diagnosis, in contrast, is a core task, where automation is resisted; Strickland \cite{strickland2019ibm} discusses a related point.  Hence, from a change management\index{change management} perspectrive (Section~\ref{sec:acceptability}), many clinician welcome AI support for reporting, but not for diagnosis.

In other words, doctors will probably be receptive to \scare{We will use AI to automate paperwork so that you have more time for careful decision-making}.  However, they may not be receptive to \scare{We will use AI to automate decision-making so that you have more time to carefully complete your paperwork}.

From a commercial evaluation perspective, the benefits of report automation (time saved by doctors) are also relatively straightforward to measure. Changes in patient outcomes can be harder to measure; for example, if we are interested in survival rate 5 years after treatment, we will have to wait five years to get this information.

From a scalability\index{scalability} perspective (Section~\ref{sec:scalability}), reporting practices differ in different healthcare organisations, which means that reporting systems are often specific to individual hospitals or even hospital units.  An exception is regulatory reporting, which of course is standardised to the requirements of the regulator; for example I have seen commercial work on using NLG to generate clinical safety reports (on pharmaceuticals) for regulatory agencies.

\subsubsection{Example: Summarising consultations}\label{sec:NoteGenerator2}
\begin{figure}
\lineacross\\
\includegraphics[scale  = 0.75]{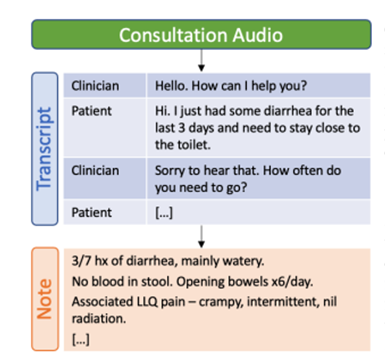}
\caption{Example of Note Generation; this is the same as Figure~\ref{fig:NoteGenerator}}
\label{fig:NoteGenerator2}     
\lineacross  
\end{figure}

\index{summarisation}\index{consultations}
An example of an NLG system used for medical reporting is Note Generator\index{Note Generator} (described in Section~\ref{IntroConsultationSummary}). It generates a summary of a doctor-patient consultation, which is entered into the patient's medical record after being checked and edited by a doctor.  An extract from Note Generator was shown in Figure~\ref{fig:NoteGenerator};  I repeat this in Figure~\ref{fig:NoteGenerator2} for convenience.

Manually writing a summary of a doctor-patient consultation can take 2-3 minutes, which is a sizeable chunk of the ten minutes allocated to consultations with general practitioners (GPs) in the UK.  Partially automating this task gives doctors more time to interact with patients, which they welcome.  Moramarco \cite{moramarcophd} analysed real-world usage of Note Generator (Section~\ref{sec:historicalcomparisons}) and
reported that doctor using Note Generator (ie, post-editing a Note Generator summary) could generate a summary in 9\% less time compared to manually writing summaries from scratch.
Summaries produced by post-editing\index{post-editing} Note Generator drafts were also more accurate\index{accuracy} and contained fewer errors\index{errors} than manually written summaries. 

\subsection{Use case: Patient information and behaviour change}\index{patient information}\index{behaviour change}
It is a truism in medicine that the best way to improve most people's health is to change their behaviour.  This includes adopting healthier life styles (diet, exercise, smoking, etc), correctly complying with medical instructions (e.g., take medication correctly), and contacting healthcare personnel when appropriate.    Indeed, it is hard to see how healthcare systems can cope with increasingly elderly populations unless people look after themselves better.

Also, modern medicine expects patients to make informed decisions about their healthcare, and informed decisions are only possible if patients understand benefits and risks\index{risk}.  This is difficult for many people, especially if probabilities are involved; many people have a weak understanding of risk and probabilities (Section~\ref{sec:riskcommunication}).

Hence there is a lot of potential in building NLG systems that directly interact with patients in order to educate people about health consequences of their behaviour, encourage behaviour change,
and/or explain important information.   I personally think that this is an area of medicine where NLG could potentially make very important contributions.

A general rule is that we can help and encourage people to do things they would like to do, but its very hard to change intentions.  For example,  looking at smoking cessation, we can give suggestions to people who want to stop smoking about how to achieve this goal; its much harder to try to convince committed smokers to stop smoking.  Similarly we can support people who want to make informed decisions by giving them appropriate information, but giving information isn't going to do much good if the patient wants to delegate all decisions to his or her doctor.

Behaviour change is difficult even for people who want to change, especially on a long-term basis.  For example, most adult smokers in the UK want to stop smoking, but quitting is still very hard.  Similarly most overweight and obese people would like to lose weight, and may achieve this temporarily (e.g. by following a diet), but find it harder to lose weight on a sustained long-term basis.

In part for the above reasons, it is very useful for NLG systems (and indeed AI systems more generally) in this space to be integrated into the broader healthcare system, so that they are just one of many tools which people can use to help them change their behaviour.  Unfortunately, differences between healthcare systems can make this difficult.

\subsubsection{Example: encouraging smoking cessation}\label{sec:stop2}
The STOP\index{STOP} system (described in Section~\ref{sec:stopsmoking}) generated personalised smoking-cessation advice for smokers, based on their answers to a smoking questionnaire.   An extract from STOP was shown in Figure~\ref{fig:stopleaflet}; I repeat this in Figure~\ref{fig:stopleaflet2} for convenience.

\begin{figure}
\lineacross\\
\textbf{Dear Ms Cameron}
\blankline\\
Thank you for taking the trouble to return the smoking questionnaire that we sent you.  It appears from your answers that although you're not planning to stop smoking in the near future, you would like to stop if it was easy.  You think it would be difficult to stop because \emph{smoking helps you cope with stress}, \emph{it is something to do when you are bored}, and \emph{smoking stops you putting on weight}.  However, you have reasons to be confident of success if you did try to stop, and there are ways of coping with the difficulties.\\
\caption{Part of a STOP leaflet; this is the same as Figure~\ref{fig:stopleaflet}}\label{fig:stopleaflet2}
\lineacross
\end{figure}

Unfortunately, a randomised controlled clinical trial showed that the STOP system was not effective, i.e. people who received a fixed letter about smoking-cessation (not produced by an NLG system) were at least as likely to stop smoking as people who received a letter produced by STOP \cite{Lennox1396,REITER200341}.

\subsubsection{Communicating risks and probabilities}\label{sec:riskcommunication}\index{risk}\index{probability}
It is often necessary in patient information systems to communicate risks and probabilities.  Unfortunately many patients in the UK. US, and elsewhere have limited numeracy\index{numeracy} skills and may not understand numbers, so risk should be communicated to them in other ways, such as with words.

Some of the challenges in communicating risk linguistically are described by Berry \cite{berry2006health,berry2004ebook,berry2011}; see also Hommes et al \cite{hommes-etal-2019-personalized}.  For example:
\begin{itemize}
\item Different people interpret descriptors such as \lingform{likely} and \lingform{common} in different ways.
\item It matters how risk is framed (risk of success vs risk of failure).
\item Is risk communicated in absolute or relative terms (\lingform{this procedure has a 5\% chance of success} vs. \lingform{this procedure has a 50\% higher chance of success than the alternative})?
\end{itemize}
Even more complexity arises if we wish to communicate the reliability of a risk, e.g. \lingform{the model says you have a 10\% chance of developing lung cancer, however the risk may be higher because the model has ignored the fact that  both of your parents died of lung cancer} \cite{sivaprasad-reiter-2024-linguistically}.

Unfortunately, while there is a large general literature on risk communication, I am not aware of much research in the NLG community on communicating risk in a way which takes these issues into account.

\subsection{Use case: Clinical decision support}\index{decision support}
NLG systems can summarise information and help doctors\index{doctors} make decisions.  This is an appealing concept in principle, but success is difficult, in part  because of the issues mentioned in Section~\ref{sec:medicalreporting}.  Doctors enjoy decision making and generally are pretty good at it;  they also have access to additional information sources not available to the NLG system (eg, observing and talking to the patient). So help is often neither wanted nor needed \cite{strickland2019ibm}.  

Safety\index{safety} issues (Section~\ref{sec:safety}) are very important in this area; mistakes could cost lives and also could lead to massive lawsuits.  Trust\index{trust} is  paramount; if doctors don't trust an NLG system (and even a single mistake can destroy trust), they will not use it.  Adding explanations, if done well, could enhance trust.

\personal{Many years ago I was very excited by using NLG as a decision support tool, by summarising key information for clinicians.  Looking back at this now, I think this may be possible technically, but getting clinicians to use such a system would be very hard.  Adoption will probably be significantly easier for systems focusing on the other use cases discussed here (reporting, behaviour change, business intelligence).}

\subsubsection{Example: ICU decision support}\label{sec:babytalk2}
Babytalk\index{Babytalk} (described in Section~\ref{sec:babytalk}) generated decision-support material, nursing shift handover reports, and parent reports for babies in a neonatal intensive care unit, using information extracted from the baby's electronic patient record.  An extract from the Babytalk systems was shown in Figure~\ref{fig:BabytalkOutputs};  I repeat the extract from the Babytalk decision-support system in Figure~\ref{fig:BabytalkOutputs2} for convenience.

\begin{figure}
\lineacross\\
\emph{Example BT45 output (extract)}:\\
By 11:00 the baby had been hand-bagged a number of times causing 2 successive bradycardias. She was successfully re-intubated after 2 attempts. The baby was sucked out twice. At 11:02 FIO2 was raised to 79\%.\\

\caption{Example outputs from Babytalk BT45\index{Babytalk!BT45} decision-support system; this is the same as shown in Figure~\ref{fig:BabytalkOutputs}}
\label{fig:BabytalkOutputs2}
\lineacross
\end{figure}

\subsection{Medical business intelligence use cases}\label{sec:medicalbi}\index{business intelligence}
Of course medicine is a huge area, and there are many other ways we can use NLG in healthcare.  In particular, NLG can be used in many healthcare contexts to provide business intelligence (BI) (Section~\ref{sec:bi}) which supports non-clinical processes such as logistics and quality control.

For example, we can use NLG BI to support logistics\index{logistics}, and help insure that equipment, medication, blood supplies, ambulances, etc are in the right place.  I am not aware of research projects in this space, but I have seen some commercial uses of NLG that support BI in healthcare.  A medical colleague once told me that she thought logistics was the best place to apply AI in healthcare, since it is extremely important, and generally raises fewer concerns about safety, acceptability to clinicians, etc.

Another use of NLG+BI is to provide quality-of-service information on healthcare, either to patients (summarising quality and outcome data about a hospital) or to doctors (summarising how well they are doing at reaching various quality targets).  Management can also use NLG+BI to analyse efficiency and outcome data.  Again I'm not aware of research projects in this area, it seems mostly of interest to the commercial NLG world.

Last but not least, we can use NLG+BI to summarise and explain public health data, especially in local areas where public health officials may lack data analysis expertise.  I've seen discussions about this but not actual projects, which is a shame.

\subsection{Safety}\label{sec:medical-safety}\index{safety}
Safety (Section~\ref{sec:safety}) is obviously very important in medical applications, especially since incorrect medical information or advice could injure or kill people.  Hallucinations\index{hallucinations} are unacceptable in texts that impact medical care, as are omissions\index{omissions} of key information.  Singhai et al \cite{singhal2023large} present a formal evaluation of a medical language model, MedPaLM\index{MedPaLM}; Tamayo-Sarver \cite{TamayoBlog} describes some of the failings he has encountered when trying to apply ChatGPT\index{ChatGPT} to real-world medical situations.  Readers should keep in mind that humans as well as AI systems make mistakes, and a workflow\index{workflow} which combines human and AI can make fewer mistakes (Section~\ref{sec:NoteGenerator2}) and hence perhaps be safer, than a workflow that uses just humans (or just AI systems).

When systems are interacting with patients, it is also essential to minimise adverse emotional\index{emotion} effects.  Of course discussing medical details and options can be inherently depressing for patients \cite{van2024potential}, but systems should not unnecessarily make people feel depressed, inadequate, etc.
As discussed in Section~\ref{sec:upsetting}, Balloccu \cite{balloccuphd} asked domain experts to analyse texts produced by ChatGPT to response to dietary struggles, and he found many cases responses were factually correct but inappropriate and indeed unsafe because of their potential emotional impact.

Balloccu's domain experts also expressed a general concern ChatGPT seemed to largely be trained on texts
from internet forums which were not written by healthcare professionals.

Other medical professionals have made similar comments to me about the outputs of large language models\index{large language models}; the text was correct but not an appropriate thing to say to a patient.  In general, they feel that health information should come from reputable medical sources, and are concerned that Internet-trained LLMs may rely on sources that are not quality-controlled, such as Reddit\index{Reddit} forums.  Certainly in my own personal experience, advice from well-meaning lay people such as friends and family (as opposed to advise from health professionals) has often been insensitive to emotional issues.

\section{Further Reading}\label{sec:furtherApplications}
The usage of NLG in real-world applications is evolving and changing very rapidly.  There are numerous commercial websites which proclaim the wonders of specific NLG products, as well as numerous consulting firms who offer advice and analysis on NLG applications.  This material is usually very well written and presented, but of course is intended to further the author's commercial objectives.  It also tends to emphasise case studies, demos, and stories instead of rigorous evaluation.

Diakopoulos et al \cite{diakopoulos2024} is an excellent survey of how large language models and NLG are being used by journalists in 2024.  While this is about journalism\index{journalism}, many of the insights about use cases, workflows, ethics, etc. are generic and apply to other applications as well; I highly recommend this survey to anyone developing NLG applications. Diakopoulos also wrote a book in 2019 on \scare{Automating the newsroom} \cite{diakopoulos2019automating}, and helps run a website on NLG in journalism (\url{https://generative-ai-newsroom.com/}); both of these are also excellent sources.

Anyone interested in using NLG or indeed any form of AI in healthcare\index{health} should read Strickland's excellent retrospective on why IBM's Watson\index{IBM Watson} system was not successful in healthcare \cite{strickland2019ibm}; I would love to see more such retrospectives!
Hüske-Kraus's 2003 survey of NLG in clinical medicine in 2003 \cite{huske2003text} is still useful;  technology has changed radically since then, but many of the broader issues have not.

Dale occasionally writes insightful articles about commercial NLG, such as \cite{Dale_2023}.

My personal blog includes many blogs about applications of NLG and the issues raised in this chapter, including:
\begin{itemize}
\item \emph{Change management}: \url{https://ehudreiter.com/2020/01/27/ai-professionals-also-focus-on-change-management/}
\item \emph{BBC Election Reporter} (Section~\ref{sec:bbcelection}: \url{https://ehudreiter.com/2019/12/23/election-results-lessons-from-a-real-world-nlg-system/}
\item \emph{Challenges in medical NLG}: \url{https://ehudreiter.com/2021/06/21/pain-points-in-health-nlg/}
\item \emph{Covid reporter} (Section~\ref{sec:covid}): \url{https://ehudreiter.com/2020/05/21/adding-narrative-to-a-covid-dashboard/}
\item \emph{Sports NLG}: \url{https://ehudreiter.com/2022/03/21/sports-nlg-commercial-vs-academic-perspective/}
\end{itemize}
I have written many other blogs about NLG applications which readers may find useful.


\backmatter

\bibliography{bookreferences}
\bibliographystyle{spmpsci.bst}

\printindex


\end{document}